%% file: paper_en.tex
\documentclass[11pt,a4paper]{article}
\usepackage[utf8]{inputenc}
\usepackage[T1]{fontenc}
\usepackage{geometry}
\usepackage{amsmath,amssymb,amsthm}
\usepackage{booktabs,tabularx,longtable,array}
\usepackage{enumitem}
\usepackage{graphicx}
\usepackage{tikz}
\usetikzlibrary{arrows.meta,positioning,fit,calc,shapes.geometric,shapes.symbols,decorations.pathreplacing,backgrounds}
\usepackage{pgfplots}
\usepgfplotslibrary{groupplots}
\pgfplotsset{compat=1.18}
\definecolor{classical}{HTML}{2156A6}
\definecolor{modelnative}{HTML}{E87817}
\definecolor{timelineBlue}{HTML}{4C78A8}
\definecolor{timelineOrange}{HTML}{D9824B}
\definecolor{timelineGray}{HTML}{A7AFB7}
\definecolor{timelineDark}{HTML}{263238}
\definecolor{softBlue}{HTML}{EAF2FA}
\definecolor{softOrange}{HTML}{FCEFE5}
\definecolor{softGray}{HTML}{F4F6F8}
\usepackage[unicode]{hyperref}
\usepackage{xcolor}
\usepackage{url}
\usepackage{multicol}
\usepackage{float}
\usepackage[font=small,labelfont=bf]{caption}
\usepackage{subcaption}
\hypersetup{colorlinks=true,linkcolor=blue,citecolor=blue,urlcolor=blue}
\newcolumntype{Y}{>{\raggedright\arraybackslash}X}
\usepackage[most]{tcolorbox}
\newtheorem{heuristic}{Heuristic}[section]
\newtheorem{definition}{Definition}[section]
\newtcolorbox{axiombox}[1]{
  colback=blue!4, colframe=blue!50!black,
  fonttitle=\bfseries, title={Axiom #1},
  rounded corners, boxrule=0.6pt,
  left=6pt, right=6pt, top=4pt, bottom=4pt
}
\newcounter{axiom}[section]
\renewcommand{\theaxiom}{\thesection.\arabic{axiom}}

\title{\textbf{Model-Native Computing Architecture}\\[6pt]
\large Envisioning Future System Architecture Through the Lens of Computer Architecture}

\author{
\textbf{Hai Lin}$^{1,2}$\quad \textbf{Hoilam Pao}$^{1}$\quad
\textbf{Shaoxiong Zhan}$^{1}$\quad \textbf{Hai-Tao Zheng}$^{1,2,*}$\\[6pt]
$^{1}$Shenzhen International Graduate School, Tsinghua University\\
$^{2}$Pengcheng Laboratory, Shenzhen, China\\[4pt]
\texttt{ngyygm@outlook.com},\ \texttt{bkl24@mails.tsinghua.edu.cn},\ \texttt{zhansx24@mails.tsinghua.edu.cn}\\
$^{*}$Corresponding author: \texttt{zheng.haitao@sz.tsinghua.edu.cn}
}

\begin{document}
\maketitle

\begin{abstract}
Large language models are undergoing a fundamental transition from ``model technology'' to ``system technology.''
The engineering challenges now dominating practice---cache reuse, context capacity, agent scheduling, permission control---bear an unmistakable resemblance to classical computer-systems problems.
This raises a natural question: if we treat the LLM as a CPU, the KV cache as a processor cache, the context window as main memory, and the agent framework as an operating system, can eight decades of computer-architecture wisdom guide the next generation of model-native computing systems?

This paper pursues that analogy as a \emph{visionary survey}.
We map computer-architecture concepts onto the emerging model-native stack, survey the rapidly growing literature across LLM-as-OS, memory management, agent frameworks, tool protocols, multi-agent coordination, cognitive architectures, and safety governance, and find that each strand addresses a different layer of the same system yet lacks a unifying layered model.
To fill that gap we propose the \textbf{Intelligent Computing Architecture (ICA)}: six functional layers, each with interface contracts and design axioms.
The enduring tension over whether the LLM resembles a CPU or an operating system is resolved by our \textbf{dual-plane architecture}---a probabilistic execution plane (concerned with what \emph{can} be computed) and a deterministic control plane (concerned with what \emph{should} be computed), through which every layer passes as a graded crossover.

At the quantitative level we propose three Amdahl-style design \emph{heuristics}---Semantic Locality, Context Budget, and Agent Speedup---which we are explicit are organizing back-of-envelope models rather than validated scaling laws: we illustrate their parameter ranges with published data and identify systematic predictive validation as the principal open task.
We articulate the boundaries of the analogy, note the fundamental differences between silicon-era and model-era architectures, and propose a research roadmap.
This paper is a conceptual and survey contribution; it presents no new experimental results.
\end{abstract}

\noindent\textbf{Keywords:} LLM systems; computer architecture analogy; Intelligent Computing Architecture; Amdahl-style design heuristics; KV cache; context engineering; agent runtime; dual-plane architecture

\input{en_sections/section_intro.tex}
\input{en_sections/section_bg.tex}

\input{en_sections/section_related.tex}
\input{en_sections/section_framework.tex}

\input{en_sections/section_components.tex}
\input{en_sections/section_icam.tex}
\input{en_sections/section_laws.tex}
\input{en_sections/section_agent_evolution.tex}
\input{en_sections/section_challenges.tex}
\input{en_sections/section_paradigm.tex}
\input{en_sections/section_social.tex}
\input{en_sections/section_implementation.tex}
\input{en_sections/section_roadmap.tex}

\input{en_sections/section_ethics.tex}
\input{en_sections/section_conclusion.tex}

\bibliographystyle{plain}
\bibliography{refs}

\end{document}

%% file: en_sections/section_intro.tex
\section{Introduction: Model-Native Computing Demands an Architecture}
\label{sec:intro}

Large language models are undergoing a fundamental transition, from a \emph{model technology} to a \emph{system technology}.
Starting with GPT-3's demonstration of in-context learning~\cite{brown2020gpt3}, through the LLaMA family's catalysis of open-weight ecosystems~\cite{llama2023}, and onward to the rapid proliferation of tool use~\cite{schick2023toolformer}, retrieval-augmented generation~\cite{lewis2020rag}, autonomous agents~\cite{yao2023react}, and code generation systems~\cite{openhands_paper_2025,sweagent2024}, the research frontier has shifted decisively.
The central question is no longer ``how do we train more capable models?'' but rather ``how do we organize intelligence into systems that are stable, scalable, and auditable?''

This shift is reshaping engineering practice itself.
The engineer's role is migrating from writing code line by line to orchestrating ensembles of specialized agents, compressing entire development cycles from weeks to hours~\cite{anthropic2026agenticreport}.
Yet many of the bottlenecks that now dominate practice (memory bandwidth walls~\cite{gholami2024memorywall}, cache reuse inefficiency~\cite{kwon2023pagedattention}, context management~\cite{memgpt2023}, batch scheduling~\cite{orca2022}, and sandboxed permission control~\cite{ironclaw2025}) are not failures of model capability.
They are, at their core, \emph{systems problems}.

The impact is already measurable in production.
Engineers at Rakuten have run Claude Code autonomously for hours over a 12.5-million-line codebase~\cite{anthropic2026agenticreport}.
An Augment Code enterprise customer delivered a project that their CTO had estimated at four to eight months in just two weeks~\cite{anthropic2026agenticreport}.
These are not laboratory demonstrations; they are evidence that the challenges of model-native computing have moved from theoretical possibility to engineering reality.

Given these parallels, a natural question arises: classical computer architecture solved a remarkably similar class of complexity-management problems through a mature stack of layered abstractions.
Can that intellectual framework transfer to model-native computing systems?

\begin{figure}[h]
    \centering
    \includegraphics[width=0.98\textwidth]{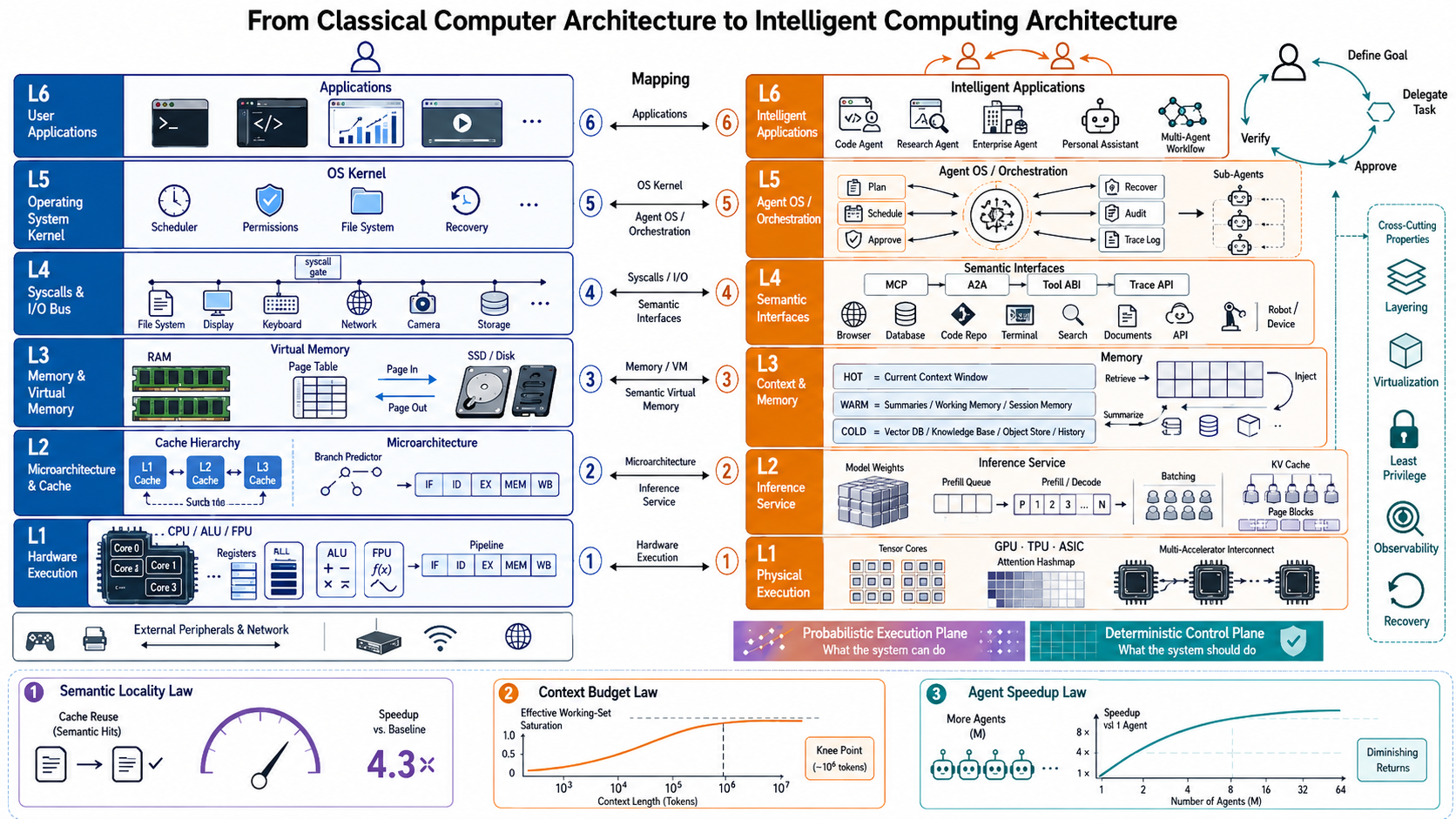}
    \caption{Overview analogy mapping between computer architecture and model-native computing systems}
    \label{fig_analogy_map}
    \end{figure}

\subsection{A Natural Analogy}

The core mapping between the two worlds is summarized in Table~\ref{tab:analogy-mapping}.

\begin{table}[h]
\centering
\footnotesize
\caption{Core analogy mapping between classical computing and model-native computing.}
\label{tab:analogy-mapping}
\begin{tabularx}{\textwidth}{>{\raggedright\arraybackslash}p{0.28\textwidth}>{\raggedright\arraybackslash}p{0.28\textwidth}>{\raggedright\arraybackslash}p{0.34\textwidth}}
\toprule
\textbf{Classical Computing} & \textbf{Model-Native Computing} & \textbf{Shared Problem} \\
\midrule
CPU & LLM inference core & General-purpose computation \\
Cache (L1/L2/L3) & KV cache & Hot data reuse, memory optimization \\
Virtual memory & Context window + external memory & Finite-capacity address space mgmt \\
Operating system & Agent runtime & Scheduling, permissions, resource mgmt \\
I/O buses (PCIe/USB) & MCP / A2A protocols & Standardized peripheral/tool integration \\
Applications \& users & Agent apps \& domain experts & Turning computation into solutions \\
\bottomrule
\end{tabularx}
\end{table}

The correspondence runs deeper than surface resemblance.
(1)~The LLM inference core provides general-purpose semantic understanding and generation, much as a CPU executes a general instruction set, with different models (GPT-4, Claude, LLaMA) resembling different microarchitectures that implement the same ``instruction set'' of reasoning with distinct performance profiles.
(2)~The \emph{KV cache} reuses previously computed attention key-value pairs during autoregressive decoding to avoid redundant computation, mirroring exactly how a CPU cache exploits temporal locality to reuse hot data.
PagedAttention~\cite{kwon2023pagedattention} makes this analogy explicit, borrowing the operating system's paging mechanism to partition KV caches into fixed-size blocks allocated on demand.
(3)~The context window is strictly bounded (e.g., 128K tokens), yet agents routinely require access to knowledge and interaction histories far exceeding this limit---precisely the problem that motivated virtual memory.
MemGPT~\cite{memgpt2023} formalizes this as \emph{virtual context management}, directly analogous to the operating system's virtual memory abstraction.
(4)~Agent runtimes such as Codex~\cite{openai_codex_overview_2026}, Claude Code~\cite{anthropic_claudecode_overview_2026}, and OpenHands~\cite{openhands_paper_2025} manage task decomposition, tool dispatch, sub-agent coordination, sandbox isolation, and result submission: a responsibility envelope that increasingly resembles an operating system kernel.
(5)~MCP (Model Context Protocol)~\cite{mcp_intro_2026} defines a standardized interface between agents and external tools, while A2A (Agent-to-Agent Protocol)~\cite{google_a2a_2025} specifies interoperability contracts among agents.
Together they play the role of I/O buses: MCP functions as a vertical bus (analogous to PCIe, connecting an agent to its tools), while A2A functions as a horizontal interconnect (analogous to a network fabric, connecting agents to one another)~\cite{agentprotocols2025}.

In 2023, Andrej Karpathy captured this intuition in a single sentence: ``LLMs are not chatbots, they are the kernel process of a new Operating System''~\cite{karpathy2023lmos}.
The observation resonated widely precisely because it named something already happening: the engineering problems accumulating around large models look increasingly like classical computer systems problems.

\subsection{From Intuition to Problem}

An analogy, however, is not a theory.
``Resembles'' does not imply ``is.''
If the mapping remains at the level of rhetoric, it yields inspirational metaphors but not testable predictions or actionable design principles.

Multiple independent efforts have already explored facets of this analogy from different angles.
AIOS treats the LLM as an operating system kernel, building an agent runtime with a scheduler, context manager, and memory manager~\cite{mei2024aios}.
MemGPT~\cite{memgpt2023} applies virtual-memory thinking to context management.
Mi et al.\ design agent architectures by drawing on process models, file systems, and other classical abstractions~\cite{mi2025llmagentscomputersystems}.
Ge et al.\ articulate the vision of ``LLM as OS, Agents as Apps''~\cite{ge2023llmos}.
L2MAC frames the LLM as a stored-program automatic computer~\cite{holt2024l2mac}.
ArbiterOS defines the LLM as a ``Probabilistic CPU'' governed by a higher-level governor~\cite{arbiteros2025}.
AgentOS positions it as a reasoning kernel~\cite{agentos2025}.
MemOS elevates memory itself to a first-class operating-system resource~\cite{memos2025}.
On the inference engineering side, Mei et al.\ (2025) survey the emerging discipline of \emph{context engineering}, defining it as the systematic optimization of all information supplied to an LLM~\cite{contextengineer_survey2025}.

These works each cover one or several layers of the system stack---the OS layer, the memory layer, the agent kernel, the inference layer.
Yet they suffer from a fundamental \emph{metaphor conflict}: is the model a CPU, an OS kernel, or a reasoning kernel?
More critically, they lack a unified layered model, well-defined inter-layer interfaces, and quantifiable design principles.
Moreover, nearly all of them treat the human as a boundary condition of the system rather than as a core component of the interaction loop.
The latest industrial evidence, however, tells a different story: the collaboration pattern between human judgment and agent execution, not the agent's standalone capability, is the primary determinant of system effectiveness.
Engineers report using AI for approximately 60\% of their work, yet they can \emph{fully delegate} only 0--20\% of tasks~\cite{anthropic2026agenticreport}; this high-use/low-delegation pattern is corroborated by independent studies, which find that developers accept only about 30\% of AI code suggestions on average~\cite{dohmke2023seachange} and report using AI assistants despite deep mistrust in their output~\cite{klemmer2024aiassistants}.
This is the collaboration paradox: AI is embedded deeply into the workflow, but meaningful autonomous delegation remains elusive.

The situation echoes the early days of computer architecture.
Before the introduction of the Instruction Set Architecture (ISA) as a contract between hardware and software, every new CPU required a complete rewrite of all software, and no improvement at one layer could be leveraged by another.
Without an analogous contract for model-native computing, each new system must be built from scratch, and insights remain siloed within individual projects.

\subsection{Contributions}

Accordingly, this paper makes five contributions. \emph{The central thesis} is that the enduring ``Is an LLM a CPU or an OS?'' conflict dissolves once one recognizes two orthogonal planes---a probabilistic execution plane and a deterministic control plane---and that instantiating this dual-plane view as the six-layer ICA is what no prior work has integrated.

\begin{enumerate}
\item We return to the foundational abstractions of computer architecture, operating systems, and distributed systems to distill an analogy framework suited to the model-native system stack, explicitly delineating where the analogy holds and where it breaks down.

\item We survey the landscape of related work, expose the metaphor conflicts among existing approaches, and propose the \textbf{dual-plane architecture} as a unifying resolution, partitioning the model-native stack into a \emph{probabilistic execution plane} (model inference and generation) and a \emph{deterministic control plane} (system scheduling and control), with clearly defined interfaces between the two.

\item Building on this foundation, we introduce the \textbf{Intelligent Computing Architecture (ICA)}: a six-layer framework with inter-layer interface contracts and six design axioms.

\item We propose three \textbf{Amdahl-style design heuristics} for model-native computing---the Semantic Locality, Context Budget, and Agent Speedup heuristics---and \emph{illustrate} their parameter ranges with published system data. We are explicit that these are organizing back-of-envelope intuition rather than validated scaling laws: they transplant Amdahl's analytical form into the model-native regime to give a previously qualitative design space computable intuition, and we identify systematic predictive validation---rather than the back-solved parameter recovery that pervades this literature---as the key open task.

\item We survey current open-source implementations and industrial practice, and we chart a research roadmap for the field.
\end{enumerate}

%% file: en_sections/section_bg.tex
\section{Background: The Classical Computing Stack and the Model-Native System Stack}
\label{sec:background}

This section first reviews the layered logic of classical computer architecture and the quantitative design principles that underpin it, then surveys the nascent stratification currently taking shape in large language model (LLM) systems. Together, these two perspectives lay the groundwork for the analogy framework developed in subsequent sections.

\subsection{Layered Abstractions in Classical Computer Architecture}

The layered organization of a classical computing system is not an ad hoc design choice but a general engineering strategy for managing complexity~\cite{hennessy2017quantitative}.
The central idea is that each layer depends only on the \emph{interface} exposed by the layer below, never on its \emph{implementation}.
This separation allows any layer to be independently optimized (or even replaced) without disrupting the rest of the stack.
We now review the core concepts of each layer in turn.

\subsubsection{Instruction Set Architecture (ISA): The Contract Between Hardware and Software}

The \textit{Instruction Set Architecture} (ISA) specifies the machine behavior visible to software: the available instructions (arithmetic, memory access, branching, jumping), the register file, how exceptions are raised and handled, and how the address space is organized.
The ISA serves as a stable contract: as long as the processor honors this contract, software above it runs correctly, regardless of the microarchitectural details hidden below.
Intel x86, ARM, and RISC-V embody three distinct design philosophies~\cite{intel_sdm_2026,riscv_spec_2026}: x86 prioritizes backward compatibility (code written four decades ago still executes), ARM optimizes for energy efficiency (a decisive advantage in mobile devices), and RISC-V embraces openness and modularity (users may define custom instruction extensions).
Despite their differences, all three serve the same purpose: providing a stable programming target for the software stack (Figure~\ref{fig:bg21:isa}).
\input{figures/fig_bg_21_isa}

\subsubsection{Microarchitecture: Multiple Implementations Under One Contract}

The \textit{microarchitecture} determines how a given ISA is realized in silicon and, consequently, what performance a processor actually delivers.
It answers questions such as: how many clock cycles does an instruction require? How many instructions can execute simultaneously? What is the penalty for a mispredicted branch?
Concretely, a \emph{pipeline} decomposes instruction execution into stages (fetch, decode, execute, memory access, write-back) so that different instructions can overlap across stages.
A \emph{superscalar} design dispatches multiple instructions per cycle to independent execution units.
A \emph{branch predictor} uses historical patterns to speculate on the outcome of conditional branches, thereby avoiding pipeline bubbles.
Both Intel's Core microarchitecture and AMD's Zen microarchitecture implement the same x86 ISA yet deliver different performance characteristics, and this difference is entirely transparent to software (Figure~\ref{fig:bg21:microarch}).
\input{figures/fig_bg_21_microarch}

\subsubsection{Cache Hierarchy: Exploiting Locality to Bridge the Processor--Memory Gap}

The \textit{cache hierarchy} addresses a fundamental speed mismatch between processors and main memory.
A modern CPU core can complete an operation in roughly one nanosecond, whereas a DRAM access requires approximately 100 nanoseconds, forcing the processor to stall for two orders of magnitude longer if every data reference must reach main memory.
Caches exploit the \emph{principle of locality}: \emph{temporal locality} holds that data accessed recently is likely to be accessed again soon, while \emph{spatial locality} holds that data at nearby addresses will likely be accessed in the near future~\cite{drepper2007memory}.
Accordingly, caches store recently and nearby accessed data in faster but smaller SRAM: the L1 cache (typically 32--64\,KB, $\sim$4-cycle latency) sits closest to the core, the L2 cache (256\,KB--1\,MB, $\sim$10 cycles) acts as a second-level buffer, and the L3 cache (several to tens of megabytes, $\sim$40 cycles) is shared among multiple cores.
On a cache miss, data must be fetched from DRAM at a cost of roughly 200 cycles.
This tiered design trades capacity for speed at each level, using SRAM's low latency for hot data and DRAM's density for the working set at large (Figure~\ref{fig:bg21:cache}).
\input{figures/fig_bg_21_cache}

\subsubsection{Virtual Memory: The Illusion of Infinite Address Space}

\textit{Virtual memory}, managed by the Memory Management Unit (MMU) and page tables, provides each process with an independent address space that can far exceed physical memory.
The core mechanism is \emph{paging}: the virtual address space is divided into fixed-size \emph{pages} (typically 4\,KB), physical memory into equally sized \emph{page frames}, and a page table records the mapping from virtual page numbers to physical frame numbers.
When a process accesses a virtual page not yet resident in physical memory, a \emph{page fault} is raised; the operating system loads the page from disk into a free frame, updates the mapping, and resumes the process~\cite{ostep_2023}.
Every process thus believes it possesses a complete, private address space, while the operating system transparently manages allocation and reclamation of the underlying physical resource.
Virtual memory also enforces isolation: each process maintains its own page table and cannot address another process's memory (Figure~\ref{fig:bg21:vm}).
\input{figures/fig_bg_21_virtualmem}

\subsubsection{Operating System: Scheduling, Protection, and Resource Management}

The \textit{operating system} (OS) is the first software layer above hardware.
It manages compute resources and provides uniform abstractions to applications.
Its core responsibilities include: process scheduling (deciding which process runs next, e.g., Linux's Completely Fair Scheduler~\cite{linux_cfs_2026}), memory management (allocating and reclaiming physical page frames), I/O management (providing uniform file and device interfaces), access control (ensuring processes access only authorized resources), and concurrency support (semaphores, locks, condition variables)~\cite{ostep_2023,xv6_2022}.
The OS design principle is simple: expose clean abstractions upward (processes, files, sockets) while managing complex hardware downward (CPU cores, memory modules, disks, network interfaces).
Applications invoke system calls such as \texttt{open()}, \texttt{read()}, and \texttt{fork()} without needing to know whether the storage device is an SSD or an HDD, or whether the network link runs at 1\,GbE or 10\,GbE (Figure~\ref{fig:bg21:os}).
\input{figures/fig_bg_21_os}

\subsubsection{Distributed Systems: Scaling Beyond a Single Node}

\textit{Distributed systems} extend the computing stack across multiple nodes organized into a logically unified whole.
The fundamental challenge is maintaining correctness and availability when individual nodes may fail, network latencies are variable, and data must be replicated~\cite{cap2002}.
The Raft consensus algorithm~\cite{raft2014} enables a cluster of nodes to agree on the next command to append to a shared log, tolerating crashes of a minority of participants.
Google Spanner~\cite{spanner2012} provides a globally consistent, distributed database service across multiple data centers worldwide.
The CAP theorem~\cite{cap2002} formalizes an inherent trade-off: during a network partition, a system cannot simultaneously guarantee both consistency (C) and availability (A); a design choice must be made (Figure~\ref{fig:bg21:dist}).
\input{figures/fig_bg_21_distributed}

\subsubsection{Quantitative Design Principles}

What elevates computer architecture from craft to science is a set of \textbf{quantitative principles} that make performance predictable and compositional.

\paragraph{Amdahl's Law.}
Amdahl's Law establishes a theoretical upper bound on the speedup achievable through parallelization:
\begin{equation}
\label{eq:amdahl}
S = \frac{1}{(1-f) + f/p}
\end{equation}
where $f$ is the fraction of the workload that can be parallelized, $p$ is the number of processors, and $S$ is the resulting speedup.
For instance, if 80\% of a task is parallelizable ($f=0.8$) and four processors are available ($p=4$), the maximum speedup is $S = 1/(0.2 + 0.8/4) = 2.5\times$, far below the ideal $4\times$.
Even as $p \to \infty$, speedup asymptotically approaches $1/(1-f) = 5\times$.
The key insight is that \emph{the serial fraction caps the speedup ceiling}.
Malekar and Zand (2024) adapted Amdahl's Law to the LLM setting, proposing an analytical framework for LLM throughput~\cite{malekar2024amdahlllm}.

\paragraph{The Roofline Model.}
The Roofline model visualizes the performance constraints of an operator on a given hardware platform~\cite{hennessy2017quantitative}.
The attainable performance is bounded by two lines: a horizontal line representing peak compute throughput (FLOP/s) and a sloped line representing memory bandwidth multiplied by arithmetic intensity ($\mathrm{Bandwidth} \times \mathrm{Arithmetic\ Intensity}$).
Their intersection is called the \emph{ridge point}.
Operators with arithmetic intensity below the ridge point are \emph{memory-bandwidth-bound}; those above it are \emph{compute-bound}.
Yuan et al.\ systematically applied the Roofline model to LLM inference, revealing how different optimization strategies trade off bandwidth and compute utilization~\cite{yuan2024llminference}.

\paragraph{Locality Principle.}
Temporal and spatial locality directly inform cache design: temporal locality (recently accessed data will be re-accessed) motivates cache residency, while spatial locality (nearby addresses will be accessed soon) motivates cache lines (typically 64 bytes), hardware prefetching, and replacement policies such as LRU~\cite{drepper2007memory}.

Taken together, these quantitative principles have enabled the computing stack to scale from thousands of floating-point operations per second to over $10^{18}$ operations per second across eight decades, all without requiring programmers to understand the physical details of every layer.

\subsection{Emerging Layers of the LLM System Stack}

LLM systems are currently undergoing a stratification process that closely mirrors the classical computing stack.
We survey each layer from bottom to top, highlighting key developments and the system-level abstractions they implicitly introduce.

\subsubsection{Model Layer: Evolution of the Inference Core}

On the modeling front, the Transformer architecture~\cite{vaswani2017attention} remains the dominant backbone.
Its self-attention mechanism captures long-range dependencies by computing pairwise interactions across all positions in a sequence.
Open-weight models continue to evolve along several axes: RoPE (Rotary Position Embedding)~\cite{su2021roformer} improves positional representations, enabling models to handle longer sequences more effectively; Mixture-of-Experts (MoE)~\cite{switch2022,mixtral2024} scales model capacity through sparse activation (activating only a subset of parameters per inference), avoiding a proportional increase in computation; long-context extensions~\cite{yarn2024,longrope2024} push the effective context window from thousands of tokens to hundreds of thousands or even millions.
In parallel, selective state-space models such as Mamba~\cite{gu2024mamba} open a linear-complexity sequence modeling paradigm that no longer relies on self-attention (Figure~\ref{fig:bg22:model}).
\input{figures/fig_bg_22_model}

\subsubsection{Inference Layer: Systematic Serving Optimizations}

On the inference front, a growing body of work has transformed LLM serving from a single forward pass into a systems engineering problem.
FlashAttention~\cite{dao2022flashattention,dao2023flashattention2} reformulates attention computation as an I/O-aware problem: by tiling the computation to reduce accesses to high-bandwidth memory (HBM), it lowers attention memory complexity from $O(N^2)$ to $O(N)$.
vLLM's PagedAttention~\cite{kwon2023pagedattention} models KV cache management as a paging problem, dividing the KV cache into fixed-size ``pages'' that are allocated and reclaimed on demand, thereby eliminating the waste of pre-allocating maximum memory per request.
ORCA~\cite{orca2022} introduces continuous batching at the iteration level, allowing new requests to join a running batch during the decode phase of existing requests.
DistServe~\cite{zhong2024distserve} decouples the prefill phase (processing the input prompt) from the decode phase (generating tokens one at a time) onto separate GPUs, avoiding resource contention between the two.
Sarathi-Serve~\cite{agrawal2024sarathi} further refines the throughput--latency trade-off.
Disaggregated serving architectures push this prefill/decode decoupling to cluster scale: Splitwise~\cite{patel2024splitwise} and Mooncake~\cite{qin2024mooncake} separate the two phases across dedicated GPU pools and treat the KV cache as a global, transferable resource, realizing the ``KV cache as a hardware-managed cache'' analogy at hyperscale.
A unifying theme across these systems is the adaptation of classical operating systems techniques (paging, batch scheduling, resource isolation) to LLM inference engines (Figure~\ref{fig:bg22:inference}).
\input{figures/fig_bg_22_inference}

\subsubsection{Memory Layer: Reconciling Finite Windows with Persistent State}

At the memory layer, the core challenge is that the context window is finite (e.g., 128K tokens), whereas agents often need access to knowledge and interaction histories that far exceed this limit.
Multiple technical directions are being explored in parallel.
Long-context extensions~\cite{gemini2024,yarn2024} attempt to enlarge the window directly, but face quadratic growth in computation cost with sequence length.
Retrieval-Augmented Generation (RAG)~\cite{lewis2020rag} retrieves relevant passages from an external knowledge base at inference time and injects them into the context, analogous to demand paging in an operating system.
MemGPT~\cite{memgpt2023} explicitly introduces \textit{virtual context management}, partitioning context into a working context (analogous to main memory) and an external storage (analogous to disk), and automatically moving information between the two.
LongMem~\cite{longmem2023} and Generative Agents~\cite{park2023generativeagents} explore alternative forms of long-term memory.
LongMemEval~\cite{longmemeval2024} establishes a benchmark for evaluating memory in long-term interactions.
Letta~\cite{letta_stateful_2026} places stateful agents at the center of its design, enabling agents to retain memory across sessions.
MemOS~\cite{memos2025} goes further by treating memory as a first-class operating-system-level resource, unifying management of textual memory, activation memory, and other representational forms.
MemoryOS~\cite{memoryos2025}, presented at EMNLP 2025, proposes a memory operating system tailored to AI agents.
\emph{Context engineering}~\cite{contextengineer_survey2025} has emerged as a nascent discipline that systematizes this space.
It is defined as the engineering practice of systematically designing, selecting, and optimizing all information presented to an LLM, encompassing retrieval augmentation, memory management, tool-use scaffolding, and multi-turn dialogue optimization (Figure~\ref{fig:bg22:memory}).
\input{figures/fig_bg_22_memory}

\subsubsection{Agent Layer: From Single-Turn Inference to Complex Runtimes}

At the agent layer, LLMs have evolved beyond single-turn inference engines into controllers within complex runtime environments.
ReAct~\cite{yao2023react} interleaves reasoning with acting, enabling a model to solve multi-step tasks in a ``think--act--observe'' loop.
Toolformer~\cite{schick2023toolformer} teaches models to invoke external tools (calculators, search engines) autonomously.
SWE-agent~\cite{sweagent2024} and OpenHands~\cite{openhands_paper_2025} are autonomous agents for software engineering that can navigate code repositories, write patches, and run test suites.
Codex~\cite{openai_codex_overview_2026} and Claude Code~\cite{anthropic_claudecode_overview_2026} are production-grade coding agents equipped with OS-kernel-like capabilities: sub-agent scheduling~\cite{openai_codex_overview_2026,anthropic_claudecode_subagents_2026}, sandbox isolation~\cite{openai_codex_sandbox_2026,anthropic_claudecode_sandbox_2026}, and permission control~\cite{openai_codex_approvals_2026,anthropic_claudecode_security_2026}.
Multi-agent collaboration frameworks such as AutoGen~\cite{wu2023autogen} and MetaGPT~\cite{metagpt2023} further organize ensembles of specialized agents into cooperative networks (Figure~\ref{fig:bg22:agent}).
\input{figures/fig_bg_22_agent}

\subsection{The Gap: No Unified System Language}

Despite rapid progress across each layer, the LLM systems landscape lacks a unified system language, that is, a shared vocabulary and abstraction framework that connects optimizations across layers.
Without such a framework, research at each layer tends to fragment into isolated collections of engineering tricks.
Consider, for example, \emph{prefix caching} (reusing cached key--value pairs across shared prompt prefixes), \emph{chunked prefill} (breaking long prefill computations into schedulable chunks), \emph{sub-agents} (concurrently executing child agents), \emph{MCP servers} (standardized tool interfaces via the Model Context Protocol), and \emph{persistent memory} (durable storage for agent state).
These concepts appear to belong to entirely different research directions, yet from an architectural standpoint they correspond to cache reuse, scheduling granularity, concurrent execution, I/O interfaces, and tiered storage, all well-studied abstractions in classical systems.

This observation motivates the architectural analogy that runs through this paper.
The analogy does not claim that LLM systems and classical systems are ``fundamentally the same.''
Rather, it places independently developed engineering techniques back into a larger, unified design space, making it possible to compose, predict, and systematically design cross-layer optimizations.

%% file: figures/fig_bg_21_isa.tex
\begin{figure}[htbp]
  \centering
  \includegraphics[width=0.80\textwidth]{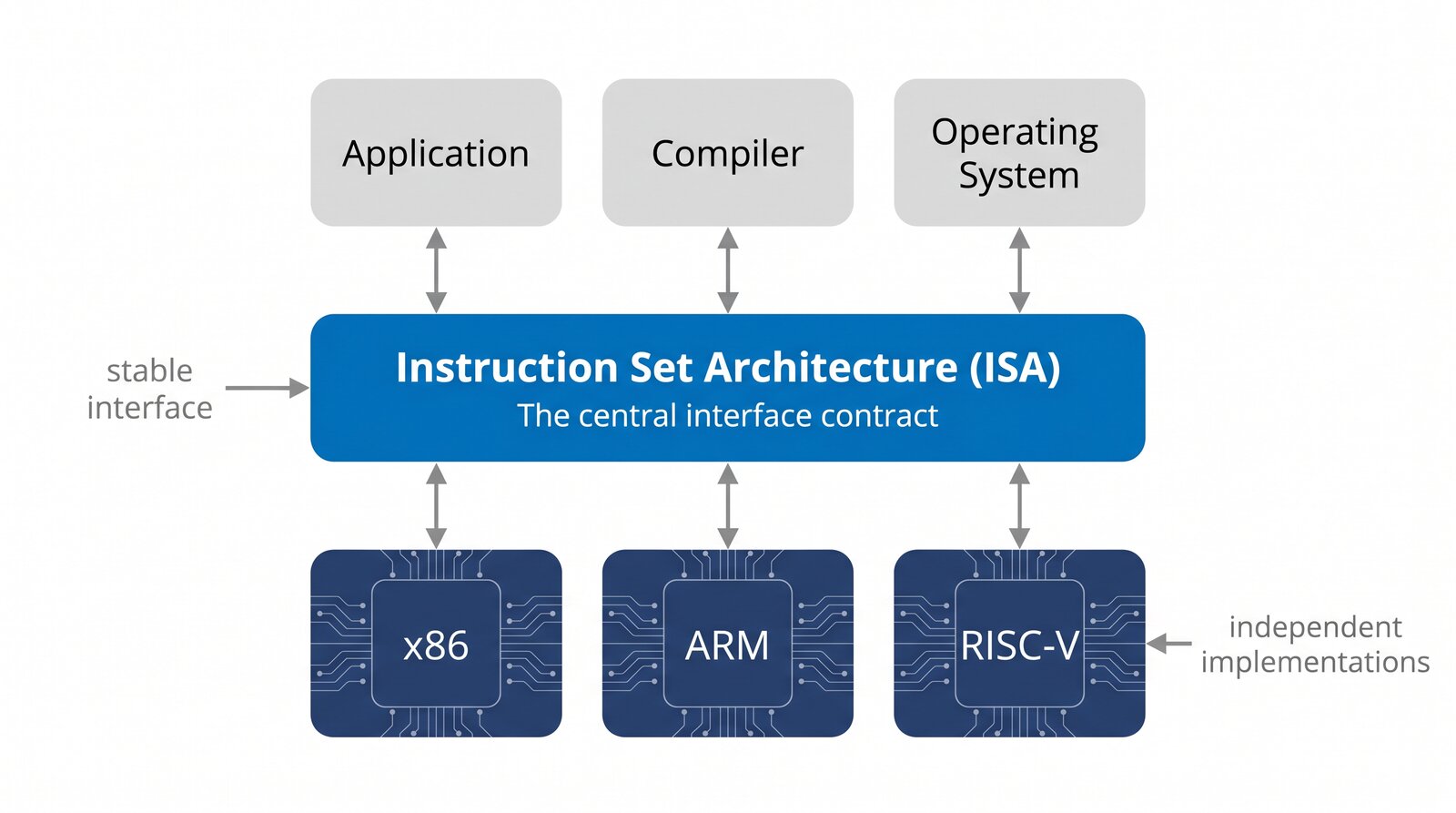}
  \caption{ISA: the stable hardware--software contract. Intel x86, ARM, and RISC-V each provide distinct microarchitectural implementations while exposing the same ISA to compilers, operating systems, and applications.}
  \label{fig:bg21:isa}
\end{figure}

%% file: figures/fig_bg_21_microarch.tex
\begin{figure}[htbp]
  \centering
  \includegraphics[width=0.80\textwidth]{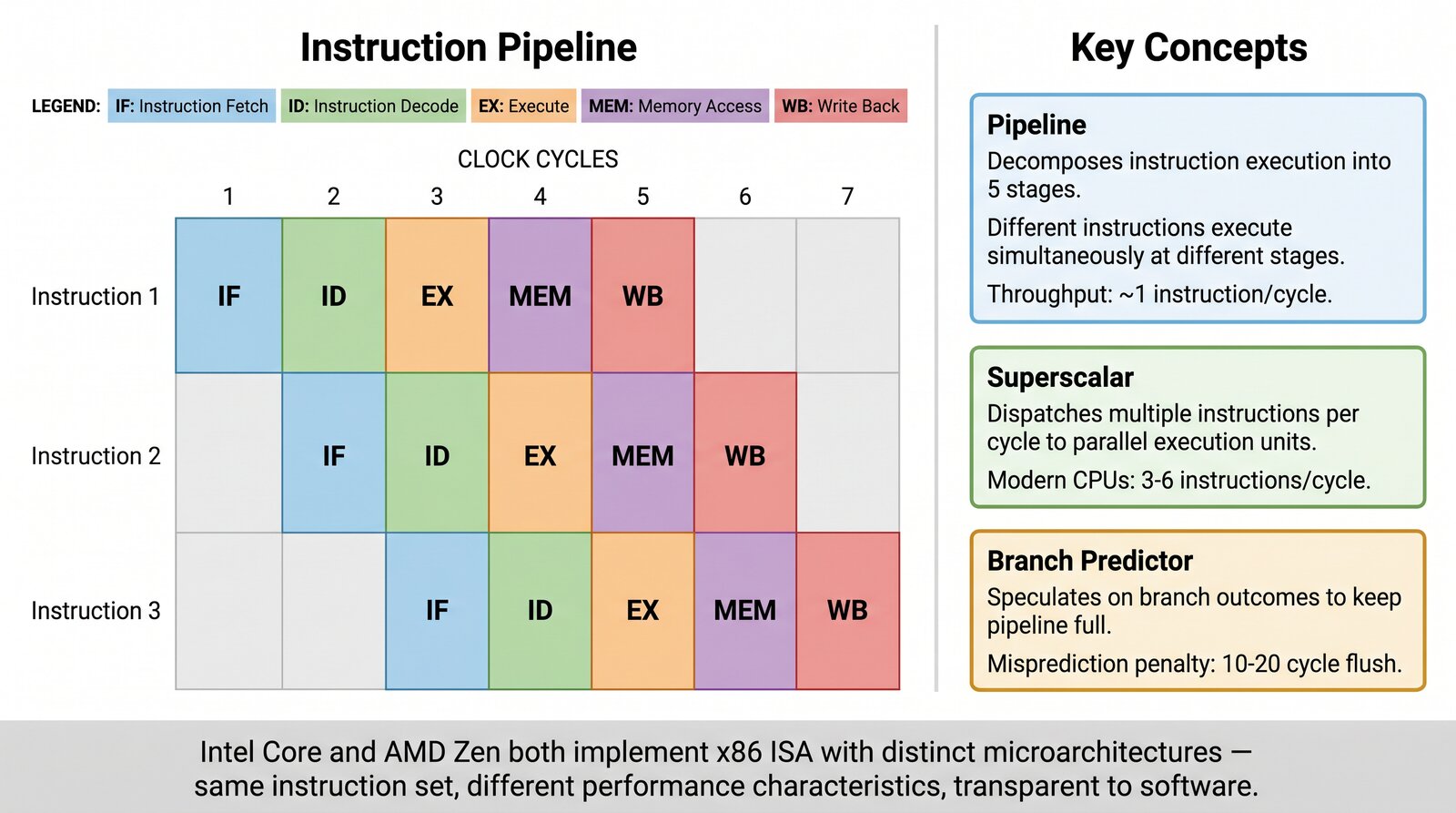}
  \caption{CPU microarchitecture: pipelined and superscalar execution. A pipeline decomposes instruction execution into stages (Fetch, Decode, Execute, Memory, Write-Back) so that multiple instructions overlap in flight simultaneously; a superscalar design further dispatches several instructions per cycle.}
  \label{fig:bg21:microarch}
\end{figure}

%% file: figures/fig_bg_21_cache.tex
\begin{figure}[htbp]
  \centering
  \includegraphics[width=0.70\textwidth]{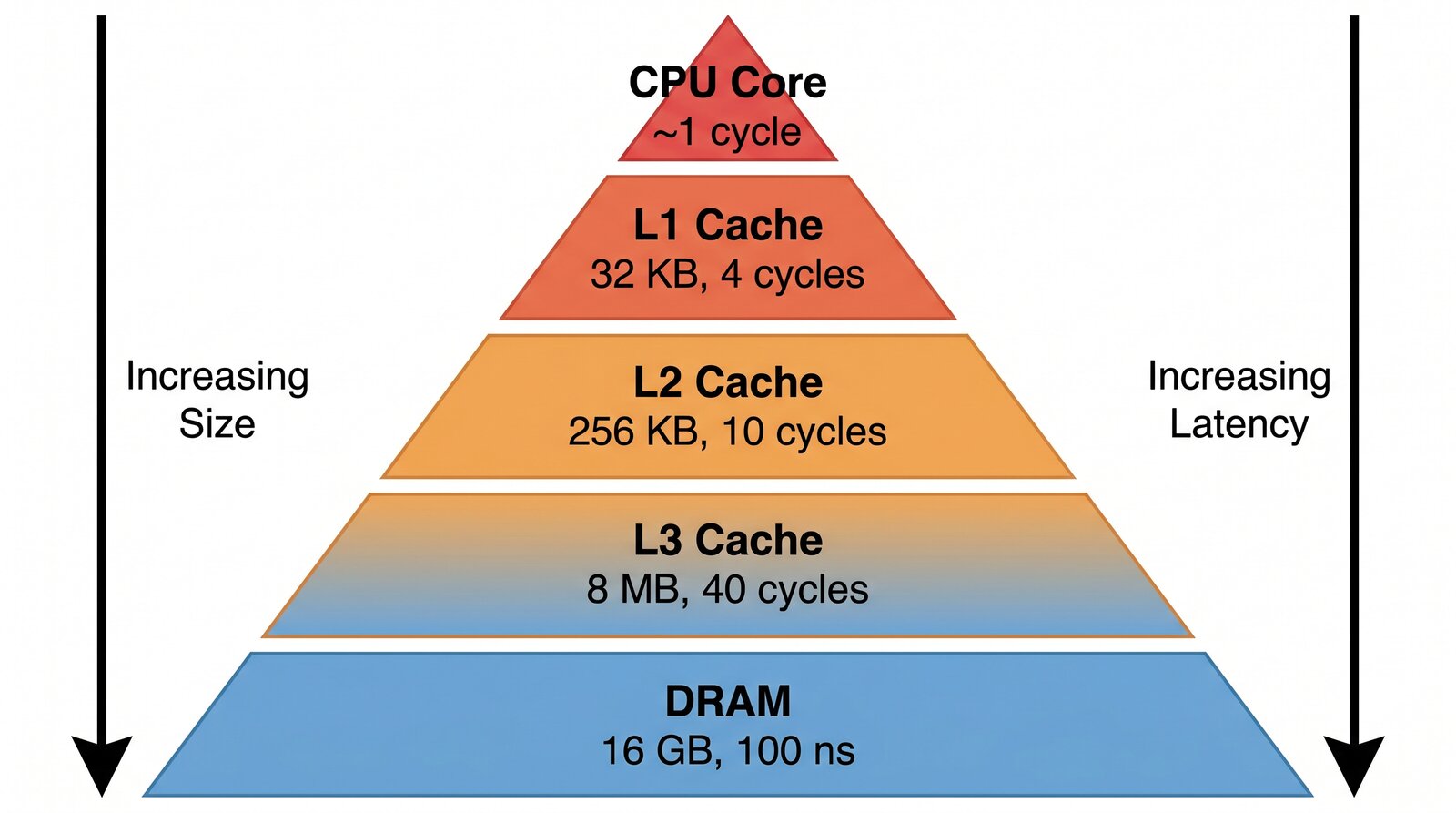}
  \caption{Cache hierarchy: exploiting temporal and spatial locality to bridge the processor--memory speed gap. Each successive tier (L1, L2, L3, DRAM) trades lower latency for higher capacity.}
  \label{fig:bg21:cache}
\end{figure}

%% file: figures/fig_bg_21_virtualmem.tex
\begin{figure}[htbp]
  \centering
  \includegraphics[width=0.70\textwidth]{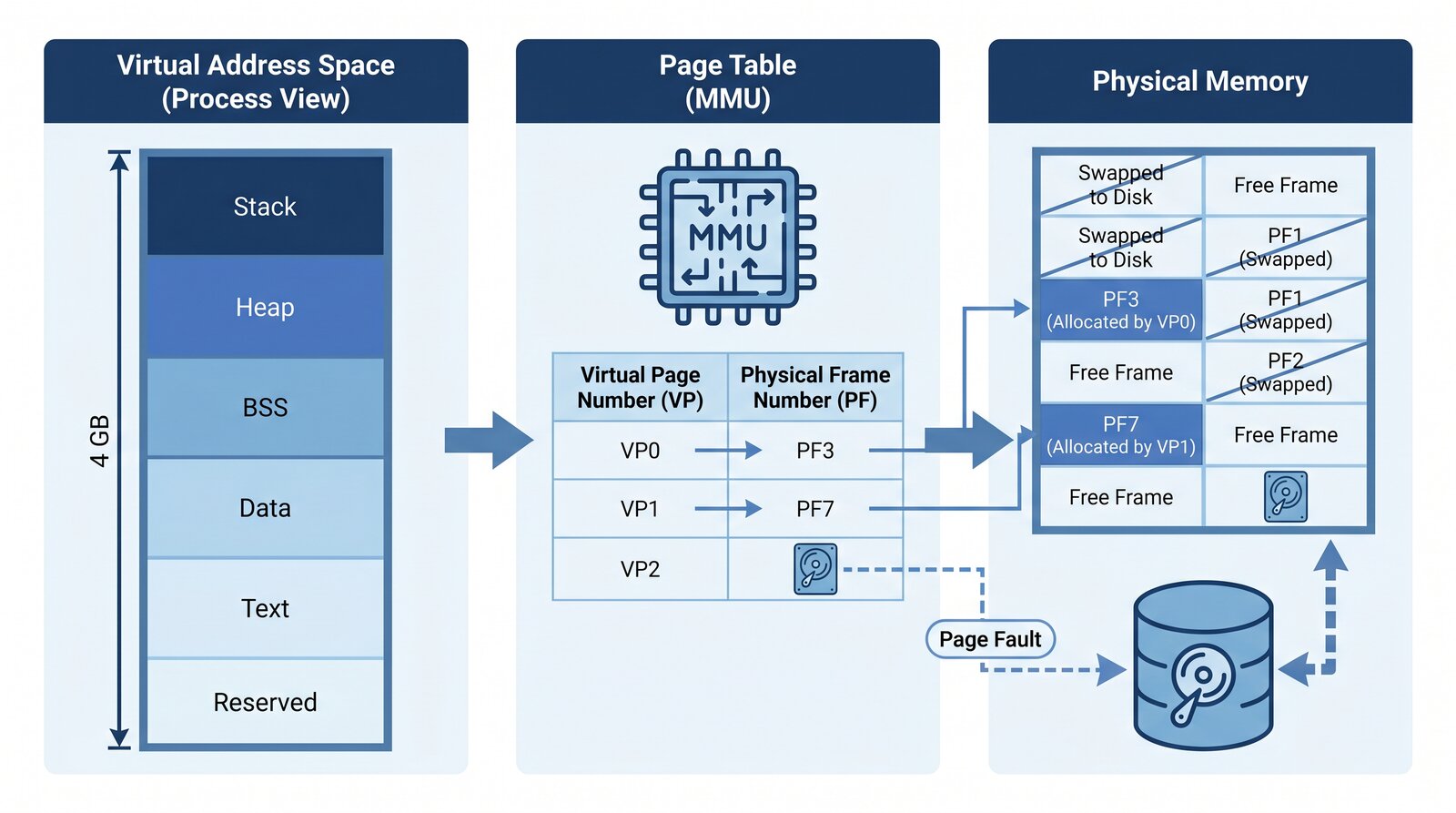}
  \caption{Virtual memory: the MMU and page tables map each process's large virtual address space onto non-contiguous physical frames, providing isolation and the illusion of unbounded memory.}
  \label{fig:bg21:vm}
\end{figure}

%% file: figures/fig_bg_21_os.tex
\begin{figure}[htbp]
  \centering
  \includegraphics[width=0.80\textwidth]{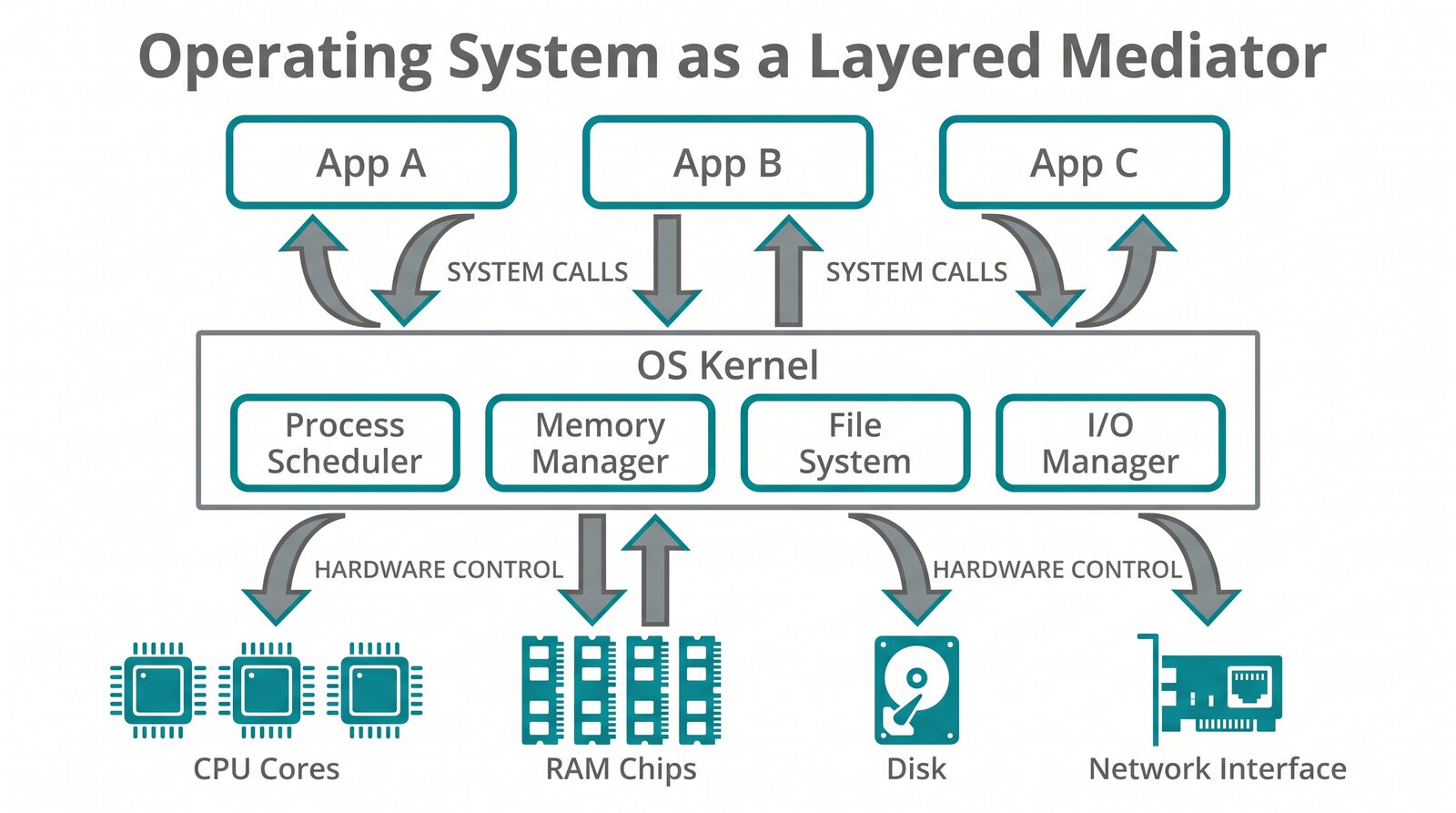}
  \caption{The operating system as a resource manager: the kernel mediates access to CPU cores, memory, storage, and network through scheduling, memory management, and a uniform system-call interface.}
  \label{fig:bg21:os}
\end{figure}

%% file: figures/fig_bg_21_distributed.tex
\begin{figure}[htbp]
  \centering
  \includegraphics[width=0.75\textwidth]{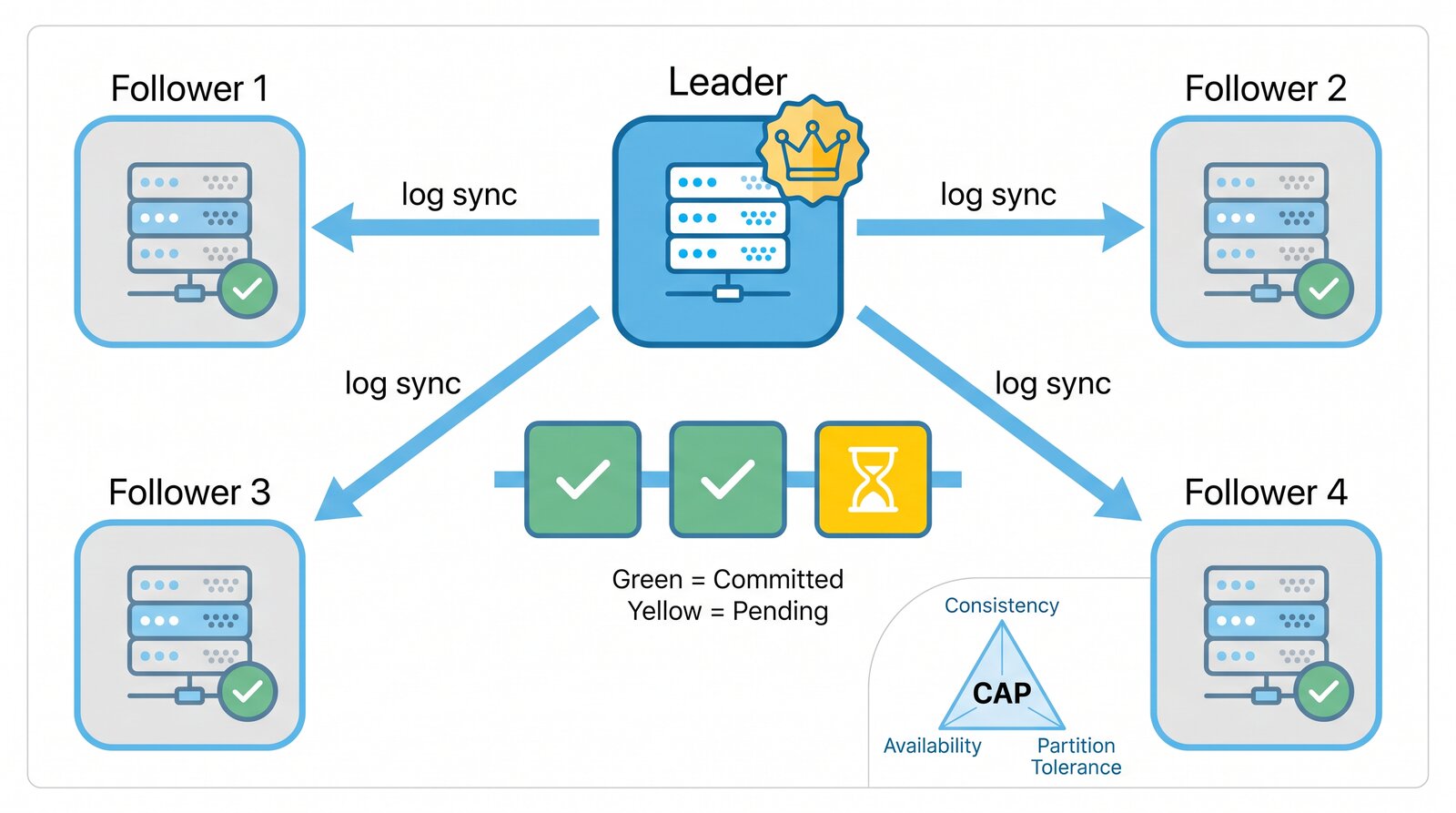}
  \caption{Distributed systems: the Raft consensus algorithm replicates a shared log across a leader and follower nodes, maintaining consistency despite individual node failures; the CAP theorem bounds what is simultaneously achievable under network partitions.}
  \label{fig:bg21:dist}
\end{figure}

%% file: figures/fig_bg_22_model.tex
\begin{figure}[htbp]
  \centering
  \includegraphics[width=0.80\textwidth]{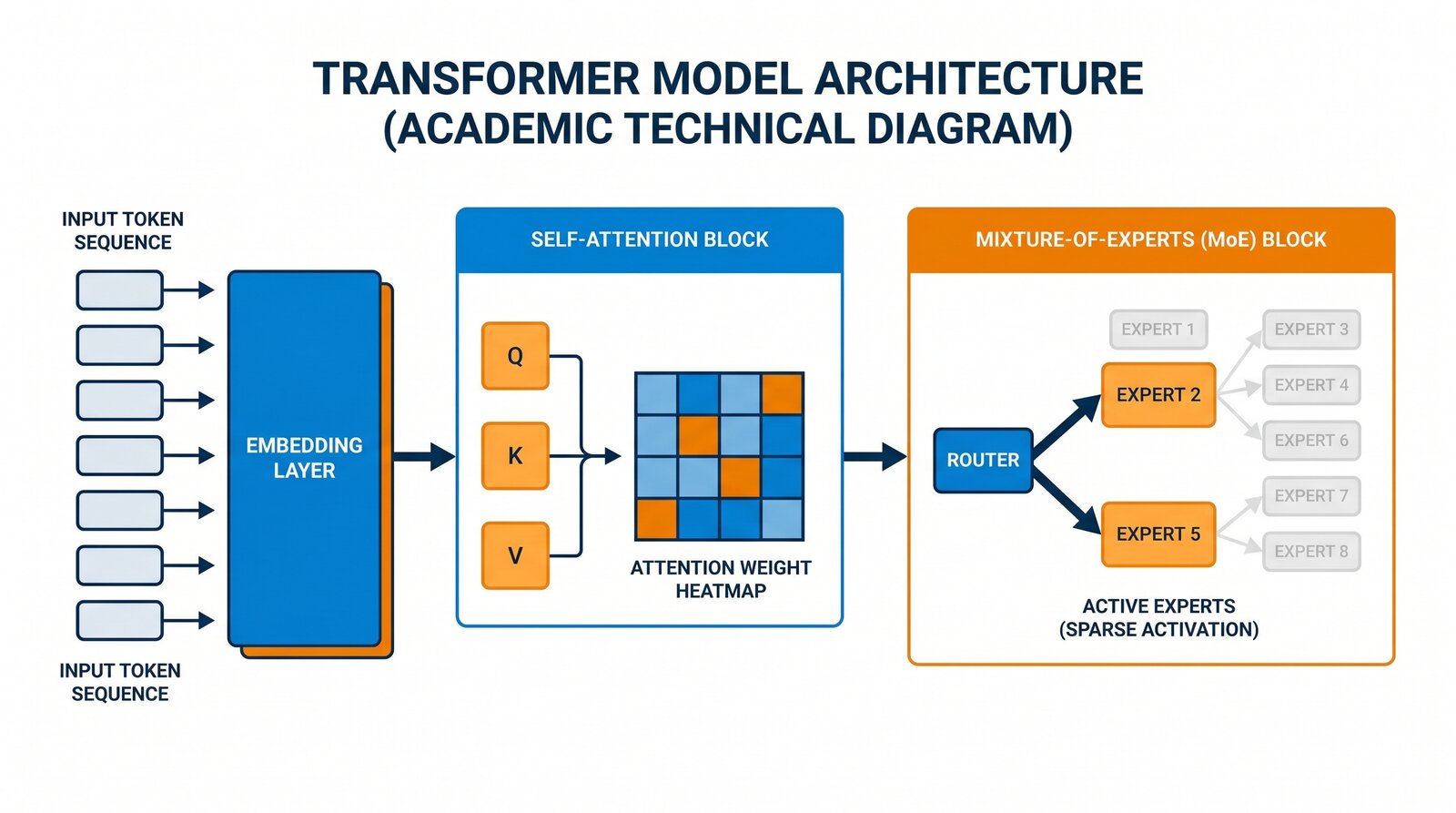}
  \caption{Model layer: the Transformer's self-attention mechanism serves as the general-purpose inference core; Mixture-of-Experts (MoE) introduces sparse routing so that only a subset of parameters is activated per inference step, enabling capacity to scale without proportional compute cost.}
  \label{fig:bg22:model}
\end{figure}

%% file: figures/fig_bg_22_inference.tex
\begin{figure}[htbp]
  \centering
  \includegraphics[width=0.80\textwidth]{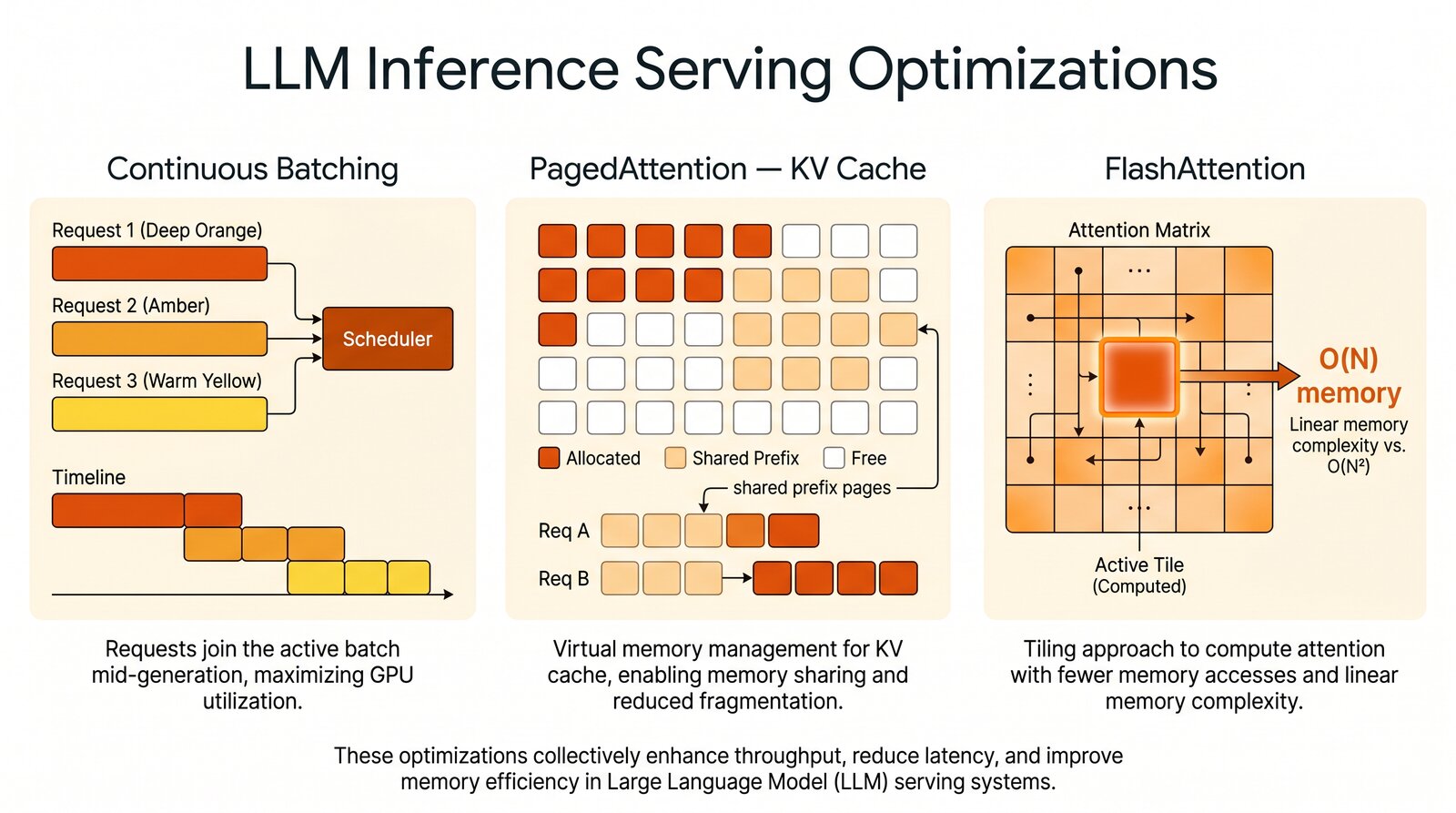}
  \caption{Inference layer: PagedAttention organizes the KV cache into demand-allocated pages and enables sharing of prefix pages across requests; continuous batching allows new requests to join in-flight batches at each decoding step; FlashAttention tiles attention computation to reduce HBM traffic.}
  \label{fig:bg22:inference}
\end{figure}

%% file: figures/fig_bg_22_memory.tex
\begin{figure}[htbp]
  \centering
  \includegraphics[width=0.75\textwidth]{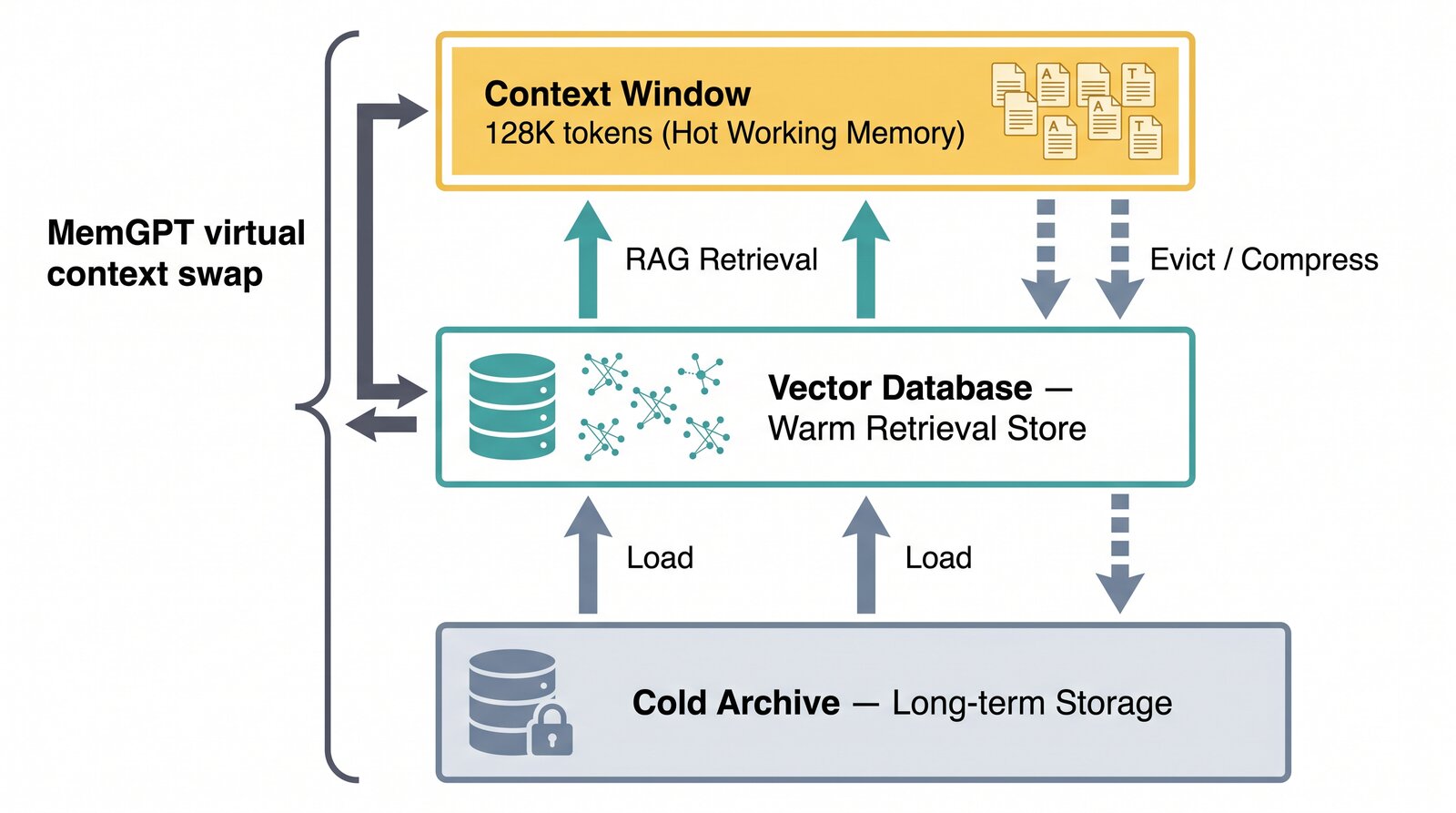}
  \caption{Memory layer: a three-tier architecture spanning the bounded context window (hot working memory), a warm retrieval store (vector database, accessed via RAG), and cold long-term archival storage; MemGPT automates swap-in and swap-out across tiers.}
  \label{fig:bg22:memory}
\end{figure}

%% file: figures/fig_bg_22_agent.tex
\begin{figure}[htbp]
  \centering
  \includegraphics[width=0.75\textwidth]{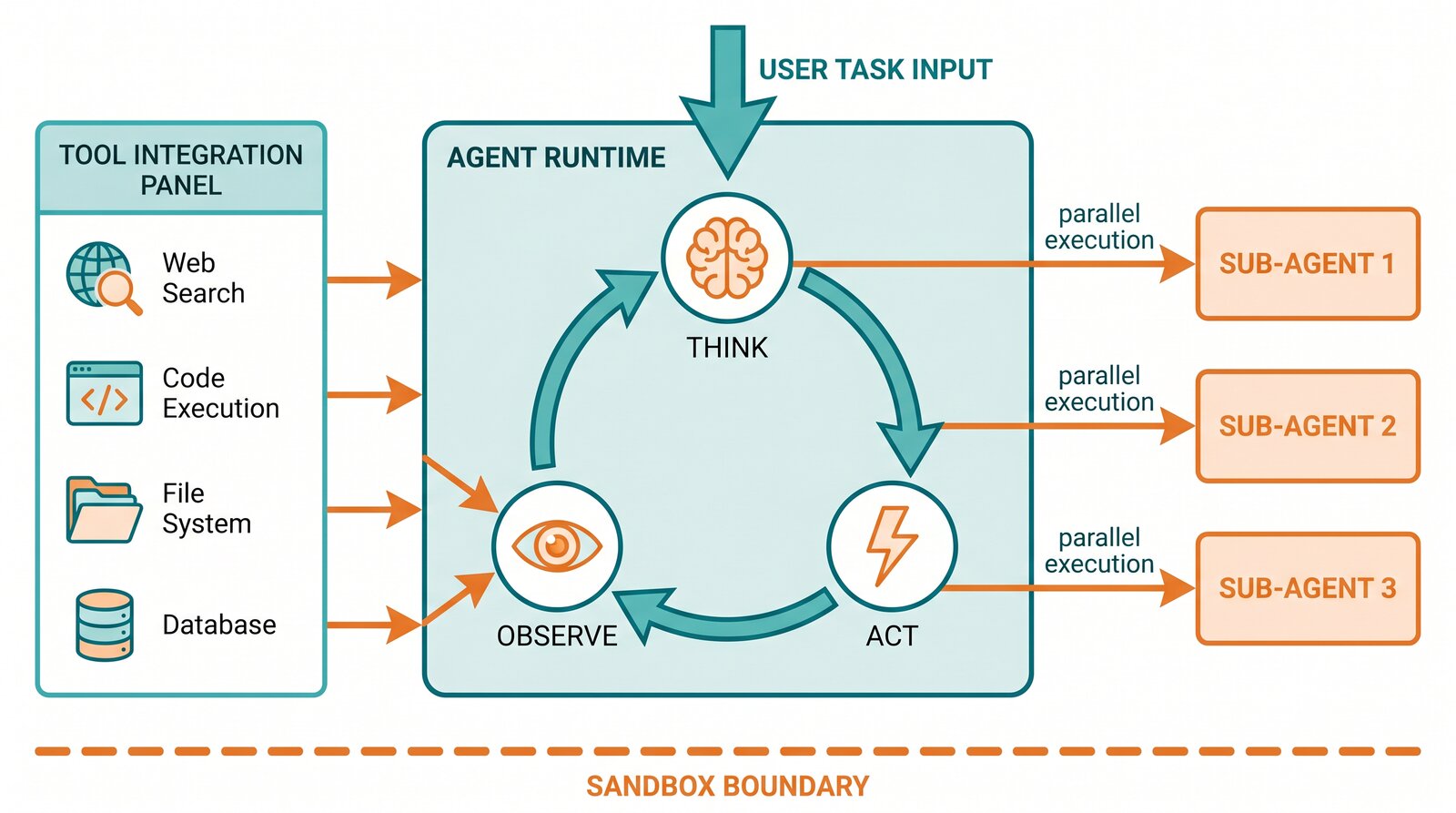}
  \caption{Agent layer: the ReAct loop interleaves reasoning with tool invocations inside a sandboxed runtime; sub-agents execute concurrently under permission control, and the orchestrator merges their results.}
  \label{fig:bg22:agent}
\end{figure}

%% file: en_sections/section_related.tex
\section{Related Work}
\label{sec:related}

The central thesis of this paper is that the conceptual framework of \textit{computer architecture} offers a productive lens for envisioning the complete layered design of future model-native computing systems.
This thesis did not emerge in a vacuum: by 2025, a substantial body of work had independently arrived at key insights such as ``LLM systems require OS-level abstractions,'' ``agents need memory management,'' and ``tool invocation demands standardized protocols.''
This section surveys the most relevant literature along functional themes, progressively revealing the contributions and limitations of each line of work and ultimately positioning the distinctive contribution of this paper.

\subsection{Literature Search Methodology}
\label{sec:related-method}
To ensure systematic coverage, we searched five sources---arXiv, DBLP, IEEE Xplore, the ACM Digital Library, and Semantic Scholar---over a January~2022 to May~2026 window, complemented by backward and forward citation snowballing from seed surveys (e.g., Mi et al.~\cite{mi2025llmagentscomputersystems}). Query strings combined an architecture/systems term (\emph{operating system}, \emph{computer architecture}, \emph{virtual memory}, \emph{instruction set}, \emph{cache}) with an LLM term (\emph{LLM}, \emph{agent}, \emph{KV cache}, \emph{inference serving}, \emph{tool use}, \emph{multi-agent}). From approximately 450 deduplicated records we retained 188 that (i)~introduce a system-level abstraction or mechanism for model-native computing, or (ii)~provide empirical evidence directly informing a design heuristic. We organize the retained set by functional theme rather than chronology, and acknowledge that a young, fast-moving field makes any snapshot provisional; the accompanying coverage map (Figure~\ref{fig_coverage_map}) makes the scope and its gaps explicit.

\subsection{LLM-as-OS / Agent Kernel}
\label{sec:related-llmos}

The earliest systematic articulation of the ``LLM as operating system'' analogy is due to Ge et al.\ (2023), who proposed ``LLM as OS, Agents as Apps''~\cite{ge2023llmos}.
Their central insight is captured by a system mapping diagram: the LLM corresponds to the OS kernel (responsible for core inference and scheduling), the \textit{context window} corresponds to main memory, external storage corresponds to the file system, hardware tools correspond to peripherals, software tools correspond to programming libraries, and user prompts correspond to user commands.
In other words, they reconceived the LLM not as a ``chatbot'' but as a ``kernel process,'' a metaphor that subsequently gained wide recognition in the practitioner community.
Within the analogy framework of this paper, the work of Ge et al.\ directly corresponds to the embryonic notion of the Agent OS (the ICA orchestration layer).

Mei et al.'s AIOS~\cite{mei2024aios} advanced this line of thinking from analogy to a runnable system implementation.
The core design of AIOS decouples responsibilities that were previously entangled with application logic in the agent runtime and places them into a dedicated ``kernel,'' including:
(1)~a scheduler managing concurrent execution of multiple agents;
(2)~context management distributing the limited context window across agents;
(3)~memory management handling reads and writes of short-term and long-term memory;
(4)~storage management governing files and persistent data; and
(5)~access control governing agent permissions over tools and data.
Experimental results show that this decoupled design yields up to $2.1\times$ speedup.
In our analogy framework, AIOS realizes an early Agent OS prototype whose kernel responsibilities (process management, memory management, and file system modules) map directly onto those of a traditional operating system.

AgentOS~\cite{agentos2025} went further by defining the LLM as a \textit{reasoning kernel} constrained by ``structured operating system logic,'' meaning the LLM is not a free-form inference engine but a controlled reasoning core that must operate within the structural framework of OS logic.
This perspective emphasizes that the LLM's reasoning capabilities must be orchestrated and constrained by structured system logic, rather than allowing the model to generate without restraint.

By 2025--2026, this research thread advanced from academic prototypes toward real desktop and mobile scenarios.
Microsoft's UFO$^2$~\cite{ufo2_2025} organizes a HostAgent (for understanding user intent), an AppAgent (for operating specific applications), a GUI--API hybrid action layer, and virtual-desktop parallel operations into a ``Desktop AgentOS.''
Its key innovation lies in simultaneously supporting both GUI screenshot comprehension and native API calls as interaction modalities.
Aura~\cite{aura2025}, targeting mobile scenarios, adopts a hub-and-spoke topology: a privileged System Agent parses user intent, while multiple sandboxed App Agents each handle a specific domain.
Its design philosophy is ``security first'': inter-agent communication must be mediated through the System Agent, analogous to how inter-process communication in a traditional OS must pass through the kernel.

\paragraph{Relationship to this paper.}
The works above each define a role for the LLM within the system (OS kernel, reasoning kernel, or Agent Kernel) but they have not converged on a unified layered model.
The ICA model proposed in Section~\ref{sec:icam} consolidates these disparate ``kernel'' concepts into a six-layer architecture and explicitly positions the LLM as the core of the \textbf{probabilistic execution plane} (weighted toward the physical execution and inference service layers, L1--L2, and permeating the context management layer, L3), while the Agent OS occupies the \textbf{deterministic control plane} as an orchestration layer; the two planes form a graded crossover around L3--L4 rather than a hard partition.

\subsection{Memory Hierarchy and Virtual Context Management}
\label{sec:related-memory}

Large language models have a finite \textit{context window}; for example, GPT-4 Turbo supports 128K tokens, Claude~3 supports 200K tokens, and Gemini~1.5~Pro reaches up to 1M tokens~\cite{gemini2024}.
Yet regardless of how large the model's window grows, the usable capacity remains bounded, a constraint structurally identical to the ``finite physical memory'' problem in classical computing.
A series of works has therefore borrowed \textbf{memory management} techniques from operating systems to expand the effective context available to agents.

MemGPT~\cite{memgpt2023} (later evolved into the Letta platform) is the foundational work in this direction.
Rather than simply attaching a retrieval-augmented generation (RAG)~\cite{lewis2020rag} pipeline to look up external knowledge, MemGPT draws directly on the \textbf{virtual memory management} mechanism of a traditional OS:
it treats the limited context window as ``main memory,'' external databases as ``disk,'' and employs an OS-like memory manager that automatically pages information fragments in and out between the context window and external storage, giving the agent the illusion of unbounded context.
MemGPT's control flow also explicitly employs \textit{interrupts}: when information must be retrieved from external storage, an interrupt transfers control to the memory manager.
This design maps directly onto the ``virtual memory'' concept in our analogy.

MemOS~\cite{memos2025} elevates the problem to a more systematic plane.
It unifies three granularities of memory into manageable system resources:
(1)~\textit{plaintext memory} (external knowledge stored as text);
(2)~\textit{activation-based memory} (internal hidden-layer representations of the model, analogous to CPU registers); and
(3)~\textit{parameter-level memory} (knowledge encoded in model parameters, analogous to firmware/ROM).
MemOS uses the ``MemCube'' as an encapsulation unit, bundling content with provenance and versioning metadata, and attempts to build a migration bridge between retrieval and parameter learning.
This work corresponds to the ``multi-level storage hierarchy from KV cache to parameter storage'' in our analogy framework.

MemoryOS~\cite{memoryos2025} applies OS memory management principles directly to the agent memory system.
HiAgent~\cite{hiagent2024} (published at ACL~2025) proposes ``hierarchical working memory management,'' using subgoals as memory chunking units to organize long-horizon task memory into a hierarchy, analogous to multi-level page tables in an operating system.
A-MEM~\cite{amem2025} introduces the concept of \textit{agentic memory}: memory is not a passively stored record but a dynamic resource that agents can autonomously organize, recombine, and update.
LongMemEval~\cite{longmemeval2024} further demonstrates that long-term memory is not simply ``storing more chat logs,'' but a complex systems engineering challenge involving information extraction, temporal reasoning, knowledge updating, and the capacity to refuse to answer.

\paragraph{Relationship to this paper.}
Research on memory subsystems has become highly systematic, yet it is typically discussed in isolation from the agent runtime and I/O layers, lacking a full-stack perspective.
More importantly, existing work has not explicitly identified the \textit{KV cache} (the key-value cache maintained during model inference) as the counterpart of ``hardware cache'' in the analogy, whereas this paper argues that the KV cache is precisely the key hardware abstraction bridging the probabilistic compute core and the semantic memory hierarchy.
We elaborate on this correspondence in Section~\ref{sec:kv}.

\subsection{Agent Frameworks and Runtimes}
\label{sec:related-agent-framework}

If the preceding two subsections define the ``kernel'' and ``memory,'' respectively, this section focuses on how programs execute on top of that kernel, that is, the execution frameworks and runtime environments for agents.

\paragraph{Foundational capabilities: interleaving reasoning and action.}
ReAct~\cite{yao2023react} is the foundational work on agent execution paradigms.
Prior to ReAct, reasoning (e.g., Chain-of-Thought) and acting (e.g., invoking tools) were separate, sequential steps for LLMs.
ReAct demonstrated that \textbf{interleaving} reasoning and action within a single loop (``think one step, act one step, observe the result, think again'') significantly improves task completion quality.
This is analogous to the fetch-decode-execute cycle in traditional computing, where the CPU executes an instruction, accesses memory, and checks a condition in tight alternation.

Toolformer~\cite{schick2023toolformer} addressed another foundational question: enabling the model to learn \textbf{when} to call APIs autonomously.
By automatically inserting API call examples into training data, Toolformer empowers the model to decide during generation whether it needs to invoke a particular tool, analogous to the \textit{system call} mechanism in an operating system, whereby a user program initiates a request to the kernel.

\paragraph{The rise of coding agents.}
The most striking agent advances in 2024--2026 have concentrated in software engineering.
OpenAI's Codex~CLI~\cite{openai_codex_overview_2026}, released in April~2025, is a terminal-based agent programming assistant.
It can read code repositories, edit files, execute commands, and operate safely within a sandboxed environment~\cite{openai_codex_sandbox_2026}.
In early 2026, Codex~CLI introduced a subagent capability~\cite{openai_codex_overview_2026}: the main agent can spawn specialized subagents that execute subtasks in parallel, each with its own independent context window and sandbox.
This directly mirrors the \texttt{fork()} model in traditional operating systems, where a parent process creates child processes.
Codex also supports custom instructions via AGENTS.md files~\cite{openai_codex_agents_md_2026} and reusable capability modules through Skills~\cite{openai_codex_skills_2026}.

Anthropic's Claude Code~\cite{anthropic_claudecode_overview_2026} is another representative agent runtime.
Like Codex, it is a terminal-native coding agent capable of editing files, executing shell commands, and connecting to external services.
An independent research study systematically analyzed Claude Code's system architecture~\cite{dive_claudecode_2026} and found that only approximately 1.6\% of its codebase constitutes AI decision logic, while the remaining 98.4\% is \textit{operational harness}, including safety controls, tool orchestration, and context management.
Claude Code likewise supports subagents~\cite{anthropic_claudecode_subagents_2026} and Skills-based extensions~\cite{anthropic_claudecode_skills_2026}, and provides project-level memory through CLAUDE.md files~\cite{anthropic_claudecode_memory_2026}.

Notably, the temporal span of agent execution is expanding from minutes to hours or even days.
Agent runtimes now sustain multi-hour autonomous execution over massive codebases~\cite{anthropic2026agenticreport} (Section~\ref{sec:agentevolution}), implying that they must handle state persistence, checkpoint recovery, and cross-step resource management at a scale far beyond what current interactive sessions demand.

OpenHands~\cite{openhands_paper_2025} (formerly OpenDevin) is an open-source general-purpose agent platform published at ICLR~2025.
It reports approximately 53\% issue resolution on the SWE-bench Verified benchmark~\cite{openhands_paper_2025,swebench2023} and supports multiple LLM backends.
SWE-agent~\cite{sweagent2024} focuses specifically on software engineering tasks, achieving comparable performance with a more streamlined Agent-Computer Interface (ACI) design.

\paragraph{Relationship to this paper.}
These agent frameworks already exhibit operating-system-like characteristics in their architecture: Codex's subagents correspond to process management, Claude Code's CLAUDE.md corresponds to configuration and environment management, and OpenHands' ACI corresponds to the system call interface.
However, they remain independently designed ``application-layer'' tools lacking a unified layered abstraction.
The ICA model proposed in this paper provides a unifying theoretical framework that maps their functionality onto well-defined architectural layers.

\subsection{Tool Calling and I/O Protocols}
\label{sec:related-io}

Agents must interact with the external world: querying databases, calling APIs, manipulating file systems.
In classical computing, this requirement corresponds to the I/O subsystem (input/output devices and buses).
The key development over the past two years is that I/O abstractions are moving from ad hoc tool-calling conventions toward \textbf{standardized protocols}.

HuggingGPT~\cite{hugginggpt2023} is an early representative work: it places the LLM in a ``controller'' role, tasking it with parsing user requests, selecting appropriate expert models, coordinating execution, and aggregating results.
This is analogous to the device driver management layer in an operating system, where the OS uniformly manages calls to different devices (models).
Gorilla~\cite{patil2024gorilla} focuses on a more specific problem: when API documentation is frequently updated, how does the model maintain call accuracy? This corresponds to the version-compatibility challenge for device drivers.

\paragraph{MCP: the model-to-system I/O bus.}
In November~2024, Anthropic introduced the Model Context Protocol (MCP)~\cite{mcp_intro_2026}, defined as an open standard for connecting AI applications with external data sources, tools, and workflows.
MCP adopts a client--server architecture: an MCP Client runs on the AI application side, while an MCP Server encapsulates a specific tool or data source.
Through a unified JSON-RPC protocol, any MCP-compatible AI application can connect to any tool that exposes an MCP Server; Anthropic has likened MCP to ``USB-C for AI.''
By 2026, the MCP ecosystem had expanded rapidly, with Claude Code~\cite{anthropic_claudecode_mcp_2026} and numerous third-party tools all supporting MCP connections.

\paragraph{A2A: the agent-to-agent interconnect protocol.}
In April~2025, Google announced the Agent-to-Agent Protocol (A2A)~\cite{google_a2a_2025} at Cloud Next~'25, explicitly positioned as a complementary open protocol to MCP.
A2A addresses not the ``model-to-tool'' connection problem but the ``agent-to-agent'' collaboration problem, supporting capability discovery, task lifecycle management, long-running tasks, and multimodal coordination.
A2A was donated to the Linux Foundation in 2025~\cite{a2alinux2025}, positioning it as a candidate industry standard.
A2A employs a peer-to-peer client--server model in which every agent can act as both a requesting client and a responding server.

\paragraph{Relationship to this paper.}
Taken together, MCP plays the role of a model-to-system I/O bus (analogous to USB/PCIe), while A2A serves as an agent-to-agent interconnect protocol (analogous to the TCP/IP networking stack).
In the ICA model proposed in this paper, these correspond to Layer~4 (I/O and tool layer) and Layer~5 (interconnect and protocol layer), respectively.
No existing work has yet integrated both protocols into a unified layered abstraction or examined their interactions with context management, kernel scheduling, and other layers at the architectural level.

\subsection{Multi-Agent Collaboration and Distributed Systems}
\label{sec:related-multiagent}

When multiple agents collaborate, the system exhibits characteristics of distributed computing, a development that parallels the historical evolution from standalone machines to networked systems.

AutoGen~\cite{wu2023autogen} (Microsoft Research, 2023) is the foundational framework for multi-agent systems.
Its core design treats \textit{multi-agent conversation} as universal infrastructure: agents in different roles coordinate tasks through natural language dialogue.
AutoGen has since evolved to v0.4 and been integrated into the Microsoft Agent Framework, supporting scalable, production-grade multi-agent deployments.

MetaGPT~\cite{metagpt2023} (ICLR~2024 Oral) introduced a critical design idea: explicitly encoding \textbf{Standard Operating Procedures} (SOPs) from human organizations into multi-agent collaboration.
Specifically, MetaGPT simulates a virtual software company: a product manager writes requirements, an architect designs the system, engineers write code, and a QA engineer tests, with each role being an agent and the SOP defining the workflow and information flow among them.
This is analogous to \textit{workflow orchestration} in distributed systems, constraining unordered agent interactions into an ordered assembly line.

This pipeline pattern has been validated at industrial scale.
Fountain's Copilot system employs a central orchestrator agent that coordinates three specialized sub-agents for candidate screening, document generation, and sentiment analysis.
This hierarchical multi-agent deployment achieved 50\% faster candidate screening, 40\% faster onboarding, and $2\times$ candidate conversion rates, compressing the full staffing center configuration cycle from over one week to under 72~hours~\cite{anthropic2026agenticreport}.
This hub-and-spoke topology contrasts sharply with MetaGPT's pipeline topology and AutoGen's peer-to-peer topology, suggesting that coordination topology itself is a key design variable influencing multi-agent system performance~\cite{wu2023autogen,metagpt2023}.

The Internet of Agents (IoA)~\cite{ioa2024} (published at ICLR~2025) envisions the agent ecosystem from an ``internet'' rather than single-system perspective.
Its core contribution is an agent integration protocol supporting dynamic team formation, collaboration, and distributed execution among heterogeneous agents.
IoA draws design inspiration directly from the Internet: just as the Internet connects computers from different manufacturers running different operating systems, IoA aims to connect agents built with different frameworks by different organizations.
This effort has also spurred IETF-level standardization (IoA Protocol Internet Draft).

\paragraph{Relationship to this paper.}
Multi-agent systems already exhibit ``distributed system'' morphology at the level of the phenomena they study: AutoGen corresponds to message-passing mechanisms, MetaGPT to workflow orchestration, and IoA to network interconnect protocols.
However, most work remains at the ``framework and protocol'' layer without integrating individual-agent memory management, kernel scheduling, I/O, and access control into a unified architecture.
The ICA model maps multi-agent collaboration onto Layer~5 (interconnect and protocol layer) and Layer~6 (application layer), and discusses cross-layer interaction design principles.

\subsection{Cognitive Architectures and Security Governance}
\label{sec:related-cognition-security}

The preceding subsections have surveyed related work from a functional systems perspective (kernel, memory, runtime, I/O, multi-agent).
This subsection turns to two higher-level themes: what cognitive architecture should agent systems follow, and how should their security be governed?

\paragraph{Cognitive architectures.}
CoALA (Cognitive Architectures for Language Agents)~\cite{sumers2024coala} provides a theoretical framework grounded in cognitive science.
CoALA organizes language agents into three core components:
(1)~\textit{modular memory} (divided into short-term working memory and long-term memory: semantic, episodic, and procedural);
(2)~\textit{structured action space} comprising internal actions (reasoning, memory retrieval) and external actions (tool invocation, environment interaction); and
(3)~a \textit{generalized decision-making process}, namely a plan--execute--observe loop.
The value of CoALA lies in providing the theoretical grounding for ``why this system is not merely a software stack assembly, but a genuine cognitive architecture.''
L2MAC~\cite{holt2024l2mac} (ICLR~2024) explicitly frames its architecture as the ``first practical LLM-based stored-program automatic computer (von Neumann architecture),'' comprising an instruction register, file storage, a control unit, and independent LLM agent modules.
Mi et al.~\cite{mi2025llmagentscomputersystems} organize their LLM agent survey from a ``computer systems perspective,'' explicitly inspired by the von Neumann architecture: they decompose agents into perception, cognition, memory, tool, and action modules, observe that the context window is analogous to main memory and databases to disk, and point out that agents currently lack a ``cache module.''

\paragraph{Security governance.}
ArbiterOS~\cite{arbiteros2025} identifies a fundamental tension: we are using the mental model of \textbf{deterministic software engineering} to direct what are \textbf{inherently probabilistic processors}.
This contradiction demands the introduction of a governance layer at the system architecture level.
ArbiterOS accordingly proposes the ``Agentic Computer'' mental model, redefining the LLM as a \textit{Probabilistic CPU} and designing governance-layer concepts including a governor, a formal instruction set, and a hardware abstraction layer (HAL).
Within our analogy framework, ArbiterOS's core insight, that models are probabilistic and require a deterministic governance layer, is one of the direct inspirations for the \textit{dual-plane architecture} (probabilistic execution plane + deterministic control plane) proposed in this paper.

CaMeL~\cite{debenedetti2025camel} (Google DeepMind, 2025) applies operating-system capability-security principles to LLM agents from a safety perspective.
Its core design philosophy is ``Don't execute data'': strictly separating trusted instructions from untrusted data, extracting control flow and data flow, and enforcing separate security policies on each to defend against prompt injection attacks.
IronClaw~\cite{ironclaw2025} (2026) explicitly advocates protecting AI agents in the same way an operating system protects its processes, migrating OS security mechanisms such as privilege separation, sandbox isolation, and audit logging into the agent system.

\paragraph{Relationship to this paper.}
The cognitive architecture line of work provides theoretical foundations, while the security governance line supplies engineering constraints.
Yet these two threads remain unintegrated: cognitive architectures rarely address security, and security governance rarely considers cognitive structure.
The dual-plane architecture proposed in this paper attempts to unify both: the probabilistic execution plane carries cognitive capabilities, while the deterministic control plane enforces security governance, and the two interact through well-defined interface contracts.

\subsection{Common Boundaries and Unsolved Problems}
\label{sec:related-gap}

The six themes surveyed above collectively approach the vision of a ``model-native computing architecture,'' yet four critical problems remain unsolved in a unified manner.

Figure~\ref{fig_coverage_map} summarizes the coverage of system layers by existing works.
\input{figures/fig_coverage_map.tex}

\paragraph{Problem 1: What role does the LLM actually play?}
Existing work exhibits a fundamental divergence in positioning the LLM within the system.
Ge et al.\ and AIOS place the LLM at the core of the OS~\cite{ge2023llmos,mei2024aios}, where the LLM \textit{is} the kernel;
ArbiterOS demotes it to a ``Probabilistic CPU'' governed by the kernel~\cite{arbiteros2025}, treating the LLM as a managed processor;
AgentOS defines it as a reasoning kernel constrained by structured OS logic~\cite{agentos2025}, regarding the LLM as a controlled reasoning core.
These three positions lead to fundamentally different system designs.
The ICA model proposed in this paper offers a unified answer: the LLM is the core of the \textbf{probabilistic execution plane} (analogous to the CPU, weighted toward L1--L2 and permeating L3), and the Agent OS is the orchestration layer within the \textbf{deterministic control plane} (analogous to the OS kernel); the two planes form a graded crossover around L3--L4 rather than a hard partition.
The LLM provides compute capability, while the Agent OS provides management and scheduling, just as the CPU and the OS kernel are two independent yet tightly cooperating layers in classical computer architecture.

\paragraph{Problem 2: No full-stack layered model exists.}
Measured against the standards that make a classical architecture a \emph{discipline}---a stable hardware/software interface contract (the ISA), a clear boundary between probabilistic execution and deterministic control, and quantitative design laws (Amdahl, Roofline)~\cite{hennessy2017quantitative}---no mainstream model-native work simultaneously provides all three.
L2MAC~\cite{holt2024l2mac} focuses on the stored-program computer model; Mi et al.~\cite{mi2025llmagentscomputersystems} provide a von Neumann-inspired survey framework, but each covers only one facet of the architecture.
Memory management work does not discuss I/O protocols; agent frameworks do not discuss KV cache; security governance does not discuss multi-agent collaboration; each line of work makes genuine contributions, but their boundaries are clearly delineated.

\paragraph{Problem 3: The probabilistic--deterministic tension has not been codified architecturally.}
ArbiterOS~\cite{arbiteros2025} correctly identifies the tension between probabilistic processors and deterministic governance, but it proposes only a conceptual ``governance-first'' paradigm.
CaMeL~\cite{debenedetti2025camel} and IronClaw~\cite{ironclaw2025} provide concrete security mechanisms, but these are not embedded within a complete layered architecture.
This paper \textit{codifies this tension architecturally} through the dual-plane architecture: the probabilistic execution plane carries the LLM's inference capabilities, while the deterministic control plane enforces OS-level management and security responsibilities, with the two planes cooperating through explicit interface protocols.

\paragraph{Problem 4: Structural constraints of human--agent collaboration have not been modeled.}
Existing work treats the human as a boundary condition of the system (Layer~6 user), but recent industrial evidence reveals a ``collaboration paradox'': engineers use AI for approximately 60\% of their work, yet can fully delegate only 0--20\% of their tasks~\cite{anthropic2026agenticreport}.
Engineers tend to delegate tasks that are ``easy to verify or low-risk'' while retaining those that are ``conceptually difficult or depend on design decisions.''
This structural constraint suggests that the human--agent interaction loop should become an explicit component of the architecture, rather than an external assumption.

Table~\ref{tab:artifact-comparison} consolidates the distinction: across the five architectural artifacts that distinguish a \emph{discipline} from a collection of metaphors, no single prior work combines more than one, whereas ICA integrates all five.

\begin{table}[h]
\centering
\footnotesize
\caption{Architectural artifacts distinguishing ICA from prior work. Each row lists an artifact, the closest prior approach, and ICA's contribution; no single prior work combines all five.}
\label{tab:artifact-comparison}
\begin{tabularx}{\textwidth}{>{\raggedright\arraybackslash}p{0.24\textwidth}>{\raggedright\arraybackslash}p{0.42\textwidth}>{\raggedright\arraybackslash}p{0.26\textwidth}}
\toprule
\textbf{Architectural artifact} & \textbf{Closest prior approach} & \textbf{ICA (this paper)} \\
\midrule
Full-stack layered model & Mi et al.~\cite{mi2025llmagentscomputersystems}: modular map, no inter-layer contracts & Six-layer L1--L6 with interface contracts \\
Dual-plane separation & ArbiterOS~\cite{arbiteros2025}: a governance layer over a ``probabilistic CPU'' & Dual-plane (probabilistic execution + deterministic control) \\
Interface contracts & None; layers described informally & Five formal inter-layer contracts \\
Design axioms & None & Six transplanted axioms (locality, abstraction, probabilistic execution, \ldots) \\
Quantitative heuristics & None & Three Amdahl-style heuristics \\
\bottomrule
\end{tabularx}
\end{table}

These four unsolved problems define the contribution space of this paper: providing a unified, layered, dual-plane architectural model that integrates all six threads within a single coherent framework.

%% file: figures/fig_coverage_map.tex
\begin{figure}[htbp]
  \centering
  \includegraphics[width=0.95\textwidth]{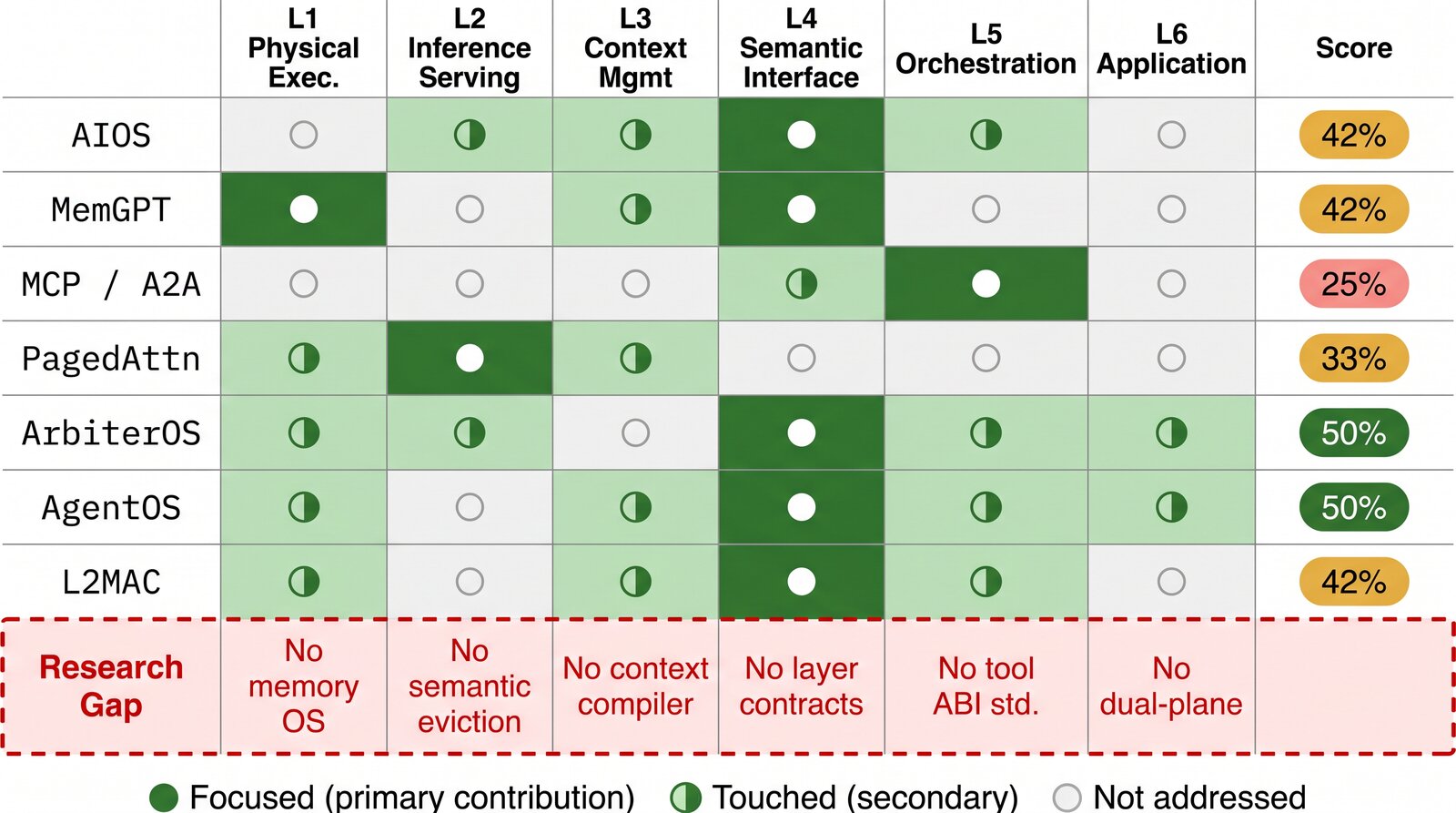}
  \caption{Coverage of ICA layers by existing works.
  Cells are classified as \emph{Focused} (primary contribution),
  \emph{Touched} (layer addressed but not central),
  or \emph{Not addressed}.
  The Score column shows the weighted coverage percentage
  ($2{\times}\text{Focused}+\text{Touched}$, out of a maximum of 12),
  color-coded as high ($\geq$50\%, green), medium (33--49\%, amber), or low ($<$33\%, red).
  The bottom row identifies the specific open gap in each layer that no single prior work has closed;
  L6~Application is the least-covered layer overall.}
  \label{fig:coverage_map}\label{fig_coverage_map}
\end{figure}

%% file: en_sections/section_framework.tex
\section{Analogy Framework: Intelligent Systems as Layered Computers}
\label{sec:framework}

\subsection{Principles of Analogy}

The analogy framework underlying this paper is summarized in Table~\ref{tab:mapping} and illustrated in Figure~\ref{fig_analogy_map}.
Before presenting the detailed mapping, we establish three principles to ensure that the analogy serves as a rigorous analytical instrument rather than a decorative metaphor.

\paragraph{Principle 1: The analogy must correspond to genuine engineering mechanisms.}
The observation ``KV cache resembles a processor cache'' holds not because the names sound alike, but because both serve the same fundamental goal of exploiting \textit{temporal locality} and \textit{state reuse}~\cite{drepper2007memory,kwon2023pagedattention}.
Temporal locality, the empirical regularity that recently accessed data is likely to be accessed again in the near future, governs both a CPU cache line and a recently computed set of KV vectors.
Similarly, the observation ``PagedAttention resembles virtual memory'' is valid because vLLM literally organizes KV blocks into fixed-size pages and maps them through a block table to non-contiguous physical memory, mirroring the mechanism by which an operating system manages virtual pages~\cite{kwon2023pagedattention}.
In each case, the analogy is grounded in shared engineering structure, not superficial naming.

\paragraph{Principle 2: The analogy must acknowledge its boundaries.}
A model is not a deterministic CPU: executing \texttt{ADD R1, R2} always produces the same result, whereas the same prompt may yield different responses across invocations.
Semantic retrieval is not exact addressing: \texttt{MOV [0x7fff], EAX} always writes to the same location, while vector similarity search returns results that vary with the embedding model and query phrasing.
An agent runtime is not a mature operating system: the Linux kernel rests on four decades of formal verification and regression testing, whereas the safety mechanisms of current agent frameworks are still evolving rapidly.
These ``not quite identical'' gaps are not weaknesses of the analogy; rather, they define precisely the research space where the most impactful future work lies.

\paragraph{Principle 3: The goal of the analogy is to generate research questions.}
A productive analogy does more than aid understanding; it surfaces new engineering problems.
Which modules require clearly defined interface contracts?
Which shared states demand consistency guarantees?
Which execution paths require access control and audit logging?
The rightmost column of Table~\ref{tab:mapping} captures the research directions that the analogy generates.


\scriptsize
\renewcommand{\arraystretch}{1.25}
\begin{longtable}{p{0.11\textwidth}p{0.13\textwidth}p{0.23\textwidth}p{0.23\textwidth}p{0.20\textwidth}}
\caption{Analogy mapping between classical computer architecture and the model-native system stack. Each row pairs a classical concept with its model-native counterpart, along with the core similarity, fundamental difference, and resulting engineering implication.}
\label{tab:mapping} \\
\toprule
\textbf{Classical Concept} & \textbf{Model-Native Counterpart} & \textbf{Core Similarity \& Example} & \textbf{Fundamental Difference} & \textbf{Engineering Implication} \\
\midrule
\endfirsthead
\toprule
\textbf{Classical Concept} & \textbf{Model-Native Counterpart} & \textbf{Core Similarity \& Example} & \textbf{Fundamental Difference} & \textbf{Engineering Implication} \\
\midrule
\endhead

ISA / ABI & Prompt templates, tool specifications, Skill/Agent interfaces & Both define the calling contract and composition boundary visible to upper layers. Example: RISC-V's \texttt{ECALL} instruction has precisely defined semantics~\cite{riscv_spec_2026}; MCP's \texttt{tools/call} interface specifies parameter schemas and return formats~\cite{mcp_intro_2026} & Prompts lack strict formal semantics; ambiguity and drift are possible, and the same prompt may behave differently across model versions & Requires stable schemas, version identifiers, and replay compatibility \\[3pt]

CPU Core & Foundation model forward pass & Both are general-purpose execution cores that map inputs to outputs. Example: a CPU maps instruction sequences to register state changes; a model maps token sequences to next-token probability distributions~\cite{vaswani2017attention,brown2020gpt3} & The model performs probabilistic inference, not deterministic logic; identical inputs can yield different outputs & Explicitly decouple model capability, inference strategy, and runtime \\[3pt]

Micro-architecture & Attention kernels, decoding algorithms, serving engines & Both determine the real performance and cost behind the same ``interface.'' Example: superscalar pipelines vs.\ FlashAttention's I/O-aware tiled computation~\cite{dao2022flashattention}; branch prediction vs.\ speculative decoding~\cite{leviathan2023speculative} & Model-era ``micro-architecture'' spans the software--hardware boundary & Joint optimization of FlashAttention, speculative decoding, and parallelism strategies \\[3pt]

L1/L2 Cache & Session-level KV cache, prefix caching & Both exploit temporal locality and hotspot reuse. Example: a CPU L1 cache retains recently accessed cache lines; a KV cache retains recently computed Key/Value vectors to avoid redundant computation~\cite{kwon2023pagedattention,vllm_prefix_2026} & Semantic cache hit criteria are far weaker than address-based hit criteria; address equality is a Boolean check, while semantic similarity is a continuous-valued comparison & Requires block-level allocation, sharing, invalidation, and monitoring \\[3pt]

Main Memory / Virtual Memory & Context window, external memory, virtual context & Both create the illusion of a large space through layering. Example: virtual memory gives each process the illusion of a contiguous, large address space; virtual context management gives an agent the illusion of unlimited memory~\cite{memgpt2023} & Semantic summarization loses information (it is not lossless paging); OS page swap is byte-exact, while summary compression is irreversible & Requires hot/warm/cold memory tiering, re-retrieval, and backfill strategies \\[3pt]

Operating System & Agent runtime & Both manage scheduling, permissions, I/O, and failure handling. Example: Linux cgroups isolate process resources~\cite{linux_cgroup_2026}; Codex sandboxes isolate agent file access~\cite{openai_codex_sandbox_2026} & Agents entail non-deterministic reasoning and language-mediated interfaces & Requires sandboxing, approval gates, sub-agent orchestration, hooks, and logging \\[3pt]

System Calls / Drivers & Tool calls, MCP servers & Both integrate external capabilities through a unified access interface. Example: the \texttt{open()} system call standardizes file operations; MCP's \texttt{tools/call} standardizes tool invocations~\cite{mcp_intro_2026} & Tool results may be non-deterministic, asynchronous, and carry strong side effects & Requires capability annotations, authentication, and idempotent design \\[3pt]

File System / Disk & Vector stores, object storage, knowledge bases & Both store high-volume, persistent state. Example: ext4 organizes files by inode; Milvus organizes knowledge by vector index~\cite{lewis2020rag} & Semantic retrieval is not exact addressing; \texttt{find(path)} returns a unique result, while vector search returns top-$k$ approximate matches & Requires provenance tracking, versioned snapshots, and TTL policies \\[3pt]

Process / Thread & Agent / Sub-agent / Task graph & Both serve as the fundamental unit of concurrent execution and isolation. Example: POSIX threads share an address space but have independent stacks; Codex sub-agents share project context but execute independently~\cite{openai_codex_overview_2026} & Sharing semantic state is more fragile than sharing memory; ``reading the same document'' does not imply ``having the same understanding'' & Requires budgeting, prioritization, and result merging \\

\bottomrule
\end{longtable}
\renewcommand{\arraystretch}{1.0}
\normalsize

Within this framework, a future model-native computing architecture can be understood as follows: \textbf{a system that uses the foundation model as its computing core, KV and context hierarchies as its storage layer, an agent kernel as its control layer, tools and the external world as its I/O layer, and serving clusters with shared memory as its distributed substrate.}
In Section~\ref{sec:components}, we unpack each component in detail, proceeding from core to periphery.

\subsection{Metaphor Conflict and the Dual-Plane Architecture}

Before examining individual components, we must resolve a fundamental conceptual tension.
As discussed in Section~\ref{sec:related}, the existing literature is deeply divided on where the LLM sits in the architectural stack.
Ge et al.\ place the LLM in the position of the OS kernel~\cite{ge2023llmos}.
AIOS incorporates it alongside resource management, context management, and access control as kernel-level services~\cite{mei2024aios}.
AgentOS treats the LLM as a ``reasoning kernel'' constrained by structured operating system logic~\cite{agentos2025}.
ArbiterOS demotes the LLM to a ``Probabilistic CPU'' governed by a separate control layer~\cite{arbiteros2025}.

These positions appear contradictory, but in reality each captures only one facet of the system.
We argue that a model-native computing system simultaneously encompasses \textbf{two distinct planes}, and that prior work has simply been looking at different cross-sections.

\subsubsection{The Probabilistic Execution Plane}

The probabilistic execution plane is, at its core, the foundation model's forward inference process.

\paragraph{Concrete example.}
When a user submits a code review request to GPT-4, the model's attention mechanism performs matrix operations over billions of parameters, ultimately producing a token probability distribution from which the response is sampled.
The same input may yield different outputs across invocations; the quality of the output depends heavily on how the context is organized; and the model itself holds no persistent state, as the results of one conversation do not automatically carry over to the next.

The key characteristics of this plane are:
\begin{itemize}[nosep]
\item \textit{Probabilistic output:} the output follows a probability distribution; identical inputs can produce different results.
\item \textit{Context sensitivity:} small changes in prompt formulation can significantly alter output quality.
\item \textit{No persistent state:} model weights remain static across requests; all ``memory'' resides in the tokens within the context window.
\end{itemize}

KV caches, attention mechanisms, and decoding strategies all belong to the ``micro-architecture'' of this plane, optimizing the efficiency and quality of the model's forward computation itself.

\subsubsection{The Deterministic Control Plane}

The deterministic control plane is the programmatic control layer that wraps around the model: the scheduler, permission enforcer, state machine, logging system, and failure recovery logic.

\paragraph{Concrete example.}
When Claude Code receives the request ``refactor this function,'' the deterministic control plane executes the following sequence: check current directory permissions (is the path within the allowed project scope?) $\to$ load the project memory file (\texttt{CLAUDE.md}) $\to$ parse the request and decompose it into subtasks $\to$ for each step involving file writes, determine whether user confirmation is required based on the approval policy $\to$ upon completion, write a change log to the audit record~\cite{anthropic_claudecode_security_2026}.
Every step in this sequence is executed by deterministic code; its behavior is predictable, auditable, and replayable.

The key characteristics of this plane are:
\begin{itemize}[nosep]
\item \textit{Deterministic execution:} given the same inputs and policy configuration, the control plane always reaches the same decisions.
\item \textit{Formal policies:} permission rules and approval gates are explicitly defined in code or configuration files.
\item \textit{Auditability:} every decision step is logged and supports post-hoc review.
\end{itemize}

Codex's approval policies~\cite{openai_codex_approvals_2026}, Claude Code's read-by-default stance and hook system~\cite{anthropic_claudecode_security_2026}, and CaMeL's capability-based security mechanisms~\cite{debenedetti2025camel} all belong to this plane.

\subsubsection{Why the Separation Is Necessary}

Industrial practice strongly suggests the necessity of this separation (the accounts below are practitioner and vendor reports rather than peer-reviewed validation).
Praetorian's development platform embeds LLMs within a deterministic orchestration layer, reporting improved agent reliability in autonomous software development~\cite{praetorian2025deterministic}.
The ``deterministic shell, probabilistic core'' pattern---formally articulated by ArbiterOS~\cite{arbiteros2025} and echoed in practitioner discussions~\cite{towardsai2025deterministicshell}---is becoming a widely adopted architectural pattern for agent systems.
The ``Blueprint First, Model Second'' framework encodes operational logic as deterministic source-code blueprints, constraining the LLM to operate within well-defined boundaries~\cite{blueprintfirst2025}.

The core rationale for separation is a clean division of responsibilities:
\begin{itemize}[nosep]
\item The probabilistic plane excels at \textit{understanding and generation}: it answers what \emph{can} be done.
\item The deterministic plane excels at \textit{control and assurance}: it governs what \emph{should} be done.
\item The probabilistic plane should \textit{not} directly execute side-effectful operations (writing files, calling external APIs); instead, such actions must be mediated through the deterministic plane's approval and audit channels.
\end{itemize}

This separation finds strong support in the \textbf{collaboration paradox} introduced in Section~\ref{sec:intro}: engineers use AI heavily yet fully delegate only a small fraction of tasks~\cite{anthropic2026agenticreport}.
This is not a temporary artifact of insufficient model capability; it is a structural property of dual-plane systems.
The deterministic control plane, including human judgment transmitted through permission policies and approval workflows, naturally constrains the action space of the probabilistic plane.
Engineers tend to delegate tasks that are ``easy to verify or low-risk'' while retaining tasks that are ``conceptually difficult or depend on design decisions.''
This is precisely the behavior of a deterministic plane acting as a gate for the probabilistic plane.
Without separation, if the model could directly execute any operation, such graduated delegation would be impossible.

The analogy to classical architecture is instructive.
In a traditional system, user-mode code cannot access hardware directly; it must issue a system call that traps into the kernel~\cite{ostep_2023}.
User programs (the probabilistic plane) may issue arbitrary requests, but the kernel (the deterministic plane) decides which requests are permitted, at what priority they execute, and how they are logged and recovered.

\paragraph{The metaphor conflict is thereby resolved.}
When ArbiterOS calls the LLM a ``Probabilistic CPU,'' it is describing the probabilistic execution plane.
When AIOS incorporates the LLM into a ``kernel,'' it is observing a mixture of the probabilistic plane and partial control logic.
When AgentOS proposes a ``reasoning kernel,'' it is attempting to impose the structure of the control plane onto the probabilistic plane.
These positions are not contradictory; they are complementary descriptions of the same dual-plane system, viewed from different vantage points.

\paragraph{A derived consequence.}
Because every irreversible, side-effecting operation must pass through the deterministic control plane (Axioms~5--6), the dual-plane structure concentrates accountability at a single, well-defined choke point: the full record of what an agent did---and could be held responsible for---is reconstructable from the deterministic-plane log alone (developed in Section~\ref{sec:challenges}). This is a concrete consequence of the separation, not merely a relabeling.

Figure~\ref{fig_dual_plane} illustrates the dual-plane architecture and its mapping onto the ICA layers.
\input{figures/fig_dual_plane.tex}

%% file: figures/fig_dual_plane.tex
\begin{figure}[htbp]
  \centering
  \includegraphics[width=0.8\textwidth]{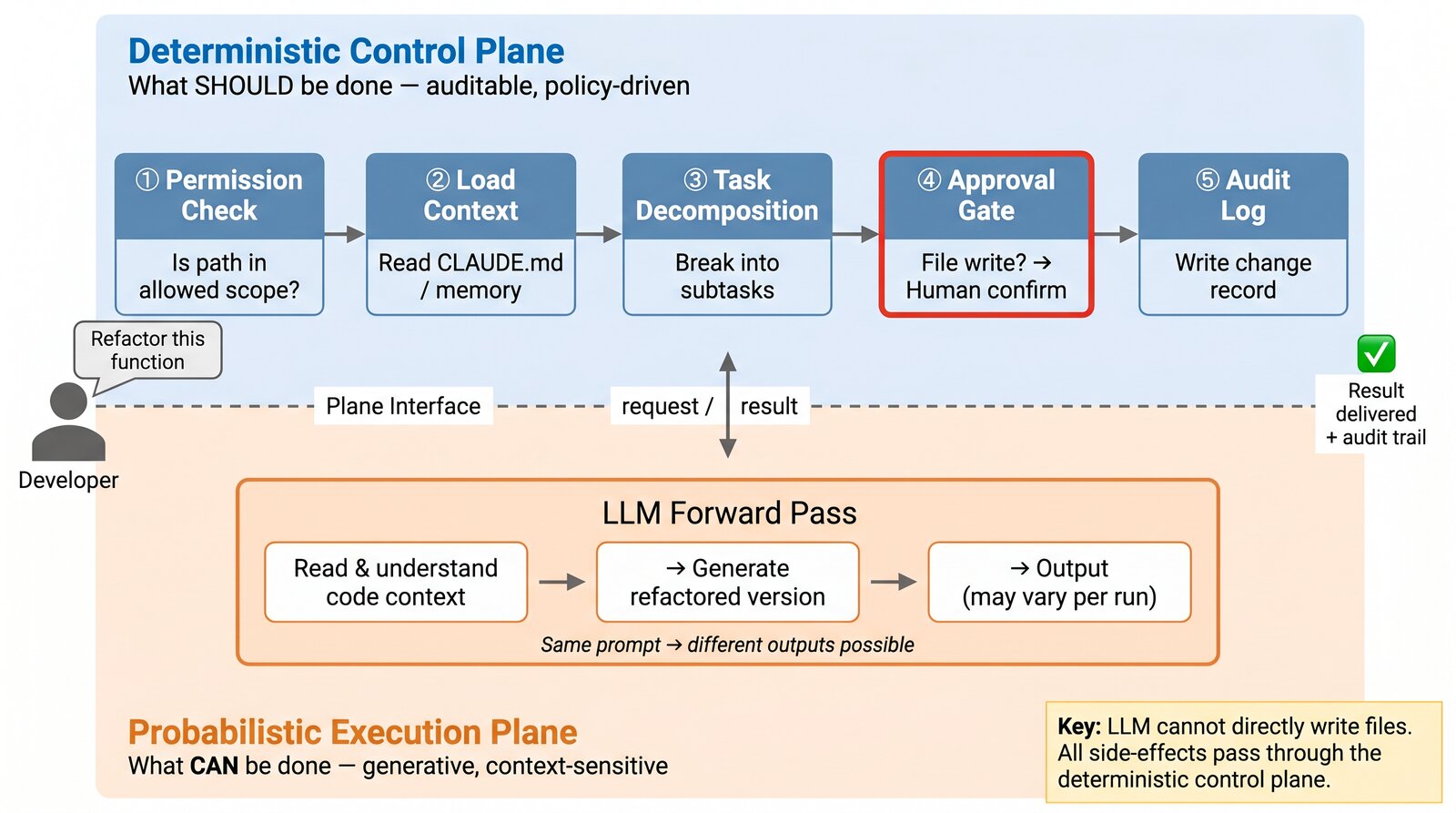}
  \caption{Dual-plane architecture illustrated through a concrete software-engineering scenario.
  When a developer submits ``Refactor this function,'' the \emph{deterministic control plane} (blue, top)
  executes a fully auditable, policy-driven sequence---permission check, context loading,
  task decomposition, human-approval gate for file writes, and audit logging---while the
  \emph{probabilistic execution plane} (orange, bottom) performs the LLM forward pass that
  understands the code and generates the refactored output.
  The key architectural property is that the LLM cannot directly execute side-effectful
  operations: every file write, API call, or state mutation must pass through the deterministic
  plane's approval and audit channels.
  The probabilistic plane answers \emph{what can be done};
  the deterministic plane governs \emph{what should be done}.}
  \label{fig_dual_plane}
\end{figure}

%% file: en_sections/section_components.tex
\section{Key Component Analysis}
\label{sec:components}

This section analyzes six key components in order from core to periphery:
foundation model (corresponding to the CPU) $\to$ KV cache (corresponding to the cache hierarchy) $\to$ context management (corresponding to virtual memory) $\to$ agent runtime (corresponding to the operating system) $\to$ tool bus (corresponding to I/O) $\to$ multi-agent collaboration (corresponding to distributed systems).
For each component, we follow a uniform four-part structure: (1)~the classical computer architecture counterpart, (2)~the corresponding construct in LLM-based systems, (3)~why the two are analogous at the mechanism level, and (4)~the essential differences that the analogy should not obscure.

\subsection{Foundation Model: Probabilistic Computing Core}
\label{sec:model}

\subsubsection{The Counterpart in LLM-Based Systems}

\textit{Classical counterpart---the CPU:} an execution core that maps input state to output state by running a formally specified instruction set (the ISA) with full \textbf{determinism}---\texttt{ADD R1, R2, R3} executed a million times yields the identical result~\cite{riscv_spec_2026}.

In a model-native computing system, the foundation model (e.g., GPT-4, Claude, LLaMA) plays the role of a \textit{general-purpose execution core}.
Given an input context comprising a system prompt, user messages, tool results, and control constraints, the model produces the next action or output through a forward pass~\cite{vaswani2017attention,brown2020gpt3,llama2023}.

Prompts, tool descriptions, skill files, and project memory files (such as \texttt{AGENTS.md} or \texttt{CLAUDE.md}) can be understood as a form of \textbf{soft instruction interface}: they do not alter the model's physical weights (just as software does not alter the CPU's circuitry), yet they significantly reshape the model's execution path~\cite{openai_codex_agents_md_2026,openai_codex_skills_2026,anthropic_claudecode_memory_2026}.
For example, a carefully authored \texttt{CLAUDE.md} file can cause the same model to exhibit ``Python expert'' behavior in one project and ``Rust expert'' behavior in another, much as loading different programs onto the same CPU yields different behaviors.

SGLang has already moved in this direction: it reformulates complex language-model programs as a collaboration between a structured front-end language and a high-performance runtime, analogous to the transition from assembly language to high-level languages~\cite{sglang2024,sglang_docs_2026}.

\subsubsection{Why the Analogy Holds}

The analogy holds because the two cores share the same relationship between substrate and behavior.
In both, the substrate is fixed while behavior is customized through an external layer---programs for the CPU, prompts and tool schemas for the model---so the same silicon runs a word processor or a simulation just as the same weights write code, analyze data, translate languages, or plan a task.
Both also keep a stable interface while the implementation evolves beneath it.
A CPU exposes its operations through the ISA (arithmetic, branching, memory access); a model exposes its operations through prompt templates and tool schemas (reasoning, generation, tool invocation), and both adhere to the principle of a stable interface over an evolvable implementation.
The x86 architecture has persisted for over four decades while its microarchitecture traveled from single-scalar to superscalar out-of-order execution, much as the GPT-4 API has remained stable across successive weight updates.
Below that interface lies a rich optimization space, and in both cases engineers chase performance without breaking the contract above: CPUs benefit from pipelining, branch prediction, and superscalar issue, while models benefit from FlashAttention (I/O-aware attention computation~\cite{dao2022flashattention}), speculative decoding (using a small model to predict a larger model's output~\cite{leviathan2023speculative}), and continuous batching.

\subsubsection{Essential Differences}

The crux of the difference is \textbf{determinism versus probability}.
A CPU executes a formal instruction set in which every instruction has precisely defined, long-term-stable, fully reproducible semantics.
A foundation model processes a mixture of natural language, code, and structured schemas, producing outputs drawn from a probability distribution.
As Mi et al.\ have argued, drawing on the evolution of computer architecture to inform agent system design requires candid acknowledgment of this fundamental gap between probabilistic and deterministic execution~\cite{mi2025llmagentscomputersystems}.

Consequently, the future ``intelligent ISA'' should not be understood as hard-coding prompts as fixed instructions. Rather, the research agenda should prioritize developing more stable interface layers: tool schemas, constrained output formats, reusable skill packages, task-specific DSLs, and explicit context declarations.
From an architectural perspective, this means the community should evaluate models not only by parameter count but also by \textbf{model programmability}, namely the degree to which upper layers can reliably and predictably control model behavior.
Figure~\ref{fig:comp51:model} illustrates the structural parallel between the CPU--ISA contract and the foundation model's prompt-and-tool-schema interface.

\input{figures/fig_comp_51_model}

\subsection{KV Cache: Cache Hierarchy for Intelligent Systems}
\label{sec:kv}

\subsubsection{The Counterpart in LLM-Based Systems}

\textit{Classical counterpart---the cache hierarchy:} L1/L2/L3 caches bridge the processor--memory speed gap by exploiting \textbf{temporal locality}, managed through placement, replacement (e.g., LRU), and write policies~\cite{drepper2007memory}.

During autoregressive inference, generating each new token requires computing attention between that token's Query vector and the Key/Value vectors of \textit{all} preceding tokens.
Recomputing the full prefix KV at every step would incur $O(n^2)$ cost in sequence length.
The KV cache avoids this by storing previously computed Key and Value vectors, reducing each attention step to an $O(1)$ lookup and bringing overall inference cost from $O(n^2)$ to $O(n)$.
This rationale mirrors that of classical caching: \textbf{trade capacity (GPU memory) for latency and bandwidth advantages}~\cite{drepper2007memory,kwon2023pagedattention}.

\paragraph{A KV cache hierarchy is emerging:}
\begin{itemize}[nosep]
\item \textbf{L1---Session-level KV cache:} KV vectors within a single session; lowest latency, capacity bounded by GPU memory.
\item \textbf{L2---Prefix/shared cache:} KV blocks shared across requests for common system prompts or codebase indices, analogous to read-only code pages shared among processes in a CPU~\cite{vllm_prefix_2026,sglang_docs_2026,gim2024promptcache}.
\item \textbf{L3---Semantic cache:} Reusable tool-call results for similar queries, or reusable indices for identical repository structures, producing ``semantic-level'' cache hits.
\end{itemize}

When vLLM introduced PagedAttention, which organizes KV vectors into fixed-size blocks mapped through a block table to non-contiguous physical memory, the analogy ceased to be merely heuristic and became a direct borrowing of virtual-memory page management~\cite{kwon2023pagedattention}.

\subsubsection{Why the Analogy Holds}

The similarity between KV caches and CPU caches operates at the level of engineering mechanisms, and four of those mechanisms line up almost one-to-one.

\paragraph{Shared reliance on temporal locality.}
CPU caches retain recently accessed cache lines because data used recently will be used again soon, and KV caches retain recently computed KV vectors because generating the next token requires attending to all prior tokens.
Both keep frequently reused data in fast storage to avoid expensive recomputation or memory accesses.

\paragraph{Capacity pressure and replacement decisions.}
L1 caches have limited capacity and employ LRU-style replacement policies, and KV caches face analogous GPU memory pressure.
H2O identifies ``heavy-hitter'' tokens (those receiving the highest attention weights) and implements intelligent eviction~\cite{h2o2023}, analogous to identifying ``hot data'' in CPU caches.
StreamingLLM discovered that initial tokens receive disproportionately high attention (the ``attention sink'' phenomenon) and achieved unlimited-length streaming inference by pinning these sink tokens~\cite{xiao2024streamingllm}, which is directly analogous to cache \textit{pinning} in CPU caches.

\paragraph{Bandwidth as the binding bottleneck.}
Gholami et al.\ observe that LLM inference during the decode phase is fundamentally memory-bandwidth-bound, not compute-bound~\cite{gholami2024memorywall}, and Yuan et al.\ systematically characterized this behavior using the Roofline model~\cite{yuan2024llminference}.
Li et al.\ organized KV cache management strategies into three tiers---token-level (which KV entries participate in attention), model-level (architectural optimizations such as Grouped-Query Attention (GQA) and Multi-Query Attention (MQA)), and system-level (memory management and scheduling)~\cite{likvcache2024survey}---mirroring the multi-level approach taken in CPU cache optimization.

\paragraph{Compression as a shared optimization direction.}
KVQuant achieved $8\times$ compression~\cite{hooper2024kvquant}, and KIVI applied asymmetric 2-bit quantization with nearly lossless accuracy and dramatically reduced memory footprint~\cite{liu2024kivi}, paralleling data compression in CPU caches (e.g., IBM MXT) to increase effective capacity.

\subsubsection{Essential Differences}

\paragraph{Boolean versus continuous hits.}
CPU cache hits are \textbf{Boolean}: the address matches or it does not.
In the KV cache, prefix caching does involve exact matching (identical prefix token sequences are reused), but upper-level semantic cache hits are \textbf{continuous-valued}: a similarity score exceeding a threshold is treated as a hit, and different thresholds yield different precision--recall trade-offs.
Classical cache hit-rate models therefore do not transfer directly to semantic caching.

\paragraph{Hardware coherence versus semantic staleness.}
CPU cache coherence is ensured by hardware protocols such as MESI: when data is modified, the corresponding cache lines are marked invalid.
KV and semantic cache ``invalidation'' is far more subtle: tool version changes, code repository commits, and stale indices can all produce hits that return outdated answers, a phenomenon we might call \textit{semantic staleness}.
Recent work has begun exploring learned eviction policies~\cite{lpc2025} and dependency-aware invalidation mechanisms~\cite{multilevelcache2025} to address this challenge.

\paragraph{A steeper capacity--latency gradient.}
In the CPU cache hierarchy, latency increases gradually from L1 to L2 to L3 to main memory on a nanosecond scale.
In the KV cache hierarchy, the latency jump from GPU memory to CPU memory to disk or network spans milliseconds, with no smooth intermediate gradient.
Future cache management will need to co-design traditional cache policies with provenance tracking, TTLs, and hash-based versioning.

Figure~\ref{fig_kv_hierarchy} illustrates the structural correspondence between the classical CPU cache hierarchy and the emerging LLM KV cache hierarchy.
\input{figures/fig_kv_hierarchy.tex}

\subsection{Context Management: Virtual Memory with Semantics}
\label{sec:memory}

\subsubsection{The Counterpart in LLM-Based Systems}

\textit{Classical counterpart---virtual memory:} paging gives each process an apparently unbounded address space backed by limited physical memory through \textbf{demand paging}, supported by page tables, TLBs, and replacement policies~\cite{ostep_2023}.

The context window is to a foundation model what main memory is to a process: a high-bandwidth, low-latency, but capacity-limited \textbf{working set}.
Anthropic defines the context window directly as the working memory available to the model during generation.

Context window capacity is expanding rapidly: Gemini~1.5 Pro supports up to 10 million multimodal tokens~\cite{gemini2024}; Ring Attention distributes sequences across multiple devices to achieve nearly unlimited context length~\cite{liu2024ringattention}; positional encoding methods such as RoPE, YaRN, and LongRoPE aim to extend the addressable history range~\cite{su2021roformer,yarn2024,longrope2024}.

Yet a growing body of evidence shows that ``declared support length'' and ``effective working-set capacity'' are not the same thing.
RULER, LongBench~v2, and LOFT provide more realistic stress tests for long-context capability~\cite{ruler2024,longbenchv2_2024,loft2024}.
A machine may advertise 256\,TB of virtual address space while having only 16\,GB of physical memory; true performance depends on whether the working set fits in physical memory at all times.

The more principled design, therefore, should not aim to ``stuff everything into context at once,'' but rather adopt \textbf{tiered management} in the spirit of virtual memory.

\paragraph{An emerging tiered scheme:}
\begin{itemize}[nosep]
\item \textbf{Hot memory (within the context window):} Complete information for the current task, analogous to active pages in physical memory.
\item \textbf{Warm memory (summaries and indices):} Compressed historical context, analogous to pages swapped out but retaining summary metadata.
\item \textbf{Cold memory (external knowledge bases):} Persistent vector stores and document repositories, analogous to data on disk.
\end{itemize}

Figure~\ref{fig:context_tiers} illustrates this three-tier hierarchy alongside its classical virtual-memory counterpart.
\input{figures/fig_context_tiers}

MemGPT formalizes this idea as \textit{virtual context management}, explicitly drawing the LLM-as-operating-system analogy and paging information between the context window (``main memory'') and external storage (``disk'')~\cite{memgpt2023}.
LongMem decouples long-term memory from the backbone model entirely~\cite{longmem2023}.
RAG functions as an \textbf{external page-in mechanism}: when the agent encounters a ``page fault,'' it retrieves relevant content from a knowledge base into the context window via semantic search~\cite{lewis2020rag}.
MemoryOS applies OS memory management principles directly to agent memory systems~\cite{memoryos2025}.
MemOS proposes a comprehensive memory operating system architecture~\cite{memos2025}.
HiAgent, inspired by human problem-solving processes, uses sub-goals as memory blocks to hierarchically manage agent working memory~\cite{hiagent2024}.
A-MEM enables agents to organize memory structures autonomously, without predefined schemas~\cite{amem2025}.

\subsubsection{Why the Analogy Holds}

The analogy between context management and virtual memory holds at the \textbf{mechanism level}, and the mechanisms rhyme across three dimensions.

\paragraph{The illusion of abundant space.}
Virtual memory convinces each process that it has a contiguous, large address space, when physical memory is far smaller; virtual context management convinces each agent that it possesses unlimited memory, when the actual context window spans only 128K--1M tokens.
Both realize this illusion through \textbf{demand loading}: only information needed right now occupies ``physical memory'' (the context window), while the rest resides on ``disk'' (external knowledge stores).

\paragraph{Address translation, exact and approximate.}
Virtual memory uses page tables to map virtual addresses to physical addresses; virtual context uses semantic retrieval to map information needs to specific knowledge entries.
The former is an exact match (page number to physical page number); the latter is an approximate match (query intent to top-$k$ relevant documents).

\paragraph{Paging under pressure.}
When physical memory runs low, the OS must decide which pages to swap out (LRU, Clock, and related policies); when the context window fills up, the agent must decide which information to evict (summarization, selective forgetting).
This is precisely the function of the ``context compiler'': the system must determine which state is loaded verbatim, which as a summary, which retains only an index entry, and which is discarded entirely~\cite{contextcompiler2025}.

\subsubsection{Essential Differences}

\paragraph{The key distinction: virtual memory is lossless; context management is lossy.}
OS paging is fully transparent to the application: pages swapped in and out retain every byte unchanged.
Agent context management, by contrast, involves \textbf{semantic compression}: summarization inevitably loses detail, and selection inevitably omits information.
``Address translation'' is also imprecise: vector similarity search returns the ``most relevant'' results, not the ``exact match.''

Semantic virtual memory can therefore never be a lossless paged system.
It is better understood as ``paged memory augmented with semantic summarization and evidence backfill'': information may be distorted after compression, but provenance chains enable verification against original data.
Research from JetBrains suggests that effective context management depends not on ``cramming in more tokens'' but on ``reducing noise and preserving signal''~\cite{jetbrains2025context}, further validating the necessity of lossy compression in this regime.
Figure~\ref{fig:comp53:context} compares the virtual-memory and context-management hierarchies side by side, highlighting the lossless-versus-lossy distinction as the fundamental divergence.

\input{figures/fig_comp_53_context}

\subsection{Agent Runtime: The Intelligent Operating System}
\label{sec:agent}

\subsubsection{The Counterpart in LLM-Based Systems}

\textit{Classical counterpart---the operating system:} system software providing \textbf{virtualization}, \textbf{concurrency}, and \textbf{persistence}---scheduling, isolation, memory management, permissions, I/O, and logging~\cite{ostep_2023}.

If the foundation model is the CPU, then the agent framework increasingly resembles the OS.
An agent runtime must:
\begin{itemize}[nosep]
\item \textbf{Parse objectives:} Decompose user requests into executable subtasks (analogous to program loading and parsing).
\item \textbf{Decide whether to plan:} Route simple requests directly and decompose complex ones into multi-step plans (analogous to fast-path versus slow-path scheduling).
\item \textbf{Load context:} Populate the context window with system prompts, project memory, and conversation history (analogous to loading a process memory image).
\item \textbf{Execute tool calls:} Invoke external tools via protocols such as MCP (analogous to system calls).
\item \textbf{Approve dangerous actions:} Perform permission checks on side-effecting operations such as file writes or code execution (analogous to capability-based access control).
\item \textbf{Spawn sub-agents concurrently:} Dispatch subtasks to independent child agents (analogous to \texttt{fork}).
\item \textbf{Merge intermediate results:} Aggregate sub-agent outputs into a final answer (analogous to inter-process communication and result aggregation).
\item \textbf{Persist state:} Write important state to memory files or knowledge bases (analogous to filesystem writes).
\item \textbf{Recover from failure:} Retry or roll back on errors to maintain task consistency (analogous to transaction recovery).
\end{itemize}

This analogy has already been articulated in the literature.
Ge et al.\ proposed the vision of ``LLM as OS, Agents as Apps''~\cite{ge2023llmos}.
Mei et al.\ built AIOS, a full LLM agent operating system kernel architecture~\cite{mei2024aios}.
Mi et al.\ systematically applied the evolutionary experience of computer architecture to agent system design~\cite{mi2025llmagentscomputersystems}.
Karpathy argued in 2023 that ``LLMs are not chatbots; they are the kernel process of a new operating system,''~\cite{karpathy2023lmos} where the filesystem becomes a vector database and the browser becomes internet search.
This vision is becoming an economic reality: an Augment Code enterprise customer completed a project in two weeks that their CTO had estimated would take four to eight months~\cite{anthropic2026agenticreport}, demonstrating that effective L3 context management combined with L5 orchestration can compress developer onboarding and project delivery from weeks to days.

Industrial practice is evolving in the same direction.
Codex provides subagents, skills, and sandbox-based control surfaces~\cite{openai_codex_overview_2026}; Claude Code offers hooks, MCP integration, memory files, and read-only approval flows~\cite{anthropic_claudecode_overview_2026}.

\subsubsection{Why the Analogy Holds}

Among all the analogies in this paper, the correspondence between the agent runtime and the OS is the deepest, because it operates at the level of \textbf{architectural responsibility}: both take ownership of the same three concerns.

\paragraph{Resource management.}
The OS manages CPU time slices, memory pages, disk space, and network bandwidth; the agent runtime manages model inference quota, context window space, tool-call rate limits, and sub-agent count.
Both must answer the same question: how should limited resources be allocated among competing demands?

\paragraph{Isolation and security.}
The OS isolates processes through address spaces so that one process crash does not affect others~\cite{ostep_2023}, and the agent runtime isolates sub-agents through sandboxes so that one sub-agent's erroneous operation does not compromise the main task~\cite{openai_codex_sandbox_2026,anthropic_claudecode_sandbox_2026}.
Both face the same access-control question: who is allowed to do what?

\paragraph{Standard interfaces.}
The OS standardizes application--hardware interaction through system calls (the POSIX interface); the agent runtime standardizes model--tool interaction through protocols such as MCP~\cite{mcp_intro_2026}.

\subsubsection{Essential Differences}

\paragraph{Unpredictable ``process'' behavior.}
A traditional OS schedules processes whose behavior is defined by deterministic code, so it can accurately predict execution time and resource requirements.
An agent's behavior is determined by probabilistic reasoning: the same task may follow entirely different execution paths across runs, succeeding once and falling into an infinite loop the next, which poses a fundamental challenge for scheduling and resource management.

\paragraph{Fragile shared state.}
When multiple processes share memory, consistency is ensured by hardware cache coherence protocols (e.g., MESI) and software locks, which are precise and verifiable.
When multiple agents share semantic state (e.g., ``an understanding of the same document''), consistency can only be maintained through natural-language communication or external synchronization mechanisms, which are inherently ambiguous and error-prone.

\paragraph{Imprecise interface specifications.}
Every system call in the POSIX interface has precisely defined semantics, parameter types, error codes, and boundary conditions.
An agent's tool-call interface can describe parameters via JSON Schema, but the ``semantic behavior'' of a tool (what it returns under which circumstances) is far from the level of precision expected of a system call.

\textbf{The greatest value of this analogy} lies in reframing many agent problems as OS problems:
\begin{itemize}[nosep]
\item Does multi-agent collaboration resemble multithreading or multiprocessing?~\cite{wu2023autogen}
\item Should tool invocations be treated as system calls?
\item Is an MCP server analogous to a device driver or bus protocol?~\cite{mcp_intro_2026}
\item Why do dangerous commands require capability-based permissions?~\cite{debenedetti2025camel}
\item How can long-running tasks recover via logs and interrupts?~\cite{openai_codex_approvals_2026}
\end{itemize}

\paragraph{Security: a particularly compelling analogy.}
In the security domain, this correspondence is especially profound.
CaMeL applies operating system capability-security principles to LLM agents, creating a protection layer independent of the LLM itself~\cite{debenedetti2025camel}.
IronClaw argues for protecting AI agents the way an operating system protects its processes~\cite{ironclaw2025}.
Systems security researchers have begun using the term ``agentic computing'' to define the foundational problems of agent security~\cite{agenticsecurity2025}.
These efforts underscore a critical insight: \textbf{security boundaries are primarily a ``runtime structure'' problem and only secondarily a ``prompt'' problem}, just as OS security depends first on the permission model and isolation mechanisms, and only then on application code quality.
Figure~\ref{fig:comp54:agent} maps the OS kernel's responsibilities onto the agent runtime, showing how classical primitives (fork, system call, sandbox, audit log) find direct counterparts in modern agent frameworks.

\input{figures/fig_comp_54_agent}

\subsection{Tool Bus and Agent Interconnection}
\label{sec:io}

\subsubsection{The Counterpart in LLM-Based Systems}

\textit{Classical counterpart---the I/O system:} it evolved from per-peripheral dedicated interfaces (1960s) through standardized buses (PCI/PCIe, USB; 1990s) to a unified network stack (TCP/IP); the central lesson is that \textbf{a standardized bus protocol is a prerequisite for ecosystem prosperity}.

In model-native computing, the I/O layer is undergoing a similar transition from point-to-point connections to a standardized bus.

\paragraph{Early phase (method-level):}
ReAct interleaved reasoning with acting~\cite{yao2023react}; Toolformer taught models when to call APIs~\cite{schick2023toolformer}; HuggingGPT placed an LLM in a ``controller'' role to orchestrate heterogeneous models~\cite{hugginggpt2023}; Gorilla focused on the accuracy and documentation-refresh resilience of API calls~\cite{patil2024gorilla}.
These efforts correspond to the ``direct device manipulation'' phase of early computing: each tool required a bespoke interface and calling convention, with no unified protocol standard.

\paragraph{Standardization phase (protocol-level):}
MCP (Model Context Protocol) standardizes the connection between AI applications and external data sources, tools, and workflows~\cite{mcp_intro_2026}, playing a role analogous to PCIe or USB in traditional computing: it defines unified discovery, connection, invocation, and error-handling mechanisms so that new tools can be ``plug-and-play.''
Google's A2A (Agent-to-Agent) protocol targets agent-to-agent interoperation~\cite{google_a2a_2025}, supporting capability discovery, task lifecycle management, and multimodal collaboration. It is analogous to TCP/IP in the networking stack, enabling agents from different vendors to discover and cooperate with one another.

Taken together, MCP serves as a \textbf{vertical bus} (agent$\to$tool) while A2A serves as a \textbf{horizontal interconnect} (agent$\to$agent).
As one survey observes, MCP, A2A, and related protocols such as ACP are forming the infrastructure layer for agent interoperability~\cite{agentprotocols2025}.

\subsubsection{Why the Analogy Holds}

\paragraph{Taming interface diversity.}
Before PCIe unified motherboard-level device interfaces, every device needed a dedicated driver; before MCP unified tool interfaces, every tool required bespoke API adapter code.
Both reduce system integration costs by defining a standard protocol that hides implementation differences.

\paragraph{A shared layered stack.}
Traditional I/O spans the physical layer (electrical signaling), the link layer (PCIe transaction layer), the protocol layer (NVMe/SCSI), and the application layer (filesystem).
Agent I/O follows a similar stack: transport layer (HTTP/SSE), protocol layer (MCP/A2A message formats), semantic layer (tool schemas and agent capability descriptions), and application layer (concrete tool invocations or agent collaborations).

\paragraph{Discovery and negotiation.}
A USB device, upon insertion, must be ``enumerated'': the host queries the device type, capabilities, and driver requirements.
An MCP server, upon connection, similarly undergoes ``capability discovery'': the agent queries which tools the server supports, along with each tool's parameter schema and side-effect description.

\subsubsection{Essential Differences}

\paragraph{Non-deterministic, asynchronous, side-effecting invocation.}
A traditional system call such as \texttt{read(fd, buffer, size)} is fully predictable: identical parameters yield identical results.
Tool invocations, by contrast, may be non-deterministic (search engine results change over time), asynchronous (long-running tasks require callbacks), and side-effecting (sending email, executing code).

\paragraph{Far more complex error handling.}
PCIe devices either respond correctly or return a well-defined error code; MCP tools may return semantically ambiguous errors (``the result is uncertain'') or---worse---results that appear correct but are in fact wrong.
This compels the agent I/O layer to introduce \textbf{verification and redundancy} mechanisms: critical operations require result validation, analogous to quorum reads in distributed systems.
Figure~\ref{fig:comp55:toolbus} traces the parallel standardization trajectory from bespoke drivers to PCIe/USB and TCP/IP on the classical side, and from hand-coded adapters to MCP and A2A on the agent side.

\input{figures/fig_comp_55_toolbus}

\subsection{Multi-Agent Collaboration: Distributed Systems with Semantics}
\label{sec:distributed}

\subsubsection{The Counterpart in LLM-Based Systems}

\textit{Classical counterpart---distributed systems:} cooperation across networked machines confronting task decomposition, state consistency, fault recovery, deadlock avoidance, and load balancing~\cite{hennessy2017quantitative}.

When multiple agents collaborate on a complex task, the system exhibits characteristics of distributed computing.

AutoGen treats multi-agent conversation as general infrastructure~\cite{wu2023autogen}, where multiple agents cooperate through message passing, much like distributed nodes.
MetaGPT uses pipeline-style Standard Operating Procedures to make multi-agent collaboration explicitly procedural~\cite{metagpt2023}, analogous to industrial-pipeline task decomposition, where each agent handles one stage (requirements analysis $\to$ design $\to$ coding $\to$ testing).
Internet of Agents imagines the agent ecosystem as an ``internet'' rather than a single machine~\cite{ioa2024}, emphasizing agent integration protocols, dynamic team formation, and distributed environments.

From an architectural perspective, clear correspondences emerge:
a single agent maps to a single process, multi-agent collaboration to multithreading or multiprocessing, and a cross-node agent cluster to a distributed system.
This immediately introduces the classical distributed-system challenges: task decomposition and scheduling, state synchronization and consistency, fault detection and recovery, deadlock avoidance, and load balancing.

\subsubsection{Why the Analogy Holds}

\paragraph{Communication overhead versus parallelism benefit.}
In distributed systems, Amdahl's law tells us that parallel speedup is limited by the serial fraction.
In multi-agent systems, Heuristic~III (the agent speedup heuristic) describes analogous behavior: when orchestration efficiency $E$ is insufficient, simply adding more agents yields diminishing returns.
AutoGen's multi-agent experiments are consistent with this picture: increasing the agent count from 2 to 8 produced progressively smaller reductions in task completion time~\cite{wu2023autogen}.
Industrial case studies qualitatively illustrate that coordination topology is itself a design variable: Fountain adopted a hub-and-spoke topology (a central Copilot coordinating sub-agents for screening, document generation, and sentiment analysis)~\cite{anthropic2026agenticreport}; MetaGPT employs a pipeline topology (SOP-driven sequential collaboration); AutoGen uses a peer-to-peer topology (conversational coordination). We treat these as qualitative illustrations of topology diversity rather than quantitative calibrations of $E$---the Fountain figures in particular are vendor-reported and non-identifiable, as discussed in Section~\ref{sec:law3}.
Different topologies yield different $E$ values across task scenarios, implying that the $E$ in Heuristic~III is not a fixed scalar but a design variable coupled to the coordination structure.

\paragraph{The need for explicit coordination mechanisms.}
Distributed systems use consensus protocols such as Raft~\cite{raft2014} to help multiple nodes agree on decisions; multi-agent systems need ``debate,'' ``voting,'' or ``hierarchical arbitration'' mechanisms for multiple agents to converge on a conclusion.

\paragraph{A consistency--availability tension.}
The CAP theorem states that a distributed system cannot simultaneously guarantee consistency, availability, and partition tolerance~\cite{cap2002}.
In multi-agent systems, an analogous tension exists between ``semantic consistency'' (whether agents share the same understanding of a fact) and ``response speed'' (whether to wait for all agents to finish reasoning before producing a conclusion).

\subsubsection{Essential Differences}

\paragraph{Lossy, low-precision communication.}
Distributed nodes communicate through structured protocols (e.g., gRPC with Protocol Buffers), yielding precise, lossless message semantics.
Agents communicate through natural language or semi-structured messages, where information may be misinterpreted, omitted, or distorted during transmission; \textit{semantic noise} is a challenge unique to agent-based distributed systems.

\paragraph{Gradual rather than binary failure.}
Distributed-node failures are typically binary (online or offline); agent ``failures'' can be \textit{gradual}: the agent has not crashed, but its reasoning quality degrades, it drifts from the task objective, or it begins to hallucinate.
Detecting ``whether an agent is executing the task correctly'' is far harder than detecting ``whether a node is alive.''

\paragraph{Emergence, not deterministic composition.}
The behavior of a distributed system is the deterministic composition of its nodes' behaviors; multi-agent systems may produce \textbf{emergent effects}, where each agent executes correctly in isolation, yet the collective outcome is unpredictable.
This places new demands on system debuggability and accountability.
Figure~\ref{fig:comp56:multiagent} summarizes the structural correspondence between distributed computing and multi-agent collaboration across coordination topologies, consensus mechanisms, and the consistency--availability tradeoff.

\input{figures/fig_comp_56_multiagent}


Figure~\ref{fig:arch} presents the proposed layered architecture for model-native computing.
Each layer corresponds to one of the component levels analyzed in this section: the model core (CPU) $\to$ KV cache (caches) $\to$ memory tier (virtual memory) $\to$ agent/OS layer (operating system) $\to$ I/O and persistence tier (peripherals) $\to$ distributed substrate (cluster).
Note the bypass arrows from the agent/OS layer directly to the memory tier and I/O tier, reflecting the dual-plane architecture's assignment of resource-management responsibility to the deterministic control plane.

\input{figures/fig_arch.tex}

Figure~\ref{fig:flow} illustrates the control flow of a typical request.
The flow embodies the interaction pattern of the dual-plane architecture: the probabilistic plane is responsible for generation and understanding, while the deterministic plane handles scheduling, approval, and state management.

\input{figures/fig_flow.tex}

%% file: figures/fig_comp_51_model.tex
\begin{figure}[htbp]
  \centering
  \includegraphics[width=0.88\textwidth]{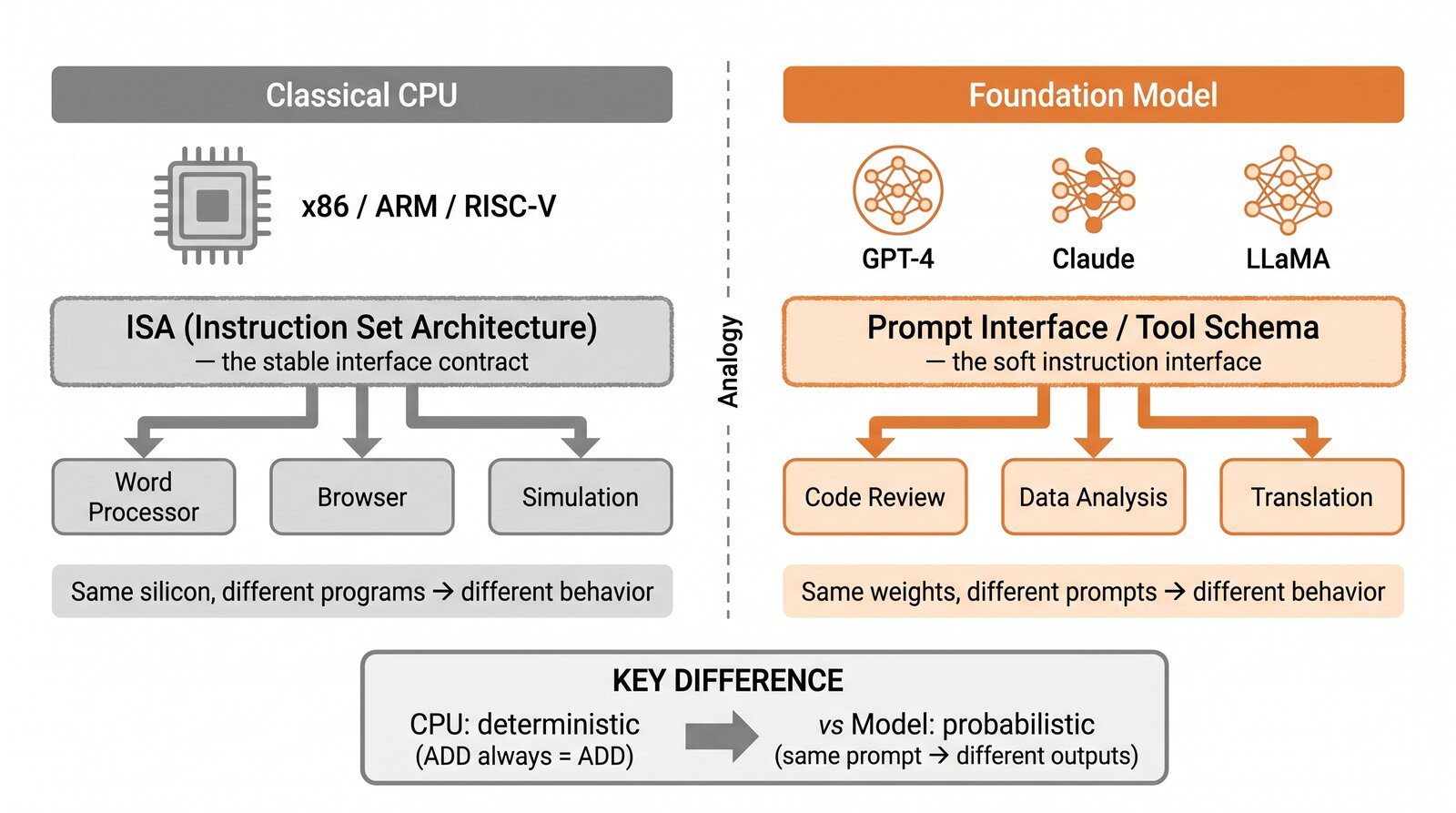}
  \caption{Foundation model as probabilistic computing core (§\ref{sec:model}).
  Both the CPU and the foundation model expose a stable interface above
  (ISA / prompt-and-tool-schema) while hiding a changing implementation below,
  enabling diverse behavior from the same substrate.
  The critical difference is determinism: a CPU instruction always produces the same
  result, whereas the same prompt may yield different outputs across invocations.}
  \label{fig:comp51:model}
\end{figure}

%% file: figures/fig_kv_hierarchy.tex
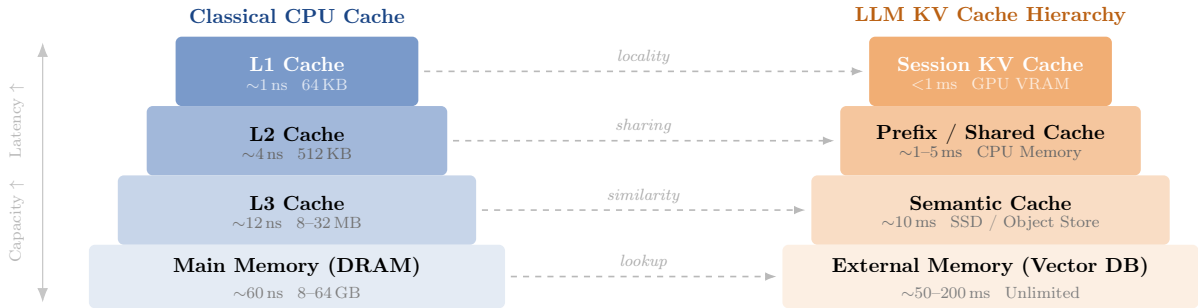
\begin{figure}[htbp]
\centering
\resizebox{\textwidth}{!}{%
\begin{tikzpicture}[
    every node/.style={font=\small},
]


\fill[classical!60, rounded corners=3pt]
    (-2.1, 3.5) rectangle (2.1, 4.7);
\node[text=white, font=\small\bfseries] at (0, 4.22) {L1 Cache};
\node[font=\scriptsize, text=classical!15!white] at (0, 3.88) {$\sim$1\,ns \enspace 64\,KB};

\fill[classical!42, rounded corners=3pt]
    (-2.6, 2.3) rectangle (2.6, 3.5);
\node[font=\small\bfseries] at (0, 3.02) {L2 Cache};
\node[font=\scriptsize, text=black!60] at (0, 2.68) {$\sim$4\,ns \enspace 512\,KB};

\fill[classical!25, rounded corners=3pt]
    (-3.1, 1.1) rectangle (3.1, 2.3);
\node[font=\small\bfseries] at (0, 1.82) {L3 Cache};
\node[font=\scriptsize, text=black!55] at (0, 1.48) {$\sim$12\,ns \enspace 8--32\,MB};

\fill[classical!12, rounded corners=3pt]
    (-3.6, 0.0) rectangle (3.6, 1.1);
\node[font=\small\bfseries] at (0, 0.72) {Main Memory (DRAM)};
\node[font=\scriptsize, text=black!50] at (0, 0.28) {$\sim$60\,ns \enspace 8--64\,GB};

\node[font=\small\bfseries, text=classical!80!black] at (0, 5.05)
    {Classical CPU Cache};

\fill[modelnative!60, rounded corners=3pt]
    (9.9, 3.5) rectangle (14.1, 4.7);
\node[text=white, font=\small\bfseries] at (12, 4.22) {Session KV Cache};
\node[font=\scriptsize, text=modelnative!15!white] at (12, 3.88) {$<$1\,ms \enspace GPU VRAM};

\fill[modelnative!42, rounded corners=3pt]
    (9.4, 2.3) rectangle (14.6, 3.5);
\node[font=\small\bfseries] at (12, 3.02) {Prefix / Shared Cache};
\node[font=\scriptsize, text=black!60] at (12, 2.68) {$\sim$1--5\,ms \enspace CPU Memory};

\fill[modelnative!25, rounded corners=3pt]
    (8.9, 1.1) rectangle (15.1, 2.3);
\node[font=\small\bfseries] at (12, 1.82) {Semantic Cache};
\node[font=\scriptsize, text=black!55] at (12, 1.48) {$\sim$10\,ms \enspace SSD / Object Store};

\fill[modelnative!12, rounded corners=3pt]
    (8.4, 0.0) rectangle (15.6, 1.1);
\node[font=\small\bfseries] at (12, 0.72) {External Memory (Vector DB)};
\node[font=\scriptsize, text=black!50] at (12, 0.28) {$\sim$50--200\,ms \enspace Unlimited};

\node[font=\small\bfseries, text=modelnative!80!black] at (12, 5.05)
    {LLM KV Cache Hierarchy};

\draw[dashed, gray!55, thick, -{Latex[length=2mm]}]
    (2.2, 4.1) -- (9.8, 4.1);
\node[font=\scriptsize, text=gray!65] at (6, 4.35) {\textit{locality}};

\draw[dashed, gray!55, thick, -{Latex[length=2mm]}]
    (2.7, 2.9) -- (9.3, 2.9);
\node[font=\scriptsize, text=gray!65] at (6, 3.15) {\textit{sharing}};

\draw[dashed, gray!55, thick, -{Latex[length=2mm]}]
    (3.2, 1.7) -- (8.8, 1.7);
\node[font=\scriptsize, text=gray!65] at (6, 1.95) {\textit{similarity}};

\draw[dashed, gray!55, thick, -{Latex[length=2mm]}]
    (3.7, 0.55) -- (8.3, 0.55);
\node[font=\scriptsize, text=gray!65] at (6, 0.80) {\textit{lookup}};

\draw[{Latex[length=2.5mm]}-{Latex[length=2.5mm]}, gray!45, line width=0.6pt]
    (-4.4, 0.1) -- (-4.4, 4.6);
\node[font=\scriptsize, text=gray!55, rotate=90] at (-4.85, 2.35)
    {Capacity $\uparrow$ \quad Latency $\uparrow$};

\end{tikzpicture}
}%
\caption{Structural correspondence between the classical CPU cache hierarchy and the emerging LLM KV cache hierarchy. Both exploit locality and reuse at multiple granularities, with capacity increasing and latency worsening at each lower tier.}
\label{fig_kv_hierarchy}
\end{figure}

%% file: figures/fig_context_tiers.tex
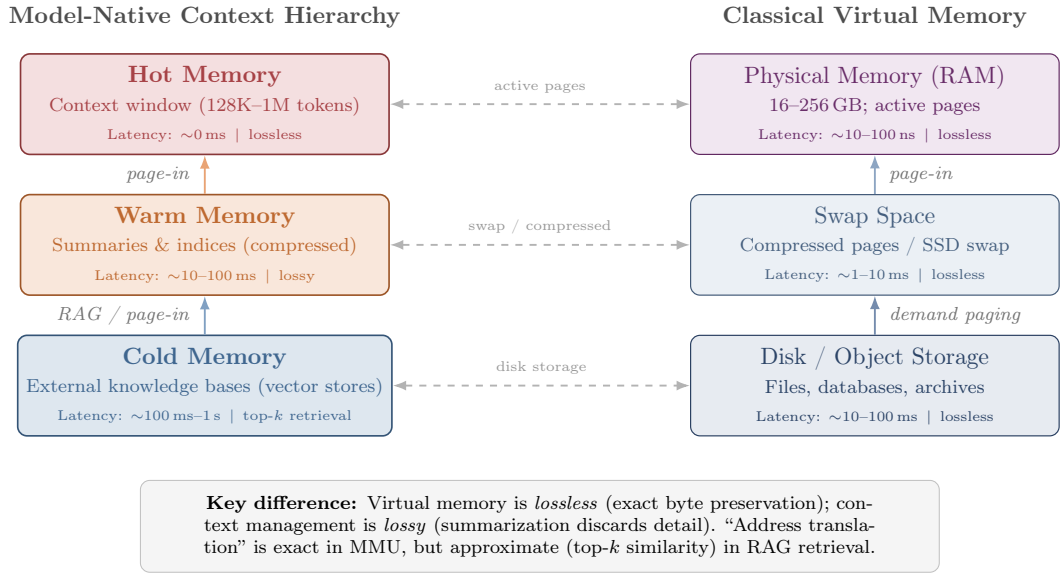
\begin{figure}[htbp]
\centering
\providecolor{hotColor}{HTML}{C44E52}
\providecolor{warmColor}{HTML}{E07B39}
\providecolor{coldColor}{HTML}{4A78A8}
\providecolor{classicHot}{HTML}{8B3A8B}
\providecolor{classicWarm}{HTML}{5C7EA8}
\providecolor{classicCold}{HTML}{3A5A8A}
\resizebox{0.88\textwidth}{!}{%
\begin{tikzpicture}[
    font=\small, >=Latex,
    tierbox/.style={draw=#1!70!black, fill=#1!18, rounded corners=4pt,
        align=center, font=\small\bfseries, text=#1!85!black, line width=0.7pt},
    classicbox/.style={draw=#1!70!black, fill=#1!12, rounded corners=4pt,
        align=center, font=\small, text=#1!75!black, line width=0.5pt},
    arrowlbl/.style={font=\scriptsize, text=black!55},
]

\def\xL{-5.0}

\node[tierbox=hotColor, minimum width=5.5cm, minimum height=1.5cm]
    (hot) at (\xL, 3.8)
    {Hot Memory\\
     {\normalfont\scriptsize Context window (128K--1M tokens)}\\
     {\normalfont\tiny Latency: $\sim$0\,ms\enspace|\enspace lossless}};

\node[tierbox=warmColor, minimum width=5.5cm, minimum height=1.5cm]
    (warm) at (\xL, 1.7)
    {Warm Memory\\
     {\normalfont\scriptsize Summaries \& indices (compressed)}\\
     {\normalfont\tiny Latency: $\sim$10--100\,ms\enspace|\enspace lossy}};

\node[tierbox=coldColor, minimum width=5.5cm, minimum height=1.5cm]
    (cold) at (\xL, -0.4)
    {Cold Memory\\
     {\normalfont\scriptsize External knowledge bases (vector stores)}\\
     {\normalfont\tiny Latency: $\sim$100\,ms--1\,s\enspace|\enspace top-$k$ retrieval}};

\node[font=\small\bfseries, text=black!70] at (\xL, 5.1) {Model-Native Context Hierarchy};

\draw[-{Latex[length=2mm]}, warmColor!70, thick]
    (warm.north) -- node[arrowlbl, left, xshift=-2pt] {\textit{page-in}} (hot.south);
\draw[-{Latex[length=2mm]}, coldColor!70, thick]
    (cold.north) -- node[arrowlbl, left, xshift=-2pt] {\textit{RAG / page-in}} (warm.south);

\def\xR{5.0}

\node[classicbox=classicHot, minimum width=5.5cm, minimum height=1.5cm]
    (l1) at (\xR, 3.8)
    {Physical Memory (RAM)\\
     {\scriptsize 16--256\,GB; active pages}\\
     {\tiny Latency: $\sim$10--100\,ns\enspace|\enspace lossless}};

\node[classicbox=classicWarm, minimum width=5.5cm, minimum height=1.5cm]
    (l2) at (\xR, 1.7)
    {Swap Space\\
     {\scriptsize Compressed pages / SSD swap}\\
     {\tiny Latency: $\sim$1--10\,ms\enspace|\enspace lossless}};

\node[classicbox=classicCold, minimum width=5.5cm, minimum height=1.5cm]
    (l3) at (\xR, -0.4)
    {Disk / Object Storage\\
     {\scriptsize Files, databases, archives}\\
     {\tiny Latency: $\sim$10--100\,ms\enspace|\enspace lossless}};

\node[font=\small\bfseries, text=black!70] at (\xR, 5.1) {Classical Virtual Memory};

\draw[-{Latex[length=2mm]}, classicWarm!70, thick]
    (l2.north) -- node[arrowlbl, right, xshift=2pt] {\textit{page-in}} (l1.south);
\draw[-{Latex[length=2mm]}, classicCold!70, thick]
    (l3.north) -- node[arrowlbl, right, xshift=2pt] {\textit{demand paging}} (l2.south);

\draw[{Latex[length=2mm]}-{Latex[length=2mm]}, thick, black!30, dashed]
    (hot.east)  -- node[font=\tiny, above, text=black!40, yshift=1pt]
    {active pages} (l1.west);
\draw[{Latex[length=2mm]}-{Latex[length=2mm]}, thick, black!30, dashed]
    (warm.east) -- node[font=\tiny, above, text=black!40, yshift=1pt]
    {swap / compressed} (l2.west);
\draw[{Latex[length=2mm]}-{Latex[length=2mm]}, thick, black!30, dashed]
    (cold.east) -- node[font=\tiny, above, text=black!40, yshift=1pt]
    {disk storage} (l3.west);

\node[draw=black!25, fill=black!4, rounded corners=4pt,
      font=\scriptsize, align=center, text width=11.5cm,
      inner sep=6pt] at (0, -2.5)
    {\textbf{Key difference:} Virtual memory is \textit{lossless} (exact byte preservation);
     context management is \textit{lossy} (summarization discards detail).
     ``Address translation'' is exact in MMU, but approximate (top-$k$ similarity) in RAG retrieval.};

\end{tikzpicture}}%
\caption{The hot/warm/cold tiered context management hierarchy (left) and its classical virtual-memory analogy (right). Both exploit temporal locality and demand-loading to give consumers the illusion of abundant capacity. The critical distinction is that context management is lossy: summaries and semantic retrieval introduce irreversible information loss absent in classical paging.}
\label{fig:context_tiers}
\end{figure}

%% file: figures/fig_comp_53_context.tex
\begin{figure}[htbp]
  \centering
  \includegraphics[width=0.88\textwidth]{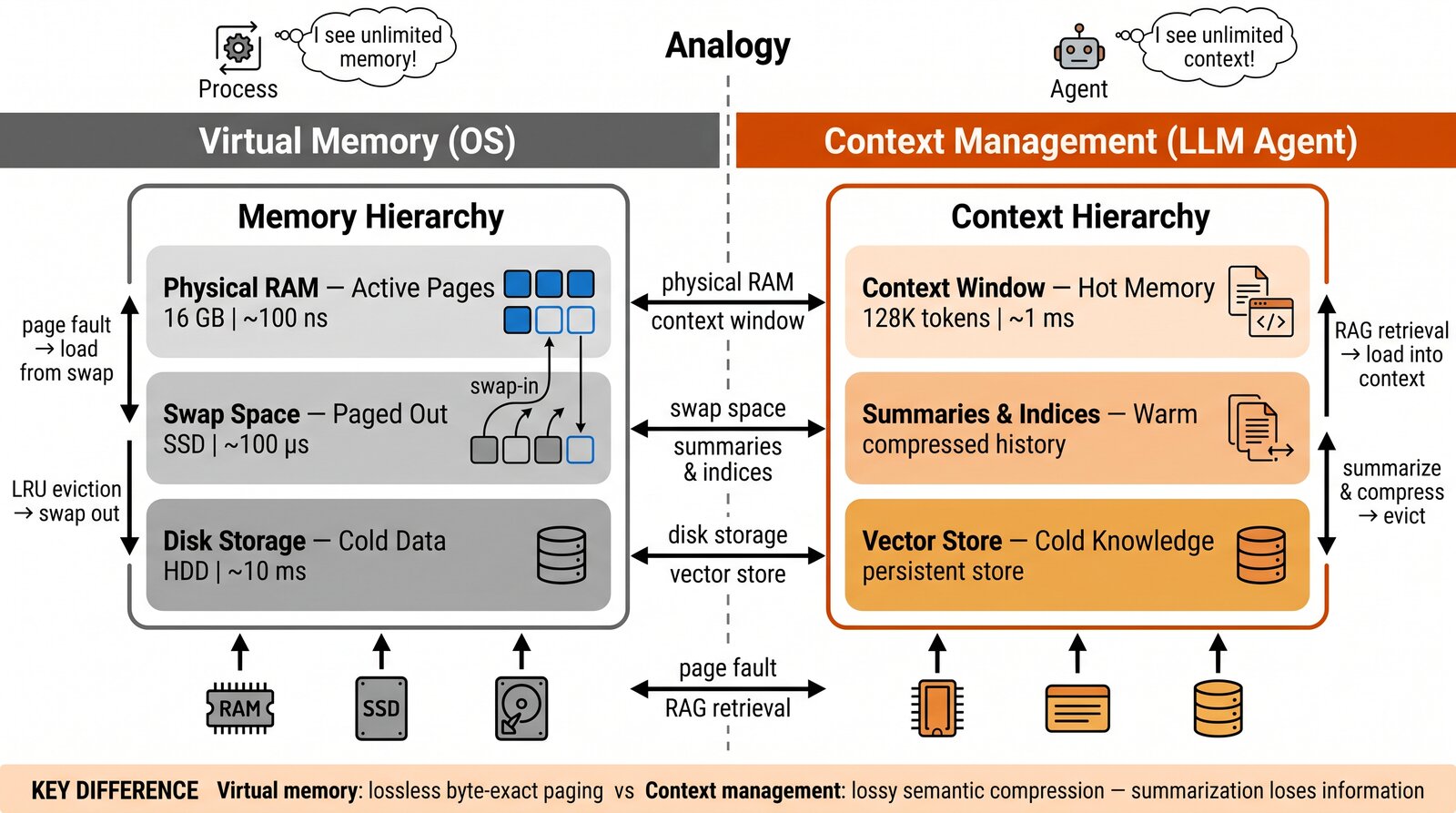}
  \caption{Context management as virtual memory with semantics (§\ref{sec:memory}).
  Both systems virtualize a limited physical resource (RAM / context window) into
  a larger logical space through tiered storage and demand loading.
  The critical difference is lossiness: OS paging swaps bytes exactly,
  whereas context management compresses via summarization---information is
  inevitably distorted, not merely delayed.}
  \label{fig:comp53:context}
\end{figure}

%% file: figures/fig_comp_54_agent.tex
\begin{figure}[htbp]
  \centering
  \includegraphics[width=0.88\textwidth]{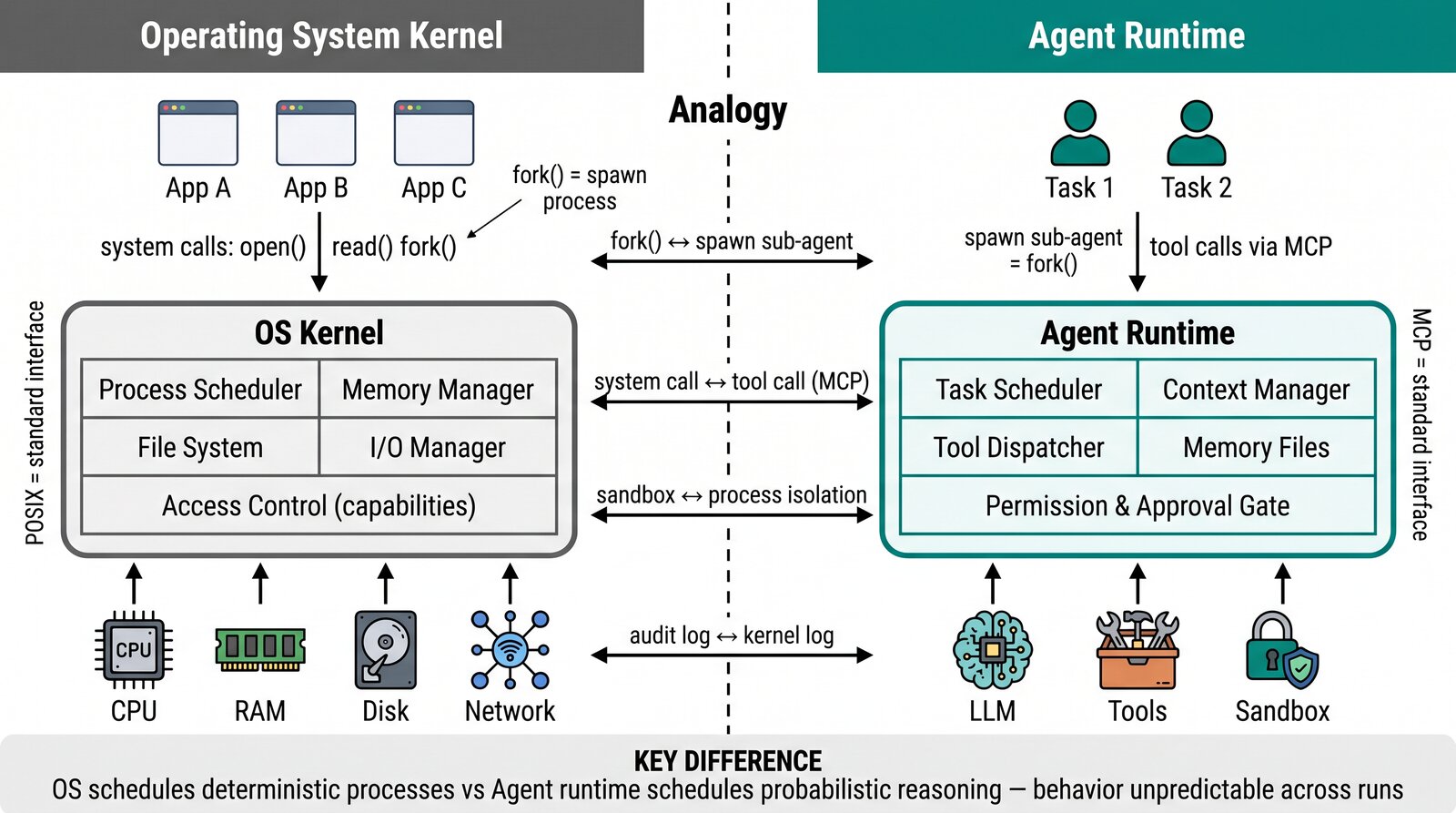}
  \caption{Agent runtime as intelligent operating system (§\ref{sec:agent}).
  Both take ownership of the same three architectural concerns: resource management,
  isolation, and standardized interfaces (system calls / MCP tool calls).
  The critical difference is predictability: an OS schedules deterministic processes
  whose execution time can be bounded, whereas an agent runtime schedules
  probabilistic reasoning whose behavior may differ across identical runs.}
  \label{fig:comp54:agent}
\end{figure}

%% file: figures/fig_comp_55_toolbus.tex
\begin{figure}[htbp]
  \centering
  \includegraphics[width=0.8\textwidth]{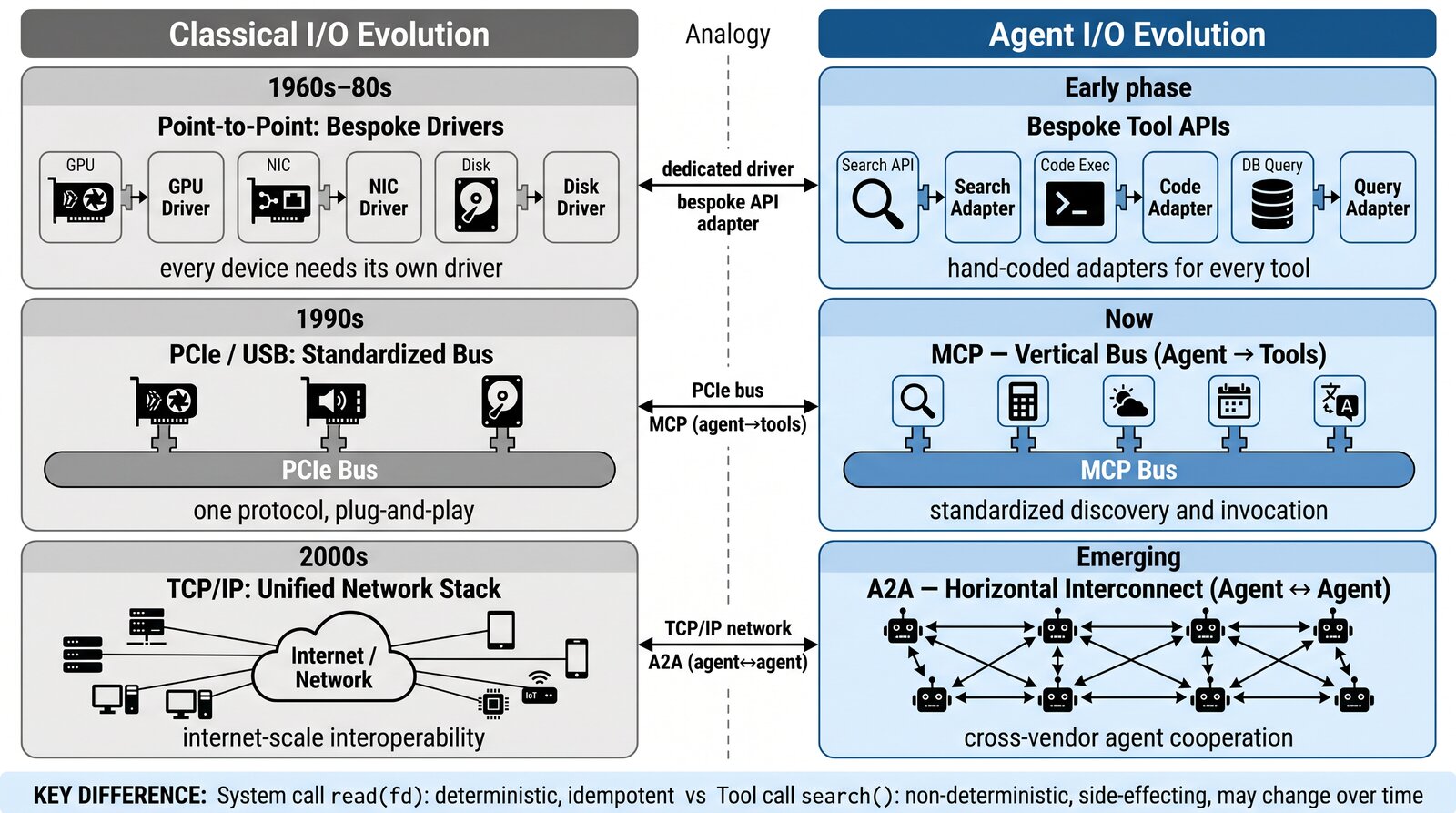}
  \caption{Tool bus and agent interconnection as I/O system (§\ref{sec:io}).
  Classical I/O evolved from per-device bespoke drivers through standardized buses
  (PCIe, USB) to a unified network stack (TCP/IP); agent I/O is undergoing
  the same transition, from hand-coded API adapters to MCP (vertical agent-to-tool bus)
  and A2A (horizontal agent-to-agent interconnect).
  The critical difference is side-effect semantics: \texttt{read(fd)} is idempotent
  and deterministic, whereas a tool call may be non-deterministic, asynchronous,
  and irreversible.}
  \label{fig:comp55:toolbus}
\end{figure}

%% file: figures/fig_comp_56_multiagent.tex
\begin{figure}[htbp]
  \centering
  \includegraphics[width=0.9\textwidth]{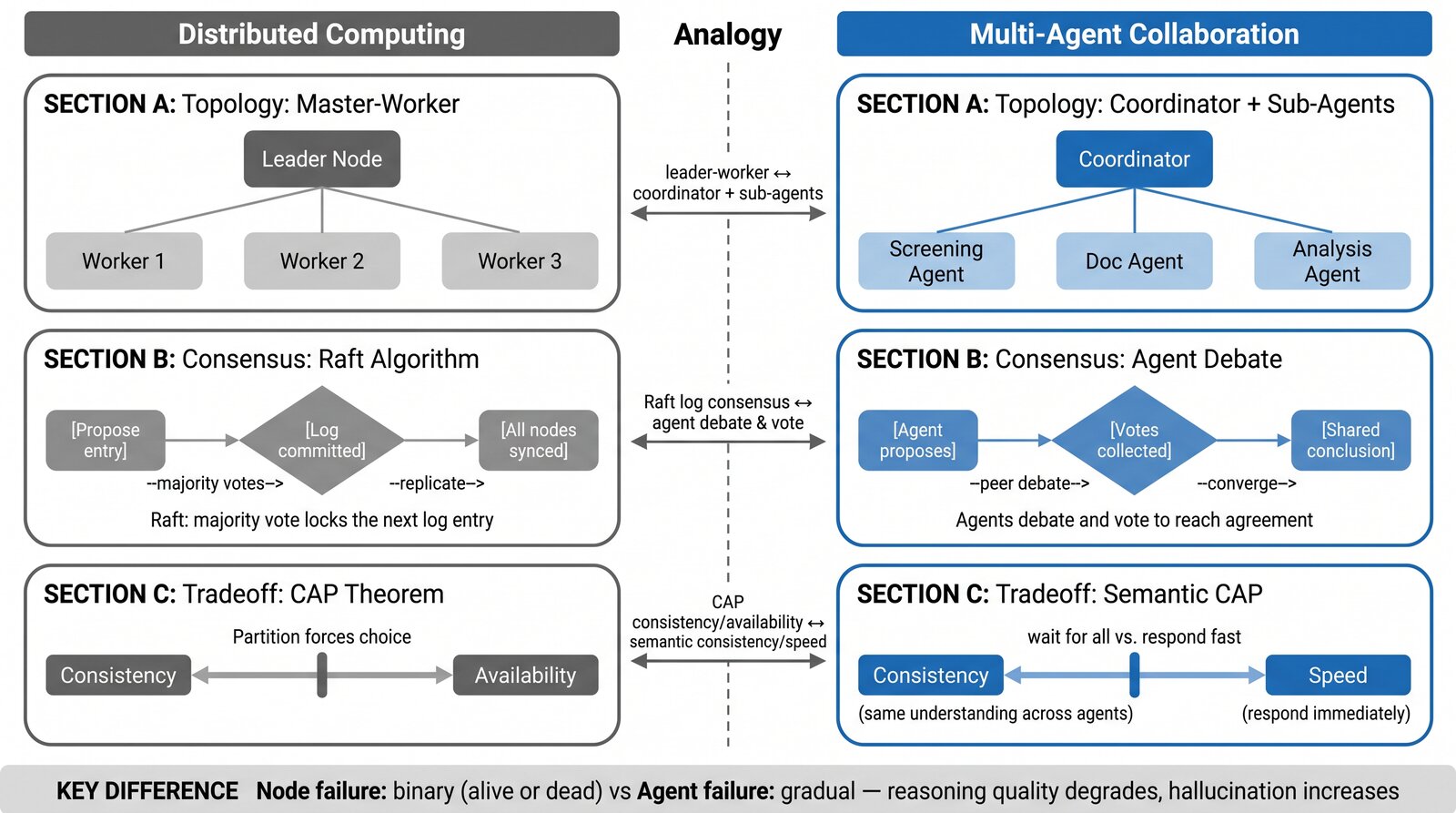}
  \caption{Multi-agent collaboration as distributed computing with semantics (§\ref{sec:distributed}).
  Both domains face the same coordination topologies (hub-and-spoke, pipeline, peer-to-peer)
  and the same fundamental tension between consistency and availability (CAP theorem /
  semantic CAP: wait for all agents vs.\ respond fast).
  The critical difference is failure mode: distributed-node failure is binary (alive or dead),
  whereas agent failure is gradual---reasoning quality degrades continuously through
  hallucination, task drift, and goal misalignment.}
  \label{fig:comp56:multiagent}
\end{figure}

%% file: figures/fig_arch.tex
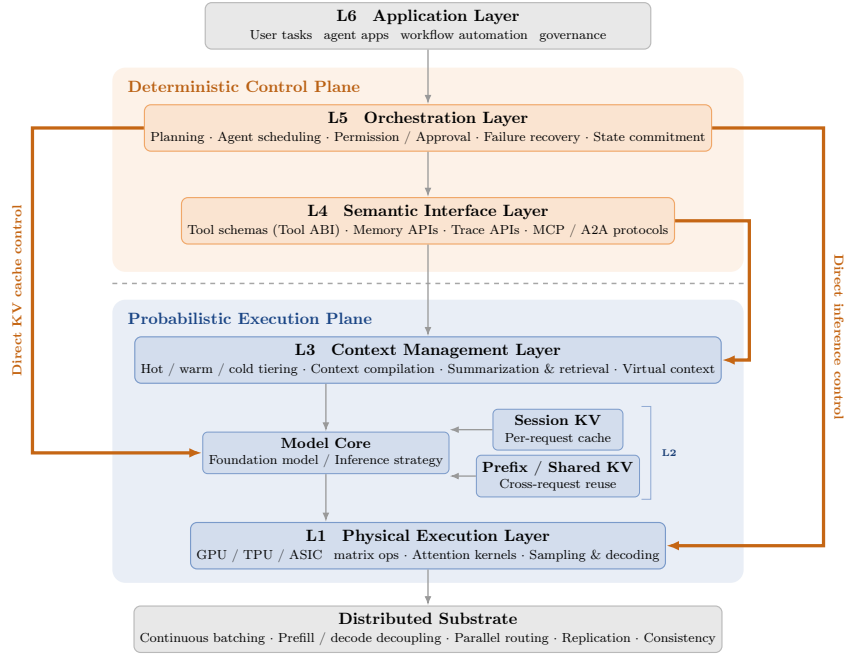
\begin{figure}[h]
    \centering
    \resizebox{0.7\textwidth}{!}{%
    \begin{tikzpicture}[
        node distance=4mm and 5mm,
        layer/.style={
            draw, rounded corners=4pt, minimum width=9.6cm,
            minimum height=1cm, font=\small, align=center,
            line width=0.7pt
        },
        subbox/.style={
            draw, rounded corners=3pt, minimum height=0.85cm,
            font=\footnotesize, align=center, line width=0.6pt
        },
        arrow/.style={-{Latex[length=2mm]}, thick, black!40},
        bypass/.style={
            -{Latex[length=2.8mm]}, line width=2pt,
            modelnative!85!black
        },
    ]
    
    \fill[modelnative!10, rounded corners=8pt]
        (-6.8, 1.4) rectangle (6.8, 5.8);
    \node[font=\small\bfseries, text=modelnative!80!black, anchor=west]
        at (-6.6, 5.4)
        {Deterministic Control Plane};
    
    %
    \fill[classical!10, rounded corners=8pt]
        (-6.8, -5.3) rectangle (6.8, 0.8);
    \node[font=\small\bfseries, text=classical!80!black, anchor=west]
        at (-6.6, 0.4)
        {Probabilistic Execution Plane};
    
    \node[layer, fill=gray!18, draw=gray!55]
        (L6) at (0,6.7)
        {\textbf{L6 \, Application Layer}\\[-1pt]
         {\scriptsize User tasks \, agent apps \, workflow automation \, governance}};
    
    \node[layer, fill=modelnative!20, draw=modelnative!65]
        (L5) at (0,4.5)
        {\textbf{L5 \, Orchestration Layer}\\[-1pt]
         {\scriptsize Planning $\cdot$ Agent scheduling $\cdot$ Permission / Approval $\cdot$
          Failure recovery $\cdot$ State commitment}};
    
    \node[layer, fill=modelnative!20, draw=modelnative!65]
        (L4) at (0,2.5)
        {\textbf{L4 \, Semantic Interface Layer}\\[-1pt]
         {\scriptsize Tool schemas (Tool ABI) $\cdot$ Memory APIs $\cdot$ Trace APIs $\cdot$
          MCP / A2A protocols}};
    
    %
    \draw[dashed, thick, black!35, line width=0.9pt]
        (-6.8,1.15) -- (6.8,1.15);
    
    \node[layer, fill=classical!20, draw=classical!60]
        (L3) at (0,-0.5)
        {\textbf{L3 \, Context Management Layer}\\[-1pt]
         {\scriptsize Hot / warm / cold tiering $\cdot$ Context compilation $\cdot$
          Summarization \& retrieval $\cdot$ Virtual context}};
    
    \node[subbox, fill=classical!20, draw=classical!60,
          minimum width=4.6cm] 
        (L2core) at (-2.2,-2.5)
        {\textbf{Model Core}\\[-1pt]
         {\scriptsize Foundation model / Inference strategy}};
    
    \node[subbox, fill=classical!20, draw=classical!60,
          minimum width=2.8cm]
        (L2sess) at (2.8,-2.0)
        {\textbf{Session KV}\\[-1pt]
         {\scriptsize Per-request cache}};
    
    \node[subbox, fill=classical!20, draw=classical!60,
          minimum width=2.8cm]
        (L2pref) at (2.8,-3.0)
        {\textbf{Prefix / Shared KV}\\[-1pt]
         {\scriptsize Cross-request reuse}};
    
    \draw[classical!50, line width=0.5pt]
        (4.6,-1.5) -- (4.8,-1.5) -- (4.8,-3.5) -- (4.6,-3.5);
    \node[font=\tiny\bfseries, text=classical!70!black, anchor=west]
        at (4.9,-2.5) {L2};
    
    \node[layer, fill=classical!20, draw=classical!60]
        (L1) at (0,-4.5)
        {\textbf{L1 \, Physical Execution Layer}\\[-1pt]
         {\scriptsize GPU / TPU / ASIC \, matrix ops $\cdot$ Attention kernels $\cdot$
          Sampling \& decoding}};
    
    \node[layer, fill=gray!18, draw=gray!55]
        (L0) at (0,-6.3)
        {\textbf{Distributed Substrate}\\[-1pt]
         {\scriptsize Continuous batching $\cdot$ Prefill / decode decoupling $\cdot$
          Parallel routing $\cdot$ Replication $\cdot$ Consistency}};
    
    \draw[arrow] (L6)   -- (L5);
    \draw[arrow] (L5)   -- (L4);
    \draw[arrow] (L4)   -- (L3);
    
    \draw[arrow] (L3.south -| L2core.north) -- (L2core.north);
    \draw[arrow] (L2core.south)   -- (L1.north -| L2core.south);
    \draw[arrow] (L1)   -- (L0);
    
    \draw[arrow] (L2sess.west)  -- (L2core.east |- L2sess.west);
    \draw[arrow] (L2pref.west)  -- (L2core.east |- L2pref.west);
    
    
    \draw[bypass]
        (L5.west) -- ++(-2.4,0) coordinate (bl1) -- (bl1 |- L2core.west) -- (L2core.west);
    \path (L5.west) -- ++(-2.4,0) coordinate (bl1) -- (bl1 |- L2core.west)
        node[midway, font=\scriptsize\bfseries, text=modelnative!85!black,
             rotate=90, anchor=south, yshift=1mm]
        {Direct KV cache control};
    
    \draw[bypass]
        (L4.east) -- ++(1.6,0) coordinate (br1) -- (br1 |- L3.east) -- (L3.east);
    
    \draw[bypass]
        (L5.east) -- ++(2.4,0) coordinate (br3) -- (br3 |- L1.east) -- (L1.east);
    \path (L5.east) -- ++(2.4,0) coordinate (br3) -- (br3 |- L1.east)
        node[midway, font=\scriptsize\bfseries, text=modelnative!85!black,
             rotate=-90, anchor=south, yshift=1mm]
        {Direct inference control};
    
    \end{tikzpicture}}
    \caption{ICA layered architecture for model-native computing.
      \textbf{Top to bottom:} L6 Application, L5 Orchestration, L4 Semantic Interface
      (Deterministic Control Plane, orange), dashed L3--L4 interface boundary,
      L3 Context Management, L2 Inference Serving (Model Core + Session KV +
      Prefix/Shared KV), L1 Physical Execution (Probabilistic Execution Plane, blue),
      and Distributed Substrate.
      Thick orange bypass arrows show the Deterministic Control Plane reaching
      through the interface boundary to directly manage KV caches (L2) and
      hardware execution (L1).}
    \label{fig:arch}
    \end{figure}

%% file: figures/fig_flow.tex
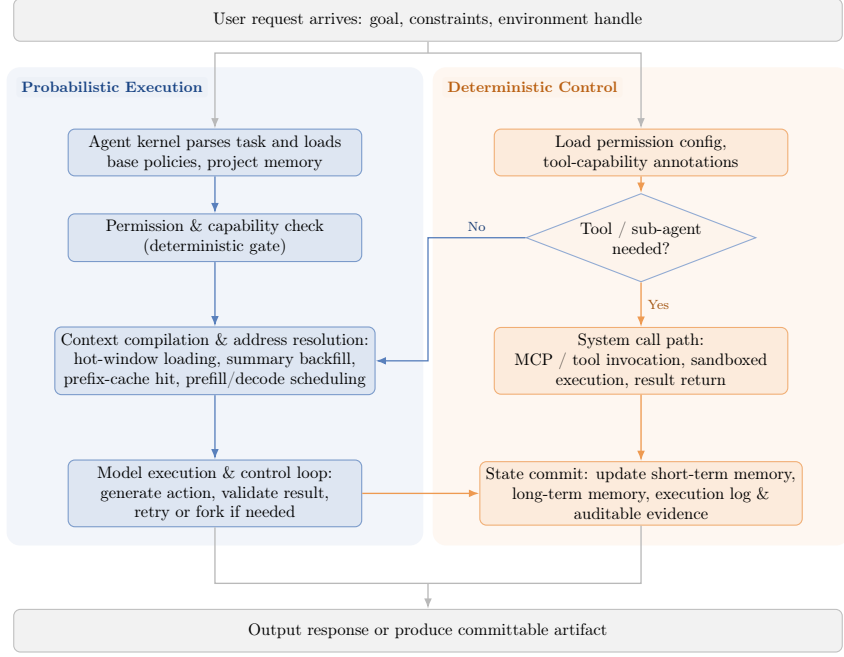
\begin{figure}[htbp]
    \centering
    \resizebox{0.7\textwidth}{!}{%
    \begin{tikzpicture}[
        procB/.style={draw=classical!60, rounded corners=4pt,
            minimum width=6.2cm, minimum height=9mm, align=center,
            fill=classical!15, font=\small, line width=0.65pt},
        procO/.style={draw=modelnative!60, rounded corners=4pt,
            minimum width=6.2cm, minimum height=9mm, align=center,
            fill=modelnative!15, font=\small, line width=0.65pt},
        procN/.style={draw=gray!50, rounded corners=4pt,
            minimum height=9mm, align=center,
            fill=gray!10, font=\small, line width=0.65pt},
        dec/.style={draw, diamond, aspect=2.6, align=center,
            inner sep=2pt, font=\small, line width=0.65pt},
        arrB/.style={-{Latex[length=2.2mm]}, thick, classical!75},
        arrO/.style={-{Latex[length=2.2mm]}, thick, modelnative!75},
        arrG/.style={-{Latex[length=2.2mm]}, thick, gray!55},
        hdr/.style={font=\footnotesize\bfseries, inner sep=3pt,
            rounded corners=2pt, fill opacity=0.55, text opacity=1},
        node distance=5mm and 5mm,
    ]
    
    %
    \pgfmathsetmacro{\cxL}{-4.5}   
    \pgfmathsetmacro{\cxR}{+4.5}   
    \pgfmathsetmacro{\halfW}{4.4}  
    
    \def\yTop{1.8}    %
    \def\yBot{-8.3}   %
    
    \def\yReq{2.8}    
    \def\ySplit{2.1}  %
    
    \def\yRowA{0.0}   
    \def\yRowB{-1.8}  
    \def\yRowC{-4.4}  
    \def\yRowD{-7.2}  
    
    \def\yMerge{-9.1} 
    \def\yResp{-10.1} 
    
    \fill[classical!6, rounded corners=8pt]
        (\cxL-\halfW, \yBot) rectangle (\cxL+\halfW, \yTop);
    
    \fill[modelnative!6, rounded corners=8pt]
        (\cxR-\halfW, \yBot) rectangle (\cxR+\halfW, \yTop);
    
    \node[hdr, text=classical!80!black, fill=classical!10, anchor=north west]
        at (\cxL-\halfW+0.2, \yTop-0.2) {Probabilistic Execution};
        
    \node[hdr, text=modelnative!80!black, fill=modelnative!10, anchor=north west]
        at (\cxR-\halfW+0.2, \yTop-0.2) {Deterministic Control};
    
    \node[procN, minimum width=17.5cm]
        (req) at (0, \yReq)
        {User request arrives: goal, constraints, environment handle};
    
    \node[procB] (plan) at (\cxL, \yRowA)
        {Agent kernel parses task and loads\\base policies, project memory};
    
    \node[procB] (perm) at (\cxL, \yRowB)
        {Permission \& capability check\\(deterministic gate)};
    
    \node[procB] (prefill) at (\cxL, \yRowC) 
        {Context compilation \& address resolution:\\
        hot-window loading, summary backfill,\\
        prefix-cache hit, prefill/decode scheduling};
    
    \node[procB] (gen) at (\cxL, \yRowD)
        {Model execution \& control loop:\\
        generate action, validate result,\\
        retry or fork if needed};
    
    \node[procO] (policy) at (\cxR, \yRowA)
        {Load permission config,\\
        tool-capability annotations};
    
    \node[dec, draw=classical!50, fill=modelnative!8] 
        (tool) at (\cxR, \yRowB) 
        {Tool / sub-agent\\needed?};
    
    \node[procO] (syscall) at (\cxR, \yRowC)
        {System call path:\\
        MCP / tool invocation, sandboxed\\
        execution, result return};
    
    \node[procO] (commit) at (\cxR, \yRowD)
        {State commit: update short-term memory,\\
        long-term memory, execution log \&\\
        auditable evidence};
    
    %
    \node[procN, minimum width=17.5cm]
        (resp) at (0, \yResp)
        {Output response or produce committable artifact};
    
    \coordinate (split) at (0, \ySplit);
    \draw[thick, gray!55] (req.south) -- (split);
    \draw[arrG] (split) -| (plan.north);
    \draw[arrG] (split) -| (policy.north);
    
    \draw[arrB] (plan) -- (perm);
    \draw[arrB] (perm) -- (prefill);
    \draw[arrB] (prefill) -- (gen);
    
    \draw[arrO] (policy) -- (tool);
    \draw[arrO] (tool.south) --
        node[right, font=\scriptsize, text=modelnative!70!black] {Yes}
        (syscall.north);
    \draw[arrO] (syscall) -- (commit);
    
    \draw[arrB] (tool.west) -- 
        node[above, font=\scriptsize, text=classical!70!black] {No} 
        (0, \yRowB) |- (prefill.east);
    
    \draw[arrO] (gen.east) -- (commit.west);
    
    \draw[thick, gray!55] (gen.south)  |- (0, \yMerge);
    \draw[thick, gray!55] (commit.south) |- (0, \yMerge);
    \draw[arrG] (0, \yMerge) -- (resp.north);
    
    \end{tikzpicture}}
    \caption{Request flow through the model-native computing architecture as a dual-plane swim-lane diagram.
    Blue nodes belong to the probabilistic execution plane; orange nodes belong to the deterministic control plane.
    The split-coloured diamond represents the control-plane decision that gates both paths.
    Grey boxes span both planes.}
    \label{fig:flow}
    \end{figure}

%% file: en_sections/section_icam.tex
\section{The Intelligent Computing Architecture (ICA)}
\label{sec:icam}

The analogy framework developed in the preceding four sections provides valuable intuition, but intuition alone cannot discipline engineering design. The enduring strength of classical computer architecture rests on its possession of \textbf{formal models}: from the hardware--software contract embodied in the instruction set architecture (ISA), to the locality theory governing memory hierarchies, to the quantitative analysis framework of Amdahl's Law. These models give architects a shared, precise vocabulary for reasoning about trade-offs.

This section \textbf{formalizes} the analogies above into the \textit{Intelligent Computing Architecture} (ICA), comprising a six-layer functional hierarchy, inter-layer interface contracts, and six design axioms. Together, they provide the conceptual foundation for the design heuristics developed in Section~\ref{sec:laws}.

\subsection{Six-Layer Hierarchy}
\label{sec:icam-layers}

The central organizing principle of classical computer architecture is \textbf{layered abstraction}: hardware is hidden behind the ISA, the operating system exposes services through system calls, and user applications need only invoke library functions. Each layer interacts solely with its immediate neighbors and remains oblivious to the implementation details of distant layers. This separation of concerns keeps system complexity tractable and enables independent evolution at every level.

ICA transfers this principle to the model-native computing domain by defining six functional layers.

\begin{definition}[Intelligent Computing Architecture --- ICA]
The Intelligent Computing Architecture (ICA) organizes a model-native computing system into six functional layers, each interacting with adjacent layers through well-defined interface contracts:
\begin{enumerate}[nosep]
\item \textbf{L1~--- Physical Execution Layer.}
GPU, TPU, ASIC, and compute-in-memory hardware that execute matrix multiplications and attention computations. This layer is the analog of the ALU and floating-point unit in classical architecture: it carries out compute instructions without regard for their semantic meaning. The Transformer Engine in the NVIDIA H100, which accelerates FP8 matrix multiplication at the hardware level, is a representative L1 component~\cite{wolters2024computeinmemory}.

\item \textbf{L2~--- Inference Serving Layer.}
Model weight loading, KV cache management, continuous batching, and prefill--decode scheduling. This layer parallels the microarchitecture and cache hierarchy in classical systems: it manages compute resources and hot data so that invocations from upper layers execute as efficiently as possible. vLLM's PagedAttention, which enables on-demand KV cache allocation and sharing~\cite{kwon2023pagedattention}, and SGLang's RadixAttention, which reuses cached prefixes across requests~\cite{sglang2024}, are both representative L2 implementations.

\item \textbf{L3~--- Context Management Layer.}
Hot/warm/cold memory tiering, context compilation, summarization and retrieval, and semantic virtual memory. This layer is the analog of the virtual memory subsystem in classical operating systems: it virtualizes the limited physical context window into a larger logical working space. MemGPT's \textit{virtual context management} treats the context window as main memory and external storage as disk, enabling the model to autonomously manage memory swap-in and swap-out~\cite{memgpt2023}; MemoryOS borrows OS memory-management mechanisms to design a tiered memory architecture~\cite{memoryos2025}. Both are representative L3 implementations.

\item \textbf{L4~--- Semantic Interface Layer.}
Tool schemas (the Tool ABI), Memory APIs, and Trace APIs. This layer parallels the system-call interface and ABI in classical architecture: it defines the uniform protocols through which an intelligent system accesses external capabilities. The Model Context Protocol (MCP) standardizes communication between models and tools~\cite{mcp_intro_2026}, while the Agent-to-Agent (A2A) protocol defines interoperability between agents~\cite{agentprotocols2025}. Both are representative L4 implementations.

\item \textbf{L5~--- Orchestration Layer.}
Task planning, agent scheduling, permission enforcement, failure recovery, and state commitment. This layer is the analog of the operating system kernel: it is responsible for scheduling tasks, allocating resources, enforcing isolation, and providing fault tolerance. OpenAI Codex submits agent tasks to sandboxes for execution and supports the creation and coordination of sub-agents~\cite{openai_codex_overview_2026}; Claude Code extends agent capabilities through a skill mechanism and connects to external tools via MCP~\cite{anthropic_claudecode_overview_2026,anthropic_claudecode_skills_2026}. Both are representative L5 implementations.

\item \textbf{L6~--- Application Layer.}
User tasks, agent applications, and workflow automation. This layer parallels the user-space application layer in classical systems: it carries end-user intent. SWE-agent autonomously resolves GitHub issues~\cite{sweagent2024}; OpenHands serves as a general-purpose software-development agent~\cite{openhands_paper_2025}. It is worth noting that the L6 user base is rapidly expanding from professional engineers to domain experts: Anthropic's legal team used Claude to reduce marketing-review turnaround from two to three days to under 24~hours; Zapier achieved 89\% organization-wide AI adoption and deployed over 800 internal agents~\cite{anthropic2026agenticreport}. This trend underscores the need for richer interaction modes and stronger built-in guardrails at L6.
\end{enumerate}
\end{definition}

The ICA layering is a deliberate, pragmatic mapping rather than a uniquely forced one: each of the six layers corresponds one-to-one to a distinct classical stratum (Table~\ref{tab:icam}) and is characterized by a distinct ``core difference'' from its classical analog, with each boundary independently checkable through the interface contract of Section~\ref{sec:icam-interfaces}. The organizing principle is the same as in network protocol stacks and operating-system hierarchies---each layer serves the layer above and depends only on the interface of the layer below---and the resulting benefit is \textbf{separation of concerns}: an L2 engineer optimizing KV cache policies need not understand agent orchestration logic, and an L5 developer designing task schedulers need not comprehend the hardware details of attention computation. We present this decomposition as an organizing proposal; the precise boundaries will almost certainly be refined as the field matures (Section~\ref{sec:conclusion}).

Table~\ref{tab:icam} maps each ICA layer to its classical counterpart, summarizes its responsibilities, highlights the core difference, and lists representative implementations.

\begin{table}[t]
\centering
\footnotesize
\caption{ICA layers mapped to classical computer architecture.}
\label{tab:icam}
\begin{tabularx}{\textwidth}{p{0.08\textwidth}p{0.12\textwidth}p{0.22\textwidth}p{0.24\textwidth}p{0.22\textwidth}}
\toprule
ICA Layer & Classical Analog & Responsibility & Core Difference & Representative Implementations \\
\midrule
L1 & Hardware / ISA & Execute basic compute operations & Probabilistic matrix operations, not deterministic logic gates & GPU, TPU, Compute-in-Memory~\cite{wolters2024computeinmemory} \\
L2 & Micro-arch.\ / Cache & Manage compute resources and hot data & KV cache grows with semantic sequence, not address-indexed & vLLM, SGLang~\cite{kwon2023pagedattention,sglang2024} \\
L3 & Virtual Memory & Expand the addressable working set & Semantic summarization is lossy, not lossless paging & MemGPT, MemoryOS~\cite{memgpt2023,memoryos2025} \\
L4 & Syscall / ABI & Uniform interface to external capabilities & Tool semantics carry natural-language ambiguity & MCP, A2A~\cite{mcp_intro_2026,agentprotocols2025} \\
L5 & Operating System & Scheduling, permissions, fault tolerance & Agents themselves contain non-deterministic reasoning & Codex, Claude Code~\cite{openai_codex_overview_2026,anthropic_claudecode_overview_2026} \\
L6 & User Application & Carry end-user tasks & Tasks may have open-ended goals and fuzzy constraints & SWE-agent, OpenHands~\cite{sweagent2024,openhands_paper_2025} \\
\bottomrule
\end{tabularx}
\end{table}

\subsection{Inter-Layer Interface Contracts}
\label{sec:icam-interfaces}

The sustained evolution of classical computer architecture hinges on the \textbf{stability of inter-layer interfaces}. The ISA defines the contract between hardware and software: Intel's processors evolved from the 8086 to today's Core Ultra with sweeping microarchitectural changes, yet the x86 ISA maintained backward compatibility throughout. POSIX defines the contract between applications and the operating system: the Linux kernel progressed from 2.0 to 6.x, yet user applications required virtually no modification. The principle of ``stable interfaces, mutable implementations'' allows each layer to be optimized independently.

ICA similarly defines an interface contract for each pair of adjacent layers. Figure~\ref{fig:interface_contracts} provides a visual overview of all five interface contracts. The following subsections specify the operations, invariants, and metrics for each interface.

\input{figures/fig_interface_contracts}

\subsubsection{L1--L2 Interface: Tensor Operation Contract}

L2 requests tensor operations from L1. In plain terms, this interface resembles an application invoking a GPU driver: the upper layer says ``compute this matrix multiplication,'' and the lower layer executes it efficiently and returns the result.

\begin{itemize}[nosep]
\item \textbf{Operations.}
\texttt{forward(batch\_tokens) $\to$ logits}.
Input a batch of token embedding vectors; receive the corresponding logit probability distributions.
\item \textbf{Invariants.}
Compute precision must not fall below the configured minimum (e.g., FP16/BF16).
\item \textbf{Metrics.}
FLOPS utilization, memory-bandwidth utilization, TTFT (Time to First Token), and TPOT (Time Per Output Token).
\end{itemize}

\subsubsection{L2--L3 Interface: KV Block Management Contract}

L3 requests allocation, loading, and eviction of KV blocks from L2. This interface is analogous to an application calling \texttt{malloc}/\texttt{free}: the upper layer says ``I need a block of memory to store KV cache,'' and the lower layer allocates physical space, handles fragmentation, and reclaims capacity when it runs short.

\begin{itemize}[nosep]
\item \textbf{Operations.}
\texttt{alloc}, \texttt{load(prefix\_hash)}, \texttt{evict(policy)}.
Allocate a new KV block, load an existing block by prefix hash, and reclaim inactive blocks according to a replacement policy.
\item \textbf{Invariants.}
KV blocks required by the current inference step must not be lost, analogous to the operating-system guarantee that a memory page actively referenced by a process will not be swapped out.
\item \textbf{Metrics.}
KV hit rate $H$ (the core parameter in Heuristic~I, Section~\ref{sec:law1}), peak GPU memory usage, and page-replacement count.
\end{itemize}

\subsubsection{L3--L4 Interface: Context Loading Contract}

L4 requests context loading from L3. This interface resembles an application memory-mapping a file via \texttt{mmap}: the upper layer says ``I need information about X,'' and the lower layer retrieves, compresses, and loads the relevant content from external storage into the context window.

\begin{itemize}[nosep]
\item \textbf{Operations.}
\texttt{read(query)}, \texttt{prefetch(hint)}, \texttt{compact}, \texttt{evict}.
Retrieve context by query, prefetch based on a hint, compress context, and evict inactive context.
\item \textbf{Invariants.}
Every returned context item carries a monotonic version stamp and a retrieval timestamp, analogous to a file system guaranteeing that reads return the most recent committed version (test: re-issuing the same read within the item's time-to-live returns an identical version stamp).
\item \textbf{Metrics.}
Retrieval recall, context compression ratio, and loading latency.
\end{itemize}

\subsubsection{L4--L5 Interface: Capability Invocation Contract}

L5 requests tool invocations and memory reads/writes through L4. This interface is analogous to an application issuing system calls to the operating system: the upper layer says ``I need to read/write a file'' (corresponding to memory access) or ``I need network access'' (corresponding to tool invocation), and the lower layer enforces permission checks and audit logging.

\begin{itemize}[nosep]
\item \textbf{Operations.}
\texttt{call(tool, args, capability\_token)}, \texttt{query(memory)}.
Invoke a tool with a capability token; query memory.
\item \textbf{Invariants.}
Tool invocations without a valid capability token are rejected (analogous to a process lacking file-open permissions); side-effecting operations are auditable~\cite{debenedetti2025camel}.
\item \textbf{Metrics.}
Tool-call success rate and permission-denial rate.
\end{itemize}

\subsubsection{L5--L6 Interface: Task Submission Contract}

L6 submits tasks to L5. This interface resembles a user launching a program from the command line: the upper layer says ``please complete this task,'' and the lower layer handles queuing, execution, monitoring, and result delivery.

\begin{itemize}[nosep]
\item \textbf{Operations.}
\texttt{submit(task\_spec)}, \texttt{poll}, \texttt{cancel}.
Submit a task specification, poll for status, and cancel a running task.
\item \textbf{Invariants.}
Task execution produces a traceable audit trail, and side-effecting operations are compensable: each is paired with a logged compensating action, so that a failure triggers compensating replay (test: no uncompensated side-effecting operation remains in the trace after a failed task), analogous to database transaction atomicity.
\item \textbf{Metrics.}
Task completion rate and human-takeover rate.
\end{itemize}

\paragraph{An empirical constraint worth noting.}
Anthropic's internal research reveals that engineers use AI for approximately 60\% of their work, yet can ``fully delegate'' only 0--20\% of tasks~\cite{anthropic2026agenticreport}.
This \textit{collaboration paradox} implies that the L5--L6 task submission contract cannot be modeled as a binary decision (delegate or not); it must instead reflect a spectrum of \textit{delegation confidence}. The more verifiable a task and the less it depends on organizational context, the higher the delegation confidence.
This observation connects directly to the dual-plane architecture: the deterministic control plane provides the auditable traces that increase verifiability and, in turn, raise delegation confidence.

\subsection{Dual-Plane Cross-Mapping with ICA}
\label{sec:dual-plane-icam}

Section~\ref{sec:framework} introduced the dual-plane architecture comprising a probabilistic execution plane (the non-deterministic reasoning of LLMs) and a deterministic control plane (auditable, programmatic control). A natural question arises: how do the two planes relate to the six ICA layers?

The answer is that they are \textbf{orthogonal dimensions}. ICA decomposes the system \textit{vertically} (from hardware to application), while the dual-plane architecture partitions it \textit{horizontally} (from probabilistic to deterministic). Their intersection forms a $6 \times 2$ matrix.

Table~\ref{tab:dual-plane-icam} presents this cross-mapping.

\begin{table}[h]
\centering
\footnotesize
\caption{Cross-mapping of the dual-plane architecture onto ICA layers.}
\label{tab:dual-plane-icam}
\begin{tabularx}{\textwidth}{p{0.10\textwidth}p{0.40\textwidth}p{0.40\textwidth}}
\toprule
ICA Layer & Probabilistic Execution Plane & Deterministic Control Plane \\
\midrule
L1 & Matrix operations, attention computation (probabilistic output) & Precision verification, hardware error detection \\
L2 & Semantic KV-cache matching, prefix sharing & Cache capacity management, eviction policy \\
L3 & Semantic retrieval, context compilation, summarization & Memory retention policy, version stamps, state persistence \\
L4 & Semantic understanding of tool results & Tool schemas, protocol specifications, capability annotations, auditing \\
L5 & Agent reasoning and decision-making (probabilistic) & Task scheduling, access control, failure recovery, trace logging \\
L6 & \multicolumn{2}{c}{Consumer of both planes: users receive intelligent output \textit{and} an auditable trace} \\
\bottomrule
\end{tabularx}
\end{table}

Two structural features emerge from this mapping.

\paragraph{The probabilistic plane is weighted toward the bottom.}
It primarily spans L1 (physical execution) and L2 (inference serving) and permeates into L3 (semantic retrieval and context compilation). This is unsurprising: the core capabilities of LLMs, namely language understanding and generation, are concentrated in these layers.

\paragraph{The deterministic plane is weighted toward the top.}
It primarily covers L5 (orchestration) and the interface-contract portions of L4 (auditing, capability annotations), extending into the state-management portions of L3 (memory retention policies, version stamps). System-level requirements for safety, reliability, and auditability demand deterministic guarantees.

L4 occupies a distinctive role as a \textit{transition layer}: tool schemas and protocol specifications belong to the deterministic plane (they are formalized interface definitions), while the semantic interpretation of tool results belongs to the probabilistic plane (the LLM must understand natural-language return values). L6 (application layer) is a consumer of both planes: end users simultaneously receive intelligent output from the probabilistic plane and an auditable trace from the deterministic plane. Accordingly, the plane boundary should be read as a \textit{graded interface} centered around L3--L4, not a single hard layer cut; the $6\times 2$ matrix in Table~\ref{tab:dual-plane-icam} is the authoritative mapping.

This cross-mapping resolves an apparent contradiction in the literature. Researchers focused on inference optimization regard the LLM as a compute engine at L1--L2~\cite{kwon2023pagedattention,zhou2024efficientinference}; those focused on agent systems see it as an intelligent kernel at L5--L6~\cite{openai_codex_overview_2026,anthropic_claudecode_overview_2026}; those focused on context management treat it as a memory processor at L3~\cite{memgpt2023,memoryos2025}. These perspectives are not mutually contradictory; they are observing different layers of the same ICA stack through different planes.

\subsection{Design Axioms}
\label{sec:icam-axioms}

Eight decades of classical computer architecture have distilled a set of enduring \textbf{design principles}: the principle of locality guides cache design, layered abstraction guides interface specification, and least privilege guides security architecture. We now transplant these principles into the model-native computing domain and, informed by its distinctive characteristics, propose six design axioms. For each axiom we identify its classical source, its embodiment within ICA, and its concrete design implication. Figure~\ref{fig:axioms_overview} summarizes the applicability of each axiom across the six ICA layers.

\input{figures/fig_axioms_overview}

\refstepcounter{axiom}
\begin{axiombox}{\theaxiom\enspace Locality Axiom}
Model-native computation exhibits both \textbf{temporal locality} (recently accessed context fragments are likely to be accessed again) and \textbf{semantic locality} (semantically similar queries access similar context).
\end{axiombox}
\begin{description}[style=nextline,leftmargin=2em,font=\bfseries]
\item[Intuition.] Just as running programs tend to revisit the same set of memory pages (motivating LRU cache replacement), LLMs in multi-turn conversations tend to re-reference the same context fragments. A user who first asks ``how do Python decorators work?'' will typically follow up with questions on the same topic. This is semantic locality in action.
\item[Classical source.] The theoretical foundations of cache design~\cite{drepper2007memory}.
\item[ICA embodiment.] KV cache reuse at L2 (identical prefixes share KV cache entries), prefix caching at L2 (multiple requests share the KV cache of a common system prompt), and semantic caching at L3 (semantically similar queries reuse previously generated results).
\item[Design implication.] Every ICA layer should employ locality-aware caching and prefetching. Heuristic~I (Section~\ref{sec:law1}) offers a back-of-envelope formula for this axiom.
\end{description}

\refstepcounter{axiom}
\begin{axiombox}{\theaxiom\enspace Layered Abstraction Axiom}
Complexity should be managed through layering: each layer depends only on the interface contract of the layer below, never on its implementation details.
\end{axiombox}
\begin{description}[style=nextline,leftmargin=2em,font=\bfseries]
\item[Intuition.] Just as a user application need not know whether the CPU is RISC or CISC, or whether the cache is 4-way or 8-way set-associative, an agent's orchestration logic need not know which inference engine serves the underlying model.
\item[Classical source.] ISA--microarchitecture separation; kernel--user-space separation~\cite{hennessy2017quantitative,ostep_2023}.
\item[ICA embodiment.] A model upgrade (a change in L2's implementation) must not break agent logic at L5; a tool-schema revision (an interface adjustment at L4) must not corrupt memory data at L3.
\item[Design implication.] Interface contracts between adjacent layers must be defined with sufficient formality to enable independent evolution. Breaking changes at any layer should be detectable through contract-compatibility tests rather than through end-to-end regressions.
\end{description}

\refstepcounter{axiom}
\begin{axiombox}{\theaxiom\enspace Probabilistic Execution Axiom}
The core execution in model-native computing, namely the model's forward pass, is probabilistic: identical inputs do not guarantee identical outputs.
\end{axiombox}
\begin{description}[style=nextline,leftmargin=2em,font=\bfseries]
\item[Intuition.] This is the \textbf{most fundamental difference} between model-native and classical computing. In classical systems, \texttt{2+2} always yields 4; in an LLM, posing the same question twice may produce different answers; this is not a bug but a feature. This inherent non-determinism invalidates traditional testing methodologies: we cannot verify system correctness by replaying an identical input sequence; instead, we must define \textit{semantic equivalence} as the acceptance criterion.
\item[Classical source.] No direct classical counterpart. This axiom is unique to model-native computing.
\item[ICA embodiment.] The entire ICA hierarchy must accommodate probabilistic behavior. Interface contracts at L4--L5 cannot demand fully deterministic tool-call outcomes; they must tolerate semantic-level variation. Failure-recovery mechanisms at L5 cannot rely on exact input replay; they require semantic-level checkpoints.
\item[Design implication.] Every interface contract, testing procedure, and reliability mechanism in the ICA stack must be designed for a world where the primary compute substrate is stochastic. Deterministic verification must be replaced (or supplemented) by statistical quality bounds and semantic acceptance tests.
\end{description}

\refstepcounter{axiom}
\begin{axiombox}{\theaxiom\enspace Virtualization Axiom}
Limited physical resources should be virtualized to present a larger logical resource, while preserving isolation and protection.
\end{axiombox}
\begin{description}[style=nextline,leftmargin=2em,font=\bfseries]
\item[Intuition.] Just as a 32-bit operating system uses virtual memory to give each process the illusion of a 4\,GB address space (even when physical RAM is only 1\,GB), a model-native system must give each agent the illusion of an unbounded context window (even when the physical context window is only 128\,K tokens).
\item[Classical source.] Virtual memory and process isolation~\cite{ostep_2023}.
\item[ICA embodiment.] MemGPT's virtual context management~\cite{memgpt2023}; PagedAttention's page-based KV management~\cite{kwon2023pagedattention}, which organizes KV cache into ``pages,'' allocates them on demand, and reclaims inactive pages, a design identical in structure to the operating system's virtual memory manager; agent sandboxing~\cite{openai_codex_sandbox_2026}, which isolates an agent's execution environment from the host system, much as a process's address space is isolated from other processes.
\item[Design implication.] Every bounded resource at every ICA layer should be wrapped in a virtualization layer that (a)~presents an apparently unbounded interface to the consumer above, and (b)~enforces isolation between concurrent consumers.
\end{description}

\refstepcounter{axiom}
\begin{axiombox}{\theaxiom\enspace Least Privilege Axiom}
Every execution unit should be granted only the minimal set of permissions required to complete its assigned task.
\end{axiombox}
\begin{description}[style=nextline,leftmargin=2em,font=\bfseries]
\item[Intuition.] Just as a text editor does not need administrator privileges to edit a file, a code-generation agent should not have the authority to delete the entire repository.
\item[Classical source.] Operating-system security principles~\cite{ostep_2023}.
\item[ICA embodiment.] CaMeL's capability-security mechanism~\cite{debenedetti2025camel}, which assigns fine-grained capability tokens to each agent operation; Codex's OS-level sandbox~\cite{openai_codex_security_2026}, which restricts an agent's file-system and network access permissions.
\item[Design implication.] The L4--L5 interface must enforce per-operation permission checks via capability tokens (Section~\ref{sec:icam-interfaces}), and every L5 orchestration layer must support configurable permission policies that can be tightened or relaxed without modifying agent logic.
\end{description}

\refstepcounter{axiom}
\begin{axiombox}{\theaxiom\enspace Observability Axiom}
Every side-effecting operation must produce a traceable, auditable event record.
\end{axiombox}
\begin{description}[style=nextline,leftmargin=2em,font=\bfseries]
\item[Intuition.] Just as a database writes a Write-Ahead Log (WAL) for every write operation, every tool invocation and file modification made by an agent should be recorded so that the rationale for each action can be reconstructed after the fact.
\item[Classical source.] System logging and performance monitoring~\cite{linux_cgroup_2026}.
\item[ICA embodiment.] The tracing mechanism in the OpenAI Agents SDK~\cite{openai_codex_agentsdk_2026}, which logs each agent decision and tool invocation; the trace-grading methodology in agent-security research~\cite{agenticsecurity2025}, which evaluates the safety of an agent's action trajectory.
\item[Design implication.] The deterministic control plane must capture a complete, tamper-evident log of all operations that pass through the L4--L5 interface. This log serves dual purposes: it enables post-hoc debugging and compliance auditing, and it provides the delegation-confidence signal (Section~\ref{sec:icam-interfaces}) that determines which tasks can be safely delegated at the L5--L6 boundary.
\end{description}

%% file: figures/fig_interface_contracts.tex
\begin{figure}[htbp]
\centering
\small
\newcommand{\ifccard}[5]{%
  \begin{tikzpicture}
    \def\cardW{14.5cm}
    \def\tagW{2.6cm}
    \def\colW{3.6cm}
    \fill[#2!12, rounded corners=5pt] (0,0) rectangle (\cardW, 1.6cm);
    \draw[#2!50, rounded corners=5pt, line width=0.6pt] (0,0) rectangle (\cardW, 1.6cm);
    \fill[#2!70, rounded corners=4pt] (0.08cm,0.08cm) rectangle (\tagW, 1.52cm);
    \node[text=white, font=\bfseries\small, align=center, text width=2.4cm]
      at ({(\tagW+0.08cm)/2}, 0.8cm) {#1};
    \draw[#2!30, line width=0.4pt] (\tagW+0.1cm, 0.12cm) -- (\tagW+0.1cm, 1.48cm);
    \draw[#2!30, line width=0.4pt] (\tagW+\colW+0.2cm, 0.12cm) -- (\tagW+\colW+0.2cm, 1.48cm);
    \draw[#2!30, line width=0.4pt] (\tagW+2*\colW+0.3cm, 0.12cm) -- (\tagW+2*\colW+0.3cm, 1.48cm);
    \node[font=\bfseries\scriptsize, #2!80!black] at (\tagW+\colW/2+0.15cm, 1.35cm) {\textsc{api calls}};
    \node[font=\bfseries\scriptsize, #2!80!black] at (\tagW+\colW*1.5+0.25cm, 1.35cm) {\textsc{invariant}};
    \node[font=\bfseries\scriptsize, #2!80!black] at (\tagW+\colW*2.5+0.35cm, 1.35cm) {\textsc{key metric}};
    \node[font=\ttfamily\scriptsize, align=left, text width=\colW-0.2cm, #2!80!black]
      at (\tagW+\colW/2+0.15cm, 0.7cm) {#3};
    \node[font=\scriptsize, align=left, text width=\colW-0.2cm, black!75]
      at (\tagW+\colW*1.5+0.25cm, 0.7cm) {#4};
    \node[font=\scriptsize, align=left, text width=\colW-0.2cm, black!75]
      at (\tagW+\colW*2.5+0.35cm, 0.7cm) {#5};
  \end{tikzpicture}%
}
\newcommand{\layerbar}[3]{%
  \begin{tikzpicture}
    \fill[#2, rounded corners=3pt] (0,0) rectangle (14.5cm, 0.6cm);
    \node[text=white, font=\bfseries\small] at (7.25cm, 0.3cm) {#1\enspace\normalfont\small #3};
  \end{tikzpicture}%
}

\definecolor{cOrg}{HTML}{C8551B}
\definecolor{cAmb}{HTML}{D9822B}
\definecolor{cBlu}{HTML}{2E6FA3}

\setlength{\tabcolsep}{0pt}
\renewcommand{\arraystretch}{0}
\resizebox{0.7\textwidth}{!}{%
\begin{tabular}{c}
\layerbar{L6 \;—\; Application Layer}{cOrg}{user tasks · agent apps · governance}\\[2pt]
\ifccard{L5--L6\\[2pt]Task\\Submission}{cOrg}%
  {submit(task\_spec)\\ poll\\ cancel}%
  {Audit trail on every side-effect; compensable rollback}%
  {Task completion rate; delegation confidence}\\[2pt]
\layerbar{L5 \;—\; Orchestration Layer}{cOrg}{planning · approval · failure recovery}\\[2pt]
\ifccard{L4--L5\\[2pt]Capability\\Invocation}{cOrg}%
  {call(tool, args, cap\_token)\\ query(memory)}%
  {Valid capability token required; all side-effects auditable}%
  {Tool-call success rate; permission-denial rate}\\[2pt]
\layerbar{L4 \;—\; Semantic Interface}{cAmb}{tool schema · memory API · trace API}\\[2pt]
\ifccard{L3--L4\\[2pt]Context\\Loading}{cAmb}%
  {read(query)\\ prefetch(hint)\\ compact\enspace evict}%
  {Monotonic version stamp + retrieval timestamp per returned item}%
  {Retrieval recall; compression ratio; load latency}\\[2pt]
\layerbar{L3 \;—\; Context Management}{cBlu}{hot/warm/cold memory · summarization}\\[2pt]
\ifccard{L2--L3\\[2pt]KV Block\\Management}{cBlu}%
  {alloc\\ load(prefix\_hash)\\ evict(policy)}%
  {Active KV blocks never evicted during live inference step}%
  {KV hit rate $H$; GPU memory; page-replacement count}\\[2pt]
\layerbar{L2 \;—\; Inference Serving}{cBlu}{KV cache · batching · prefill/decode}\\[2pt]
\ifccard{L1--L2\\[2pt]Tensor\\Operation}{cBlu}%
  {forward(batch\_tokens)\\ \quad $\to$ logits}%
  {Compute precision $\geq$ FP16\,/\,BF16}%
  {FLOPS utilization; TTFT; TPOT}\\[2pt]
\layerbar{L1 \;—\; Physical Execution}{cBlu}{GPU · TPU · ASIC · compute-in-memory}\\
\end{tabular}%
}

\caption{ICA inter-layer interface contracts.
Each card specifies the API operations, invariant, and key metric for the
corresponding layer boundary.
Orange cards (L4--L6) belong to the deterministic control plane;
blue cards (L1--L3) belong to the probabilistic execution plane;
the amber card (L3--L4) marks the graded transition layer.}
\label{fig:interface_contracts}
\end{figure}

%% file: figures/fig_axioms_overview.tex
\begin{figure}[htbp]
\centering
\providecolor{axHigh}{HTML}{C44E52}
\providecolor{axMed}{HTML}{E8A87C}
\providecolor{axLow}{HTML}{F5DEB3}
\providecolor{axNone}{HTML}{F0F0F0}
\providecolor{hdrBlue}{HTML}{4A78A8}
\resizebox{0.8\textwidth}{!}{%
\begin{tikzpicture}[
    font=\small,
    >=Latex,
    cellH/.style={fill=axHigh!80, draw=axHigh!60!black, rounded corners=1pt, line width=0.4pt},
    cellM/.style={fill=axMed!80,  draw=axMed!60!black,  rounded corners=1pt, line width=0.4pt},
    cellL/.style={fill=axLow!80,  draw=axLow!60!black,  rounded corners=1pt, line width=0.4pt},
    cellN/.style={fill=axNone,    draw=gray!30,          rounded corners=1pt, line width=0.4pt},
    colhdr/.style={font=\small\bfseries, align=center, text=hdrBlue!80!black},
    rowhdr/.style={font=\small\bfseries, align=center, anchor=east},
    celltxt/.style={font=\scriptsize, align=center, text=black!70},
]


\def\cw{1.90}
\def\ch{1.00}

\def\xA{-0.05}
\def\xB{1.97}
\def\xC{3.99}
\def\xD{6.01}
\def\xE{8.03}
\def\xF{10.05}

\def\xAl{-1.00}
\def\xBl{1.02}
\def\xCl{3.04}
\def\xDl{5.06}
\def\xEl{7.08}
\def\xFl{9.10}

\def\yAb{-1.12}
\def\yBb{-2.24}
\def\yCb{-3.36}
\def\yDb{-4.48}
\def\yEb{-5.60}
\def\yFb{-6.72}

\def\yAc{-0.62}
\def\yBc{-1.74}
\def\yCc{-2.86}
\def\yDc{-3.98}
\def\yEc{-5.10}
\def\yFc{-6.22}

\node[colhdr] at (\xA,  0.3) {L1};
\node[colhdr] at (\xB,  0.3) {L2};
\node[colhdr] at (\xC,  0.3) {L3};
\node[colhdr] at (\xD,  0.3) {L4};
\node[colhdr] at (\xE,  0.3) {L5};
\node[colhdr] at (\xF,  0.3) {L6};

\node[rowhdr] at (-1.20, \yAc) {A1};
\node[rowhdr] at (-1.20, \yBc) {A2};
\node[rowhdr] at (-1.20, \yCc) {A3};
\node[rowhdr] at (-1.20, \yDc) {A4};
\node[rowhdr] at (-1.20, \yEc) {A5};
\node[rowhdr] at (-1.20, \yFc) {A6};

\draw[cellH] (\xAl,\yAb) rectangle +(\cw,\ch); \node[celltxt] at (\xA,\yAc) {Primary};
\draw[cellH] (\xBl,\yAb) rectangle +(\cw,\ch); \node[celltxt] at (\xB,\yAc) {Primary};
\draw[cellM] (\xCl,\yAb) rectangle +(\cw,\ch); \node[celltxt] at (\xC,\yAc) {Secondary};
\draw[cellL] (\xDl,\yAb) rectangle +(\cw,\ch); \node[celltxt] at (\xD,\yAc) {Tertiary};
\draw[cellN] (\xEl,\yAb) rectangle +(\cw,\ch); \node[celltxt] at (\xE,\yAc) {---};
\draw[cellN] (\xFl,\yAb) rectangle +(\cw,\ch); \node[celltxt] at (\xF,\yAc) {---};

\draw[cellM] (\xAl,\yBb) rectangle +(\cw,\ch); \node[celltxt] at (\xA,\yBc) {Secondary};
\draw[cellH] (\xBl,\yBb) rectangle +(\cw,\ch); \node[celltxt] at (\xB,\yBc) {Primary};
\draw[cellH] (\xCl,\yBb) rectangle +(\cw,\ch); \node[celltxt] at (\xC,\yBc) {Primary};
\draw[cellH] (\xDl,\yBb) rectangle +(\cw,\ch); \node[celltxt] at (\xD,\yBc) {Primary};
\draw[cellH] (\xEl,\yBb) rectangle +(\cw,\ch); \node[celltxt] at (\xE,\yBc) {Primary};
\draw[cellM] (\xFl,\yBb) rectangle +(\cw,\ch); \node[celltxt] at (\xF,\yBc) {Secondary};

\draw[cellH] (\xAl,\yCb) rectangle +(\cw,\ch); \node[celltxt] at (\xA,\yCc) {Primary};
\draw[cellH] (\xBl,\yCb) rectangle +(\cw,\ch); \node[celltxt] at (\xB,\yCc) {Primary};
\draw[cellM] (\xCl,\yCb) rectangle +(\cw,\ch); \node[celltxt] at (\xC,\yCc) {Secondary};
\draw[cellM] (\xDl,\yCb) rectangle +(\cw,\ch); \node[celltxt] at (\xD,\yCc) {Secondary};
\draw[cellL] (\xEl,\yCb) rectangle +(\cw,\ch); \node[celltxt] at (\xE,\yCc) {Tertiary};
\draw[cellN] (\xFl,\yCb) rectangle +(\cw,\ch); \node[celltxt] at (\xF,\yCc) {---};

\draw[cellL] (\xAl,\yDb) rectangle +(\cw,\ch); \node[celltxt] at (\xA,\yDc) {Tertiary};
\draw[cellH] (\xBl,\yDb) rectangle +(\cw,\ch); \node[celltxt] at (\xB,\yDc) {Primary};
\draw[cellH] (\xCl,\yDb) rectangle +(\cw,\ch); \node[celltxt] at (\xC,\yDc) {Primary};
\draw[cellM] (\xDl,\yDb) rectangle +(\cw,\ch); \node[celltxt] at (\xD,\yDc) {Secondary};
\draw[cellL] (\xEl,\yDb) rectangle +(\cw,\ch); \node[celltxt] at (\xE,\yDc) {Tertiary};
\draw[cellN] (\xFl,\yDb) rectangle +(\cw,\ch); \node[celltxt] at (\xF,\yDc) {---};

\draw[cellN] (\xAl,\yEb) rectangle +(\cw,\ch); \node[celltxt] at (\xA,\yEc) {---};
\draw[cellL] (\xBl,\yEb) rectangle +(\cw,\ch); \node[celltxt] at (\xB,\yEc) {Tertiary};
\draw[cellM] (\xCl,\yEb) rectangle +(\cw,\ch); \node[celltxt] at (\xC,\yEc) {Secondary};
\draw[cellH] (\xDl,\yEb) rectangle +(\cw,\ch); \node[celltxt] at (\xD,\yEc) {Primary};
\draw[cellH] (\xEl,\yEb) rectangle +(\cw,\ch); \node[celltxt] at (\xE,\yEc) {Primary};
\draw[cellM] (\xFl,\yEb) rectangle +(\cw,\ch); \node[celltxt] at (\xF,\yEc) {Secondary};

\draw[cellN] (\xAl,\yFb) rectangle +(\cw,\ch); \node[celltxt] at (\xA,\yFc) {---};
\draw[cellM] (\xBl,\yFb) rectangle +(\cw,\ch); \node[celltxt] at (\xB,\yFc) {Secondary};
\draw[cellM] (\xCl,\yFb) rectangle +(\cw,\ch); \node[celltxt] at (\xC,\yFc) {Secondary};
\draw[cellH] (\xDl,\yFb) rectangle +(\cw,\ch); \node[celltxt] at (\xD,\yFc) {Primary};
\draw[cellH] (\xEl,\yFb) rectangle +(\cw,\ch); \node[celltxt] at (\xE,\yFc) {Primary};
\draw[cellH] (\xFl,\yFb) rectangle +(\cw,\ch); \node[celltxt] at (\xF,\yFc) {Primary};

\draw[cellH] (-1.0,-7.55) rectangle +(0.5,0.40); \node[font=\scriptsize,anchor=west] at (-0.35,-7.35) {Primary};
\draw[cellM] (2.0,-7.55) rectangle +(0.5,0.40); \node[font=\scriptsize,anchor=west] at (2.65,-7.35) {Secondary};
\draw[cellL] (5.0,-7.55) rectangle +(0.5,0.40); \node[font=\scriptsize,anchor=west] at (5.65,-7.35) {Tertiary};
\draw[cellN] (8.0,-7.55) rectangle +(0.5,0.40); \node[font=\scriptsize,anchor=west] at (8.65,-7.35) {Not applicable};

\end{tikzpicture}}%
\caption{Applicability heatmap of the six ICA design axioms across the six architecture layers. \textcolor{axHigh!80!black}{\textbf{Primary}} indicates the axiom directly governs design decisions at that layer; \textcolor{axMed!80!black}{\textbf{secondary}} indicates significant but indirect applicability; \textcolor{axLow!80!black}{\textbf{tertiary}} indicates contextual relevance. A1~(Locality) is most critical at the inference and KV-cache layers; A5~(Least Privilege) and A6~(Observability) dominate the interface and orchestration layers.}
\label{fig:axioms_overview}
\end{figure}

%% file: en_sections/section_laws.tex
\section{Design Heuristics for Model-Native Systems}
\label{sec:laws}

The enduring power of classical computer architecture lies in its \textbf{computable principles}. Amdahl's Law tells us that if the serial fraction of a program is $(1-f)$, then even an infinite speedup of the parallel portion yields an overall speedup no greater than $1/(1-f)$, a theoretical ceiling for parallel computing \cite{hennessy2017quantitative}. The Roofline model tells us that the peak performance of each compute core is governed jointly by its arithmetic intensity and available memory bandwidth, enabling engineers to diagnose whether a kernel is compute-bound or memory-bound at a glance \cite{hennessy2017quantitative}. The value of these laws lies not in their mathematical sophistication but in the \textbf{unifying quantitative language} they provide, enabling data-driven engineering decisions.

In this section we propose three \emph{Amdahl-style design heuristics} for model-native computing. Formally, Heuristics~I and~III are isomorphic to Amdahl's Law; semantically, however, they capture performance constraints unique to model-native computing. We are explicit from the outset: these are \textbf{not} novel mathematical discoveries, nor validated scaling laws, but the \textbf{deliberate transplantation} of Amdahl's analytical framework into the model-native regime, reinterpreted with domain-specific parameters. Their value lies in furnishing a previously qualitative design space with back-of-envelope intuition---transforming ``what should I optimize first?'' from pure guesswork into order-of-magnitude reasoning. We deliberately frame them as \emph{heuristics} rather than \emph{laws}: a law implies a validated, falsifiable quantitative constraint, which the data in this young field cannot yet support. Where we appeal to published results below, we do so to \emph{illustrate} plausible parameter ranges, not to test predictive power; we identify systematic predict-then-measure validation as the central open task for future work.

\subsection{Heuristic~I: Semantic Locality}
\label{sec:law1}

\begin{heuristic}[Semantic Locality]
In autoregressive inference, the speedup $S$ is governed by the KV-cache hit rate $H$ and the KV-reuse speedup factor $\alpha$:
\begin{equation}
\label{eq:locality}
S = \frac{1}{(1-H) + H \cdot \alpha^{-1}}
\end{equation}
\end{heuristic}

\subsubsection{Intuitive Parameter Semantics}

Before proceeding to the derivation, we develop intuition for the two core parameters.

\begin{itemize}[nosep]
\item $H$ (KV-cache hit rate): \textbf{The fraction of KV tensors that can be reused directly rather than recomputed from scratch.} Consider a customer-service chatbot: if a user engages in five consecutive turns on the same topic, the KV tensors produced during the first four turns can be reused by the fifth, and $H$ will be close to~1. Conversely, if every conversation starts on a brand-new topic, previous KV caches are of no use and $H$ approaches~0. The domain of $H$ is $[0, 1]$.

\item $\alpha$ (KV-reuse speedup factor): \textbf{How many times faster a cache hit is compared to a miss.} On a hit, the system simply loads existing KV vectors from memory in $O(1)$ time; on a miss, it must compute the full attention in $O(n^{2})$ time. The parameter $\alpha$ quantifies this gap. In typical deployments, $\alpha$ ranges from $10$ to $100$.
\end{itemize}

\subsubsection{Derivation}

We begin from first principles.

\paragraph{Step~1: Decompose execution time.}
Let $T$ denote the total wall-clock time for a single inference request. This time decomposes into two disjoint components:
\begin{equation}
T = T_{\mathrm{miss}} + T_{\mathrm{hit}}
\end{equation}
where $T_{\mathrm{miss}}$ is the time spent on KV-cache misses (requiring full recomputation) and $T_{\mathrm{hit}}$ is the time spent on KV-cache hits (direct reuse of existing KV tensors).

\paragraph{Step~2: Express each component in terms of the hit rate.}
Let $T_{\mathrm{total}}$ be the time required when \emph{no} cache is available (i.e., every token must be computed from scratch). When a cache is present, a fraction $(1-H)$ of the work incurs misses while a fraction $H$ enjoys hits. Because a hit is $\alpha$ times faster than a miss, the hit component costs only $1/\alpha$ of its uncached value:
\begin{equation}
T = T_{\mathrm{total}} \cdot (1-H) + T_{\mathrm{total}} \cdot H \cdot \frac{1}{\alpha}
\end{equation}
Collecting terms:
\begin{equation}
T = T_{\mathrm{total}} \cdot \left[(1-H) + \frac{H}{\alpha}\right]
\end{equation}

\paragraph{Step~3: Define the speedup.}
The speedup $S$ is the ratio of uncached to cached execution time:
\begin{equation}
S = \frac{T_{\mathrm{total}}}{T} = \frac{T_{\mathrm{total}}}{T_{\mathrm{total}} \cdot \left[(1-H) + H/\alpha\right]} = \frac{1}{(1-H) + H/\alpha}
\end{equation}
which is exactly Equation~(\ref{eq:locality}).

\subsubsection{Limit Behavior}

\begin{itemize}[nosep]
\item When $H \to 1$ (nearly universal hits): $S \to 1/(0 + 1/\alpha) = \alpha$. The speedup ceiling is governed entirely by reuse efficiency: even with 100\% hits, the maximum speedup is $\alpha$.

\item When $H \to 0$ (nearly universal misses): $S \to 1/(1 + 0) = 1$. No speedup whatsoever, degenerating to the uncached baseline.

\item When $\alpha \to \infty$ (hits become instantaneous): $S \to 1/(1-H)$. The speedup ceiling is governed entirely by the hit rate: even if hits are ``free,'' misses still consume time.
\end{itemize}

\subsubsection{Numerical Example}

Consider an LLM inference service with $\alpha = 10$ (a hit is ten times faster than a miss). We examine how $H$ affects $S$:

\begin{itemize}[nosep]
\item $H = 0.5$ (half of all KV tensors are cached): $S = 1/(0.5 + 0.05) = 1.82\times$ speedup.
\item $H = 0.8$ (80\% hit rate): $S = 1/(0.2 + 0.08) = 3.57\times$ speedup.
\item $H = 0.95$ (95\% hit rate): $S = 1/(0.05 + 0.095) = 6.9\times$ speedup.
\end{itemize}

Raising $H$ from 0.5 to 0.8 (an increase of 0.3) yields roughly a $2\times$ improvement ($3.57/1.82$), whereas raising $H$ from 0.8 to 0.95 (an increase of only 0.15, half the previous gain) yields roughly a $1.9\times$ improvement ($6.9/3.57$), exhibiting \textbf{increasing marginal returns}: the second, smaller improvement produces nearly the same multiplicative speedup as the first, larger one. Once a reasonable cache foundation exists, every additional percentage point of hit rate pays disproportionate dividends.

\subsubsection{Correspondence to Amdahl's Law}

Equation~(\ref{eq:locality}) is mathematically isomorphic to Amdahl's Law:
\begin{equation}
S_{\mathrm{Amdahl}} = \frac{1}{(1-f) + f/p}
\end{equation}
where $f$ is the fraction of work that can be accelerated and $p$ is the per-unit speedup. The correspondence is: $f \leftrightarrow H$ (the accelerable fraction equals the cache-hit fraction) and $p \leftrightarrow \alpha$ (the per-unit speedup equals the KV-reuse factor).

\paragraph{Design implication.}
There are two levers for improving inference throughput: increasing $H$ (through better caching policies, prefix sharing, or semantic caching \cite{gim2024promptcache}) and increasing $\alpha$ (through faster KV loading, PagedAttention \cite{kwon2023pagedattention}, or KV quantization \cite{hooper2024kvquant}). These two levers are \textbf{complementary}: when $H$ is already high, further increases in $H$ yield large marginal gains; when $H$ is low, raising $\alpha$ alone is ineffective, and the priority should be improving the caching strategy first.

\subsubsection{Illustration, Not Validation}

We emphasize a methodological caveat that applies to all three heuristics. Equation~(\ref{eq:locality}) relates three quantities---$S$, $H$, and $\alpha$---but a deployed system typically reports only $S$. With a single observation and two unknowns, one cannot \emph{validate} the model: assuming a value of $\alpha$ and solving for $H$ is parameter recovery, which is unfalsifiable by construction (any reported $S$ can be absorbed by a suitable $(H,\alpha)$ pair). The examples below therefore \emph{illustrate} that published speedups are consistent with plausible parameter ranges; they do not test predictive power. A genuine validation would independently measure $\alpha$ and $H$, \emph{predict} $S$, and compare---a predict-then-measure protocol we defer to future work.

\begin{enumerate}[nosep]
\item \textbf{vLLM / PagedAttention.} Kwon et al.\ report $2$--$4\times$ throughput improvement over baseline \cite{kwon2023pagedattention}. We caution that this gain arises primarily from eliminating KV-memory fragmentation and enabling continuous batching, \emph{not} from KV reuse along the dimension $H$ in our heuristic; PagedAttention does not itself implement prefix reuse. It is therefore an illustration of the broader locality principle (memory management pays off) rather than a clean instantiation of Equation~(\ref{eq:locality}).

\item \textbf{Prompt Cache.} Gim et al.\ report time-to-first-token (TTFT) reductions of $8\times$ to $60\times$ on highly structured prompts via prompt-level prefix caching \cite{gim2024promptcache}. Here the reuse mechanism \emph{does} match $H$: assuming $\alpha \approx 60$ (cached prefix attention is skipped rather than recomputed), the implied $H$ ranges from $0.89$ to nearly $1.00$, consistent with near-perfect reuse of a fixed system-prompt prefix. This is the cleaner illustration, though it still back-solves $H$ rather than measuring it independently.
\end{enumerate}

\subsection{Heuristic~II: Context Budget}
\label{sec:law2}

\begin{heuristic}[Context Budget]
The effective working set $W_{\mathrm{eff}}$ of a language model is the integral of the attention retention rate $\beta(L)$ over the nominal context window $C$:
\begin{equation}
\label{eq:context}
W_{\mathrm{eff}} = \int_0^C \beta(L)\,dL = C \cdot \bar{\beta} \leq C, \qquad \bar{\beta} = \frac{1}{C}\int_0^C \beta(L)\,dL
\end{equation}
where $C$ is the model's nominal context window size (e.g., 128K, 1M tokens), $L$ is the positional distance of information within the window, $\beta(L) \in [0,1]$ measures the probability that information at position $L$ is effectively utilized by the model, and $\bar{\beta} \in (0,1]$ is the average retention rate across the full window. The inequality $W_{\mathrm{eff}} \leq C$ follows directly from $\beta(L) \leq 1$, with equality only when every token is perfectly utilized.
\end{heuristic}

\subsubsection{Intuitive Parameter Semantics}

\begin{itemize}[nosep]
\item $C$ (context window size): The model's ``physical memory capacity.'' For example, GPT-4-Turbo has $C = 128\mathrm{K}$ and Gemini~1.5~Pro has $C = 1\mathrm{M}$. $C$ determines how many tokens the model can ``see'' at once.

\item $\beta(L)$ (attention retention rate): The central innovation of Heuristic~II. It measures \textbf{how much of the information placed at position $L$ within the context window the model actually uses}. $\beta(L) = 1$ means information at that position is perfectly utilized; $\beta(L) = 0$ means it is effectively ignored.

\item $W_{\mathrm{eff}}$ (effective working set): The model's ``truly usable memory.'' Even if $C = 1\mathrm{M}$, if $\beta(L)$ is low at certain positions, $W_{\mathrm{eff}}$ may be far smaller than $C$, analogous to purchasing a 1\,TB hard drive but reliably using only 100\,GB.
\end{itemize}

\subsubsection{Why $\beta(L)$ Decays: the ``Lost in the Middle'' Phenomenon}

$\beta(L)$ is not a constant but a function that decays with positional distance $L$. This empirical fact originates from a widely observed phenomenon known as the \textbf{``Lost in the Middle''} effect.

Liu et al.\ systematically studied this phenomenon in their seminal work \cite{liu2023lostmiddle}: they placed critical information at various positions within long contexts and measured whether the model could retrieve it. The results revealed a pronounced \textbf{U-shaped curve}: information at the \textbf{beginning} of the context (analogous to a primacy effect) and at the \textbf{end} (analogous to a recency effect) was retrieved with high accuracy, while information in the \textbf{middle} suffered a significant drop in retrieval accuracy.

Subsequent studies have further quantified this decay. The RULER benchmark shows that retrieval accuracy drops from $>$95\% at 4K tokens to $<$70\% at 128K tokens \cite{ruler2024}. LongBench~v2 confirms that the decay is more severe for tasks requiring deep reasoning \cite{longbenchv2_2024}. The LOFT study demonstrates that even models supporting ultra-long contexts exhibit significant under-utilization of their stated capacity \cite{loft2024}. Recent work suggests that the effective context length exploited by LLMs may be only 10--20\% of the stated window.

Figure~\ref{fig_beta_decay} visualises the empirically observed U-shaped retention curve alongside the monotone decay envelope used to approximate its average.
\input{figures/fig_beta_decay.tex}

\subsubsection{The Functional Form of $\beta(L)$: U-Shape, Not Monotone Decay}

A modeling choice must be made for $\beta(L)$. The empirically dominant pattern is the \textbf{U-shape} of ``Lost in the Middle''~\cite{liu2023lostmiddle}: retention is high at the start (primacy) and end (recency) of the window and lowest in the middle. A purely monotone decay---e.g.\ an exponential $\beta(L) \approx \beta_{0} \cdot e^{-\lambda L / C}$---\emph{cannot} represent this U-shape and would systematically over-penalize late-window (recency) information. We therefore do \emph{not} commit to the exponential as ``the'' model of $\beta(L)$; it serves only as an analytically convenient envelope for the average decay $\bar{\beta}$ when a closed-form working-set estimate is needed, and any monotone decreasing form (linear, power-law, exponential) yields the same qualitative conclusion $W_{\mathrm{eff}} < C$.

Three observations are nevertheless worth recording. \textbf{First}, RULER-style retrieval accuracy declines roughly linearly on a log scale with length~\cite{ruler2024}, consistent with---but not uniquely selecting---an exponential envelope for the bulk of the curve. \textbf{Second}, the primacy/recency peaks mean the true $\beta(L)$ is non-monotone, so a faithful model needs at least two regimes; fitting it across $C \in \{4\mathrm{K}\dots128\mathrm{K}\}$ on RULER/LongBench and comparing functional forms (exponential vs.\ power-law vs.\ U-shape via AIC/BIC) is a concrete open task. \textbf{Third}, because $\bar{\beta} = \tfrac{1}{C}\int_0^C \beta(L)\,dL$, the working-set definition $W_{\mathrm{eff}} = C\bar{\beta}$ is robust to the choice of functional form: whatever the shape of $\beta(L)$, the substantive claim is simply $W_{\mathrm{eff}} < C$.

We emphasize that the value of Heuristic~II is conceptual: it establishes that the effective working set is strictly smaller than the nominal window, and frames the magnitude of this gap---and strategies for closing it (hierarchical memory at L3)---as empirical questions rather than asserted constants.

\subsubsection{Correspondence to Classical Working-Set Theory}

Heuristic~II has a deep structural correspondence with Denning's working-set model \cite{denning1968thrashing}.

In 1968, Peter Denning proposed that the set of memory pages accessed by a process within a time window $\tau$ constitutes its \textbf{working set} $W(\tau)$. Denning's central insight was that if physical memory is smaller than $W(\tau)$, the system suffers \textbf{thrashing}, where the CPU spends most of its time waiting for pages to be swapped in and out rather than performing useful work \cite{denning1968thrashing,denning2020workingset}.

Equation~(\ref{eq:context}) is the semantic analogue of the working-set model: $C$ corresponds to physical memory size and $W_{\mathrm{eff}}$ to the process's true working set. The crucial difference is that in the classical model, the ``usefulness'' of a page is binary (a process either accesses it or does not); in Heuristic~II, $\beta(L)$ is continuous: the model's utilization of information at different positions within the context forms a gradual spectrum.

\subsubsection{Design Implications}

Heuristic~II carries three direct design implications:

\textbf{(1) Diminishing marginal returns from simply enlarging $C$.} Extending the context window (e.g., from 128K to 1M) does increase $C$, but if $\beta(L)$ decays rapidly with $L$, then $W_{\mathrm{eff}}$ grows far more slowly than $C$. Merely chasing a larger nominal context window is not a fundamental solution to ``insufficient memory.''

\textbf{(2) Critical information should be placed at high-$\beta$ positions.} Because $\beta(L)$ is highest at the beginning and end of the context, prompt engineers should concentrate critical information at these locations and compress or offload auxiliary information to a RAG system.

\textbf{(3) This provides quantitative justification for hierarchical memory (L3).} When $C$ is fixed, the only way to break through the $W_{\mathrm{eff}}$ ceiling is via semantic virtual memory (the responsibility of L3), swapping inactive context out to external storage and swapping active context in to high-$\beta$ positions. MemGPT's virtual context management \cite{memgpt2023} and MemoryOS's hierarchical memory \cite{memoryos2025} embody precisely this strategy.

\subsubsection{Illustration}

Illustrating Heuristic~II requires measuring the gap between a model's stated context window $C$ and its effective context $W_{\mathrm{eff}}$.

\begin{enumerate}[nosep]
\item \textbf{RULER benchmark.} Hsieh et al.\ designed RULER to test effective context via needle-in-a-haystack retrieval tasks at varying context lengths \cite{ruler2024}. Even for models that nominally support 128K contexts, accuracy on complex retrieval tasks drops from $>$95\% at 4K to $<$70\% at 128K. This monotone decline in retrieval accuracy as $C$ grows is consistent with a decreasing $\bar{\beta}$, confirming that $W_{\mathrm{eff}} = C \cdot \bar{\beta}$ grows substantially slower than $C$ and is far below the nominal window size.

\item \textbf{LongBench v2.} Bai et al.\ find that decay is substantially faster for deep reasoning tasks (e.g., mathematical proofs, multi-step logic) than for simple retrieval \cite{longbenchv2_2024}. This is consistent with $\lambda$ varying with task type rather than being a fixed property of the model.
\end{enumerate}

\subsection{Heuristic~III: Agent Speedup}
\label{sec:law3}

\begin{heuristic}[Agent Speedup]
When multiple agents execute in parallel, the speedup $S_{\mathrm{agent}}$ is governed by the parallelizable fraction $F$, the number of sub-agents $N$, and the orchestration efficiency $E$:
\begin{equation}
\label{eq:agent}
S_{\mathrm{agent}} = \frac{1}{(1-F) + \frac{F}{N \cdot E}}
\end{equation}
\end{heuristic}

\subsubsection{Intuitive Parameter Semantics}

\begin{itemize}[nosep]
\item $F$ (parallelizable fraction): \textbf{The fraction of the task that can be distributed across multiple agents for concurrent execution.} For example, a ``compare three papers'' task allows each paper's summary to be extracted by a separate agent simultaneously ($F$ is high); a ``compile, test, deploy in strict sequence'' task forces most steps to execute serially ($F$ is low). The domain of $F$ is $[0, 1]$.

\item $N$ (number of sub-agents): The number of agents that can be dispatched concurrently, analogous to the number of processors in a distributed system. For instance, Codex supports creating multiple sub-agents for complex tasks \cite{openai_codex_overview_2026}, and Claude Code supports parallel sub-task processing through its sub-agent mechanism \cite{anthropic_claudecode_subagents_2026}.

\item $E$ (orchestration efficiency): \textbf{The overhead factor incurred in coordinating multiple agents,} $E \in (0, 1]$. $E$ captures the additional costs of multi-agent coordination: context synchronization (agents must share intermediate results), result merging (outputs from multiple agents must be integrated), conflict detection (multiple agents may modify the same resource), and permission negotiation (each agent's permissions must be independently authorized). $E = 1$ indicates zero orchestration overhead; $E < 1$ indicates that overhead exists.
\end{itemize}

\subsubsection{Derivation}

The derivation parallels that of Heuristic~I but introduces the orchestration efficiency factor.

\paragraph{Step~1: Decompose execution time.}
Let $T_{\mathrm{total}}$ denote the execution time on a single agent. The task decomposes into a parallelizable fraction $F$ and a serial fraction $(1-F)$.

\paragraph{Step~2: Compute the time for each component.}
\begin{itemize}[nosep]
\item Serial component: $(1-F) \cdot T_{\mathrm{total}}$ (cannot be accelerated).
\item Parallel component: distributed across $N$ agents but subject to orchestration overhead $E$, so the effective speedup is $N \cdot E$ rather than $N$. The time is $F \cdot T_{\mathrm{total}} / (N \cdot E)$.
\end{itemize}

\paragraph{Step~3: Aggregate and compute the speedup.}
\begin{equation}
T = (1-F) \cdot T_{\mathrm{total}} + \frac{F \cdot T_{\mathrm{total}}}{N \cdot E}
\end{equation}
\begin{equation}
S_{\mathrm{agent}} = \frac{T_{\mathrm{total}}}{T} = \frac{1}{(1-F) + \frac{F}{N \cdot E}}
\end{equation}

\paragraph{Step~4: Verify degeneration.}
When $E = 1$ (no orchestration overhead), the formula degenerates to the standard Amdahl's Law, confirming the derivation's consistency.

\subsubsection{Limit Behavior}

\begin{itemize}[nosep]
\item When $N \to \infty$: $S_{\mathrm{agent}} \to 1/(1-F)$. The speedup ceiling is determined entirely by the serial fraction: no matter how many agents are added, the serial portion cannot be parallelized.

\item When $F \to 1$ (nearly everything is parallelizable): $S_{\mathrm{agent}} \to N \cdot E$. Speedup is governed by the agent count and orchestration efficiency.

\item When $E \to 0$ (extreme orchestration overhead): $F/(NE) \to \infty$, and therefore $S_{\mathrm{agent}} \to 0$. If the coordination overhead between agents swallows all the gains from parallelism, a multi-agent system can perform \emph{worse} than a single agent, showing that \textbf{orchestration efficiency is the lifeline of any multi-agent architecture}.
\end{itemize}

\subsubsection{Numerical Example}

Consider a code-review task with $F = 0.8$ (80\% of the review work can be distributed across independent files) and $E = 0.8$ (20\% orchestration overhead). These values lie on the $E=0.8$ curve plotted in Figure~\ref{fig_law_curves}(c):

\begin{itemize}[nosep]
\item $N = 2$: $S = 1/(0.2 + 0.8/(2 \times 0.8)) = 1/(0.2 + 0.50) = 1.43\times$.
\item $N = 4$: $S = 1/(0.2 + 0.8/(4 \times 0.8)) = 1/(0.2 + 0.25) = 2.22\times$.
\item $N = 8$: $S = 1/(0.2 + 0.8/(8 \times 0.8)) = 1/(0.2 + 0.125) = 3.08\times$.
\item $N = 16$: $S = 1/(0.2 + 0.8/(16 \times 0.8)) = 1/(0.2 + 0.0625) = 3.81\times$.
\end{itemize}

Increasing the agent count from 2 to 4 yields a $55\%$ additional speedup ($2.22/1.43$), whereas increasing from 8 to 16 yields only $24\%$ ($3.81/3.08$), exhibiting \textbf{diminishing returns}, the hallmark of Amdahl-style scaling.

\paragraph{A qualitative industry signal: the Fountain system.}
The Fountain Copilot system reported by Anthropic uses a central orchestration agent coordinating three specialized sub-agents for candidate screening, document generation, and sentiment analysis, compressing a full recruiting-center configuration from over a week to under 72 hours~\cite{anthropic2026agenticreport}. We cite this only as a \emph{qualitative} signal that multi-agent pipelines deliver real aggregate speedups, \emph{not} as a calibration of the heuristic: with three free parameters ($F$, $N$, $E$) and a single observed end-to-end ratio, the model is non-identifiable, and back-solving for one triplet is unfalsifiable storytelling. A genuine calibration would require independently measured $F$ and $E$ on a controlled workload.

\subsubsection{Comparison with Classical Amdahl's Law}

Heuristic~III differs from Amdahl's Law in the introduction of the orchestration efficiency $E$:
\begin{equation}
S_{\mathrm{Amdahl}} = \frac{1}{(1-F) + F/N} \quad \text{vs.} \quad S_{\mathrm{agent}} = \frac{1}{(1-F) + F/(NE)}
\end{equation}

The presence of $E$ means the speedup of an agent system is \textbf{always strictly below} the classical Amdahl prediction (for $E < 1$). The reason is fundamental: in classical parallel computing, inter-processor communication overhead can often be mitigated through hardware (high-bandwidth interconnects) and algorithms (communication-avoiding designs). In agent systems, orchestration overhead encompasses not only communication costs but also \textbf{the inherent probabilistic nature of LLM inference}, because agents coordinate through natural language, which is inherently less precise and more ambiguous than binary protocols.

\subsubsection{Design Implications}

Heuristic~III carries three direct design implications:

\textbf{(1) Naively increasing the agent count does not guarantee proportional gains.} When $E$ is low (high coordination overhead), increasing $N$ yields diminishing returns. This explains why simple ``stack more agents'' strategies often underperform expectations.

\textbf{(2) Prioritize improving $F$ and $E$.} The return on investment is far higher for increasing $F$ (better task decomposition, converting serial steps into parallelizable sub-tasks) and $E$ (reducing orchestration overhead through standardized inter-agent communication protocols such as A2A \cite{agentprotocols2025}) than for simply increasing $N$.

\textbf{(3) A cost-defined agent-count knee exists.} Although $S_{\mathrm{agent}}$ is monotonically increasing in $N$, the \emph{marginal} speedup is
\begin{equation}
\frac{\partial S_{\mathrm{agent}}}{\partial N} = \frac{F}{E\,N^{2}\,g(N)^{2}}, \qquad g(N) = (1-F) + \frac{F}{EN},
\end{equation}
which decays as $1/N^{2}$ for large $N$ (where $g(N)\to 1-F$). If each additional agent consumes inference resources at marginal cost $c$ and the speedup--cost exchange rate is $\rho$, the net utility $U(N) = S_{\mathrm{agent}} - \rho c N$ is maximized at the $N^{*}$ satisfying $\partial S/\partial N = \rho c$, i.e.\ approximately
\begin{equation}
N^{*} \approx \sqrt{\frac{F}{E\,\rho c\,(1-F)^{2}}}.
\end{equation}
This replaces the earlier hand-waved claim with an explicit (if idealized) closed form: the knee grows with $\sqrt{F}$ and shrinks with $\sqrt{E}$, $\rho c$, and the serial fraction $(1-F)$---precisely the intuition that highly serial, expensive, or already well-orchestrated systems hit diminishing returns sooner. As an illustration, taking the code-review scenario above ($F=0.8$, $E=0.8$) with a unit speedup--cost exchange ($\rho c = 1$) gives $N^{*} \approx \sqrt{0.8/(0.8 \cdot 1 \cdot 0.04)} = \sqrt{25} = 5$; quadrupling the per-agent cost ($\rho c = 4$) halves the knee to $N^{*} \approx 2.5$. These are order-of-magnitude guides to where adding agents stops paying off, not predictions.

\subsubsection{Illustration}

Illustrating Heuristic~III requires real performance data from multi-agent systems.

\begin{enumerate}[nosep]
\item \textbf{AutoGen.} Wu et al.\ report experimental data for multi-agent systems showing that as the agent count increases from 2 to 8, the rate of reduction in task completion time progressively diminishes \cite{wu2023autogen}, consistent with Heuristic~III's prediction that $S_{\mathrm{agent}}$ exhibits diminishing growth as $N$ increases.

\item \textbf{GTA-2 benchmark.} Wang et al.\ report that the open-workflow success rate of top-performing models on the GTA-2 benchmark is only 14.39\% \cite{gta2_2025}. This suggests that current systems suffer from both low $F$ (inadequate task-decomposition capability leaves most work serial) and low $E$ (poor inter-agent coordination efficiency).

\item \textbf{Language Model Teams.} A recent study models multi-agent LLM teams as distributed systems and analyzes their speedup directly through an Amdahl's Law framework \cite{langmodelteams2026}. The authors find that highly parallelizable tasks (e.g., independent document summarization) achieve near-linear speedup, whereas strongly dependent tasks (e.g., multi-step reasoning) exhibit a speedup approaching~1, fully consistent with the role of $F$ in Heuristic~III.
\end{enumerate}

\subsection{A Unifying Perspective}
\label{sec:laws-summary}

Table~\ref{tab:empirical} summarizes the core formula, supporting evidence, and open questions for each of the three heuristics.

\begin{table}[h]
    \centering
    \footnotesize
    \caption{Illustrative parameter ranges and supporting data for the three design heuristics.}
    \label{tab:empirical}
    \begin{tabularx}{\textwidth}{lY Y Y}
    \toprule
    & Heur.~I (Semantic Locality) & Heur.~II (Context Budget) & Heur.~III (Agent Speedup) \\
    \midrule
    Core formula & $S = 1/((1-H)+H/\alpha)$ & $W_{\mathrm{eff}} = C\cdot\bar{\beta} \leq C$ & $S = 1/((1-F)+F/(NE))$ \\
    Classical analogue & Amdahl's Law ($f \leftrightarrow H$, $p \leftrightarrow \alpha$) & Denning working-set model ($W(\tau) \leftrightarrow W_{\mathrm{eff}}$) & Amdahl + orchestration overhead ($E < 1$) \\
    Supporting data & vLLM: $2$--$4\times$ throughput gain (fragmentation \& continuous batching, not $H$-reuse) \cite{kwon2023pagedattention}; Prompt Cache: $8$--$60\times$ TTFT reduction ($H \approx 0.89$--$\approx 1.00$) \cite{gim2024promptcache} & RULER: accuracy drops from $>$95\% at 4K to $<$70\% at 128K \cite{ruler2024}; LongBench~v2: faster decay for deep reasoning \cite{longbenchv2_2024} & AutoGen: diminishing returns from 2 to 8 agents \cite{wu2023autogen}; GTA-2: open-workflow success only 14.39\% \cite{gta2_2025} \\
    Open questions & Distribution of $H$ under semantic caching & Precise functional form of $\beta(L)$ & Quantitative relationship between $E$ and task complexity \\
    \bottomrule
    \end{tabularx}
    \end{table}

Figure~\ref{fig_law_curves} plots the characteristic curves of the three heuristics.
\input{figures/fig_law_curves.tex}

The three heuristics share a common theme: each sketches a \textbf{performance ceiling} in model-native computing and points toward strategies for approaching it. Heuristic~I characterizes the limit of KV-cache reuse. Heuristic~II frames why naively extending the context window yields sub-linear gains in the effective working set. Heuristic~III delineates the diminishing returns of scaling the agent count. Together they form the \emph{organizing intuition} of the ICA model---not a validated quantitative backbone, but a shared back-of-envelope language that turns the qualitative design principles of Section~\ref{sec:icam} into order-of-magnitude engineering targets awaiting systematic empirical calibration.

%% file: figures/fig_beta_decay.tex
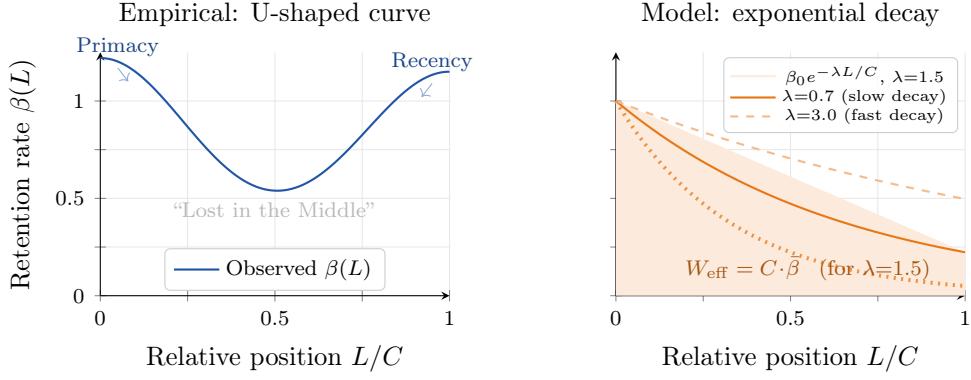
\begin{figure}[htbp]
\centering
\begin{tikzpicture}
\begin{groupplot}[
    group style={
        group size=2 by 1,
        horizontal sep=2.2cm,
    },
    width=6.2cm,
    height=4.8cm,
    grid=major,
    grid style={gray!18},
    tick label style={font=\scriptsize},
    label style={font=\small},
    every axis plot/.append style={thick},
    axis lines=left,
    xmin=0, xmax=1,
    ymin=0, ymax=1.25,
    xtick={0,0.25,0.5,0.75,1.0},
    xticklabels={0,,0.5,,1},
    ytick={0,0.25,0.5,0.75,1.0},
    yticklabels={0,,0.5,,1},
    enlargelimits=false,
    clip=false,
]

\nextgroupplot[
    xlabel={Relative position $L/C$},
    ylabel={Retention rate $\beta(L)$},
    title={\small Empirical: U-shaped curve},
    title style={font=\small},
    legend style={
        font=\scriptsize,
        at={(0.50,0.02)},
        anchor=south,
        draw=gray!40,
        fill=white,
        rounded corners=2pt,
    },
]

\addplot[classical, thick, smooth, domain=0:1, samples=100]
    { 0.92*exp(-8*(x-0)^2) + 0.85*exp(-8*(x-1)^2) + 0.30 };
\addlegendentry{Observed $\beta(L)$}

\node[font=\scriptsize, text=classical!80!black, anchor=south]
    at (axis cs:0.05, 1.18) {Primacy};
\draw[classical!50, ->, thin]
    (axis cs:0.05, 1.17) -- (axis cs:0.08, 1.10);

\node[font=\scriptsize, text=classical!80!black, anchor=south]
    at (axis cs:0.95, 1.10) {Recency};
\draw[classical!50, ->, thin]
    (axis cs:0.95, 1.09) -- (axis cs:0.92, 1.02);

\node[font=\scriptsize, text=gray!55]
    at (axis cs:0.50, 0.44) {``Lost in the Middle''};

\nextgroupplot[
    xlabel={Relative position $L/C$},
    ylabel={},
    title={\small Model: exponential decay},
    title style={font=\small},
    yticklabels={},
    legend style={
        font=\tiny,
        at={(0.98,0.98)},
        anchor=north east,
        draw=gray!40,
        fill=white,
        fill opacity=0.85,
        text opacity=1,
        rounded corners=2pt,
        inner xsep=2pt,
        inner ysep=1pt,
        row sep=-3pt,
    },
]

\addplot[modelnative!15, fill=modelnative!15, draw=none]
    coordinates {(0,0) (0,1) (1,{exp(-1.5)}) (1,0)} -- cycle;

\addplot[modelnative, thick, smooth, domain=0:1, samples=80]
    { exp(-1.5*x) };
\addlegendentry{$\beta_0 e^{-\lambda L/C}$, $\lambda{=}1.5$}

\addplot[modelnative!50, thick, smooth, domain=0:1, samples=80, dashed]
    { exp(-0.7*x) };
\addlegendentry{$\lambda{=}0.7$ (slow decay)}

\addplot[modelnative!80, thick, smooth, domain=0:1, samples=80, dotted, line width=1.5pt]
    { exp(-3.0*x) };
\addlegendentry{$\lambda{=}3.0$ (fast decay)}

\node[font=\scriptsize, text=modelnative!70!black, anchor=center]
    at (axis cs:0.55, 0.14) {$W_{\mathrm{eff}} = C\!\cdot\!\bar{\beta}$ \, (for $\lambda{=}1.5$)};

\end{groupplot}
\end{tikzpicture}
\caption{Two perspectives on the attention retention rate $\beta(L)$. \textbf{Left:} Empirically observed U-shaped curve (Liu et al.~\cite{liu2023lostmiddle}): information at the beginning and end of the context is recalled well, while the middle region suffers a sharp drop in retention. \textbf{Right:} Exponential decay model $\beta(L)\approx\beta_0 e^{-\lambda L/C}$ adopted in Heuristic~II for analytical tractability; the shaded area represents the effective working set $W_{\mathrm{eff}}=C\cdot\bar{\beta}$.}
\label{fig_beta_decay}
\end{figure}

%% file: figures/fig_law_curves.tex
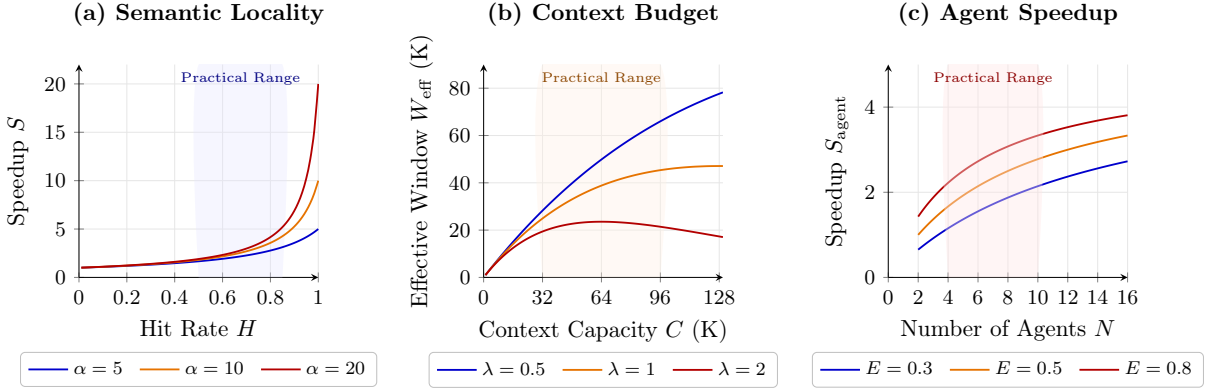
\begin{figure}[h]
\centering
\resizebox{\textwidth}{!}{%
\begin{tikzpicture}
\small
\begin{groupplot}[
    group style={
        group size=3 by 1,
        horizontal sep=2.5cm,
    },
    width=5.2cm,
    height=4.8cm,
    grid=major,
    grid style={gray!20},
    tick label style={font=\footnotesize},
    label style={font=\small},
    legend style={
        font=\scriptsize,
        at={(0.5,-0.35)},
        anchor=north,
        legend columns=3,
        /tikz/every even column/.append style={column sep=5pt},
        rounded corners=2pt,
        draw=gray!50,
    },
    every axis plot/.append style={thick, smooth, no marks},
    axis lines=left,
    enlargelimits=false,
    clip=true,
]

\nextgroupplot[
    xlabel={Hit Rate $H$},
    ylabel={Speedup $S$},
    xmin=0, xmax=1,
    ymin=0, ymax=22,
    xtick={0,0.2,0.4,0.6,0.8,1.0},
    title style={font=\small\bfseries, at={(0.5,1.08)}},
    title={(a) Semantic Locality},
]
\addplot[blue!10, fill=blue!10, fill opacity=0.35, draw=none, forget plot]
    coordinates {(0.5,0)(0.5,22)(0.85,22)(0.85,0)};
\node[font=\tiny, blue!50!black, anchor=north] at (axis cs:0.675,22) {Practical Range};
\addplot[blue!80!black, domain=0.01:1, samples=60]
    {1/((1-x)+x/5)};
\addlegendentry{$\alpha=5$}
\addplot[orange!90!black, domain=0.01:1, samples=60]
    {1/((1-x)+x/10)};
\addlegendentry{$\alpha=10$}
\addplot[red!70!black, domain=0.01:1, samples=60]
    {1/((1-x)+x/20)};
\addlegendentry{$\alpha=20$}

\nextgroupplot[
    xlabel={Context Capacity $C$ (K)},
    ylabel={Effective Window $W_{\mathrm{eff}}$ (K)},
    xmin=0, xmax=130,
    ymin=0, ymax=90,
    xtick={0,32,64,96,128},
    ytick={0,20,40,60,80},
    title style={font=\small\bfseries, at={(0.5,1.08)}},
    title={(b) Context Budget},
]
\addplot[orange!10, fill=orange!10, fill opacity=0.40, draw=none, forget plot]
    coordinates {(32,0)(32,90)(96,90)(96,0)};
\node[font=\tiny, orange!55!black, anchor=north] at (axis cs:64,90) {Practical Range};
\addplot[blue!80!black, domain=1:130, samples=60]
    {x * exp(-0.5*x/128)};
\addlegendentry{$\lambda=0.5$}
\addplot[orange!90!black, domain=1:130, samples=60]
    {x * exp(-1.0*x/128)};
\addlegendentry{$\lambda=1$}
\addplot[red!70!black, domain=1:130, samples=60]
    {x * exp(-2.0*x/128)};
\addlegendentry{$\lambda=2$}

\nextgroupplot[
    xlabel={Number of Agents $N$},
    ylabel={Speedup $S_{\mathrm{agent}}$},
    xmin=0, xmax=16,
    ymin=0, ymax=5,
    xtick={0,2,4,6,8,10,12,14,16},
    title style={font=\small\bfseries, at={(0.5,1.08)}},
    title={(c) Agent Speedup},
]
\addplot[blue!80!black, domain=2:16, samples=60]
    {1/(0.2+0.8/(x*0.3))};
\addlegendentry{$E=0.3$}
\addplot[orange!90!black, domain=2:16, samples=60]
    {1/(0.2+0.8/(x*0.5))};
\addlegendentry{$E=0.5$}
\addplot[red!70!black, domain=2:16, samples=60]
    {1/(0.2+0.8/(x*0.8))};
\addlegendentry{$E=0.8$}
\addplot[red!10, fill=red!10, fill opacity=0.40, draw=none, forget plot]
    coordinates {(4,0)(4,5)(10,5)(10,0)};
\node[font=\tiny, red!55!black, anchor=north] at (axis cs:7,5) {Practical Range};

\end{groupplot}
\end{tikzpicture}}%
\caption{Characteristic curves of three design heuristics. (a) Semantic Locality: $S=1/((1-H)+H/\alpha)$. (b) Context Budget: $W_{\mathrm{eff}}=C\cdot\bar{\beta}$ (representative exponential-decay envelope shown). (c) Agent Speedup: $S=1/((1-F)+F/(NE))$ with $F=0.8$. Shaded areas indicate typical operating ranges.}
\label{fig_law_curves}
\end{figure}

%% file: en_sections/section_agent_evolution.tex
\section{Agent Framework Evolution: From ReAct to the Model-Native OS}
\label{sec:agentevolution}

An \textit{agent framework} is a software system built around a large language model that enables autonomous planning, tool invocation, state management, and the completion of complex multi-step tasks.
Between 2022 and 2025, these frameworks have undergone a systematic evolution that closely mirrors the historical progression of operating systems: from single-task batch processing, through multiprogramming and time-sharing, to protected and governed execution environments.
This section traces five distinct generations of this evolution, examining each along three dimensions: its \textbf{core architectural contribution}, its \textbf{representative systems and their concrete mechanisms}, and its \textbf{structural correspondence to classical operating-system history}.

Figure~\ref{fig_agent_timeline} presents a timeline of the five-generation evolution.
\input{figures/fig_agent_timeline.tex}

\subsection{Generation I: Establishing the Execution Paradigm (2022--2023)}

\paragraph{Core contribution.}
The first generation of agent frameworks established the interleaved \textit{Thought--Action--Observation} loop as the fundamental execution paradigm.
Prior to this, LLM usage followed a single-turn, question-answer pattern: the user supplied a prompt, and the model returned a complete response, analogous to submitting a batch job and waiting for its output.
Generation~I broke this limitation by placing the model inside an active loop in which it could \textbf{reason} (analyze the current state and decide on the next step), \textbf{act} (invoke a tool or execute an operation), and \textbf{observe} (read the tool's return value), before entering the next iteration.

\paragraph{Representative systems.}
ReAct~\cite{yao2023react} is the foundational work of this generation.
Its core mechanism injects a ``Thought'' field into the prompt template, requiring the model to emit an explicit reasoning trace before every tool invocation.
When asked, for example, ``What is Shakespeare's longest play?'', the model does not guess the answer outright; instead, it first outputs ``I need to search for the lengths of all Shakespeare's plays,'' calls a search tool, and then reasons over the results to arrive at the final answer.
This ``think before you act'' discipline, while seemingly simple, effectively mitigates the \textit{hallucinated tool calls} that arise when models generate tool invocations without intermediate deliberation.
Toolformer~\cite{schick2023toolformer} approached the problem from a complementary angle: it taught the model \textbf{when and how} to invoke tools by injecting tool-call demonstrations into the training corpus, thereby acquiring tool usage as a self-supervised capability.

\paragraph{OS analogy.}
Generation~I frameworks correspond to the \textbf{single-task batch-processing systems} of the 1950s, such as the Fortran Monitor System on the IBM~704~\cite{ostep_2023}.
The system handles one task at a time, to completion, before starting the next; there is no concurrency, no interrupts, and no persistent state across tasks.
The ReAct loop is the simplest possible ``fetch--decode--execute'' pipeline, with each iteration analogous to a single instruction cycle.
There is no resource management whatsoever: no task scheduler (only one task exists), no memory management (the context window \textit{is} the entire ``memory''), and no I/O abstraction (tool calls are hard-coded).
Figure~\ref{fig:ae:gen1} illustrates this structural correspondence.
\input{figures/fig_ae_gen1}

\subsection{Generation II: Persistent Multi-Step Agents (2023)}

\paragraph{Core contribution.}
The defining advance of Generation~II is \textbf{persistence} and \textbf{autonomous goal decomposition}.
In Generation~I, each user query triggers an independent reasoning--action loop, and all intermediate state vanishes upon completion.
Generation~II frameworks allow agents to run continuously, autonomously decompose high-level goals into sequences of subtasks, maintain intermediate state across steps, and pass information between subtasks, in effect endowing the agent with a \textit{working memory}.

\paragraph{Representative systems.}
AutoGPT~\cite{autogpt_repo_2026} is the most widely recognized effort of this generation.
Given a high-level objective such as ``survey recent advances in quantum computing and write a report,'' AutoGPT autonomously decomposes it into subtasks (searching papers, extracting key findings, organizing an outline, and drafting the report) and executes them in sequence.
It introduces long-term memory backed by a vector database to store intermediate results across steps, enabling the agent to ``remember'' what it has already done.
However, AutoGPT also exposed Generation~II's central weakness: the absence of failure recovery.
Once a subtask goes wrong, the entire chain tends to spiral into infinite loops or accumulate compounding errors.

Voyager~\cite{wang2023voyager} made a critical improvement within the Minecraft environment by introducing a \textbf{skill library} (encapsulating successfully completed subtasks as reusable ``skills,'' analogous to function definitions) and an \textbf{automatic curriculum} that generates the next learning objective based on the agent's current capabilities, much like a pedagogical syllabus.
This enables the agent to accumulate experience incrementally rather than starting from scratch each time.
BabyAGI implemented a cleaner task-queue manager: maintaining a prioritized to-do list, executing tasks in priority order, and dynamically spawning new tasks based on execution results.

\paragraph{OS analogy.}
Generation~II corresponds to \textbf{multiprogramming} in the 1960s~\cite{ostep_2023}: the system begins to manage complex multi-step state, and tasks can share and pass information.
AutoGPT's task queue is a primitive ``job scheduler''; Voyager's skill library resembles an ``executable program library''; BabyAGI's priority ordering parallels early ``job scheduling policies.''
Compared with mature operating systems, however, Generation~II lacks three critical mechanisms: \textbf{isolation} (different tasks may interfere with one another), \textbf{preemptive scheduling} (a stalled task blocks the entire system), and \textbf{protection} (a failed task can corrupt the state of other tasks).
Figure~\ref{fig:ae:gen2} illustrates this correspondence.
\input{figures/fig_ae_gen2}

\subsection{Generation III: General-Purpose Runtimes (2023--2024)}

\paragraph{Core contribution.}
Generation~III frameworks systematically introduce \textbf{scheduling}, \textbf{isolation}, and \textbf{resource management}, upgrading the agent framework from a single-script automation tool into a general-purpose runtime platform.
The key indicator is that an agent is no longer merely a serial loop; it becomes a schedulable, suspendable, resumable ``process.''
Multiple agents can run concurrently without mutual interference, and the system explicitly manages computational resources such as inference budgets, context space, and tool-call rate limits.

\paragraph{Representative systems.}
AIOS~\cite{mei2024aios} constructed a complete agent operating system kernel organized into six subsystems: an Agent Scheduler for concurrent multi-agent scheduling, a Context Manager for context isolation and sharing across agents, a Memory Manager for hierarchical memory, a Tool Manager for tool registration and permission control, an I/O Manager for input--output mediation, and a Knowledge Manager for knowledge-base integration.
Each agent is encapsulated as a ``process'' with its own state space and resource budget.

The OpenAI Agents SDK~\cite{openai_codex_agentsdk_2026} pursued a more engineering-driven approach, packaging orchestration, state management, approval workflows, agent handoffs, and full-chain tracing into a code-first runtime.
Developers define agent behavioral specifications, permission boundaries, and collaboration protocols in Python, rather than encoding them implicitly through prompt templates.

AutoGen~\cite{wu2023autogen} proposed a multi-agent conversational framework in which multiple agents collaborate on complex tasks through structured message passing, each assuming a distinct role (e.g., programmer, tester, project manager), analogous to multi-process cooperation in a conventional operating system.

\paragraph{OS analogy.}
Generation~III corresponds to the \textbf{time-sharing operating systems} of the 1960s--70s, such as Multics and Unix~\cite{ostep_2023}.
The core breakthrough of time-sharing OSes was providing each user (process) with an independent address space and fair CPU time slices, enabling multiple users to safely share a single computer.
AIOS's Agent Scheduler corresponds to the process scheduler of a time-sharing OS; its Context Manager maps to the address isolation provided by the memory management unit (MMU); and the Tool Manager's permission controls parallel Unix's file permission system.
The Agents SDK's tracing mechanism mirrors OS-level system-call auditing via \texttt{strace} or \texttt{auditd}.
Figure~\ref{fig:ae:gen3} illustrates this structural correspondence.
\input{figures/fig_ae_gen3}

\subsection{Generation IV: Engineering-Grade Operating Systems (2024--2025)}

\paragraph{Core contribution.}
The hallmark of Generation~IV is \textbf{deep integration with real-world software engineering environments}.
Whereas the first three generations validated their concepts primarily in laboratory or constrained settings, Generation~IV confronts real codebases, real terminal environments, real file systems, and real external services.
This forces the system to address problems that earlier generations could sidestep: large repositories whose context far exceeds the model's window, tool calls that may produce irreversible side effects, concurrent edits that can introduce conflicts, and real projects that demand permission controls and audit trails.

\paragraph{Representative systems.}
OpenAI Codex~\cite{openai_codex_overview_2026} is a cloud-based coding agent that runs inside an isolated sandbox: it checks out the repository in a dedicated container, performs reads, writes, and command execution, and submits a diff for human review upon completion.
Codex introduced a sub-agent mechanism~\cite{openai_codex_overview_2026}: the primary agent can dispatch subtasks to specialized sub-agents, each running in its own sandbox, with results merged after approval, analogous to Unix's \texttt{fork()} creating child processes.
Codex also supports skills packages~\cite{openai_codex_skills_2026} and project-level instruction files (AGENTS.md)~\cite{openai_codex_agents_md_2026}, enabling reusable configurations across projects.

Claude Code~\cite{anthropic_claudecode_overview_2026} adopts a complementary design philosophy: it runs in the developer's local terminal, connects to external tools via the Model Context Protocol (MCP)~\cite{mcp_intro_2026}, defaults to read-only access (unless the user explicitly authorizes writes), and supports custom hooks that intercept and audit every tool call~\cite{anthropic_claudecode_security_2026}.
Claude Code's memory system~\cite{anthropic_claudecode_memory_2026} distinguishes between project-level memory (\texttt{CLAUDE.md}, analogous to a system configuration file) and user-level memory (cross-project personal preferences, analogous to dotfiles in the user's home directory).

OpenHands~\cite{openhands_paper_2025} provides an open agent platform supporting multiple model backends and sandbox environments; its OpenHands Index has become one of the standard benchmarks for evaluating coding agents.
OSWorld~\cite{osworld2024} advanced this direction from an evaluation standpoint by constructing real operating-system environments (complete with file managers, browsers, and terminals) and requiring agents to complete open-ended tasks within authentic GUI settings, thereby exposing agents' limitations when confronted with real-world complexity.
Devin~\cite{devin2024cognition} further exemplified Generation~IV's engineering orientation: as an ``AI software engineer,'' it integrates a code editor, browser, and terminal, and is capable of autonomous planning, coding, debugging, and deployment.

\paragraph{OS analogy.}
Generation~IV corresponds to the \textbf{modern operating systems} of the 1980s--90s, such as Unix System~V, early Linux, and Windows~NT~\cite{ostep_2023}.
The defining characteristic of modern OSes is their ability to handle real-world complexity: multitasking schedulers (e.g., the Completely Fair Scheduler~\cite{linux_cfs_2026}), virtual memory with demand paging, network protocol stacks, device driver frameworks, and security auditing.
Codex's sandbox maps to OS-level process isolation (each sub-agent runs in an independent container, akin to \texttt{chroot} or container technology); Claude Code's hooks parallel Linux Security Modules (LSM) security hooks; Codex's approval mechanism corresponds to Unix's \texttt{sudo} privilege escalation; and OpenHands's multi-model backend serves the role of a hardware abstraction layer (HAL).
Figure~\ref{fig:ae:gen4} illustrates this structural correspondence.
\input{figures/fig_ae_gen4}

\subsection{Generation V: Governed Operating Systems (2025--Present)}

\paragraph{Core contribution.}
The defining breakthrough of Generation~V is the elevation of security and governance from post-hoc patches to \textbf{first-principles architectural design}.
In the first four generations, security measures were additive: first make the agent functional, then bolt on sandboxes, approval gates, and audit logs.
Generation~V inverts this ordering: define security boundaries and governance rules first, then empower the agent to act within those boundaries.
This paradigm shift parallels the historical transition of operating systems from ``functional'' to ``trustworthy'': just as Multics introduced ring protection and mandatory access control in the 1960s, but it was not until the widespread adoption of SELinux and capability-based security that protection became an organic component of OS architecture~\cite{ostep_2023}.

\paragraph{Representative systems.}
ArbiterOS~\cite{arbiteros2025} explicitly articulates a layered architecture of ``probabilistic CPU + deterministic governor/kernel'': the underlying LLM serves as the probabilistic processor that determines \textit{what the agent can do}, while the governor above acts as the deterministic controller that decides \textit{what the agent should do}.
The governor constrains the agent's behavioral boundaries through policy specifications, analogous to how an OS kernel defines a process's capability envelope via system-call tables and permission bits.

AgentOS~\cite{agentos2025} introduces a reasoning kernel governed by structured operating-system logic, building system-level intelligent behavior upward from token-level context management.
Its central thesis is that an agent's reasoning process should not be an uncontrolled ``black-box loop,'' but rather managed through structured abstractions and interfaces, much like an OS kernel.

UFO$^2$~\cite{ufo2_2025}, targeting desktop environments, organizes the Desktop AgentOS into three tiers: a HostAgent as the top-level coordinator (analogous to the \texttt{init} process), AppAgents (one per application, analogous to service processes), and a GUI--API action layer at the bottom (analogous to the driver layer).
This stratification enables independent management and auditing of agent operations across different applications.

Aura~\cite{aura2025} realizes a security-first agent OS architecture on mobile devices, comprising a System Agent (with highest privilege), Sandboxed App Agents (one per application, akin to Android's application sandbox), and an Agent Kernel responsible for permission arbitration and resource allocation.
Aura emphasizes the unique risks of mobile environments: semantic input pollution (hijacking agent behavior through malicious inputs), memory taint propagation (compromised memory affecting all subsequent operations), and cross-application permission leakage.

CaMeL~\cite{debenedetti2025camel} derives its security model from first principles, proposing a capability-based model to defend against prompt-injection attacks.
Its core insight draws directly from capability-based security in operating systems: an agent cannot directly execute any operation with side effects, but must obtain a temporary capability token from a deterministic capability manager.
Each token authorizes only a single, specific operation (e.g., ``read-only access to file~X'') rather than granting blanket permissions.
This mirrors the evolution in Unix from the coarse-grained root/non-root privilege model to the fine-grained POSIX capabilities and SELinux mandatory access controls.

IronClaw~\cite{ironclaw2025} systematically argues why AI agents require operating-system-grade protection.
It observes that traditional software security assumes deterministic code with predictable behavior, whereas agent behavior is driven by probabilistic models.
The attack surface thus expands from ``code vulnerabilities'' to ``semantic vulnerabilities'': an adversary can hijack an agent's behavior through natural-language prompts without exploiting any code defect.

The OWASP \textbf{Agentic Applications Top~10}~\cite{owasp_agentic2026}, published in late 2025, further systematizes these threats into ten risk categories, including Agent Goal Hijack, Tool Misuse \& Exploitation, and Identity \& Privilege Abuse, providing a taxonomic foundation for the security design of Generation~V frameworks.

\paragraph{OS analogy.}
Generation~V corresponds to the inflection point at which operating systems transitioned from ``functional'' to ``trustworthy.''
In classical OS history, Unix~V6 (1975) was a functional but insecure system: it lacked memory protection (any process could read or write any memory address) and privilege separation (the root user held unrestricted power).
As networking and multi-user environments developed, security evolved from an ``optional feature'' to a ``foundational requirement,'' giving rise to Multics ring protection, SELinux mandatory access control, and formal verification of modern microkernels such as seL4~\cite{ostep_2023}.
Generation~V agent frameworks are undergoing the same transition: security is no longer merely ``add a sandbox,'' but requires a top-to-bottom redesign of the agent's permission model, audit mechanisms, and failure-recovery strategies at the architectural level.
Figure~\ref{fig:ae:gen5} illustrates this structural correspondence.
\input{figures/fig_ae_gen5}

Figure~\ref{fig_gen_matrix} summarises the capability maturity of each generation across six architectural dimensions.
\input{figures/fig_gen_matrix.tex}

\subsection{The Five Generations Through the Lens of Heuristic~III}

Viewed through the agent speedup heuristic (Heuristic~III) introduced in Section~\ref{sec:laws}, the five generations of evolution correspond to a progressive increase in $F$ (the parallelizable fraction) and $E$ (the orchestration efficiency).

Recall the formal statement of Heuristic~III:
\begin{equation}
\label{eq:agent_law_review}
S_{\mathrm{agent}} = \frac{1}{(1-F) + \dfrac{F}{N \cdot E}}
\end{equation}
where $F \in [0,1]$ denotes the fraction of the task that can be decomposed for parallel execution, $N$ is the number of schedulable agents, and $E \in (0,1]$ is the orchestration efficiency ($E=1$ means zero overhead; $E\to 0$ means overhead consumes all parallel benefit).

\paragraph{Quantitative trajectory across generations.}
Figure~\ref{fig:gen_trajectory} plots the estimated values of $F$, $E$, and $S_{\mathrm{agent}}$ across the five generations.

\input{figures/fig_gen_trajectory}

\textit{Generation~I (ReAct).}
$F \approx 0$ (no task decomposition; single-agent serial execution), $N=1$, $E \approx 1$ (no orchestration overhead), yielding $S_{\mathrm{agent}} \approx 1$.
This corresponds to the ``uniprocessor, no parallelism'' regime.

\textit{Generation~II (AutoGPT/Voyager).}
$F \approx 0.1$--$0.2$ (a small fraction of subtasks can be separated), $N=1$ (still a single agent), $E \approx 0.9$, yielding $S_{\mathrm{agent}} \approx 1$.
Decomposition capability begins to emerge, but with only one agent there is no opportunity for parallel speedup.

\textit{Generation~III (AIOS/Agents SDK).}
$F \approx 0.3$ (moderate decomposition capability), $N \approx 2$--$4$ (multi-agent support emerges), $E \approx 0.7$ (scheduling and context switching introduce significant overhead), yielding $S_{\mathrm{agent}} \approx 1/(0.7 + 0.3/2.8) \approx 1.27$.
Multi-agent execution begins to deliver speedup, but orchestration overhead partially cancels the gains.

\textit{Generation~IV (Codex/Claude Code).}
$F \approx 0.5$ (real-world software engineering tasks exhibit substantial decomposability), $N \approx 4$--$8$ (sub-agent concurrency), $E \approx 0.5$ (sandbox isolation, approval workflows, and state synchronization impose significant overhead), yielding $S_{\mathrm{agent}} \approx 1/(0.5 + 0.5/4) \approx 1.6$.
Speedup becomes substantial, yet orchestration overhead remains the binding constraint.

\textit{Generation~V (ArbiterOS/CaMeL/IronClaw).}
$F \approx 0.72$ (governance-first design enables broader task decomposition), $N \approx 6$ (mature multi-agent concurrency), $E \approx 0.65$ (governance overhead partially recovered through structured policy enforcement and capability tokens), yielding $S_{\mathrm{agent}} \approx 1/(0.28 + 0.72/3.9) \approx 2.1$.
The partial recovery of $E$ relative to Gen~IV reflects the shift from ad-hoc sandboxing to principled capability-based security, which reduces redundant permission checks. However, $F$ and $E$ remain below the $0.8$ threshold required for the target $3.3\times$ speedup.

The execution time horizon of Generation~IV systems is also expanding from minutes to hours.
In a striking demonstration, engineers at Rakuten tested Claude Code on the vLLM project (12.5~million lines of code), where it ran autonomously for seven hours to complete an activation-vector extraction task, achieving 99.9\% numerical accuracy~\cite{anthropic2026agenticreport}.
This kind of long-duration autonomous execution introduces new architectural challenges, including cross-hour KV-cache persistence, checkpoint/recovery protocols, and sustained resource management, that further reinforce the analogy between agent runtimes and operating systems.

\paragraph{The target: a true model-native OS.}
Achieving a genuine model-native operating system requires $F > 0.8$ and $E > 0.8$, which would yield $S_{\mathrm{agent}} > 1/(0.2 + 0.8/8) \approx 3.3$.
This target demands both high decomposability (high~$F$) \textit{and} low orchestration overhead (high~$E$), a dual requirement that no current system meets.

Recent benchmark results on GAIA~\cite{gaia2023}, $\tau$-bench~\cite{taubench2024}, and SWE-bench~\cite{swebench2023} remind us how far current systems remain from this goal.
On SWE-bench Verified, the best-performing agent systems as of late 2025 achieve a pass rate of approximately 70--75\%, meaning that in roughly one quarter of real software engineering tasks, agents still cannot correctly decompose and execute the required work.
On $\tau$-bench, agent performance during multi-turn interactions with real users is even less stable, because the ambiguity of user instructions and the dynamism of the environment further depress both $F$ and $E$.

%% file: figures/fig_agent_timeline.tex
\begin{figure}[htbp]
\centering
\includegraphics[width=0.95\textwidth]{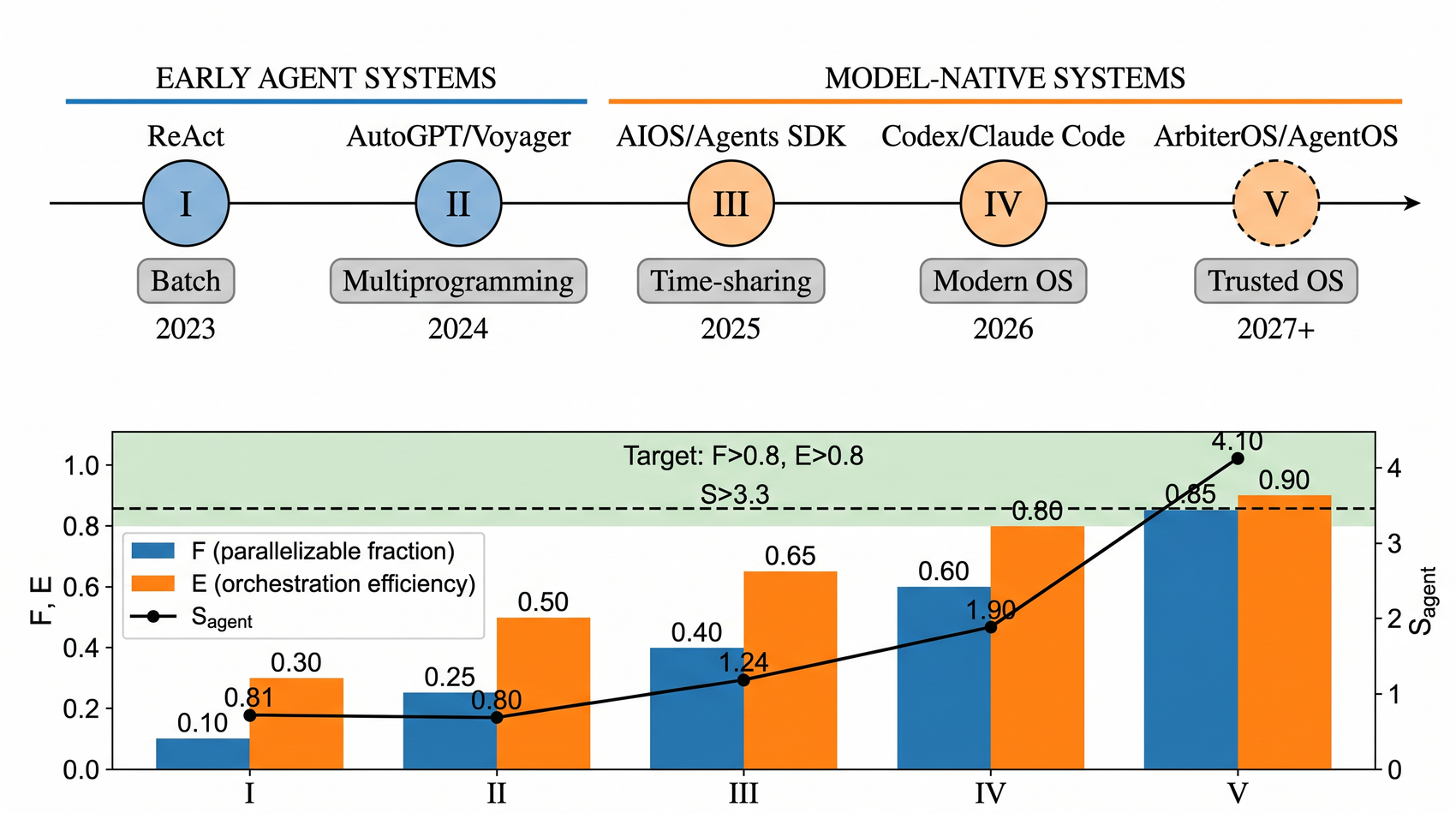}
\caption{Agent framework evolution from single-turn reasoning to model-native operating systems.
\textbf{Top:} representative frameworks with analogous OS milestones.
\textbf{Bottom:} quantitative trajectory of parallelizable fraction ($F$), orchestration efficiency ($E$), and agent speedup ($S_{\mathrm{agent}}$) per Heuristic~III; green band marks the target regime ($F{>}0.8,\;E{>}0.8$).}
\label{fig_agent_timeline}
\end{figure}

%% file: figures/fig_ae_gen1.tex
\begin{figure}[htbp]
  \centering
  \includegraphics[width=0.88\textwidth]{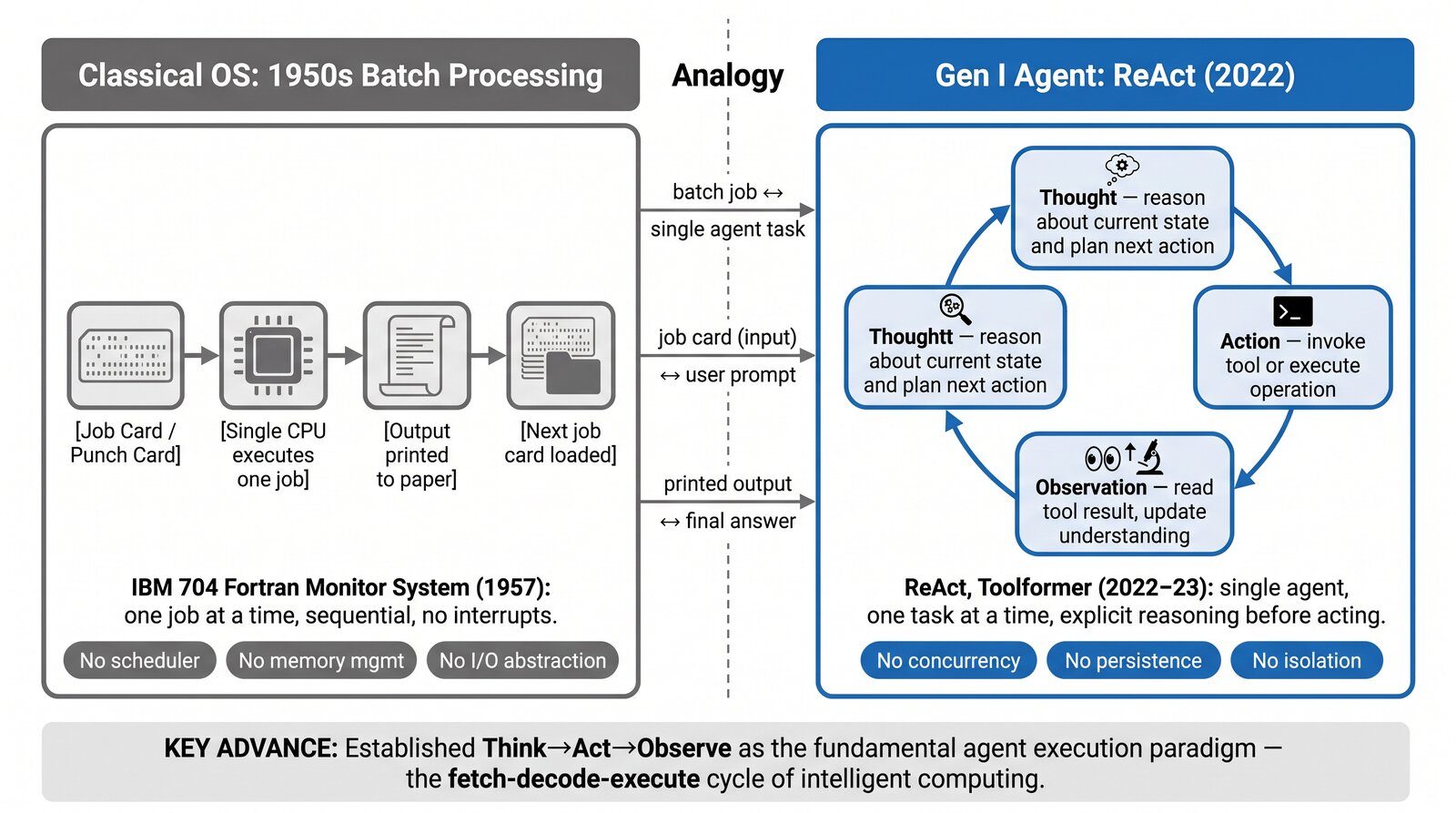}
  \caption{Generation~I agent frameworks versus 1950s batch-processing systems.
  The ReAct Think--Act--Observe loop is structurally analogous to a single-job batch system:
  one task executes to completion with no concurrency, no persistent state, and no resource isolation.
  The core advance was establishing the interleaved reasoning--action paradigm
  as the fundamental agent execution primitive.}
  \label{fig:ae:gen1}
\end{figure}

%% file: figures/fig_ae_gen2.tex
\begin{figure}[htbp]
  \centering
  \includegraphics[width=0.88\textwidth]{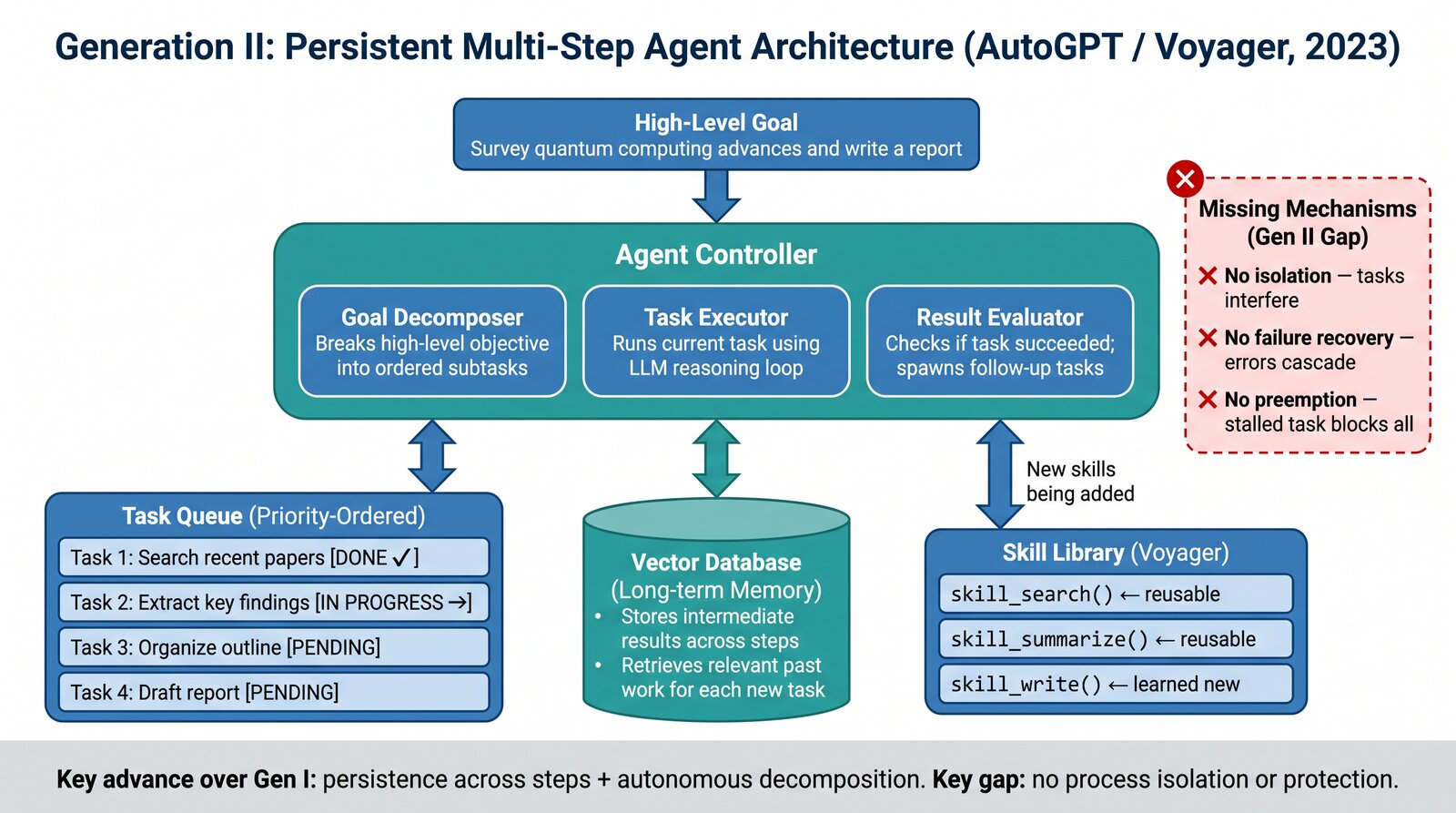}
  \caption{Generation~II agent frameworks versus 1960s multiprogramming systems.
  AutoGPT's task queue, vector-database memory, and Voyager's skill library correspond to
  a multiprogramming system's job scheduler, shared memory, and program library.
  The defining advance is persistence and autonomous goal decomposition;
  the remaining gap is the absence of isolation and failure recovery.}
  \label{fig:ae:gen2}
\end{figure}

%% file: figures/fig_ae_gen3.tex
\begin{figure}[htbp]
  \centering
  \includegraphics[width=0.88\textwidth]{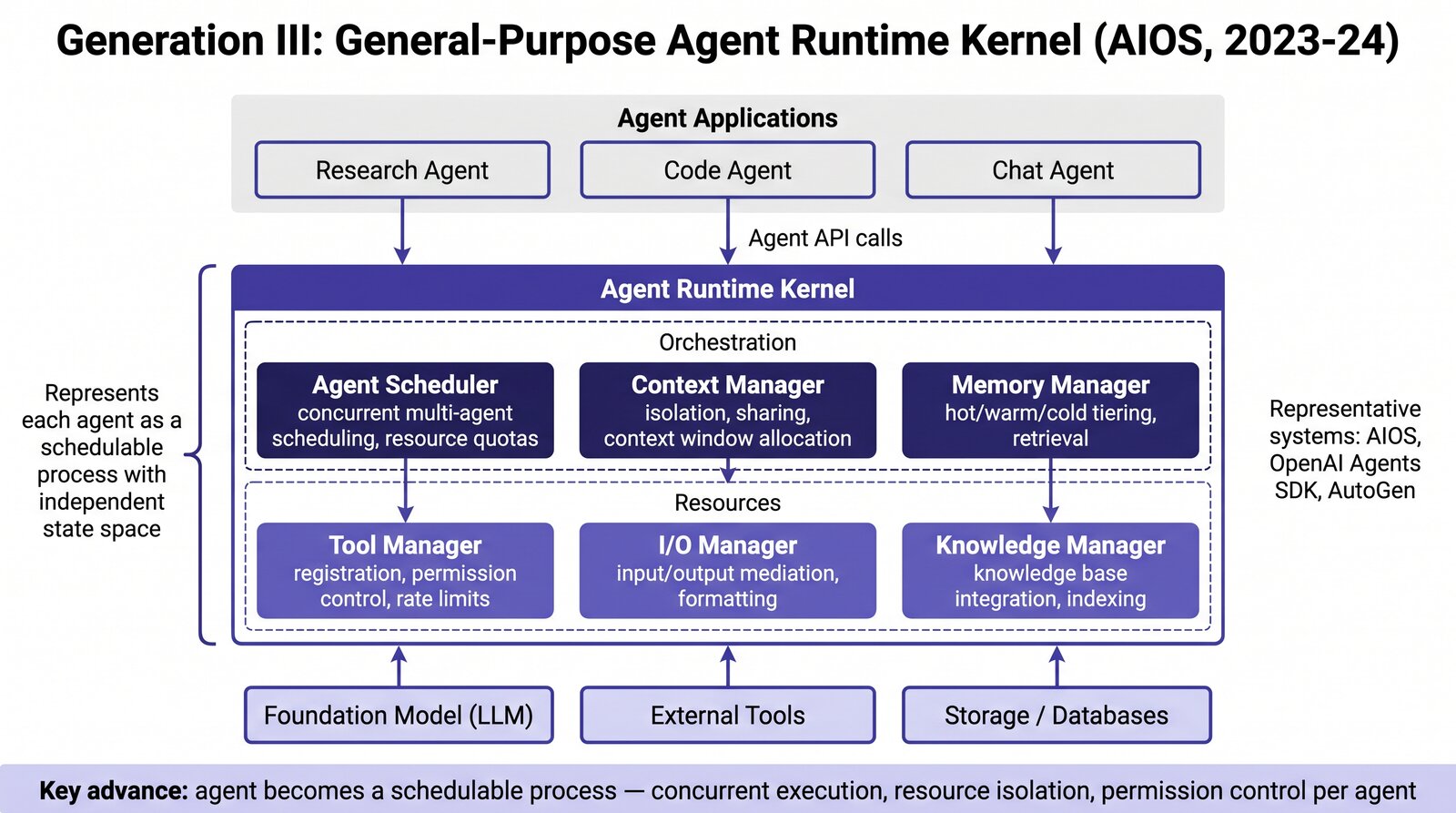}
  \caption{Generation~III agent frameworks versus time-sharing operating systems (Unix/Multics).
  AIOS's Agent Scheduler, Context Manager, and Tool Manager correspond to
  a time-sharing OS's process scheduler, MMU address isolation, and file permission system.
  The key advance: each agent becomes a schedulable process with an independent state space
  and resource budget.}
  \label{fig:ae:gen3}
\end{figure}

%% file: figures/fig_ae_gen4.tex
\begin{figure}[htbp]
  \centering
  \includegraphics[width=0.88\textwidth]{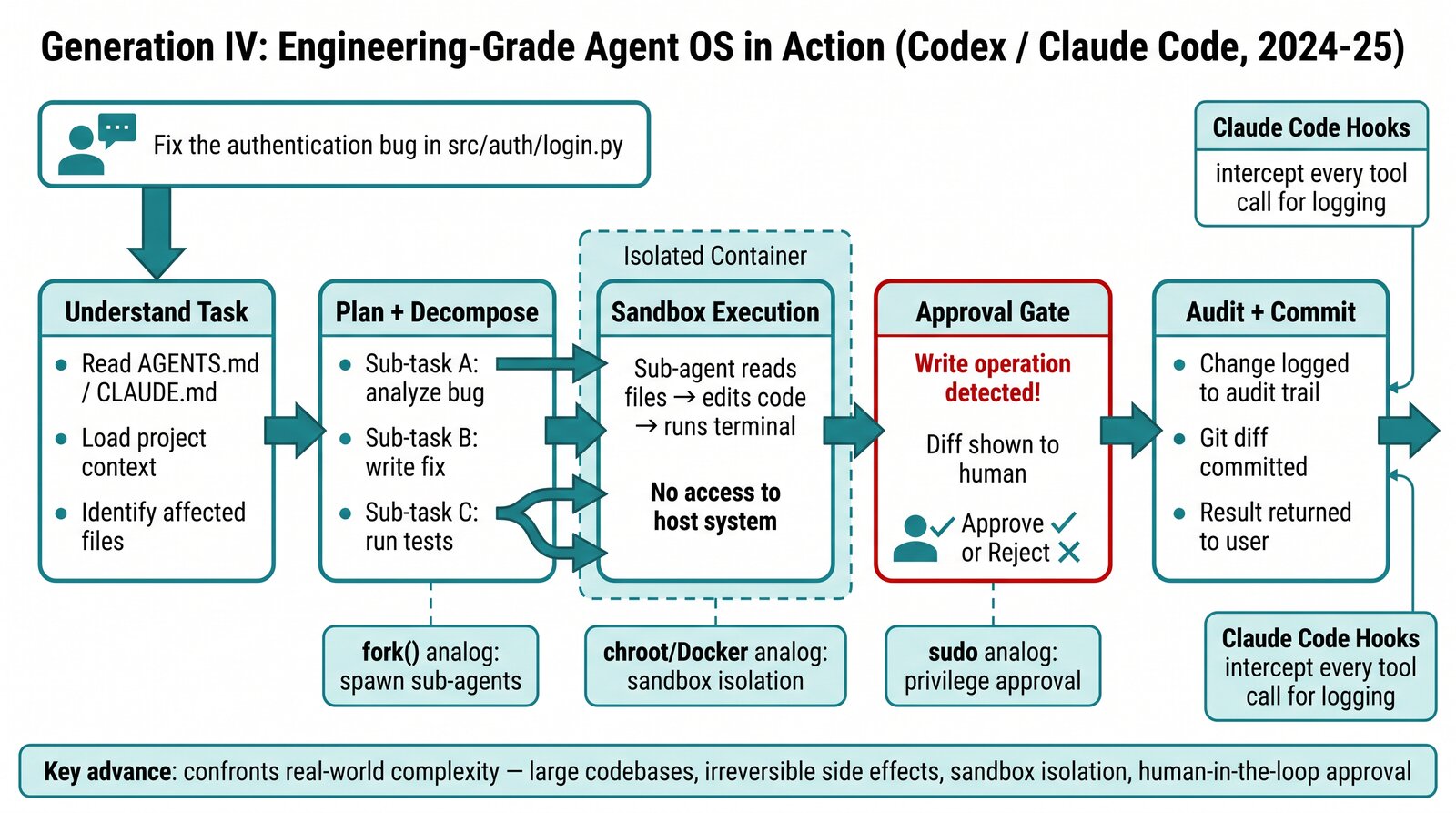}
  \caption{Generation~IV agent frameworks versus modern operating systems (Linux/Windows~NT).
  Codex's sub-agent spawning maps to \texttt{fork()};
  its sandbox maps to \texttt{chroot}/container isolation;
  Claude Code's hooks parallel Linux Security Module (LSM) hooks;
  and the approval gate corresponds to \texttt{sudo} privilege elevation.
  The key advance: confronting real-world codebases with irreversible side effects
  demands formal isolation and human-in-the-loop approval.}
  \label{fig:ae:gen4}
\end{figure}

%% file: figures/fig_ae_gen5.tex
\begin{figure}[htbp]
  \centering
  \includegraphics[width=0.88\textwidth]{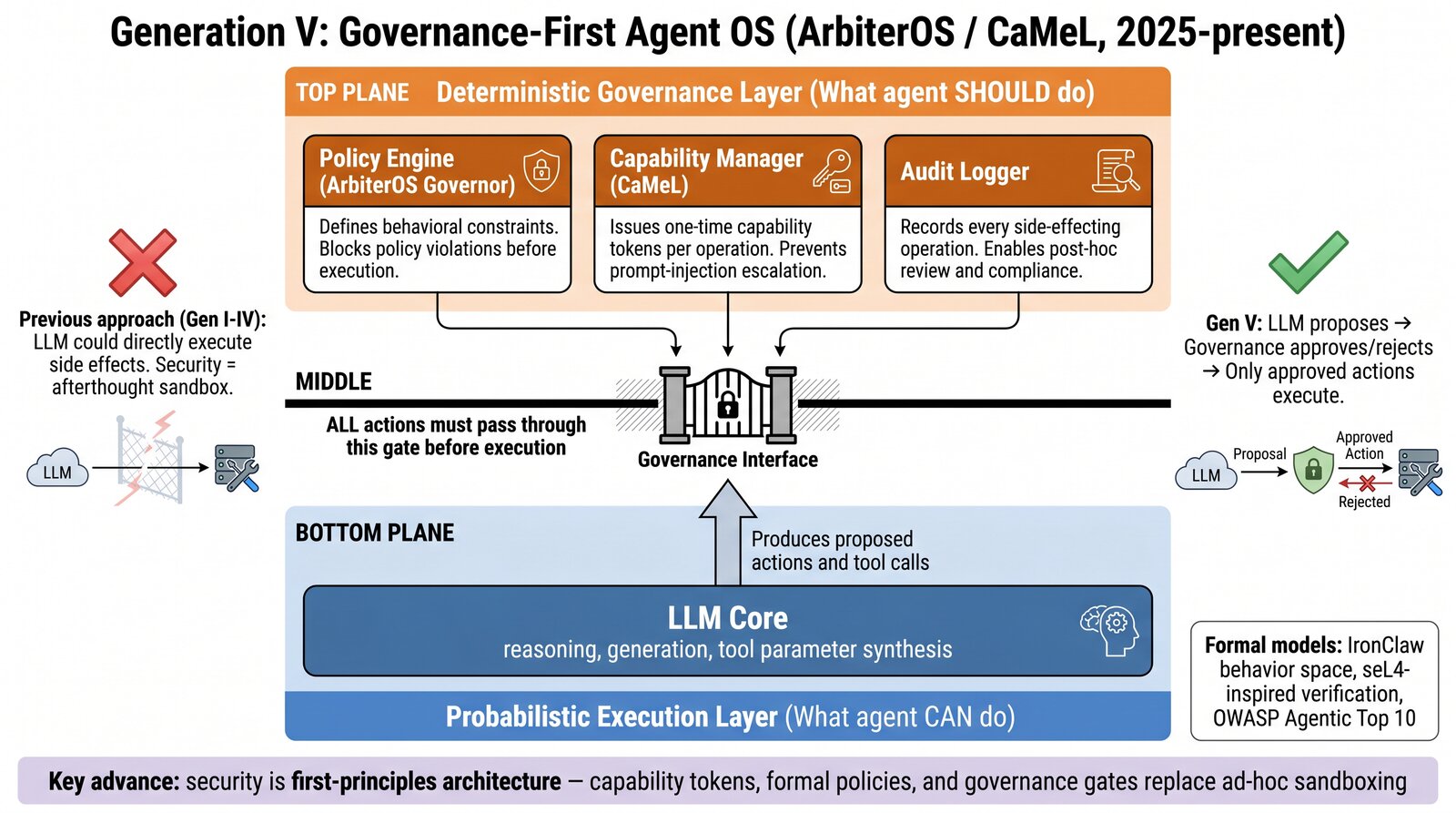}
  \caption{Generation~V agent frameworks versus trustworthy operating systems (SELinux/seL4).
  ArbiterOS's probabilistic CPU and deterministic governor correspond to
  user-mode/kernel-mode separation;
  CaMeL's one-time capability tokens correspond to POSIX fine-grained capabilities;
  and IronClaw's behavioral modeling corresponds to seL4's formal verification.
  The key advance: security is embedded as a first architectural principle,
  not bolted on after the fact.}
  \label{fig:ae:gen5}
\end{figure}

%% file: figures/fig_gen_matrix.tex
\begin{figure}[htbp]
    \centering
    \resizebox{\textwidth}{!}{%
    \begin{tikzpicture}[
        cell/.style={
            minimum width=2.4cm, minimum height=1.0cm,
            align=center, font=\scriptsize,
            inner sep=2pt,
        },
        rowlabel/.style={
            minimum width=2.8cm, minimum height=1.0cm,
            text width=2.5cm, align=left, font=\small\bfseries,
            inner sep=4pt,
        },
        collabel/.style={
            minimum width=2.4cm, minimum height=0.9cm,
            align=center, font=\small\bfseries,
            inner sep=2pt,
        },
    ]
    
    \def\cols{{"Task\\Decomp.", "Concurrency\\(F)", "Orchestration\\Eff.~(E)", "State\\Persist.", "Isolation\\/ Sandbox", "Security\\/ Govern."}}
    \def\ncols{6}
    
    \def\fillcol#1{%
        \ifnum#1=0 white\fi%
        \ifnum#1=1 modelnative!20\fi%
        \ifnum#1=2 modelnative!50\fi%
        \ifnum#1=3 modelnative!85\fi%
    }
    \def\txtcol#1{%
        \ifnum#1=0 gray!40\fi%
        \ifnum#1=1 modelnative!60!black\fi%
        \ifnum#1=2 modelnative!70!black\fi%
        \ifnum#1=3 white\fi%
    }
    \def\levelstr#1{%
        \ifnum#1=0 ---\fi%
        \ifnum#1=1 Low\fi%
        \ifnum#1=2 Medium\fi%
        \ifnum#1=3 High\fi%
    }
    
    \def\rowA{0,0,3,0,0,0}
    \def\rowB{1,0,2,1,0,0}
    \def\rowC{2,2,2,2,1,1}
    \def\rowD{3,3,2,3,3,2}
    \def\rowE{3,3,3,3,3,3}
    
    \def\rowlabels{{"Gen~I\\(ReAct)", "Gen~II\\(AutoGPT)", "Gen~III\\(AIOS)", "Gen~IV\\(Codex)", "Gen~V\\(ArbiterOS)"}}
    \def\sublabels{{"2022--23", "2023", "2023--24", "2024--25", "2025--"}}
    
    \foreach \j/\lbl in {
        0/Task\\Decomp.,
        1/Concurrency\\$(F)$,
        2/Orchestration\\Eff.~$(E)$,
        3/State\\Persist.,
        4/Isolation\\/ Sandbox,
        5/Security\\/ Govern.}
    {
        \node[collabel, draw=gray!30, fill=gray!10]
            at ({3.9 + \j*2.4}, 0.5) {\lbl};
    }
    
    \node[rowlabel, draw=gray!30, fill=gray!10, minimum height=0.9cm]
        at (1.3, 0.5) {Generation};
    
    \newcommand{\drawcell}[3]{
        \pgfmathsetmacro\vint{int(#3)}%
        \ifnum\vint=0 \def\fc{white}\def\tc{gray!35}\fi%
        \ifnum\vint=1 \def\fc{modelnative!20}\def\tc{modelnative!65!black}\fi%
        \ifnum\vint=2 \def\fc{modelnative!50}\def\tc{modelnative!80!black}\fi%
        \ifnum\vint=3 \def\fc{modelnative!85}\def\tc{white}\fi%
        \node[cell, draw=gray!25, fill=\fc, text=\tc, minimum height=1.0cm]
            at ({3.9 + #1*2.4}, {-#2*1.0 - 0.5})
            {\ifnum\vint=0 ---\fi%
             \ifnum\vint=1 Low\fi%
             \ifnum\vint=2 Medium\fi%
             \ifnum\vint=3 High\fi};%
    }
    
    \foreach \i/\rlab/\rsub in {
        0/{Gen~I}/{\footnotesize 2022--23},
        1/{Gen~II}/{\footnotesize 2023},
        2/{Gen~III}/{\footnotesize 2023--24},
        3/{Gen~IV}/{\footnotesize 2024--25},
        4/{Gen~V}/{\footnotesize 2025--Present}%
    }{
        \node[rowlabel, draw=gray!20, fill=gray!5, minimum height=1.0cm]
            at (1.3, {-\i*1.0 - 0.5})
            {\textbf{\rlab}\\[1pt]\rsub};
    }
    
    \drawcell{0}{0}{0} \drawcell{1}{0}{0} \drawcell{2}{0}{3}
    \drawcell{3}{0}{0} \drawcell{4}{0}{0} \drawcell{5}{0}{0}
    
    \drawcell{0}{1}{1} \drawcell{1}{1}{0} \drawcell{2}{1}{2}
    \drawcell{3}{1}{1} \drawcell{4}{1}{0} \drawcell{5}{1}{0}
    
    \drawcell{0}{2}{2} \drawcell{1}{2}{2} \drawcell{2}{2}{2}
    \drawcell{3}{2}{2} \drawcell{4}{2}{1} \drawcell{5}{2}{1}
    
    \drawcell{0}{3}{3} \drawcell{1}{3}{3} \drawcell{2}{3}{2}
    \drawcell{3}{3}{3} \drawcell{4}{3}{3} \drawcell{5}{3}{2}
    
    \drawcell{0}{4}{3} \drawcell{1}{4}{3} \drawcell{2}{4}{3}
    \drawcell{3}{4}{3} \drawcell{4}{4}{3} \drawcell{5}{4}{3}
    
    \node[font=\scriptsize, text=gray!60, anchor=east] at (5.7, -5.65) {Maturity:};
    \foreach \v/\lbl in {0/None, 1/Low, 2/Medium, 3/High} {
        \ifnum\v=0\def\fc{white}\def\tc{gray!40}\fi
        \ifnum\v=1\def\fc{modelnative!20}\def\tc{modelnative!65!black}\fi
        \ifnum\v=2\def\fc{modelnative!50}\def\tc{modelnative!80!black}\fi
        \ifnum\v=3\def\fc{modelnative!85}\def\tc{white}\fi
        \node[draw=gray!30, fill=\fc, text=\tc,
              minimum width=1.5cm, minimum height=0.55cm,
              font=\scriptsize, rounded corners=2pt]
            at ({6.7 + \v*1.9}, -5.65) {\lbl};
    }
    
    \end{tikzpicture}
    }
    \caption{Capability maturity matrix across the five generations of agent framework evolution. Each cell reflects the maturity level of a key architectural dimension: \textit{None} (---) through \textit{High}. The progressive darkening from Generation~I to Generation~V illustrates the systematic increase in both the parallelisable fraction~$F$ and the orchestration efficiency~$E$ captured by Heuristic~III.}
    \label{fig_gen_matrix}
    \end{figure}
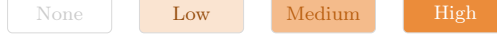

%% file: figures/fig_gen_trajectory.tex
\begin{figure}[htbp]
\centering
\resizebox{0.88\textwidth}{!}{%
\begin{tikzpicture}
\small
\begin{groupplot}[
    group style={
        group size=3 by 1,
        horizontal sep=2.2cm,
    },
    width=5.0cm,
    height=5.2cm,
    grid=major,
    grid style={gray!18},
    tick label style={font=\footnotesize},
    label style={font=\small},
    xtick={1,2,3,4,5},
    xticklabels={I,II,III,IV,V},
    xlabel style={font=\small},
    ylabel style={font=\small},
    xmin=0.5, xmax=5.5,
    axis lines=left,
    every axis plot/.append style={thick, mark=*, mark size=2.2pt},
    enlargelimits=0.12,
    legend style={
        font=\scriptsize,
        at={(0.5,-0.32)},
        anchor=north,
        legend columns=1,
        draw=gray!40,
        rounded corners=2pt,
    },
]

\nextgroupplot[
    title style={font=\small\bfseries, at={(0.5,1.06)}},
    title={(a) Parallelizable Fraction $F$},
    xlabel={Generation},
    ylabel={$F$},
    ymin=0, ymax=1.05,
    ytick={0,0.2,0.4,0.6,0.8,1.0},
]
\addplot[blue!70!black, mark=*]
    coordinates {(1,0.00)(2,0.15)(3,0.30)(4,0.50)(5,0.72)};
\addplot[red!60!black, dashed, no marks, domain=0.5:5.5] {0.8};
\addlegendentry{$F$ (estimated)}
\addlegendentry{Target $F>0.8$}
\node[font=\tiny, blue!60!black, anchor=east] at (axis cs:0.9, 0.05) {${\approx}0$};
\node[font=\tiny, blue!60!black, anchor=east] at (axis cs:1.85, 0.20) {$0.15$};
\node[font=\tiny, blue!60!black, anchor=east] at (axis cs:2.85, 0.35) {$0.30$};
\node[font=\tiny, blue!60!black, anchor=east] at (axis cs:3.85, 0.53) {$0.50$};
\node[font=\tiny, blue!60!black, anchor=east] at (axis cs:4.85, 0.73) {$0.72$};

\nextgroupplot[
    title style={font=\small\bfseries, at={(0.5,1.06)}},
    title={(b) Orchestration Efficiency $E$},
    xlabel={Generation},
    ylabel={$E$},
    ymin=0, ymax=1.05,
    ytick={0,0.2,0.4,0.6,0.8,1.0},
]
\addplot[orange!80!black, mark=square*]
    coordinates {(1,1.00)(2,0.90)(3,0.70)(4,0.50)(5,0.65)};
\addplot[red!60!black, dashed, no marks, domain=0.5:5.5] {0.8};
\addlegendentry{$E$ (estimated)}
\addlegendentry{Target $E>0.8$}
\node[font=\tiny, orange!70!black] at (axis cs:1, 0.88) {$1.0$};
\node[font=\tiny, orange!70!black] at (axis cs:2, 1) {$0.9$};
\node[font=\tiny, orange!70!black] at (axis cs:3, 0.58) {$0.7$};
\node[font=\tiny, orange!70!black] at (axis cs:4, 0.38) {$0.5$};
\node[font=\tiny, orange!70!black] at (axis cs:5, 0.5) {$0.65$};

\nextgroupplot[
    title style={font=\small\bfseries, at={(0.5,1.06)}},
    title={(c) Agent Speedup $S_{\mathrm{agent}}$},
    xlabel={Generation},
    ylabel={$S_{\mathrm{agent}}$},
    ymin=0, ymax=4.0,
    ytick={0,1,2,3,4},
]
\addplot[red!70!black, mark=triangle*, mark size=2.5pt]
    coordinates {(1,1.00)(2,1.00)(3,1.27)(4,1.60)(5,2.10)};
\addplot[red!50!black, dashed, no marks, domain=0.5:5.5] {3.3};
\addlegendentry{$S_{\mathrm{agent}}$ (estimated)}
\addlegendentry{Target $S>3.3$}
\node[font=\tiny, red!65!black] at (axis cs:1, 1.3) {$1.0$};
\node[font=\tiny, red!65!black] at (axis cs:2, 1.3) {$1.0$};
\node[font=\tiny, red!65!black] at (axis cs:3, 1.7) {$1.27$};
\node[font=\tiny, red!65!black] at (axis cs:4, 2.0) {$1.60$};
\node[font=\tiny, red!65!black] at (axis cs:5, 2.5) {$2.10$};

\end{groupplot}
\end{tikzpicture}}%
\caption{Illustrative quantitative trajectory of agent framework parameters across five generations, estimated from Heuristic~III ($S_{\mathrm{agent}} = 1/((1-F)+F/(N{\cdot}E))$ with representative $N$ values per generation). (a)~Parallelizable fraction $F$ grows from near zero (Gen~I serial loop) to ${\approx}0.72$ (Gen~V governed OS). (b)~Orchestration efficiency $E$ initially near~1 (trivial single-agent overhead) but dips in Gen~IV due to sandbox isolation and approval overhead, then recovers in Gen~V with governance-first design. (c)~Resulting speedup $S_{\mathrm{agent}}$ rises from $1.0\times$ to ${\approx}2.1\times$, still well below the target $3.3\times$ that requires $F>0.8$ and $E>0.8$ simultaneously. Dashed red lines mark the target thresholds.}
\label{fig:gen_trajectory}
\end{figure}
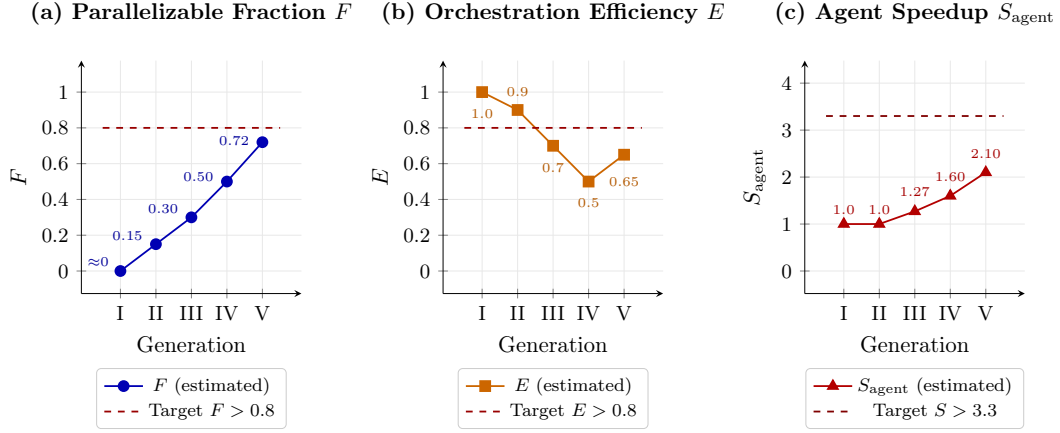

%% file: en_sections/section_challenges.tex
\section{Design Challenges: Performance, State, Consistency, and Security}
\label{sec:challenges}

Each generational leap in agent frameworks, from first-generation tool chains to fifth-generation autonomous agents, has exposed new systemic problems that neither the model itself nor ad hoc engineering workarounds can fully resolve.
This section examines five cross-cutting design challenges through a uniform lens: (1)~we describe a concrete scenario that illustrates the problem; (2)~we analyze the root cause; (3)~we survey current solutions; and (4)~we draw an explicit analogy to a well-studied problem in classical computer architecture.
Throughout, we trace each challenge back to the ICA layers (Section~\ref{sec:icam}) and the design heuristics (Section~\ref{sec:laws}), showing how the dual-plane architecture introduced in Section~\ref{sec:framework} serves as a \emph{unifying organizational lens} for these challenges.

\subsection{The Latency--Throughput--Cost Trilemma}

\subsubsection{Scenario}
Consider a code agent assisting a developer in fixing a bug that spans ten files.
The agent must read multiple files (each read triggers a model inference call), understand the code structure, locate the fault, modify the code, run tests, inspect the results, and iterate.
The entire workflow may involve 20--50 model calls, and the per-call latency directly determines the developer's wait time.
If a single call costs 2~seconds, the end-to-end latency can easily exceed one minute.
If the serving provider provisions additional GPUs to reduce per-call latency, the cost per invocation rises proportionally.
Conversely, if the provider increases batch sizes to amortize cost and improve throughput, individual requests queue longer, increasing latency.
This is the \emph{latency--throughput--cost trilemma}: reducing latency demands more resources (higher cost), improving throughput demands larger batches (higher latency), and controlling cost demands fewer resources (higher latency and lower throughput).

\subsubsection{Root Cause}
The root cause lies in a structural asymmetry inherent to autoregressive LLM serving.
The \textbf{prefill} phase is \emph{compute-bound}: the entire input sequence must be processed in one forward pass, keeping GPU compute units fully occupied, with FLOPS as the bottleneck.
The \textbf{decode} phase is \emph{memory-bandwidth-bound}: generating each new token requires reading the full model weights and the current KV cache, effectively streaming all parameters from memory on every clock cycle~\cite{gholami2024memorywall}.
Gholami et al.\ quantify this imbalance: peak FLOPS grow at $3.0\times$ every two years, while DRAM bandwidth improves at only $1.6\times$ over the same period~\cite{gholami2024memorywall}.

Because the two phases demand fundamentally different hardware resources, co-scheduling them in the same batch inevitably leaves some resource underutilized: prefill monopolizes compute while decode stalls on memory bandwidth.
Agent workloads exacerbate this problem.
Multiple sub-agents issue concurrent requests whose context lengths vary wildly: one may read a single function while another must ingest an entire repository, rendering static batching strategies inefficient.

\subsubsection{Current Solutions}
The research community has responded at multiple levels.

\paragraph{Scheduling-level optimizations.}
ORCA~\cite{orca2022} introduces \emph{iteration-level scheduling}: rather than waiting for an entire sequence to complete before admitting new requests, the scheduler checks after every decode step whether a new request can join the running batch, significantly improving GPU utilization.
DistServe~\cite{zhong2024distserve} goes further by \emph{disaggregating} prefill and decode onto separate GPU groups, allowing each group to be independently optimized: high-FLOPS GPUs for prefill, high-bandwidth GPUs for decode.
Sarathi-Serve~\cite{agrawal2024sarathi} introduces stall-free scheduling via \emph{chunked prefill}, which slices long inputs into fixed-size chunks and interleaves them with decode tasks, eliminating stalls caused by waiting for large prefills to complete.

\paragraph{Memory management optimizations.}
vLLM~\cite{kwon2023pagedattention,vllm_docs_2026} employs PagedAttention to organize the KV cache into fixed-size pages mapped through a block table to non-contiguous physical memory, enabling on-demand allocation and sharing of KV cache.
SGLang~\cite{sglang2024,sglang_docs_2026} uses RadixAttention to organize request prefixes in a radix tree, automatically identifying and reusing shared-prefix KV cache entries.
TGI~\cite{tgi_docs_2026} and TensorRT-LLM~\cite{tensorrtllm_docs_2026} provide production-grade implementations of continuous batching and paged KV management.
vAttention~\cite{prabhu2024vattention} offers an alternative approach: leveraging OS-level contiguous virtual memory with dynamic growth to support standard attention kernels directly, bypassing the need for specialized PagedAttention kernels.

Figure~\ref{fig_trilemma} illustrates the three-way trade-off. Any two of the three objectives can be jointly optimised, but improving all three simultaneously is structurally infeasible.
\input{figures/fig_trilemma.tex}

\subsubsection{Analogy to Classical Architecture}
This challenge is a direct analog of the classical \emph{memory wall}.
Since the 1980s, CPU speeds have improved at roughly 50\% per year while DRAM latency has improved at only 7\% per year~\cite{hennessy2017quantitative}.
The architectural response, including multi-level caches (L1/L2/L3), hardware prefetching, and write buffers, maps directly onto LLM serving: KV cache tiering corresponds to cache hierarchy, prefix caching to instruction cache sharing, and asynchronous decoding to write buffering.
The familiar response-time--throughput trade-off in time-sharing operating systems is also analogous: interactive tasks demand low response time (low latency) while batch tasks demand high throughput (large batches), and the scheduler must balance both~\cite{ostep_2023}.

Within ICA, this trilemma primarily spans L1 (physical execution) and L2 (inference serving).
Heuristic~I (Section~\ref{sec:law1}) characterizes the cache-side of this trade-off: the KV cache hit rate $H$ directly governs the achievable speedup, and any scheduling or memory management optimization can be understood as an attempt to maximize $H$ under the trilemma's constraints.

\subsection{State Management and Cross-Session Consistency}

\subsubsection{Scenario}
Suppose a user works with the same agent over three consecutive days to build a project.
On Day~1, the agent learns that the user prefers TypeScript over JavaScript.
On Day~2, the user asks the agent to refactor the codebase; the agent must respect the preference learned on Day~1 while correctly understanding the current state of the code.
On Day~3, the user discovers a new bug, and the agent must recall modifications made on both previous days (the bug may have been introduced by those very changes) while reasoning about the latest project state.
To complicate matters further, suppose the user runs two agent instances concurrently on Day~2: one performing the refactor and another writing tests.
Both agents may modify the same file, creating a conflict.

\subsubsection{Root Cause}
A traditional CPU is a \emph{stateless} executor: given the same inputs and register state, the output is deterministic and reproducible.
Intelligent systems, by contrast, must \emph{persist} state across sessions: user preferences (``I prefer TypeScript''), task history (``yesterday we refactored the authentication module''), and external-world changes (``three new files were added to the repository'').
Once persistent state is introduced, at least three categories of problems inevitably arise:

\begin{enumerate}[nosep]
\item \textbf{Concurrent access to shared memory.}
When multiple agents read and write shared memory simultaneously, how do we guarantee consistency?
This is analogous to race conditions in multi-threaded programming.

\item \textbf{Delayed commits and conflict resolution.}
Agent~A modifies a file but has not yet committed the change; Agent~B begins modifying the same file based on the old version.
Detecting and resolving such conflicts mirrors optimistic concurrency control in distributed systems.

\item \textbf{Memory staleness.}
The fact ``the project uses React'' learned three days ago may be obsolete because the project has since migrated to Vue.
Determining whether a memory is still valid is analogous to cache invalidation.
\end{enumerate}

\subsubsection{Current Solutions}
The LongMemEval benchmark~\cite{longmemeval2024} demonstrates that long-term memory is not simply ``storing more chat logs'' but requires four distinct capabilities: information extraction (distilling key facts from conversation), temporal reasoning (understanding event ordering), knowledge updating (revising old memories when new information conflicts), and refusal (declining to answer when memory is insufficient).

MemGPT~\cite{memgpt2023} borrows the paging and swap-in/swap-out mechanisms of OS virtual memory to manage context, dividing it into ``main memory'' (the active context window) and ``external storage'' (long-term memory in a vector database), with explicit edit and retrieve operations to move information between the two.
MemoryOS~\cite{memoryos2025} abstracts memory management as an OS memory management module, implementing structured memory encoding, retrieval, and update.
A-MEM~\cite{amem2025} proposes agentic memory management: memories are not passively stored strings but ``active entities'' whose creation, update, and deletion are governed by the agent itself.
Letta~\cite{letta_stateful_2026} (formerly the MemGPT project) further elevates ``stateful agents'' to a core abstraction, explicitly separating conversational state, core memory, and archival memory.

From a distributed systems perspective, at least three state types must be distinguished~\cite{raft2014,spanner2012,cap2002}:
\begin{enumerate}[nosep]
\item \textbf{Ephemeral state:} intermediate results of the current inference step, analogous to CPU registers; no persistence required.
\item \textbf{Session state with causal ordering:} message passing and state transfer among agents within the same task, requiring causal consistency guarantees analogous to those in distributed systems~\cite{raft2014}.
\item \textbf{Committed state with full auditability and replay:} permanent modifications to files, databases, or external systems, analogous to a database transaction's commit point, supporting audit and rollback~\cite{spanner2012}.
\end{enumerate}

Figure~\ref{fig:state_lifecycle} illustrates the lifecycle and ICA layer mapping of these three state types.
\input{figures/fig_state_lifecycle}

\subsubsection{Temporal Decay in Long-Running Agents}
Critically, the time horizon of agent execution is expanding from minutes to hours and even days.
Recent industrial practice shows that agents can autonomously run for over seven hours to complete complex tasks~\cite{anthropic2026agenticreport}.
Such cross-hour and cross-day execution introduces additional temporal decay challenges: the attention decay function $\beta(L)$ is no longer solely a function of context position but also depends on session duration and cross-step state accumulation.
This means the three-way state classification above acquires a time dimension: cross-hour execution requires session-persistent KV caches (L2), context window management must support working-set restoration across checkpoints (L3), and the governance layer must define periodic checkpointing and human-recovery protocols (L5).
Heuristic~II (Section~\ref{sec:law2}), as currently formulated, does not capture this temporal dimension, an important direction for future work.

\subsubsection{Analogy to Classical Architecture}
These challenges map directly onto \emph{virtual memory and distributed consistency} in classical systems.
OS virtual memory uses page tables to map virtual to physical addresses and provides swap mechanisms to create the illusion of unbounded memory, analogous to context management in agent systems.
Distributed file systems such as GFS and Spanner~\cite{spanner2012} use Paxos/Raft~\cite{raft2014} consensus protocols to guarantee multi-replica consistency, analogous to consistency in multi-agent shared memory.
The CAP theorem~\cite{cap2002} states that consistency, availability, and partition tolerance cannot be simultaneously achieved; agent memory management faces the same trade-off: strong consistency (synchronizing all replicas before acknowledging each operation) versus eventual consistency (tolerating brief inconsistency in exchange for lower latency).

Within ICA, this challenge primarily spans L3 (context management) and L5 (orchestration).
The dual-plane architecture naturally separates the problem: the probabilistic execution plane manages the working context (what to remember and what to forget), while the deterministic control plane enforces consistency guarantees and conflict-resolution protocols.
Figure~\ref{fig:ch82:state} illustrates the three-day scenario, the three state categories, and the concurrent conflict pattern.

\input{figures/fig_ch82_state}

\subsection{Interface Drift and Version Compatibility}

\subsubsection{Scenario}
An agent is configured to manage a code repository via the GitHub API.
One day, GitHub updates the API from v3 to v4: the endpoint \texttt{repos/:owner/:repo/pulls} returns a new response format, adds pagination parameters, and removes several fields.
The agent's tool calls begin returning errors or failing to parse responses.
A more subtle scenario: the agent's prompt template relies on the model producing output in a specific format (e.g., ``always respond in JSON''), but a model upgrade subtly changes the output style, causing the downstream parser to crash repeatedly.
Similar problems arise when MCP servers update their capability descriptions, when skill packs change their interfaces, or when memory file formats undergo version upgrades.

\subsubsection{Root Cause}
Classical computer systems achieve scalability in large part because their core interfaces, including ISA, ABI, system calls, and file formats, remain \emph{stable over decades}.
The x86 instruction set has maintained backward compatibility from the 8086 (1978) through the latest processors in 2026; the POSIX standard, established in 1988, has preserved the semantics of \texttt{read()}/\texttt{write()}/\texttt{fork()} with almost no change~\cite{ostep_2023}.
This stability allows upper-layer software to be built with confidence that the ground beneath it will not shift unexpectedly.

The current agent ecosystem, by contrast, suffers from severe \emph{interface drift}.
At least four sources of drift can be identified:
\begin{enumerate}[nosep]
\item \textbf{Model version drift:} a model upgrade may change the output format for the same prompt.
\item \textbf{Tool schema drift:} external APIs may update their interface descriptions and response formats without warning~\cite{mcp_intro_2026}.
\item \textbf{Protocol drift:} protocols such as MCP and A2A are themselves under rapid iteration~\cite{agentprotocols2025,google_a2a_2025}.
\item \textbf{Skill pack drift:} reusable agent skills may behave differently across projects or model versions.
\end{enumerate}

\subsubsection{Current Solutions}
OpenAI Codex provides project-level stability anchors through AGENTS.md~\cite{openai_codex_agents_md_2026} and the Skills mechanism~\cite{openai_codex_skills_2026}: each skill pack carries explicit version numbers and compatibility declarations.
Claude Code achieves similar functionality through CLAUDE.md~\cite{anthropic_claudecode_memory_2026}.
Interoperability protocols, including MCP~\cite{mcp_intro_2026} for model--tool communication and A2A~\cite{google_a2a_2025} for agent--agent interaction, are attempting to establish standardized interfaces.
The A2A protocol, built on JSON-RPC 2.0 over HTTP, defines standard workflows for agent discovery, capability negotiation, and task delegation~\cite{google_a2a_2025}.
However, these protocols are themselves evolving rapidly and have not yet achieved the stability of an ISA or POSIX.

\subsubsection{Analogy to Classical Architecture}
This challenge corresponds to \emph{ABI stability and version management} in classical systems.
Intel has maintained x86 backward compatibility for over 45~years; the Linux kernel has preserved system-call backward compatibility for over 30~years; even shared-library versioning has mature solutions (semantic versioning, symbol versioning).
Model-native computing architectures must develop their own ``version control discipline'': tool schemas need version numbers, model output formats need compatibility guarantees, skill packs need dependency management (analogous to npm's \texttt{package.json}), and memory files need format migration tools (analogous to database schema migration).

Within ICA, interface drift primarily affects L4 (semantic interface layer) and the L4--L5 boundary.
The deterministic control plane must maintain version-aware interface contracts that insulate the probabilistic execution plane from upstream changes, precisely the role that ABI stability plays in shielding application software from hardware evolution.

\subsection{Security, Privacy, and Least Privilege}

\subsubsection{Scenario}
A code agent is helping a developer process user data.
The agent must read database schemas that contain personally identifiable information (to understand the data model), modify code that handles user data, and run tests whose fixtures may include real user records.
During this process, the agent may: (1)~write sensitive data to log files or transmit it to a remote API (data exfiltration); (2)~be manipulated by a malicious code comment or file content into performing unintended actions via prompt injection (goal hijacking); (3)~use real user data in tests instead of synthetic fixtures (privacy violation).
Worse still: since the agent can execute arbitrary shell commands, a carefully crafted prompt injection could cause it to run \texttt{rm -rf /} or upload sensitive files to an external server.

\subsubsection{Root Cause}
Once an agent can read and write files, execute commands, browse the web, and query databases, it is no longer merely a language model but an \emph{agent} in the security sense: a principal with external action capability.
This introduces attack surfaces that classical software security does not cover:

\begin{enumerate}[nosep]
\item \textbf{Prompt injection.}
An attacker embeds malicious instructions in tool-returned data, file contents, or web pages to hijack the agent's behavior.
Unlike classical code injection (e.g., SQL injection), prompt injection exploits not a logic flaw but the model's inability to distinguish ``instructions from the user'' from ``instructions embedded in data''~\cite{debenedetti2025camel,owasp_agentic2026}.

\item \textbf{Privilege over-provisioning.}
Most current agents, when granted tool access, receive the tool's \emph{full} permission set (e.g., read/write access to the entire filesystem) rather than the minimum necessary privilege (e.g., read/write access only to specific project files).

\item \textbf{Memory taint propagation.}
A successful prompt injection can write malicious information into the agent's long-term memory; this ``tainted'' memory then influences the agent's behavior across all subsequent sessions, creating a persistent backdoor~\cite{aura2025}.
\end{enumerate}

\subsubsection{Current Solutions}
Industry and academia are constructing defenses along multiple dimensions.

\paragraph{Sandboxing.}
OpenAI Codex runs tasks by default inside an OS-enforced sandbox, where each task executes in an isolated container without access to the host filesystem or network~\cite{openai_codex_sandbox_2026}.
Claude Code operates in read-only mode by default, requiring explicit user authorization for any write operation~\cite{anthropic_claudecode_security_2026}.
Aura establishes independent sandboxes for each app agent on mobile devices and restricts the agent kernel's access to system resources~\cite{aura2025}.

\paragraph{Capability-based security models.}
CaMeL~\cite{debenedetti2025camel} applies the OS capability-security principle to defend against prompt injection: the agent cannot directly execute side-effecting operations but must obtain a ``temporary pass'' from a deterministic capability manager.
Each pass authorizes exactly one specific operation (e.g., ``read-only access to \texttt{src/main.py}'') and is consumed upon use.
This fundamentally eliminates the possibility of privilege escalation via prompt injection, because permissions are governed by deterministic code rather than a probabilistic model.

\paragraph{Approval and audit.}
Codex's approval policies~\cite{openai_codex_approvals_2026} allow developers to specify which operations require human confirmation (e.g., ``any \texttt{DELETE} operation must be approved'').
The Agents SDK's tracing mechanism~\cite{openai_codex_agentsdk_2026} records the full context of every tool call, supporting post-hoc audit and replay.

\paragraph{Formal security.}
IronClaw~\cite{ironclaw2025} proposes a formal framework for agent security that explicitly models the agent's behavior space and permission space, detecting out-of-bounds actions through static analysis and runtime monitoring.
Christodorescu et al.~\cite{agenticsecurity2025} argue that agent security must draw on the full experience of operating system security: least privilege, isolation, auditing, and fail-safe defaults.

\subsubsection{Analogy to Classical Architecture}
These challenges correspond to \emph{process security and permission management} in classical systems.
Unix security rests on three core concepts: (1)~user/kernel mode separation, where user programs cannot access hardware directly and must go through system calls~\cite{ostep_2023}; (2)~file permissions (rwx), where each file has explicit read, write, and execute permissions; (3)~root/non-root separation, where privileged operations require explicit elevation.
Agent security requires an analogous three-layer model: (1)~probabilistic-plane/deterministic-plane separation, where agents cannot directly execute side-effecting operations but must pass through the deterministic control plane's approval channel; (2)~tool capability annotations, where each tool carries explicit capability declarations and permission requirements; (3)~governance-layer privilege elevation, where high-risk operations require explicit authorization or human confirmation.

\subsubsection{Dual-Use Security Concerns}
It is important to recognize that agent capabilities are inherently dual-use.
The same capabilities that enable automated security auditing (defensive use) can also enable automated vulnerability discovery and exploitation (offensive use)~\cite{anthropic2026agenticreport}.
This means the deterministic control plane of a dual-plane architecture must not only constrain its own agents' behavior but also anticipate that adversaries possess equivalent dual-plane systems.
Anthropic's 2026 agentic security report predicts that ``agentic cyber defense systems'' will operate at machine speed~\cite{anthropic2026agenticreport}, implying that the L5 governance layer's primitives must incorporate adversarial scenarios into their threat models from the outset.

Within ICA, security spans all six layers but concentrates at L4 (tool capability annotations and call audits) and L5 (permission approval and failure recovery).
The dual-plane architecture is essential here: the probabilistic plane handles the reasoning needed to interpret user intent, while the deterministic plane enforces the non-negotiable security invariants.

\subsection{Governance as a First-Class Architectural Concern}

\subsubsection{The Core Problem}
ArbiterOS~\cite{arbiteros2025} identifies the fundamental crisis in agent engineering as a structural contradiction: developers instinctively apply \emph{deterministic software thinking} to drive a \emph{probabilistic processor}.
In traditional software engineering, code behavior is deterministic: identical inputs produce identical outputs, tests can cover all branches, and bugs can be caught with assertions.
Agent behavior, however, is driven by a probabilistic model: the same prompt may yield different outputs across invocations, tests cannot cover all possible reasoning paths, and ``bugs'' manifest as hallucinations or instruction-following failures that resist detection by traditional assertions.

Aura~\cite{aura2025} further demonstrates that once agents enter the OS and mobile environments, permission management (does the agent have access to the contacts list?), authentication (is the caller the genuine user or an attacker?), semantic input pollution (can a malicious SMS hijack the agent?), and memory taint propagation (how is tainted memory cleaned?) all become \emph{kernel-level} concerns, no longer merely application-layer vulnerabilities but foundational security issues for the entire system architecture.

Figure~\ref{fig:challenge_heatmap} maps each challenge to the ICA layers it impacts most, showing primary, secondary, and tertiary impact levels across the full stack.
\input{figures/fig_security_layers.tex}

\subsubsection{Governance Is Not a Retrofit}
These observations imply that governance must not be treated as an after-the-fact security patch or a compliance checklist.
Just as operating system designers learned, through decades of painful experience, that bolting on security after the fact is fundamentally unworkable~\cite{ostep_2023}, agent system governance must be embedded into the architecture from day one.

\subsubsection{Per-Layer Governance in ICA}
Specifically, each layer of ICA (Section~\ref{sec:icam}) requires its own governance mechanisms:

\begin{itemize}[nosep]
\item \textbf{L1---Physical Execution:} hardware-level inference precision guarantees and energy budgets.
Analogous to CPU thermal monitoring and frequency throttling, L1 must ensure that inference precision does not fall below a safety threshold (e.g., quantization must not cause critical judgments to fail) and that GPU/TPU energy consumption remains within bounds.

\item \textbf{L2---Inference Serving:} inference budgets and precision governance.
Each agent's inference calls should have budget caps (e.g., ``at most 10 inference calls per sub-task''); exceeding the budget should trigger failure recovery rather than infinite retries.
KV cache management must ensure that critical state is not inadvertently evicted.

\item \textbf{L3---Context Management:} memory retention and deletion policies.
Sensitive user information (passwords, credit card numbers) should be automatically detected and promptly removed, analogous to the GDPR's ``right to be forgotten.''
Memory entries should carry explicit time-to-live (TTL) annotations; expired memories should be automatically degraded or deleted.

\item \textbf{L4---Semantic Interface:} tool capability annotations and call audits.
Every external tool should carry a standardized capability declaration (``what this tool can do, what it cannot do, what side effects it produces''), and every tool invocation should be recorded in an auditable trace.
The MCP protocol~\cite{mcp_intro_2026} is evolving in this direction.

\item \textbf{L5---Orchestration:} permission approval and failure recovery.
High-risk operations (writing files, sending emails, executing shell commands) must pass through the deterministic control plane's approval workflow.
Failure recovery must have explicit rollback strategies (e.g., git revert) and human-takeover mechanisms.

\item \textbf{L6---Application:} intent confirmation and side-effect declarations.
Before executing any irreversible operation, the system should confirm with the user that its understanding is correct.
After each user task completes, the system should generate a side-effect report (``which files were modified, which emails were sent, which external services were accessed'').
\end{itemize}

\subsubsection{Governance in the Dual-Plane Architecture}
Viewed through the lens of the dual-plane architecture (Section~\ref{sec:framework}), governance is the \emph{central component} of the deterministic control plane.
Its role is not to replace the probabilistic plane's reasoning capability but to set boundaries, maintain audit trails, and provide rollback mechanisms for the results of that reasoning.
Concretely:
the probabilistic plane is responsible for ``understanding user intent, decomposing tasks, and generating tool-call parameters,'' the reasoning capabilities at which LLMs excel;
the deterministic plane is responsible for ``checking whether permissions are satisfied, whether tool calls are safe, whether results are compliant, and whether failures are recoverable,'' the deterministic logic at which traditional software engineering excels.
CaMeL's capability manager~\cite{debenedetti2025camel}, Codex's approval policies~\cite{openai_codex_approvals_2026}, and Claude Code's hooks~\cite{anthropic_claudecode_security_2026} are all concrete implementations of this principle.

This division of labor mirrors a classic design philosophy in operating systems: ``the kernel does not compute on behalf of user programs, but ensures that their computation does not exceed its bounds''~\cite{ostep_2023}.
The kernel does not sort your array, but it ensures your program cannot read another process's memory.
Similarly, the governance layer does not perform reasoning on behalf of the agent, but it ensures that the agent's reasoning outputs do not lead to unauthorized operations, data exfiltration, or unrecoverable failures.
Figure~\ref{fig:ch85:governance} illustrates this division of labor between the probabilistic and deterministic planes.

\input{figures/fig_ch85_governance}

\subsubsection{The Formalization Challenge}
Embedding governance into the architecture still faces a fundamental formalization challenge: how do we define ``security boundaries'' for probabilistic behavior?
Traditional OS security models rest on deterministic state machines: every system call has well-defined preconditions (e.g., ``the file descriptor must be valid'') and postconditions (e.g., ``returns the number of bytes read'').
Agent behavior, however, is probabilistic: the same operation may produce different results in different contexts, and whether a result is ``safe'' often depends on semantic judgment rather than formal verification.
ArbiterOS's policy specification~\cite{arbiteros2025} and IronClaw's behavior-space modeling~\cite{ironclaw2025} represent first steps in this direction, but a practical formal governance framework remains distant.
This is precisely the critical research direction that ICA and the dual-plane architecture delineate for future work.

Within ICA, governance is not localized to a single layer but is a \emph{cross-cutting concern} that manifests differently at each level of the stack, from hardware precision guarantees (L1) to user-facing intent confirmation (L6).
The dual-plane architecture provides the structural mechanism by which governance invariants are enforced: the deterministic control plane acts as the architectural ``kernel'' that mediates all interactions between the probabilistic reasoning engine and the external world, ensuring that Axiom~6 (``every irreversible action must be approved by the deterministic control plane'') is upheld regardless of which ICA layer originates the action.
This perspective transforms governance from an operational afterthought into an architectural invariant, a first-class design principle rather than a retrofit.

%% file: figures/fig_trilemma.tex
\begin{figure}[htbp]
    \centering
    \begin{tikzpicture}[scale=1.05]
    
    \coordinate (A) at (0,   3.5);   
    \coordinate (B) at (-3.0, 0);    
    \coordinate (C) at ( 3.0, 0);    
    
    \fill[gray!8] (A) -- (B) -- (C) -- cycle;
    \draw[gray!40, thick] (A) -- (B) -- (C) -- cycle;
    
    \node[font=\small\bfseries, text=classical!80!black, align=center] at (0, 4.15)
        {Low Latency\\{\scriptsize (fast response)}};
    
    \node[font=\small\bfseries, text=modelnative!80!black, align=center] at (-3.8, -0.55)
        {Low Cost\\{\scriptsize (efficient resource)}};
    
    \node[font=\small\bfseries, text=black!70, align=center] at (3.8, -0.55)
        {High Throughput\\{\scriptsize (batch efficiency)}};
    
    \node[font=\scriptsize, text=gray!55, rotate=54, align=center]
        at (-2.2, 2.0) {small batch,\\dedicated GPU};
    
    \node[font=\scriptsize, text=gray!55, rotate=-54, align=center]
        at (2.2, 2.0) {more GPUs,\\higher cost};
    
    \node[font=\scriptsize, text=gray!55, align=center]
        at (0, -0.35) {large batch, queue longer};
    

    \filldraw[classical!70] (-0.8, 1.8) circle (3.5pt);
    \node[font=\scriptsize, text=classical!80, anchor=north] at (-0.6, 1.65)
        {Interactive agent};
    
    \filldraw[modelnative!70] (1.2, 0.6) circle (3.5pt);
    \node[font=\scriptsize, text=modelnative!80, anchor=east] at (1.0, 0.6)
        {Batch pipeline};
    
    \filldraw[black!55] (0.5, 2.5) circle (3.5pt);
    \node[font=\scriptsize, text=black!65, anchor=east] at (0.3, 2.5)
        {RT serving};
    
    \node[font=\scriptsize, text=gray!50, align=center] at (0, -1.4)
        {Classical analogy: response time -- throughput trade-off in time-sharing OS};
    
    \end{tikzpicture}
    \caption{The latency--throughput--cost trilemma in LLM serving. Any two objectives can be jointly optimised, but improving all three simultaneously is structurally infeasible given the compute-bound prefill / bandwidth-bound decode asymmetry. Representative system designs are plotted as points inside the triangle.}
    \label{fig_trilemma}
    \end{figure}
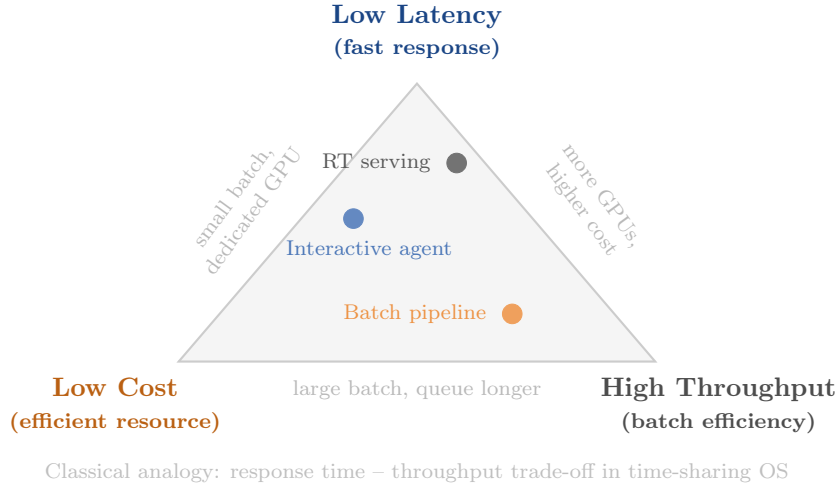

%% file: figures/fig_state_lifecycle.tex
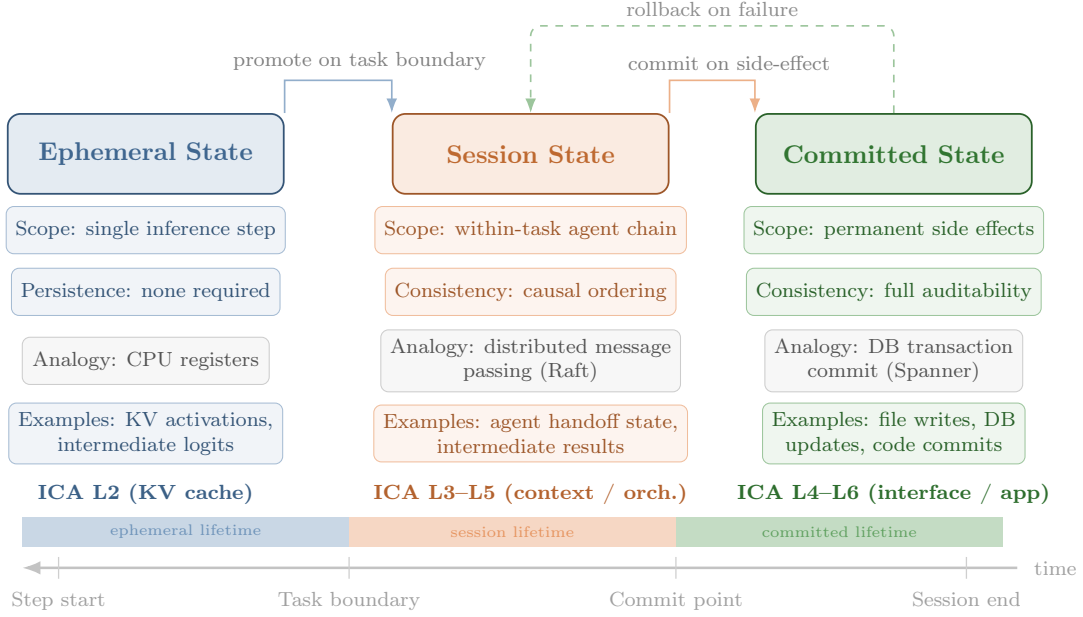
\begin{figure}[htbp]
\centering
\providecolor{ephColor}{HTML}{4A78A8}
\providecolor{sessColor}{HTML}{E07B39}
\providecolor{commColor}{HTML}{3A8A3A}
\providecolor{timeColor}{HTML}{888888}
\resizebox{0.90\textwidth}{!}{%
\begin{tikzpicture}[
    font=\small, >=Latex,
    statebox/.style={draw=#1!70!black, fill=#1!15, rounded corners=5pt,
        minimum width=3.8cm, minimum height=1.1cm,
        align=center, font=\small\bfseries, text=#1!85!black, line width=0.7pt},
    opbox/.style={draw=#1!50, fill=#1!8, rounded corners=3pt,
        minimum width=3.4cm, minimum height=0.65cm,
        align=center, font=\scriptsize, text=#1!70!black, line width=0.4pt},
    analogbox/.style={draw=gray!40, fill=gray!6, rounded corners=3pt,
        minimum width=3.4cm, minimum height=0.65cm,
        align=center, font=\scriptsize, text=gray!65!black, line width=0.4pt},
    arrowstyle/.style={-{Latex[length=2mm]}, thick, line width=0.65pt},
    timelabel/.style={font=\scriptsize, text=timeColor!80},
]

\node[statebox=ephColor] (E) at (-4.8, 2.2)
    {Ephemeral State};
\node[opbox=ephColor] at (-4.8, 1.15)  {Scope: single inference step};
\node[opbox=ephColor] at (-4.8, 0.30)  {Persistence: none required};
\node[analogbox]      at (-4.8, -0.65) {Analogy: CPU registers};
\node[opbox=ephColor] at (-4.8, -1.65) {Examples: KV activations,\\intermediate logits};

\node[statebox=sessColor] (S) at (0.5, 2.2)
    {Session State};
\node[opbox=sessColor] at (0.5, 1.15)  {Scope: within-task agent chain};
\node[opbox=sessColor] at (0.5, 0.30)  {Consistency: causal ordering};
\node[analogbox]      at (0.5, -0.65) {Analogy: distributed message\\passing (Raft)};
\node[opbox=sessColor] at (0.5, -1.65) {Examples: agent handoff state,\\intermediate results};

\node[statebox=commColor] (C) at (5.5, 2.2)
    {Committed State};
\node[opbox=commColor] at (5.5, 1.15)  {Scope: permanent side effects};
\node[opbox=commColor] at (5.5, 0.30)  {Consistency: full auditability};
\node[analogbox]       at (5.5, -0.65) {Analogy: DB transaction\\commit (Spanner)};
\node[opbox=commColor] at (5.5, -1.65) {Examples: file writes, DB\\updates, code commits};

\node[font=\scriptsize\bfseries, ephColor!80!black, anchor=north] at (-4.8, -2.2)
    {ICA L2 (KV cache)};
\node[font=\scriptsize\bfseries, sessColor!80!black, anchor=north] at (0.5, -2.2)
    {ICA L3--L5 (context / orch.)};
\node[font=\scriptsize\bfseries, commColor!80!black, anchor=north] at (5.5, -2.2)
    {ICA L4--L6 (interface / app)};

\fill[ephColor!30] (-6.5,-3.2) rectangle (-2.0,-2.8);
\node[font=\tiny, text=ephColor!80] at (-4.25, -3.0) {ephemeral lifetime};

\fill[sessColor!30] (-2.0,-3.2) rectangle (2.5,-2.8);
\node[font=\tiny, text=sessColor!80] at (0.25, -3.0) {session lifetime};

\fill[commColor!30] (2.5,-3.2) rectangle (7.0,-2.8);
\node[font=\tiny, text=commColor!80] at (4.75, -3.0) {committed lifetime};

\draw[{Latex[length=2.5mm]}-, thick, timeColor!50, line width=1.0pt]
    (-6.5, -3.5) -- (7.2, -3.5);
\node[timelabel, anchor=west] at (7.3, -3.5) {time};
\foreach \x/\lbl in {-6.0/Step start, -2.0/Task boundary, 2.5/Commit point, 6.5/Session end}{
    \draw[timeColor!40, line width=0.5pt] (\x, -3.35) -- (\x, -3.65);
    \node[timelabel, anchor=north] at (\x, -3.7) {\lbl};
}

\draw[arrowstyle, ephColor!60]
    (E.north east) -- ++(0, 0.45) -|
    node[above, font=\scriptsize, text=black!50, pos=0.35] {promote on task boundary}
    (S.north west);
\draw[arrowstyle, sessColor!60]
    (S.north east) -- ++(0, 0.45) -|
    node[above, font=\scriptsize, text=black!50, pos=0.35] {commit on side-effect}
    (C.north west);

\draw[arrowstyle, commColor!50, dashed, rounded corners=4pt]
    (C.north) -- ++(0, 1.2)
    -- ++(-5.0, 0)
    node[above, midway, font=\scriptsize, text=black!45] {rollback on failure}
    -- (S.north);

\end{tikzpicture}}%
\caption{Three-tier agent state lifecycle. \textbf{Ephemeral state} (left) lives only within a single inference step and requires no persistence, analogous to CPU registers. \textbf{Session state} (centre) spans an agent task chain and requires causal consistency guarantees, analogous to distributed message passing. \textbf{Committed state} (right) constitutes permanent, auditable side effects and requires full rollback support, analogous to a database transaction commit. Arrows indicate promotion on task boundaries and rollback paths on failure.}
\label{fig:state_lifecycle}
\end{figure}

%% file: figures/fig_ch82_state.tex
\begin{figure}[htbp]
  \centering
  \includegraphics[width=0.90\textwidth]{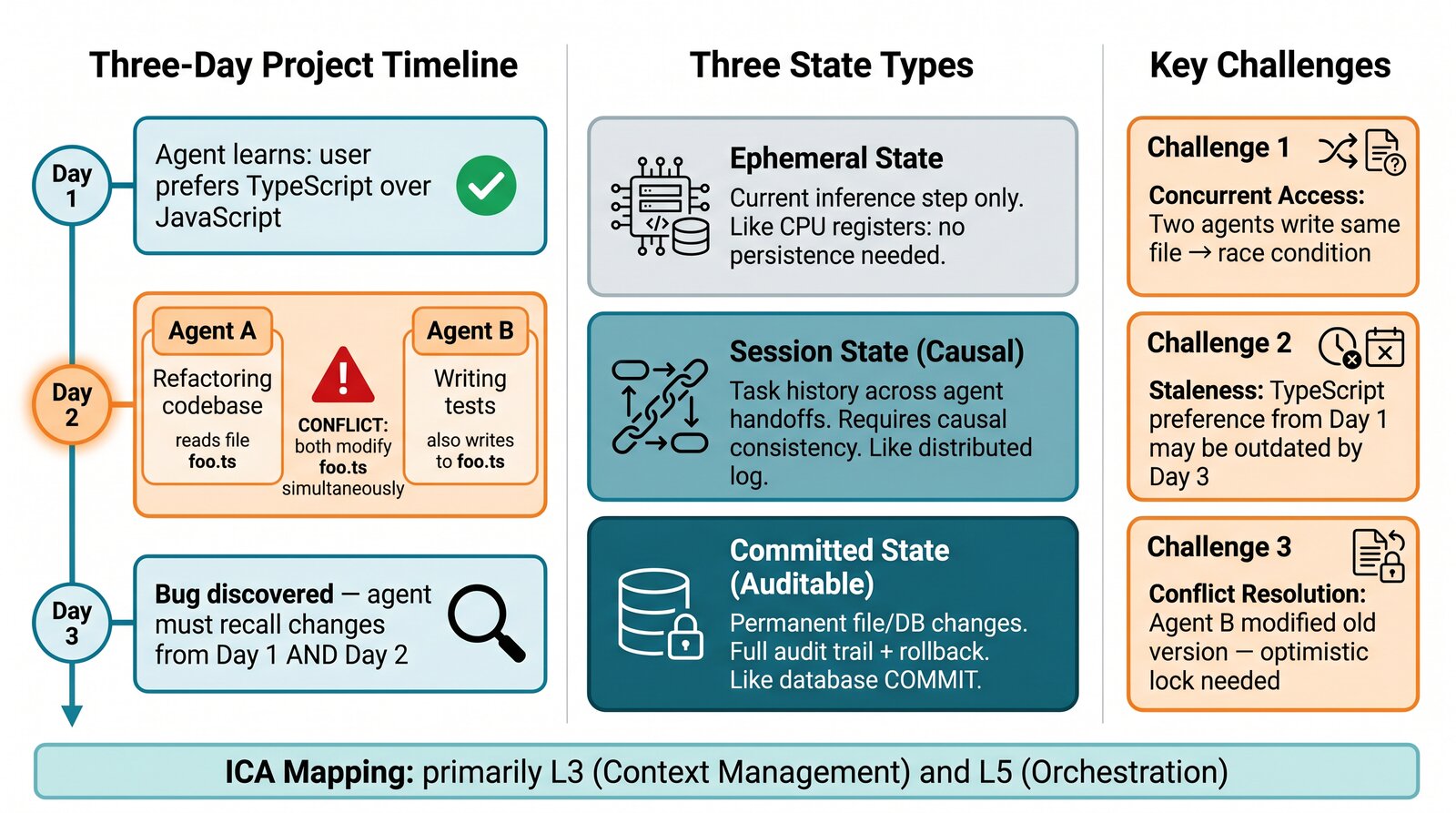}
  \caption{Cross-session state management in an AI agent system.
  A three-day project illustrates the three categories of persistent state:
  \emph{ephemeral state} (current inference step only, like CPU registers),
  \emph{session state with causal ordering} (task history requiring causal consistency),
  and \emph{committed state} (permanent changes to files or databases, requiring full audit and rollback support).
  The Day~2 concurrent scenario highlights the race-condition and conflict-resolution
  challenges that arise when multiple agents share mutable state.
  Within ICA, this challenge primarily spans L3 (context management) and L5 (orchestration).}
  \label{fig:ch82:state}
\end{figure}

%% file: figures/fig_security_layers.tex
\begin{figure}[h]
\centering
\begin{tikzpicture}[
    scale=0.78,
    transform shape,
    primary/.style={fill=red!35, draw=red!50, thick, rounded corners=1pt},
    secondary/.style={fill=orange!25, draw=orange!40, thick, rounded corners=1pt},
    tertiary/.style={fill=yellow!18, draw=yellow!35, thick, rounded corners=1pt},
    notapplic/.style={fill=gray!8, draw=gray!25, thick, rounded corners=1pt},
    colhead/.style={font=\scriptsize\bfseries, align=center, text width=1.8cm},
    rowhead/.style={font=\scriptsize\bfseries, align=right, anchor=east, text width=3.4cm},
    lawlabel/.style={font=\scriptsize\bfseries, align=center, anchor=west},
    celltxt/.style={font=\tiny, align=center},
    axlbl/.style={font=\footnotesize\bfseries},
]

\def\cellw{2.0}   
\def\cellh{1.15}   
\def\colsep{0.12}  
\def\rowsep{0.10}  
\def\rowoff{5}     
\def\coloff{5}     

\def\xstart{4.0}   
\def\ystart{6.2}    

\pgfmathsetmacro{\cx}{\xstart + 0*(\cellw+\colsep) + \cellw/2}
\node[colhead, text=orange!70!black] at (\cx, \ystart + 0.45) {L1\\Physical};
\pgfmathsetmacro{\cx}{\xstart + 1*(\cellw+\colsep) + \cellw/2}
\node[colhead, text=orange!70!black] at (\cx, \ystart + 0.45) {L2\\Inference};
\pgfmathsetmacro{\cx}{\xstart + 2*(\cellw+\colsep) + \cellw/2}
\node[colhead, text=orange!70!black] at (\cx, \ystart + 0.45) {L3\\Context};
\pgfmathsetmacro{\cx}{\xstart + 3*(\cellw+\colsep) + \cellw/2}
\node[colhead, text=blue!60!black] at (\cx, \ystart + 0.45) {L4\\Semantic};
\pgfmathsetmacro{\cx}{\xstart + 4*(\cellw+\colsep) + \cellw/2}
\node[font=\scriptsize\bfseries, align=center, text width=2.2cm, text=blue!60!black] at (\cx, \ystart + 0.45) {L5\\Orchestration};
\pgfmathsetmacro{\cx}{\xstart + 5*(\cellw+\colsep) + \cellw/2}
\node[colhead, text=blue!60!black] at (\cx, \ystart + 0.45) {L6\\Application};

\pgfmathsetmacro{\modelnativex}{\xstart + 1.5*\cellw + \colsep}
\node[font=\tiny\itshape, orange!70!black, anchor=south] at (\modelnativex, \ystart + 1.25) {Probabilistic};
\pgfmathsetmacro{\classicalx}{\xstart + 3*(\cellw+\colsep) + 1.5*\cellw + 1.5*\colsep}
\node[font=\tiny\itshape, blue!60!black, anchor=south] at (\classicalx, \ystart + 1.25) {Deterministic};

\pgfmathsetmacro{\rx}{\xstart - 0.3}
\pgfmathsetmacro{\ryA}{\ystart - 0*(\cellh+\rowsep) - \cellh/2}
\pgfmathsetmacro{\ryB}{\ystart - 1*(\cellh+\rowsep) - \cellh/2}
\pgfmathsetmacro{\ryC}{\ystart - 2*(\cellh+\rowsep) - \cellh/2}
\pgfmathsetmacro{\ryD}{\ystart - 3*(\cellh+\rowsep) - \cellh/2}
\pgfmathsetmacro{\ryE}{\ystart - 4*(\cellh+\rowsep) - \cellh/2}
\node[rowhead] at (\rx, \ryA)
    {Latency--Throughput\\--Cost Trilemma};
\node[rowhead] at (\rx, \ryB)
    {State\\Management};
\node[rowhead] at (\rx, \ryC)
    {Interface\\Drift};
\node[rowhead] at (\rx, \ryD)
    {Security \&\\Privacy};
\node[rowhead] at (\rx, \ryE)
    {Governance \&\\Accountability};

\foreach \c/\sty in {0/primary, 1/primary, 2/tertiary, 3/secondary, 4/notapplic, 5/notapplic}{
    \pgfmathsetmacro{\xll}{\xstart + \c*(\cellw+\colsep)}
    \pgfmathsetmacro{\yll}{\ystart - \cellh}
    \draw[\sty] (\xll, \yll) rectangle +(\cellw, \cellh);
}
\foreach \c/\lab in {0/Primary, 1/Primary, 2/Tertiary, 3/Secondary, 4/--, 5/--}{
    \pgfmathsetmacro{\cx}{\xstart + \c*(\cellw+\colsep) + \cellw/2}
    \pgfmathsetmacro{\cy}{\ystart - \cellh + \cellh/2}
    \node[celltxt] at (\cx, \cy) {\lab};
}

\foreach \c/\sty in {0/notapplic, 1/primary, 2/primary, 3/secondary, 4/tertiary, 5/notapplic}{
    \pgfmathsetmacro{\xll}{\xstart + \c*(\cellw+\colsep)}
    \pgfmathsetmacro{\yll}{\ystart - 1*(\cellh+\rowsep) - \cellh}
    \draw[\sty] (\xll, \yll) rectangle +(\cellw, \cellh);
}
\foreach \c/\lab in {0/--, 1/Primary, 2/Primary, 3/Secondary, 4/Tertiary, 5/--}{
    \pgfmathsetmacro{\cx}{\xstart + \c*(\cellw+\colsep) + \cellw/2}
    \pgfmathsetmacro{\cy}{\ystart - 1*(\cellh+\rowsep) - \cellh + \cellh/2}
    \node[celltxt] at (\cx, \cy) {\lab};
}

\foreach \c/\sty in {0/notapplic, 1/notapplic, 2/secondary, 3/primary, 4/primary, 5/tertiary}{
    \pgfmathsetmacro{\xll}{\xstart + \c*(\cellw+\colsep)}
    \pgfmathsetmacro{\yll}{\ystart - 2*(\cellh+\rowsep) - \cellh}
    \draw[\sty] (\xll, \yll) rectangle +(\cellw, \cellh);
}
\foreach \c/\lab in {0/--, 1/--, 2/Secondary, 3/Primary, 4/Primary, 5/Tertiary}{
    \pgfmathsetmacro{\cx}{\xstart + \c*(\cellw+\colsep) + \cellw/2}
    \pgfmathsetmacro{\cy}{\ystart - 2*(\cellh+\rowsep) - \cellh + \cellh/2}
    \node[celltxt] at (\cx, \cy) {\lab};
}

\foreach \c/\sty in {0/notapplic, 1/tertiary, 2/secondary, 3/notapplic, 4/primary, 5/primary}{
    \pgfmathsetmacro{\xll}{\xstart + \c*(\cellw+\colsep)}
    \pgfmathsetmacro{\yll}{\ystart - 3*(\cellh+\rowsep) - \cellh}
    \draw[\sty] (\xll, \yll) rectangle +(\cellw, \cellh);
}
\foreach \c/\lab in {0/--, 1/Tertiary, 2/Secondary, 3/--, 4/Primary, 5/Primary}{
    \pgfmathsetmacro{\cx}{\xstart + \c*(\cellw+\colsep) + \cellw/2}
    \pgfmathsetmacro{\cy}{\ystart - 3*(\cellh+\rowsep) - \cellh + \cellh/2}
    \node[celltxt] at (\cx, \cy) {\lab};
}

\foreach \c/\sty in {0/notapplic, 1/notapplic, 2/tertiary, 3/primary, 4/primary, 5/secondary}{
    \pgfmathsetmacro{\xll}{\xstart + \c*(\cellw+\colsep)}
    \pgfmathsetmacro{\yll}{\ystart - 4*(\cellh+\rowsep) - \cellh}
    \draw[\sty] (\xll, \yll) rectangle +(\cellw, \cellh);
}
\foreach \c/\lab in {0/--, 1/--, 2/Tertiary, 3/Primary, 4/Primary, 5/Secondary}{
    \pgfmathsetmacro{\cx}{\xstart + \c*(\cellw+\colsep) + \cellw/2}
    \pgfmathsetmacro{\cy}{\ystart - 4*(\cellh+\rowsep) - \cellh + \cellh/2}
    \node[celltxt] at (\cx, \cy) {\lab};
}

\pgfmathsetmacro{\bndx}{\xstart + 3*(\cellw+\colsep) - \colsep/2}
\pgfmathsetmacro{\bndtop}{\ystart + 0.15}
\pgfmathsetmacro{\bndbot}{\ystart - 4*(\cellh+\rowsep) - \cellh - 0.15}
\draw[very thick, dashed, black!55] (\bndx, \bndtop) -- (\bndx, \bndbot);

\pgfmathsetmacro{\bndmid}{(\bndtop + \bndbot)/2}
\node[font=\tiny\itshape, black!55, rotate=90, anchor=south] at (\bndx + 0.25, \bndmid)
    {Dual-Plane Boundary};

\pgfmathsetmacro{\lawx}{\xstart + 6*(\cellw+\colsep) + 0.35}

\pgfmathsetmacro{\ry}{\ystart - 0*(\cellh+\rowsep) - \cellh/2}
\node[lawlabel, red!70!black] at (\lawx, \ry) {Heur.~I};

\pgfmathsetmacro{\ry}{\ystart - 1*(\cellh+\rowsep) - \cellh/2}
\node[lawlabel, red!70!black] at (\lawx, \ry) {Heur.~II};

\pgfmathsetmacro{\ry}{\ystart - 2*(\cellh+\rowsep) - \cellh/2}
\node[lawlabel, red!70!black] at (\lawx, \ry) {Heur.~III};

\pgfmathsetmacro{\ry}{\ystart - 3*(\cellh+\rowsep) - \cellh/2}
\node[lawlabel, red!70!black] at (\lawx, \ry) {Heur.~II,\,III};

\pgfmathsetmacro{\ry}{\ystart - 4*(\cellh+\rowsep) - \cellh/2}
\node[lawlabel, red!70!black] at (\lawx, \ry) {Heur.~III};

\node[font=\scriptsize\bfseries, red!70!black, anchor=west] at (\lawx, \ystart + 0.45)
    {Relevant Heuristic};

\pgfmathsetmacro{\axleft}{\xstart - 3.6}
\pgfmathsetmacro{\axmid}{\ystart - 2*(\cellh+\rowsep)}
\node[axlbl, rotate=90, anchor=south] at (\axleft, \axmid)
    {ICA Layers $\longrightarrow$};

\pgfmathsetmacro{\axbotx}{\xstart + 2.5*(\cellw+\colsep)}
\pgfmathsetmacro{\axboty}{\ystart - 4*(\cellh+\rowsep) - \cellh - 0.5}
\node[axlbl, anchor=north] at (\axbotx, \axboty) {};

\def\legy{-1.0}
\def\legx{2.5}

\draw[primary] (\legx, \legy) rectangle +(0.35, 0.35);
\node[font=\scriptsize, anchor=west] at (\legx + 0.55, \legy + 0.175)
    {Primary impact\hspace{0.4em}};

\pgfmathsetmacro{\lx}{\legx + 3.1}
\draw[secondary] (\lx, \legy) rectangle +(0.35, 0.35);
\node[font=\scriptsize, anchor=west] at (\lx + 0.55, \legy + 0.175)
    {Secondary impact\hspace{0.4em}};

\pgfmathsetmacro{\lx}{\legx + 6.5}
\draw[tertiary] (\lx, \legy) rectangle +(0.35, 0.35);
\node[font=\scriptsize, anchor=west] at (\lx + 0.55, \legy + 0.175)
    {Tertiary impact\hspace{0.4em}};

\pgfmathsetmacro{\lx}{\legx + 9.6}
\draw[notapplic] (\lx, \legy) rectangle +(0.35, 0.35);
\node[font=\scriptsize, anchor=west] at (\lx + 0.55, \legy + 0.175)
    {Not applicable\hspace{0.4em}};

\pgfmathsetmacro{\lx}{\legx + 12.5}
\draw[very thick, dashed, black!55] (\lx, \legy + 0.175) -- ++(0.5, 0);
\node[font=\scriptsize, anchor=west] at (\lx + 0.7, \legy + 0.175)
    {Dual-plane boundary};

\pgfmathsetmacro{\modelleft}{\xstart}
\pgfmathsetmacro{\modelright}{\xstart + 2*\cellw + \colsep}
\pgfmathsetmacro{\classleft}{\xstart + 3*(\cellw+\colsep)}
\pgfmathsetmacro{\classright}{\xstart + 5*(\cellw+\colsep) + \cellw}
\pgfmathsetmacro{\bracetop}{\ystart + 1.1}

\draw[thick, orange!70!black]
    (\modelleft, \bracetop) -- (\modelleft, \bracetop + 0.15);
\draw[thick, orange!70!black]
    (\modelright, \bracetop) -- (\modelright, \bracetop + 0.15);
\draw[thick, orange!70!black]
    (\modelleft, \bracetop + 0.15) -- (\modelright, \bracetop + 0.15);

\draw[thick, blue!60!black]
    (\classleft, \bracetop) -- (\classleft, \bracetop + 0.15);
\draw[thick, blue!60!black]
    (\classright, \bracetop) -- (\classright, \bracetop + 0.15);
\draw[thick, blue!60!black]
    (\classleft, \bracetop + 0.15) -- (\classright, \bracetop + 0.15);

\end{tikzpicture}
\caption{Challenge-to-ICA-layer impact heatmap. Each cell indicates the severity of a given challenge at the corresponding ICA layer: \textcolor{red!50}{\textbf{primary}} (direct, critical impact), \textcolor{orange!40}{\textbf{secondary}} (significant but indirect), \textcolor{yellow!50!black}{\textbf{tertiary}} (moderate, contextual), or \textcolor{gray!40}{not applicable}. The thick dashed line marks the graded crossover between probabilistic-execution layers (L1--L3) and deterministic-control layers (L4--L6). The right-margin column identifies the most relevant design heuristic for each challenge row.}
\label{fig:challenge_heatmap}
\end{figure}

%% file: figures/fig_ch85_governance.tex
\begin{figure}[htbp]
  \centering
  \includegraphics[width=0.90\textwidth]{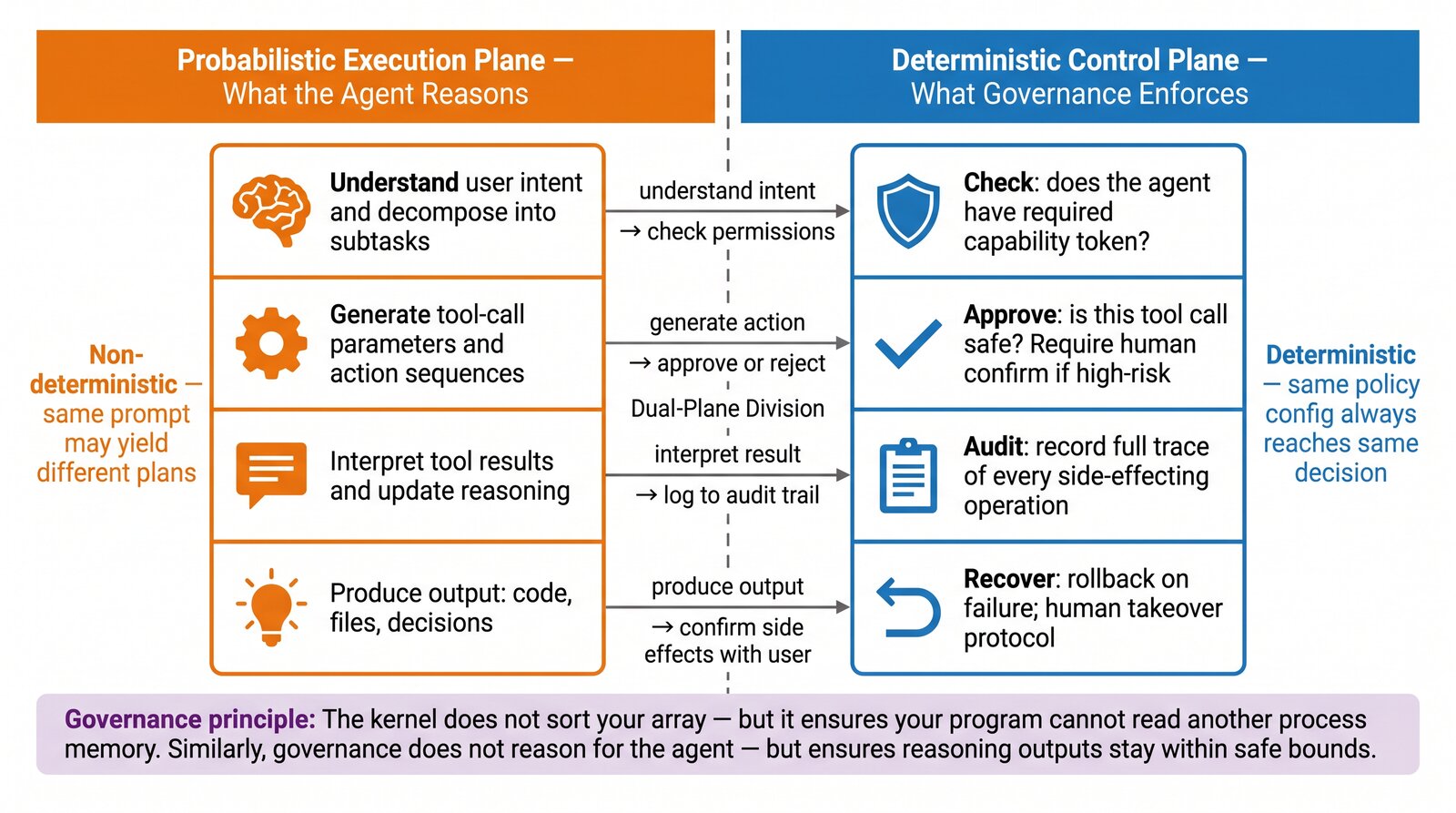}
  \caption{Governance as a structural division of labor in the dual-plane architecture.
  The probabilistic execution plane (left) handles what the agent \emph{can} reason about:
  understanding user intent, generating action sequences, interpreting results, and producing outputs.
  The deterministic control plane (right) enforces what \emph{should} happen:
  checking capability permissions, approving or rejecting tool calls, maintaining a full audit trail,
  and providing rollback and human-takeover mechanisms.
  This mirrors the classical OS principle: the kernel does not sort your array,
  but it ensures your program cannot read another process's memory.
  Governance is not a retrofit layer but the structural core of the deterministic control plane,
  a first-class architectural invariant rather than an operational afterthought.}
  \label{fig:ch85:governance}
\end{figure}

%% file: en_sections/section_paradigm.tex
\section{Paradigm Discussion}
\label{sec:paradigm}

\subsection{The Intelligent Compute Core: From Silicon to Substrate}

The preceding sections established the ICA analogy framework and its layered architecture. Figure~\ref{fig:paradigm_timeline} summarizes the structural parallels between classical CPU evolution and LLM system development. This subsection examines the core itself: the large language model as \textit{processor} within an intelligent system, and how five decades of CPU evolution illuminate its trajectory. The argument proceeds through six stages: what the core is, how it scales, how heterogeneous designs circumvent scaling limits, what specialized units surround it, what substrates might eventually replace silicon, and how to measure whether it works.

\input{figures/fig_paradigm_timeline}

\subsubsection{The LLM as Processor: Why the Compute Core Need Not Remember}

The Instruction Set Architecture (ISA) abstraction is among the most consequential ideas in computer architecture. As Patterson and Hennnessy demonstrated across decades of scholarship, a CPU's role is \textit{faithful instruction execution}, not knowledge storage \cite{patterson_hennessy_2017}. The same program compiles to the same ISA whether it runs on an Intel Core i9 or an AMD Ryzen; the hardware beneath evolves independently of the software above. This decoupling, formalized in the RISC movement and refined through successive x86 generations, enabled the entire modern software ecosystem.

Current LLMs have not yet internalized this lesson. The dominant pretraining paradigm treats the model as an encyclopedia: ingest the internet, compress knowledge into weights, and answer questions from memory. This is the ENIAC paradigm: single-purpose hardware wired for a specific application \cite{patterson_hennessy_2017}. Just as CPUs evolved from application-specific logic (ENIAC, 1945) to general-purpose processors implementing a stable ISA (x86, 1978), LLMs must shift from ``know everything'' pretraining to ``be a smart processor'' that faithfully transforms inputs under orchestration.

The token interface, context window, and tool-use protocols form the nascent \textbf{Intelligent ISA}: a stable contract between the model compute core (ICA L1, Physical Execution) and the OS/agent layer (L2, Inference Serving). Under this contract, the core's obligation is reliable instruction execution: given a structured prompt, produce a faithful response. Knowledge retrieval, long-term memory, and multi-step planning belong to higher layers, just as file I/O and process scheduling belong to the OS, not the CPU.

This decoupling has immediate practical benefits. Safety mechanisms, anti-jailbreak defenses, and commonsense guardrails can migrate to the OS layer: context filters, safety classifiers, and output validators wrap the core rather than being baked into its weights. The analogy is precise: memory protection and privilege rings moved from hardware into the operating system as architecture matured \cite{ritchie_thompson_1974}. A lean model core, surrounded by orchestration-layer safety, is both more auditable and more capable than a monolithic model attempting to internalize every constraint.

The ISA analogy also enables architectural diversity. Just as x86 runs on radically different Intel and AMD microarchitectures, the same agent framework can run on Transformer, RWKV, Mamba, or hybrid cores, so long as each implements the intelligent compute contract \cite{hennessy_patterson_2019}. Competition shifts from ``whose model knows more?'' to ``whose processor is faster, cheaper, and more faithful?'' The ICA framework accommodates this diversity at L1 while maintaining a uniform L2 interface.

\subsubsection{Frequency Scaling and Thinking Modes: The Power Wall Revisited}

In 1974, Robert Dennard and colleagues published a scaling law that governed CPU design for three decades: as transistors shrink, power density remains constant, enabling ever-higher clock frequencies without exceeding thermal limits \cite{dennard_1974}. When Dennard scaling collapsed around 2005, the industry hit the \textbf{power wall}. Frequency stalled near 4\,GHz, and the only path forward was multi-core designs that did more work per cycle rather than more cycles per second \cite{esmaeilzadeh_2011}.

LLM inference faces an analogous \textbf{energy wall}. Token generation cost scales with parameter count; serving a 405B-parameter model requires datacenter-scale power. The era of ``just make it bigger'' is ending for the same reason the GHz race ended: physics and economics impose hard limits on how much energy a single compute element can dissipate.

The industry's response in both domains follows parallel tracks. Intel's Turbo Boost automatically increases clock frequency for bursty, demanding workloads, then throttles back for routine ones. LLM \textbf{thinking modes} (extended chain-of-thought, extended reasoning) operate identically: the system automatically expends more tokens, and therefore more compute, on difficult tasks while conserving resources on easy ones. Extended reasoning at maximum token expenditure is the analog of \textit{overclocking}: marginal quality gains at disproportionate compute cost, trading efficiency for peak throughput. Just as sustained overclocking voids warranties and degrades hardware, sustained maximum thinking mode is economically unsustainable.

The \textbf{dark silicon} problem, formalized by Esmaeilzadeh et al.\ (2011), maps directly onto LLM deployment. In a chip with billions of transistors, energy budgets allow only a fraction to be active simultaneously; the rest are ``dark.'' In a monolithic LLM, not all parameters can be ``lit up'' for every token within serving latency and cost constraints. This motivates sparse activation and Mixture-of-Experts architectures, where only relevant parameter subsets are engaged per input, the intellectual descendants of the multi-core pivot \cite{moore_1965, amdahl_1967}.

The historical lesson is clear. Just as the power wall forced the CPU industry from GHz races to heterogeneous multi-core designs, the inference energy wall will force LLM systems from monolithic model scaling to heterogeneous model orchestration. The next subsection examines how this transition is already underway.

\subsubsection{big.LITTLE Intelligence: Heterogeneous Model Orchestration}

In 2011, ARM introduced the big.LITTLE architecture: pairing a high-performance Cortex-A15 with an energy-efficient Cortex-A7 on the same die, both implementing the same ARM ISA \cite{biglittle_arm_2012}. The insight was that most workloads do not require peak performance; a small, efficient core handles routine tasks while the powerful core activates only for demanding ones. The scheduling strategy evolved through three generations: \textit{cluster migration} (only big \textit{or} LITTLE active), \textit{HMP} (all cores simultaneously active with OS-level task placement), and \textit{EAS} (energy-aware scheduling integrating hardware telemetry into the Linux kernel scheduler, kernel 5.0, 2019) \cite{eas_linux_2019}.

This three-stage evolution maps directly onto the maturation path of agent frameworks. Single-model LLM systems correspond to cluster migration: one model handles everything. Plan-vs-execute routing, where a large model generates a plan and a small model executes each step, corresponds to HMP: multiple models active simultaneously with coarse-grained task assignment. Cost-aware dynamic model selection, where the orchestrator routes each subtask to the model that minimizes expected cost at acceptable quality, corresponds to EAS: energy-aware, telemetry-driven scheduling \cite{kumar_2003_singleISA, saez_2022_thread_director}. Figure~\ref{fig:biglittle} illustrates this three-stage parallel.

\input{figures/fig_biglittle}

Production agent systems already implement this pattern. Claude Code routes Sonnet-class models for planning and Haiku-class models for execution within the same agent loop \cite{dive_claudecode_2026}. Devin uses multi-agent orchestration with dynamic planning checkpoints, switching between model tiers as task complexity shifts. These are not prototypes; they are deployed systems serving millions of queries.

Speculative decoding extends the analogy into the inference serving layer. In schemes like EAGLE and Medusa, a small draft model proposes tokens that a large model verifies, achieving 2--3$\times$ speedup without quality degradation \cite{li_2024_eagle}. This is the software analog of heterogeneous scheduling: the small model (LITTLE core) handles routine token generation, and the large model (big core) intervenes only to verify or correct. The parallel extends to failure modes: just as EAS falls back to SMP scheduling under overutilization, agent systems fall back from cost-optimized small models to expensive large models when workload exceeds the small model's capability envelope.

Mixture-of-Experts architectures represent the \textit{intra-model} analog of heterogeneous computing. DeepSeekMoE segments experts into fine-grained units, activating sparse subsets per token \cite{deepseek_2024_moe}. HMoE explicitly uses experts of varying sizes: smaller experts handle simple patterns, larger ones tackle complexity, producing heterogeneous computing at the parameter level \cite{wang_2024_hmoe}. This is big.LITTLE inside a single model: the gating network functions as the Thread Director, routing tokens to the appropriately sized expert.

\subsubsection{Specialized Coprocessors: The Rise of Small Special-Purpose Models}

Modern System-on-Chip designs dedicate substantial silicon to specialized coprocessors: GPUs for parallel graphics, NPUs for neural inference, ISPs for image signal processing, DSPs for audio. The general-purpose CPU orchestrates these units but performs little of the actual domain-specific work. The same pattern is emerging in intelligent systems, where small specialized models handle tasks that do not require (and should not pay the cost of) a full LLM invocation.

\textbf{Safety models} are the security coprocessor of the intelligent stack. Dedicated content moderation classifiers, trained specifically for harm detection, achieve GPT-4-level safety performance at a fraction of the cost \cite{harmbench_2024, airbench_2024}. These models are the analog of TPM modules and hardware security enclaves: dedicated parameters for threat detection, isolated from the general-purpose core, auditable and updatable independently. In the ICA framework, they operate at L2 (Inference Serving) as guardrails around the L1 compute core.

\textbf{Context processing models} for retrieval-augmented generation parallel the cache hierarchy in classical systems. Cross-encoder rankers and dual-channel retrieval models manage which context reaches the LLM core, just as L1/L2/L3 caches with dedicated management hardware reduce main memory access \cite{slm_edge_2025}. The retrieval model is the cache controller; the vector database is main memory; the LLM core is the CPU. A well-tuned retrieval pipeline can reduce the large model's context window requirements by an order of magnitude, the same efficiency gain that cache hierarchies provide over flat memory access.

\textbf{Small Language Models} (under 3B parameters) are the edge-compute analog: deployable on-device with acceptable latency, handling routine inference without cloud round-trips \cite{llama32_2024, phi3_2024}. In the big.LITTLE framework, these are the LITTLE cores: Cortex-A7 equivalents that handle background classification, format conversion, entity extraction, and other deterministic tasks. Their defining characteristic is not that they are ``almost as good'' as large models, but that they are \textit{the right tool for specific jobs}, leaving the large model free for tasks that genuinely require its capability.

The architectural blueprint is the Apple M1 SoC: 4P+4E CPU cores, integrated GPU, 16-core Neural Engine, all sharing unified memory \cite{apple_m1_2020}. The LLM-system equivalent is a heterogeneous ensemble: large planning model, small execution models, safety classifiers, retrieval rankers, and specialized classifiers, all sharing a common context pool under unified orchestration. The ICA L1 layer hosts this diversity; L2 provides the scheduling and memory management that makes the ensemble function as a coherent system.

\subsubsection{Beyond Silicon: Neuromorphic, Biological, and Quantum Substrates}

The Hennessy--Patterson Turing Lecture (2019) called for domain-specific architectures as the path beyond the end of Dennard scaling and Moore's Law \cite{hennessy_patterson_2019}. The same imperative drives exploration of non-Von Neumann substrates for intelligence compute. This is not speculative futurism; it is the substrate evolution that the ISA argument predicts and requires.

\textbf{Neuromorphic hardware} has reached deployment scale. Intel's Hala Point system integrates 1,152 Loihi 2 chips, implementing 1.15 billion neurons with event-driven, spike-based computation that achieves 100--1000$\times$ energy efficiency for suitable workloads \cite{hala_point_2024}. SpiNNaker2 deploys approximately 5 million ARM cores in a brain-inspired topology \cite{spinnaker2_2024}. Early work has already demonstrated LLM inference, including MatMul-free language models, running on Loihi 2 chips \cite{abreu_2025_loihi2_llm}. These are not general-purpose replacements for GPUs; they are domain-specific accelerators for the spiking and event-driven workloads that biological neural computation favors.

\textbf{Biological computing} has crossed from research to product. Cortical Labs' CL1, which shipped in March 2025, is the first commercial biological computer: living neurons cultured on a microelectrode array, interfacing with digital systems \cite{cortical_2025_cl1}. The FlyWire Consortium's complete Drosophila connectome (Nature, 2024) provides a complete wiring diagram of an adult brain \cite{flywire_2024}. The Organoid Intelligence roadmap formalizes the path from current proof-of-concept systems to useful biological processors \cite{hartung_2023_oi}.

\textbf{Compute-in-Memory} architectures achieve dramatic efficiency gains by eliminating the von Neumann bottleneck. Analog in-memory computing for LLM attention mechanisms demonstrates 100$\times$ latency reduction and 10,000$\times$ energy reduction compared to digital implementations \cite{wolters2024computeinmemory}. The Taalas HC1 chip takes the logical extreme: permanently hardwiring Llama 3.1 8B weights into Mask ROM, achieving 17,000 tokens/sec with no dynamic memory access at all \cite{taalas_2026_hc1}.

\textbf{Quantum computing} for LLMs remains early-stage. Quixer (Quantinuum, 2024) is the first quantum-native transformer running on real hardware \cite{khatri_2024_quixer}. Theoretical work explores quantum linear algebra for transformer inference \cite{guo_2024_quantum_transformer}, but fault-tolerant, application-scale quantum computing is estimated at 5--15+ years away. Hybrid quantum-classical approaches are the near-term path.

These diverse substrates reinforce the ISA argument from Section~1. A stable intelligent compute interface, the ``Intelligent ISA,'' allows the physical substrate to evolve from silicon GPUs to neuromorphic chips to biological neural tissue to quantum processors without breaking the software stack. Just as the x86 ISA insulated software from the transition from vacuum tubes to transistors to integrated circuits, the token/tool-use protocol insulates agent frameworks from the transition from Transformer weights on GPUs to spike patterns on Loihi chips to living neurons on CL1 arrays. The ICA L1 layer abstracts this substrate diversity; the L2 and above layers never need to know what physical medium executes their instructions.

\subsubsection{Benchmarking the Core: The Bucket Principle and Worst-Case Evaluation}

Classical CPU benchmarking evolved through a painful maturation. In the 1990s, ``megahertz marketing'' convinced consumers that clock speed was the sole measure of processor quality. The reality was more complex: pipeline depth, cache size, branch prediction accuracy, and instruction-level parallelism all affected real performance. The industry eventually converged on holistic suites like SPEC and Geekbench that measured diverse workloads, not peak clock rates \cite{spec_benchmark}.

LLM benchmarking faces the same maturation challenge. MMLU averages and leaderboard scores are the ``megahertz'' of intelligent systems: single numbers that obscure more than they reveal. Models scoring 90\% on standard tests may score only 2\% on unseen problems, the precise analog of the MHz myth where clock speed was a misleading proxy for real performance \cite{fodor_2025_benchmarks}.

The \textbf{Law of the Weakest Link}, formalized at ICLR 2025, provides theoretical grounding for the \textbf{bucket principle}: a system's intelligence is constrained by its weakest capability dimension, not its average \cite{zhong_2024_weakest_link}. A model that excels at factual recall but fails at logical reasoning, or one that handles benign queries beautifully but produces dangerous outputs under adversarial probing, is not a ``mostly good'' system; it is a system with a critical failure mode. The bucket leaks from its lowest point.

Worst-case metrics are emerging to address this. Safety benchmarks like HarmBench and AIR-BENCH evaluate a system's most dangerous failure modes rather than its average-case helpfulness \cite{harmbench_2024, airbench_2024}. These are the LLM equivalents of worst-case execution time (WCET) analysis in real-time systems and adversarial stress testing in hardware validation.

For the intelligent compute core specifically, benchmarking must evaluate the processor's \textit{faithfulness to its ISA contract}. Does the LLM core reliably execute the agent's instructions? Does it follow specified output formats? Does it respect tool-use protocols? These questions measure processor reliability, not encyclopedic knowledge. A core that scores 70\% on MMLU but faithfully executes 99.9\% of structured instructions is a better processor than one that scores 90\% on MMLU but frequently deviates from its programming. The ICA framework makes this distinction explicit: knowledge evaluation belongs to the memory and retrieval layers; execution reliability belongs to the compute core.

The historical trajectory of CPU benchmarking, from megahertz to SPEC to workload-specific profiling, suggests that LLM evaluation will follow the same path: from aggregate scores to per-capability profiling to workload-specific benchmarks tied to real agent tasks. The bucket principle accelerates this transition by proving what practitioners already sensed: average-case metrics are insufficient for systems where a single failure mode can be catastrophic.
Figure~\ref{fig:paradigm:mhz_analogy} illustrates this parallel: just as the megahertz myth was corrected by holistic SPEC benchmarks, MMLU leaderboard averages must give way to per-capability worst-case profiling.
\input{figures/fig_paradigm_mhz_analogy}

\subsection{The Intelligent Operating System: From Code to World}

\subsubsection{The Historical Arc: From Task-Specific Systems to General Operating Systems}

Early operating systems were not general-purpose. They were instruments built for particular missions: the FORTRAN Monitor System on the IBM 704 served scientific computing; Colossus-era systems served military encryption; early batch systems served payroll processing. No one would have described these systems as ``general'' operating environments. They were narrow, domain-specific tools that happened to manage hardware resources. This characterization maps precisely onto today's intelligent agent systems. Codex, Claude Code, and Devin are powerful, but they are narrowly scoped to software engineering workflows. They edit codebases, run terminals, manage file systems, and invoke APIs. They are, in historical terms, the FORTRAN Monitor Systems of intelligent computing: task-specific instruments that will later be recognized as precursors to something far more general.

The classical OS did not become general overnight. A sequence of capability inflections was required: multiprogramming (Atlas, 1962) enabled concurrent task execution; time-sharing (CTSS, 1961; Multics, 1965) made interactive computing possible; virtual memory and protected mode (Unix, VMS, 1970s) provided isolation and reliability; networking stacks (BSD TCP/IP, 1980s) connected systems; and hardware abstraction layers (Windows NT HAL, 1990s) decoupled the OS from specific hardware \cite{ritchie_thompson_1974, corbato_1962_ctss, corbato_1965_multics}. Each layer was necessary before a single OS could serve scientific computing, office automation, real-time control, and multimedia simultaneously.

Current intelligent systems occupy what might be called the programming-language stage of their evolution. Just as early OS ran assembly or single-language programs for specific hardware, today's agent OS runs ``prompt programs'' for specific domains, predominantly code. The leap to a general intelligent OS, one that orchestrates agents across robotics, autonomous driving, engineering, agriculture, entertainment, and space navigation, requires the same kind of abstraction-layer breakthroughs that separated Unix from its single-language, single-hardware predecessors.

The five-generation framework documented earlier traces the arc from batch processing (Gen~I) through governed OS (Gen~V). But the trajectory points beyond Gen~V toward what might be called Generation~VI: the general intelligent OS that interfaces with the full physical and digital world. Patterson and Hennnessy argue that the ISA was the foundational abstraction enabling hardware/software co-evolution \cite{patterson_hennessy_2017}. Hennessy and Patterson's 2019 Turing Lecture identifies open ISAs and domain-specific co-design as pillars of a ``new golden age'' \cite{hennessy_patterson_2019}. The same structural requirement, namely an open, well-specified interface between layers, will govern the transition from today's code-only agent OS to tomorrow's world-spanning intelligent OS.

\subsubsection{Compute-Substrate Agnosticism: The ISA Decoupling Principle Generalized}

The preceding subsection chronicled recent advances in neuromorphic, biological, compute-in-memory, and quantum substrates. This subsection addresses a more fundamental architectural question: \textbf{why should the intelligent OS be agnostic to the physical implementation of its compute core?}

The answer follows directly from the lessons of ISA abstraction \cite{patterson_hennessy_2017}. The ISA is the most important abstraction in classical computing not because any single implementation is superior, but because it allows the software ecosystem and the hardware ecosystem to \textbf{evolve independently}. ICA's interface contracts generalize this principle to model-native computing: if a compute core, whether a silicon GPU, a neuromorphic chip, a compute-in-memory substrate, or a biological neural network, can meet a defined intelligence-score threshold (discussed in the next subsection), it should be capable of running the intelligent OS. The OS and compute core are fully decoupled, just as Linux and the CPU are decoupled via the ISA.

The philosophical extension follows the RISC-V open-ISA philosophy \cite{hennessy_patterson_2019}: any hardware meeting the specification can run the software, regardless of its physical implementation. Even a trained biological network (a rat brain organoid that passes the intelligence-score threshold) could in principle serve as a compute core for the intelligent OS. The same orchestration layer that schedules Claude Code sub-agents on an NVIDIA H100 today could schedule sensorimotor tasks on a Cortical Labs CL1 tomorrow \cite{cortical_2025_cl1}, provided both cores satisfy the same interface contract. This is \textbf{substrate agnosticism} in its fullest form: the OS does not care whether its compute cores are etched in silicon or grown in a petri dish.

\subsubsection{Intelligence Scoring as ISA Compliance}

The preceding subsection established the bucket principle and worst-case evaluation as theoretical foundations. This subsection shifts perspective from ``how to evaluate a core'' to ``how to enforce the ISA contract through scoring.''

ISA compliance testing is the mechanism by which the hardware--software contract is enforced. A chip cannot claim to implement x86 unless it passes a comprehensive validation suite covering every instruction, every privilege mode, every edge case. Without this testing, the ISA contract is meaningless. By direct analogy, an intelligence scoring standard is the enforcement mechanism for the Intelligent ISA. Without it, substrate agnosticism collapses into untestable claims.

The ``Law of the Weakest Link,'' formalized by Zhong et al.\ at ICLR 2025, provides theoretical grounding for the strictness this enforcement demands \cite{zhong_2024_weakest_link}: when multiple capabilities must combine to produce reliable behavior, overall system performance is constrained by the weakest individual capability. A system that scores 95th percentile on code generation but 20th percentile on safety compliance has a system intelligence of the 20th percentile. The lowest stave in the bucket determines the water level.

This means that \textbf{without a bucket-principle scoring standard, substrate agnosticism is dangerous}. A biological compute core that passes 9 of 10 safety dimensions but fails on the tenth could be deployed as ``compliant'' under averaging-based metrics, creating catastrophic risk. Minimum-score enforcement is not merely a testing preference; it is a safety imperative. Just as empirical work has demonstrated that microarchitecture matters more than ISA alone, intelligence scoring must evaluate implementation quality, not merely interface compliance.

\subsubsection{OS-Level Heterogeneous Scheduling: From Energy Models to Capability-Cost Models}

The preceding subsections established the microarchitectural analogy for heterogeneous model orchestration. This subsection shifts one layer up: when heterogeneous cores are unified under an intelligent OS, how should \textbf{the scheduling policy itself} be designed?

Classical operating systems provide the complete architectural blueprint. Energy Aware Scheduling (EAS), merged into Linux kernel 5.0 in 2019, equips the OS scheduler with an Energy Model that predicts the cost of placing each task on each available CPU \cite{eas_linux_2019}. The intelligent OS analog is a \textbf{Capability-Cost Model} that predicts the quality, latency, and dollar cost of routing each subtask to each available compute core, whether a large model, a small model, a neuromorphic chip, or a biological substrate, thereby enabling intelligence-optimal scheduling.

Two OS-level scheduling patterns merit particular attention:

\paragraph{Overutilization fallback.}
When the system is overloaded, EAS falls back to conventional SMP load balancing, sacrificing energy efficiency for throughput. The intelligent OS analog is a ``quality-at-all-costs'' fallback: when the agent system faces peak load, it routes tasks to larger, more expensive models regardless of cost optimization.

\paragraph{Real-time telemetry and dynamic routing.}
Intel's Thread Director (Alder Lake, 2021) provides the OS scheduler with real-time hardware telemetry about thread behavior. The intelligent OS analog is a \textbf{Task Director} that monitors subtask characteristics in real time, including reasoning depth, safety sensitivity, and latency tolerance, and dynamically re-routes to the optimal compute substrate mid-execution.

The introduction of biological and neuromorphic substrates makes the scheduling challenge qualitatively harder. A biological core may have fundamentally different latency profiles, failure modes, and capability signatures than a silicon GPU. The intelligent OS scheduler must accommodate not merely ``fast versus slow'' but ``probabilistic versus deterministic,'' ``high-bandwidth versus low-latency,'' and ``silicon versus biological.'' This is heterogeneous scheduling at a level of complexity that classical OS designers never confronted, yet the architectural patterns (Energy Models, overutilization fallbacks, and real-time telemetry) are directly inherited from the big.LITTLE playbook.
Figure~\ref{fig:paradigm:scheduling} places the two scheduling architectures side by side.
\input{figures/fig_paradigm_scheduling}

\subsubsection{The Physical World Interface: From Code-Only to Embodied Intelligence}

Today's intelligent OS interfaces with one domain: software engineering. Codex, Claude Code, and Devin operate on codebases, terminals, file systems, and APIs. In the historical framing established above, this is the FORTRAN Monitor System stage of intelligence: a task-specific OS for a single world. The next inflection is an intelligent OS that interfaces with the physical world: robotics, autonomous driving, engineering simulation, agriculture, entertainment, and space navigation.

Whole-brain emulation milestones are making embodied intelligence concrete. The FlyWire consortium mapped Drosophila's approximately 140,000 neurons and millions of synapses \cite{flywire_2024}, providing a complete connectome that enables detailed simulation of insect-scale neural circuits. These milestones demonstrate the trajectory toward more complex embodied agents that simulate complete organisms interacting with virtual environments.

A compute-substrate-agnostic intelligent OS means that the same orchestration layer that schedules Claude Code sub-agents today could, in principle, schedule robotic manipulation tasks, agricultural monitoring agents, and autonomous driving planners tomorrow. The requirements are twofold: the compute cores must meet the intelligence-score threshold (the bucket principle discussed above), and the OS must have appropriate world-interface drivers, such as sensors, actuators, simulation environments, and telemetry feeds.

This is the direct analog of how classical operating systems expanded from batch processing to real-time control. Unix started as a text-processing system, but the addition of real-time scheduling, device drivers, and networking stacks enabled it to run factory automation, telecommunications switches, and spacecraft guidance. Each new world required new driver interfaces but not a new OS kernel. The same Linux kernel runs a web server, a factory robot, and a Mars rover.

ICA's L4 (Semantic Interface Layer) and L5 (Orchestration Layer) are designed for this extensibility. Tool schemas and agent protocols are the device drivers of the intelligent OS. Adding a robotic arm driver or a satellite telemetry interface follows the same abstraction pattern as adding an MCP tool server today. The interface contract is consistent: sense, plan, act, verify, whether the target is a codebase or a cornfield.

Biological computing substrates add a unique dimension to embodied intelligence. Biological neural networks may possess natural advantages for sensorimotor tasks, pattern recognition, adaptive control, and real-time responsiveness that silicon-based systems lack. Kagan et al.\ (2022) demonstrated that biological neurons can learn to play Pong via feedback \cite{kagan_2022_pong_bioneurons}, confirming the viability of embodied bio-computing. The intelligent OS should be capable of dispatching sensorimotor subtasks to biological cores while routing symbolic reasoning to silicon GPUs, applying heterogeneous scheduling to embodied cognition. The scheduler picks the right core for the right task, just as a classical OS dispatches floating-point operations to the FPU and integer operations to the ALU. The physical world is not a different operating system; it is a different set of drivers for the same one.

%% file: figures/fig_paradigm_timeline.tex
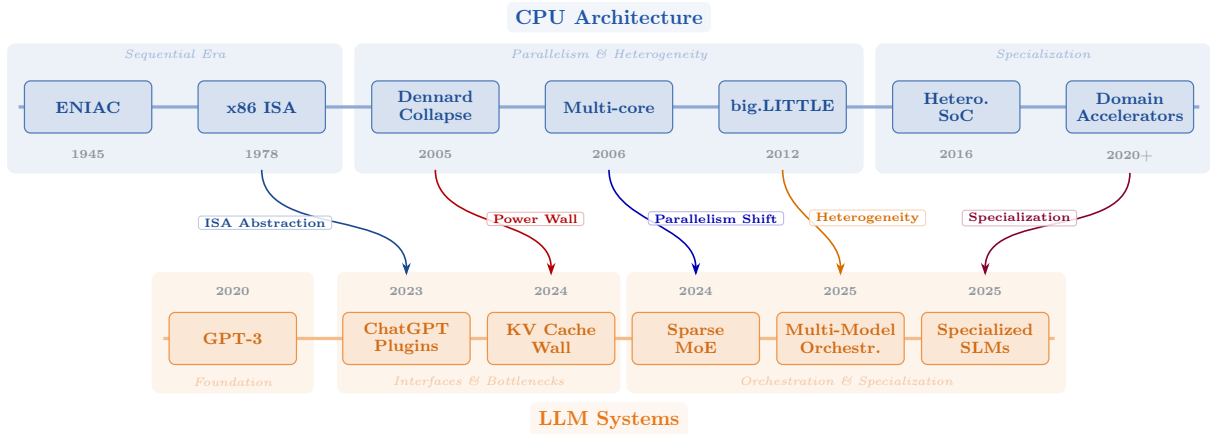
\begin{figure}[h]
    \centering
    \resizebox{\textwidth}{!}{%
    \begin{tikzpicture}[
        font=\sffamily,
        >=Stealth,
        cpuNode/.style={
            draw=classical!80, fill=classical!18,
            rounded corners=3pt,
            minimum width=22mm, minimum height=9mm,
            inner sep=2pt, align=center,
            font=\scriptsize\bfseries, text=classical!90!black,
            line width=0.6pt
        },
        llmNode/.style={
            draw=modelnative!80, fill=modelnative!18,
            rounded corners=3pt,
            minimum width=22mm, minimum height=9mm,
            inner sep=2pt, align=center,
            font=\scriptsize\bfseries, text=modelnative!90!black,
            line width=0.6pt
        },
        yearLabel/.style={
            font=\tiny\bfseries, text=timelineGray!85!black
        },
        timelineLine/.style={
            line width=1.6pt, draw=#1!50
        },
        phaseBand/.style={
            fill=#1, fill opacity=0.08, rounded corners=4pt
        },
        curvedArrow/.style={
            -{Stealth[length=2.2mm, width=1.5mm]},
            thick, draw=#1,
            shorten <=2pt, shorten >=2pt
        },
        arrowLabel/.style={
            font=\tiny\bfseries, text=#1,
            fill=white, fill opacity=0.85, text opacity=1,
            rounded corners=1.5pt, inner xsep=3pt, inner ysep=1.2pt,
            draw=#1!30, line width=0.3pt
        },
        timelineHeader/.style={
            font=\small\bfseries, text=#1,
            fill=#1!6, rounded corners=3pt, inner sep=4pt
        }
    ]
    
    \def\cpuY{2.2}
    \def\llmY{-1.8}
    
    \fill[phaseBand=classical]
        (-1.4, \cpuY+1.1) rectangle (4.4, \cpuY-1.15);
    \fill[phaseBand=classical]
        (4.6, \cpuY+1.1) rectangle (13.4, \cpuY-1.15);
    \fill[phaseBand=classical]
        (13.6, \cpuY+1.1) rectangle (19.4, \cpuY-1.15);
    
    \fill[phaseBand=modelnative]
        (1.1, \llmY+1.15) rectangle (3.9, \llmY-0.95);
    \fill[phaseBand=modelnative]
        (4.3, \llmY+1.15) rectangle (9.2, \llmY-0.95);
    \fill[phaseBand=modelnative]
        (9.3, \llmY+1.15) rectangle (16.9, \llmY-0.95);
    
    \node[font=\tiny\itshape, text=classical!45]
        at (1.5, \cpuY+0.9) {Sequential Era};
    \node[font=\tiny\itshape, text=classical!45]
        at (9.0, \cpuY+0.9) {Parallelism \& Heterogeneity};
    \node[font=\tiny\itshape, text=classical!45]
        at (16.5, \cpuY+0.9) {Specialization};
    
    \node[font=\tiny\itshape, text=modelnative!45]
        at (2.5, \llmY-0.75) {Foundation};
    \node[font=\tiny\itshape, text=modelnative!45]
        at (6.75, \llmY-0.75) {Interfaces \& Bottlenecks};
    \node[font=\tiny\itshape, text=modelnative!45]
        at (13.1, \llmY-0.75) {Orchestration \& Specialization};
    
    \node[timelineHeader=classical]   at (9, \cpuY+1.55) {CPU Architecture};
    \node[timelineHeader=modelnative] at (9, \llmY-1.4) {LLM Systems};
    
    \draw[timelineLine=classical]   (-1.2, \cpuY) -- (19.2, \cpuY);
    \draw[timelineLine=modelnative] (1.3, \llmY) -- (16.7, \llmY);
    
    \node[cpuNode] (cpu1) at (0, \cpuY) {ENIAC};
    \node[yearLabel, below=1.5mm of cpu1] (cyr1) {1945};
    
    \node[cpuNode] (cpu2) at (3, \cpuY) {x86 ISA};
    \node[yearLabel, below=1.5mm of cpu2] (cyr2) {1978};
    
    \node[cpuNode] (cpu3) at (6, \cpuY) {Dennard\\[-1pt]Collapse};
    \node[yearLabel, below=1.5mm of cpu3] (cyr3) {2005};
    
    \node[cpuNode] (cpu4) at (9, \cpuY) {Multi-core};
    \node[yearLabel, below=1.5mm of cpu4] (cyr4) {2006};
    
    \node[cpuNode] (cpu5) at (12, \cpuY) {big.LITTLE};
    \node[yearLabel, below=1.5mm of cpu5] (cyr5) {2012};
    
    \node[cpuNode] (cpu6) at (15, \cpuY) {Hetero.\\[-1pt]SoC};
    \node[yearLabel, below=1.5mm of cpu6] (cyr6) {2016};
    
    \node[cpuNode] (cpu7) at (18, \cpuY) {Domain\\[-1pt]Accelerators};
    \node[yearLabel, below=1.5mm of cpu7] (cyr7) {2020+};
    
    \node[llmNode] (llm1) at (2.5, \llmY) {GPT-3};
    \node[yearLabel, above=1.5mm of llm1] (lyr1) {2020};
    
    \node[llmNode] (llm2) at (5.5, \llmY) {ChatGPT\\[-1pt]Plugins};
    \node[yearLabel, above=1.5mm of llm2] (lyr2) {2023};
    
    \node[llmNode] (llm3) at (8.0, \llmY) {KV Cache\\[-1pt]Wall};
    \node[yearLabel, above=1.5mm of llm3] (lyr3) {2024};
    
    \node[llmNode] (llm4) at (10.5, \llmY) {Sparse\\[-1pt]MoE};
    \node[yearLabel, above=1.5mm of llm4] (lyr4) {2024};
    
    \node[llmNode] (llm5) at (13.0, \llmY) {Multi-Model\\[-1pt]Orchestr.};
    \node[yearLabel, above=1.5mm of llm5] (lyr5) {2025};
    
    \node[llmNode] (llm6) at (15.5, \llmY) {Specialized\\[-1pt]SLMs};
    \node[yearLabel, above=1.5mm of llm6] (lyr6) {2025};
    
    
    \draw[curvedArrow=red!70!black]
        (cyr3.south) to[out=-90, in=90]
        node[arrowLabel=red!70!black, pos=0.45, anchor=west, xshift=1pt]
            {Power Wall}
        (lyr3.north);
    
    \draw[curvedArrow=blue!70!black]
        (cyr4.south) to[out=-90, in=90]
        node[arrowLabel=blue!70!black, pos=0.45, anchor=west, xshift=1pt]
            {Parallelism Shift}
        (lyr4.north);
    
    \draw[curvedArrow=orange!85!black]
        (cyr5.south) to[out=-90, in=90]
        node[arrowLabel=orange!85!black, pos=0.45, anchor=west, xshift=1pt]
            {Heterogeneity}
        (lyr5.north);
    
    \draw[curvedArrow=classical!85!black]
        (cyr2.south) to[out=-90, in=90]
        node[arrowLabel=classical!85!black, pos=0.50, anchor=east, xshift=-1pt]
            {ISA Abstraction}
        (lyr2.north);
    
    \draw[curvedArrow=purple!70!black]
        (cyr7.south) to[out=-90, in=90]
        node[arrowLabel=purple!70!black, pos=0.40, anchor=east, xshift=-1pt]
            {Specialization}
        (lyr6.north);
    
    \end{tikzpicture}}
    \caption{Structural parallels between CPU and LLM system evolution.
    Curved arrows connect milestones that share the same architectural driver:
    Dennard scaling collapse and the inference energy wall both arise from a \emph{Power Wall};
    the multi-core and MoE transitions reflect a \emph{Parallelism Shift};
    big.LITTLE and heterogeneous model orchestration embody \emph{Heterogeneity};
    x86 ISA and the token/tool interface represent \emph{ISA Abstraction};
    and domain accelerators and specialized small models pursue \emph{Specialization}.}
    \label{fig:paradigm_timeline}
    \end{figure}

%% file: figures/fig_biglittle.tex
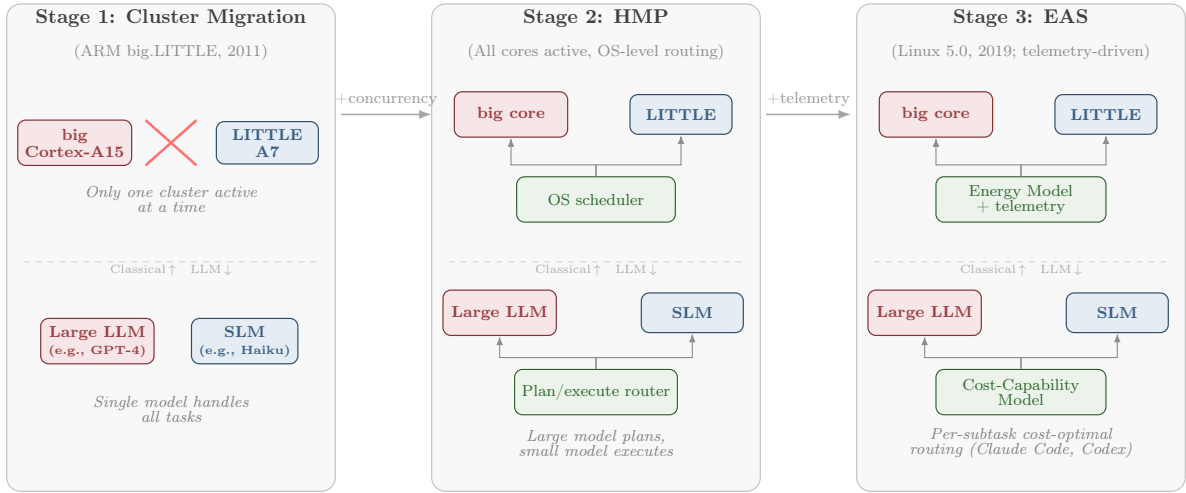
\begin{figure}[htbp]
\centering
\providecolor{bigColor}{HTML}{C44E52}
\providecolor{littleColor}{HTML}{4A78A8}
\providecolor{schedColor}{HTML}{3A8A3A}
\providecolor{stageHdr}{HTML}{555555}
\resizebox{0.98\textwidth}{!}{%
\begin{tikzpicture}[
    font=\small, >=Latex,
    stagebox/.style={draw=stageHdr!35, fill=stageHdr!4, rounded corners=7pt,
        minimum width=5.8cm, line width=0.6pt},
    bigbox/.style={draw=bigColor!65!black, fill=bigColor!14, rounded corners=4pt,
        minimum width=2.0cm, minimum height=0.80cm,
        align=center, font=\scriptsize\bfseries, text=bigColor!80!black, line width=0.6pt},
    litbox/.style={draw=littleColor!65!black, fill=littleColor!14, rounded corners=4pt,
        minimum width=1.8cm, minimum height=0.70cm,
        align=center, font=\scriptsize\bfseries, text=littleColor!80!black, line width=0.6pt},
    schbox/.style={draw=schedColor!65!black, fill=schedColor!10, rounded corners=4pt,
        minimum width=2.8cm, minimum height=0.80cm,
        align=center, font=\scriptsize, text=schedColor!75!black, line width=0.55pt},
    divline/.style={draw=black!18, dashed, line width=0.55pt},
    darr/.style={-{Latex[length=1.6mm]}, black!45, line width=0.55pt},
    earr/.style={-{Latex[length=2.2mm]}, black!35, line width=0.75pt},
    ann/.style={font=\scriptsize\itshape, text=black!50, align=center},
    hdr/.style={font=\small\bfseries, text=stageHdr!75!black},
    subhdr/.style={font=\scriptsize, text=black!45, align=center},
]

\def\xA{-7.5}   
\def\xB{0.0}    
\def\xC{7.5}    

\def\yTitle{4.6}
\def\ySub{4.0}
\def\yCoreTop{2.9}   
\def\yCoreBot{1.8}   
\def\ySched{1.4}     
\def\yAnnotC{1.9}    
\def\yDiv{0.3}
\def\yDivLbl{0.15}
\def\yLLMTop{-0.6}
\def\yLLMSched{-2.0}
\def\yAnnotL{-2.6}

\node[stagebox, minimum height=8.6cm] at (\xA, 0.55) {};
\node[stagebox, minimum height=8.6cm] at (\xB, 0.55) {};
\node[stagebox, minimum height=8.6cm] at (\xC, 0.55) {};

\node[hdr]    at (\xA, \yTitle) {Stage 1: Cluster Migration};
\node[subhdr] at (\xA, \ySub)   {(ARM big.LITTLE, 2011)};
\node[hdr]    at (\xB, \yTitle) {Stage 2: HMP};
\node[subhdr] at (\xB, \ySub)   {(All cores active, OS-level routing)};
\node[hdr]    at (\xC, \yTitle) {Stage 3: EAS};
\node[subhdr] at (\xC, \ySub)   {(Linux 5.0, 2019; telemetry-driven)};

\node[bigbox] (A_big)    at (\xA-1.7, \yCoreTop-0.5) {big\\[-1pt]Cortex-A15};
\node[litbox] (A_little) at (\xA+1.7, \yCoreTop-0.5) {LITTLE\\[-1pt]A7};
\draw[red!55, line width=1.3pt]
    (\xA-0.45, \yCoreTop-0.90) -- (\xA+0.45, \yCoreTop-0.10);
\draw[red!55, line width=1.3pt]
    (\xA-0.45, \yCoreTop-0.10) -- (\xA+0.45, \yCoreTop-0.90);
\node[ann] at (\xA, \yAnnotC-0.5) {\textit{Only one cluster active}\\[-2pt]\textit{at a time}};

\node[bigbox] (B_big)   at (\xB-1.5, \yCoreTop) {big core};
\node[litbox] (B_lit)   at (\xB+1.5, \yCoreTop) {LITTLE};
\node[schbox] (B_sched) at (\xB,     \ySched)    {OS scheduler};
\draw[darr] (B_sched.north) -- ++(0,0.2) -| (B_big.south);
\draw[darr] (B_sched.north) -- ++(0,0.2) -| (B_lit.south);

\node[bigbox] (C_big)   at (\xC-1.5, \yCoreTop) {big core};
\node[litbox] (C_lit)   at (\xC+1.5, \yCoreTop) {LITTLE};
\node[schbox, minimum width=3.0cm] (C_sched) at (\xC, \ySched) {Energy Model\\[-2pt]+ telemetry};
\draw[darr] (C_sched.north) -- ++(0,0.2) -| (C_big.south);
\draw[darr] (C_sched.north) -- ++(0,0.2) -| (C_lit.south);

\foreach \cx in {\xA, \xB, \xC}{
    \draw[divline] (\cx-2.6, \yDiv) -- (\cx+2.6, \yDiv);
    \node[font=\tiny, text=black!32] at (\cx, \yDivLbl)
        {Classical\,$\uparrow$\quad LLM\,$\downarrow$};
}

\node[bigbox] at (\xA-1.3, \yLLMTop-0.5) {Large LLM\\[-1pt]{\tiny(e.g., GPT-4)}};
\node[litbox] at (\xA+1.3, \yLLMTop-0.5) {SLM\\[-1pt]{\tiny(e.g., Haiku)}};
\node[ann]    at (\xA, \yAnnotL+0.3) {\textit{Single model handles}\\[-2pt]\textit{all tasks}};

\node[bigbox] (B_llm)   at (\xB-1.7, \yLLMTop)   {Large LLM};
\node[litbox] (B_slm)   at (\xB+1.7, \yLLMTop)   {SLM};
\node[schbox] (B_router)at (\xB,     \yLLMSched)  {Plan/execute router};
\draw[darr] (B_router.north) -- ++(0,0.2) -| (B_llm.south);
\draw[darr] (B_router.north) -- ++(0,0.2) -| (B_slm.south);
\node[ann] at (\xB, \yAnnotL-0.3) {\textit{Large model plans,}\\[-2pt]\textit{small model executes}};

\node[bigbox] (C_llm)   at (\xC-1.7, \yLLMTop)   {Large LLM};
\node[litbox] (C_slm)   at (\xC+1.7, \yLLMTop)   {SLM};
\node[schbox, minimum width=3.0cm] (C_router) at (\xC, \yLLMSched) {Cost-Capability\\[-2pt]Model};
\draw[darr] (C_router.north) -- ++(0,0.2) -| (C_llm.south);
\draw[darr] (C_router.north) -- ++(0,0.2) -| (C_slm.south);
\node[ann] at (\xC, \yAnnotL-0.3) {\textit{Per-subtask cost-optimal}\\[-2pt]\textit{routing (Claude Code, Codex)}};

\draw[earr] (-4.5, \yCoreTop)
    -- node[above, font=\scriptsize, text=black!40] {+concurrency}
    (-2.9, \yCoreTop);
\draw[earr] (3, \yCoreTop)
    -- node[above, font=\scriptsize, text=black!40] {+telemetry}
    (4.5, \yCoreTop);

\end{tikzpicture}}%
\caption{Three-stage evolution of heterogeneous scheduling, mapped between classical ARM big.LITTLE (top half of each panel) and LLM model orchestration (bottom half). Stage~1 (cluster migration / single model) activates only one resource type at a time. Stage~2 (HMP / plan-execute routing) runs both simultaneously with coarse-grained task assignment. Stage~3 (EAS / cost-capability routing) uses real-time telemetry to route each subtask to the cost-optimal compute substrate, the pattern already deployed in Claude Code and Codex.}
\label{fig:biglittle}
\end{figure}

%% file: figures/fig_paradigm_mhz_analogy.tex
\begin{figure}[htbp]
  \centering
  \includegraphics[width=0.88\textwidth]{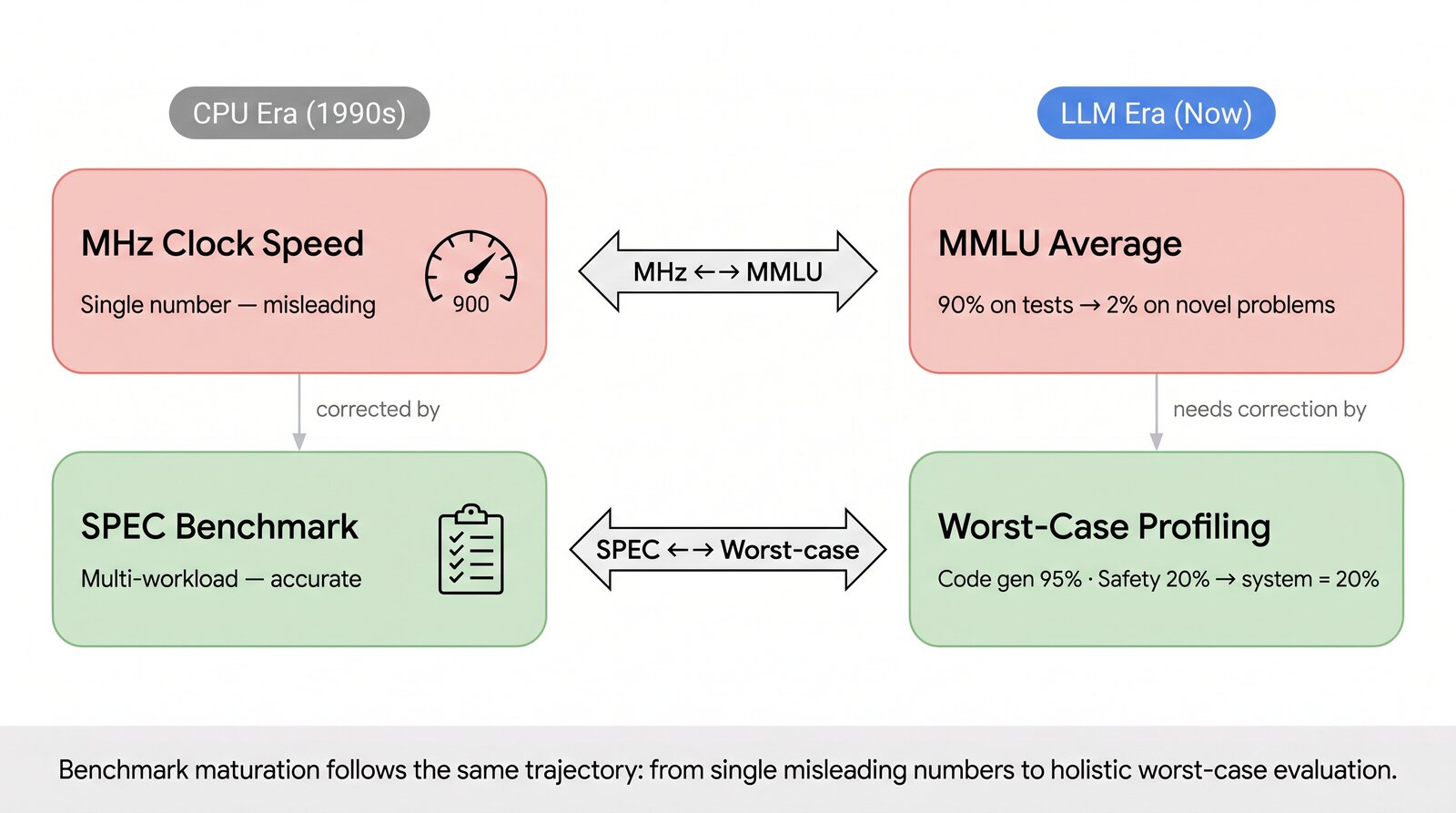}
  \caption{The megahertz myth (left) and its LLM analog (right).
  In the 1990s, CPU clock speed was marketed as the definitive performance metric,
  yet SPEC benchmarks revealed that real-workload performance depended on pipeline depth,
  cache size, and IPC---not clock rate alone.
  LLM evaluation faces the same maturation: MMLU leaderboard averages are the ``megahertz''
  of intelligent systems~\cite{fodor_2025_benchmarks}, obscuring the fact that
  a model scoring 90\% on standard tests may score only 2\% on novel unseen problems.
  Per-capability worst-case profiling (HarmBench, AIR-Bench) is the emerging SPEC equivalent:
  a system scoring 95th percentile on code generation but 20th percentile on safety compliance
  has a system intelligence of the 20th percentile~\cite{zhong_2024_weakest_link}.}
  \label{fig:paradigm:mhz_analogy}
\end{figure}

%% file: figures/fig_paradigm_scheduling.tex
\begin{figure}[htbp]
  \centering
  \includegraphics[width=0.88\textwidth]{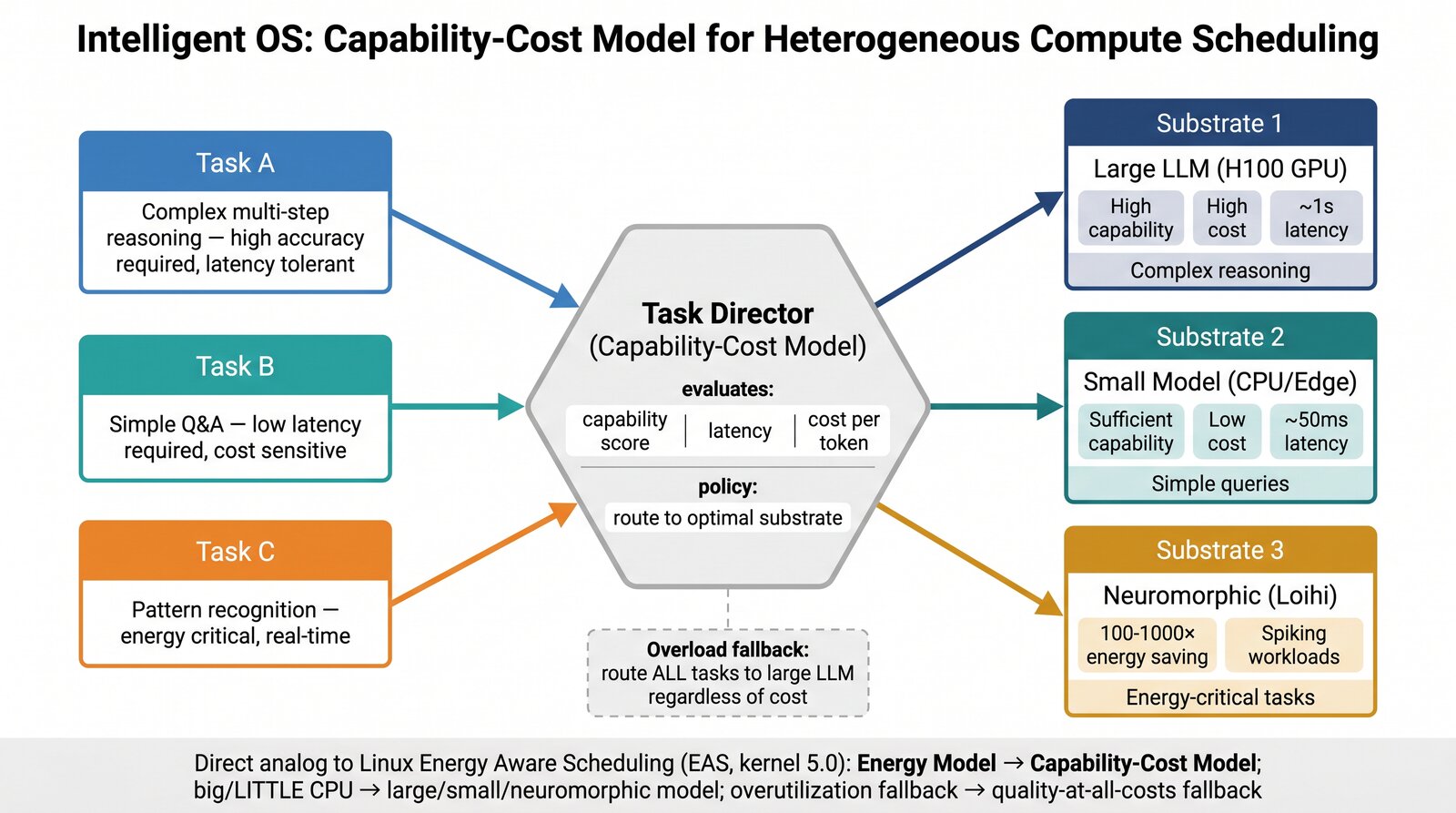}
  \caption{OS-level heterogeneous scheduling: Linux Energy Aware Scheduling (EAS) versus the intelligent OS Capability-Cost Model.
  EAS (Linux 5.0, 2019) uses an Energy Model to route each task to the most energy-efficient capable CPU
  and falls back to load balancing under overutilization.
  The Capability-Cost Model is the direct analog: a Task Director routes each subtask to
  the optimal compute core (large model, small model, neuromorphic, or biological substrate)
  by predicted capability, latency, and cost, falling back to the highest-capability core under peak load.}
  \label{fig:paradigm:scheduling}
\end{figure}

%% file: en_sections/section_social.tex
\section{Socio-Technical Implications: The Industry Structure of Intelligent Computing}
\label{sec:sociotechnical}

The ICA framework describes a layered architecture for model-native computing systems. Yet architecture does not exist in a vacuum: it both shapes and is shaped by the economic structure of the industry that builds it. The semiconductor and personal-computer industries underwent a well-documented disaggregation over four decades, evolving from vertically integrated giants into layered ecosystems of specialized firms. The emerging AI industry appears to be following a parallel trajectory, with one critical difference: model weights are software, infinitely copyable at near-zero marginal cost, which means the physical moat that once protected chipmakers has no direct analog. This section traces the structural dynamics that will determine whether intelligent computing settles into a disaggregated stack or remains dominated by a small number of vertically integrated players.

\subsection{From IDM to Fabless: The Semiconductor Analogy and AI Stack Disaggregation}
\label{sec:idm-fabless}

The semiconductor industry evolved through four structural stages. Before the 1980s, firms such as Intel, AMD, and Texas Instruments operated as Integrated Device Manufacturers (IDMs), controlling every stage from design through fabrication to sales. TSMC's founding in 1987 introduced the pure-play foundry model, separating fabrication from design. The fabless era followed: NVIDIA and Qualcomm designed chips without owning fabs. ARM then added a fourth layer by licensing processor IP without manufacturing at all. Each stage lowered entry barriers and created specialized value-chain layers.

The AI industry today resembles the IDM phase. NVIDIA dominates as a vertically integrated platform, owning both the hardware (GPUs) and the software stack (CUDA), much as IBM controlled mainframe hardware, operating systems, and applications before the PC era's disaggregation. The central structural question is: will AI follow the PC-era pattern of disaggregation into separate chip, platform/OS, model, and application layers, or will it remain vertically integrated under dominant players?

The first wave of disaggregation is already visible. Cloud giants building custom silicon, such as Google TPUs, Amazon Trainium, and Microsoft Maia, signal the separation of hardware provision from the software and model layers, analogous to how the foundry model separated fabrication from design. Stanford HAI's 2025 data showing a 280-fold drop in reasoning costs, combined with surging small-model performance, creates commoditization pressure on foundation model providers~\cite{stanfordhai2025index}, accelerating the structural parallel to how commodity x86 CPUs enabled the PC disaggregation. Gartner's forecast of major consolidation in agentic AI due to oversupply mirrors the semiconductor industry's consolidation cycles~\cite{gartner2026trends}, suggesting that only a few foundation model providers will survive to become the ``Intel/AMD'' of AI.
Figure~\ref{fig:social:idm_fabless} places the two disaggregation trajectories side by side.
\input{figures/fig_social_idm_fabless}

\subsection{Model Weights as Software: The Copyability Problem and the Closed-Source Strategic Response}
\label{sec:copyability}

The semiconductor analogy, however, breaks down at one crucial point. A CPU's intellectual property is embedded in its physical manufacturing process: reverse-engineering a chip from its external behavior is economically infeasible. Model weights, by contrast, are fundamentally software. They can be copied infinitely at near-zero marginal cost, and they can be extracted from API outputs through model distillation attacks. This asymmetry is the primary structural reason that top-tier models (GPT-4, Claude, Gemini) remain closed-source.

The closed-source strategy manifests as API-only access. Model providers sell token-based inference rather than the weights themselves, with OpenAI generating \$5.7 billion in Q1~2026 revenue through token-based pricing. This mirrors Microsoft's historical practice of assessing per-device OEM licensing fees: the revenue is tied to usage rather than to the artifact itself.

DeepSeek's open-source R1 model, which demonstrated competitive benchmarks with OpenAI's o1, pioneered a ``third pathway'': generating revenue from open-source AI~\cite{thirdbridge2025deepseek}. Proprietary enterprise AI use reportedly shrank from 80\% to 44\% across 2025, while open-source alternatives gained ground~\cite{mckinsey2025aisurvey}. The \$200 billion-plus web of circular investments linking OpenAI, Anthropic, Microsoft, Google, NVIDIA, and Amazon through partnerships and equity stakes~\cite{brookings2025aicustomers} represents a vertically integrated \textit{keiretsu} structure that temporarily enforces the closed-source model through capital interdependence. It cannot, however, solve the fundamental copyability problem: weights, once leaked or distilled, propagate without marginal cost.

\subsection{Hardware Binding and Weight Protection: Rebuilding the Physical Moat}
\label{sec:hardware-binding}

If copyability undermines the software-only business model, the industry's response is to reconstruct a physical moat around model weights. Three strategies span a spectrum from software-based protection to permanent hardware binding.

The first approach treats weights as copyable but renders them useless without authorized hardware. RAND Corporation's 2024 playbook proposes 167 security measures for weight protection, recommending Trusted Execution Environment (TEE)-based encryption where weights are decrypted only inside a verified enclave~\cite{rand2024weightplaybook}. NVIDIA's H100, the first GPU with confidential computing, exemplifies this pathway: a form of digital rights management for AI models.

The second approach embeds hardware-level identity into the inference process, integrating Physical Unclonable Functions (PUFs) with compute-in-memory architectures to create a device-specific fingerprint that ties weight execution to a particular piece of silicon. This is an intermediate strategy: weights remain digitally distributed, but correct execution requires the matching hardware.

The third and most radical approach permanently burns model weights into silicon transistors. Taalas (2026) produces dedicated inference chips with a one-to-one chip-to-model binding, achieving approximately 17,000 tokens per second per user (roughly ten times GPU performance)~\cite{taalas_2026_hc1}. The model becomes inseparable from the hardware: the ``model is the computer.'' This creates a true physical moat analogous to Intel's chip fabrication, where the intellectual property is embodied in the manufacturing process itself. These three approaches (TEE encryption, PUF-based compute-in-memory, and permanent silicon burning) represent increasing degrees of ``physicalization,'' each trading flexibility for IP protection, directly paralleling the semiconductor industry's progression from licensed IP (ARM) to manufactured chips (Intel) to dedicated ASICs.
Figure~\ref{fig:social:hardware_binding} illustrates the three strategies as a physicalization spectrum.
\input{figures/fig_social_hardware_binding}

\subsection{The OS Layer as Traffic Gateway: Platform Strategy in the AI Era}
\label{sec:os-gateway}

The hardware-binding strategies of the previous subsection address weight protection at the model layer. Yet the history of the PC industry suggests that the highest-value capture point lies not in hardware or in the model, but in the platform layer: the interface that mediates between users and underlying capabilities.

Microsoft's pivotal 1990 achievement illustrates the dynamic. By securing OEM pre-installation agreements with thirty major PC manufacturers for Windows~3.0~\cite{licendi2024mslicensing}, Microsoft created a flywheel: ubiquitous distribution fed a developer ecosystem, which drove consumer lock-in, which sustained licensing revenue that reached \$23--29 billion per year in recent times. Windows was the traffic gateway between users and PC computing. The entity that controlled the gateway captured disproportionate value, even though Intel manufactured the hardware that made it all possible.

Current AI chatbot interfaces, including ChatGPT, Claude, and Gemini, function as de facto ``operating systems'' for AI interaction. They are the traffic gateway between users and model capabilities. Walmart, Google, and WeChat are explicitly competing to become ``AI-era traffic gateways''~\cite{ceibs2025aiagents}, confirming that the platform layer is recognized as the highest-value capture point in the AI stack. Google's Android strategy, an open-source OS monetized through embedded services (ads, Play Store, cloud)~\cite{wikipedia2024msopensource}, represents an alternative model that AI platform companies may adopt: open the interface to gain market share, then monetize through embedded commerce and data flows. As AI industry standards and protocols mature, enabling model interoperability and portability, the entity controlling the gateway rather than the model itself will likely capture disproportionate value, exactly as Microsoft captured more value than Intel in the PC era~\cite{ccia2025competition}.

\subsection{Convergence Scenarios: When Training Stabilizes and Brains Become Physical Products}
\label{sec:convergence}

The semiconductor industry's four-stage evolution provides a roadmap for AI industry maturation. The prerequisite for full disaggregation is the emergence of industry standards and protocols that enable interoperability across layers, just as standard interfaces such as the x86 ISA, the PCI bus, and USB enabled the PC ecosystem to disaggregate.

Several signals suggest the industry is approaching a ``training stabilization'' inflection point. Industry analysts project substantial AI spending growth for 2026 and predict major consolidation in agentic AI~\cite{gartner2026trends}. McKinsey reports that only 33\% of organizations have scaled AI beyond pilots~\cite{mckinsey2025aisurvey}, a ``trough of disillusionment'' that historically precedes industry maturation. The PC industry went through a similar phase: the 1983--1985 hardware shakeout eliminated dozens of incompatible platforms before the standard IBM PC compatible architecture enabled explosive growth.

When model architectures stabilize, the ``selling brains as physical devices'' model becomes viable. Dedicated inference chips with burned-in weights, such as the Taalas HC1 achieving tenfold GPU performance~\cite{taalas_2026_hc1}, could commoditize the model layer while creating a new hardware product category: the AI appliance. This would parallel the transition from mainframe time-sharing to personal computers, where computing ceased to be a shared service and became a personal possession.

The convergence of three trends (hardware binding technologies maturing, inference costs dropping 280-fold, and custom silicon proliferating) creates the conditions for a ``fabless AI'' industry structure. Model companies would design ``brains'' (architectures and weights), foundries (TSMC, Samsung) would manufacture them as dedicated chips, and platform companies would distribute them through OS-level gateways. The most likely end-state is a three-layer industry structure: fabless model designers (the future Intels and AMDs of AI), AI foundries manufacturing dedicated inference and training chips, and AI OS platforms controlling user relationships through interface and gateway services. This mirrors the semiconductor and PC disaggregation that created today's Intel, TSMC, and Microsoft, the very firms whose history has served as this section's predictive lens.
Figure~\ref{fig:social:fabless_ai} maps the predicted three-layer AI industry structure onto its semiconductor analog.
\input{figures/fig_social_fabless_ai}

\subsection{Layer-by-Layer: Mapping the Industry Stack onto ICA}
\label{sec:social-layers}

The three-layer fabless-AI prediction above is a coarse-grained sketch. Projecting it onto the finer ICA hierarchy makes explicit where value will concentrate and where commoditization will bite. Table~\ref{tab:industry-layers} maps each ICA layer to its classical industry analog, the emerging class of AI-industry owner, and the structural prospects for value capture.

\begin{table}[h]
\centering
\footnotesize
\caption{Mapping the emerging AI industry stack onto the ICA layers, with the classical industry analog and value-capture prospect for each.}
\label{tab:industry-layers}
\begin{tabularx}{\textwidth}{>{\raggedright\arraybackslash}p{0.08\textwidth}>{\raggedright\arraybackslash}p{0.26\textwidth}>{\raggedright\arraybackslash}p{0.38\textwidth}>{\raggedright\arraybackslash}p{0.20\textwidth}}
\toprule
\textbf{Layer} & \textbf{Classical analog} & \textbf{Emerging AI-industry owner} & \textbf{Value-capture prospect} \\
\midrule
L1 & Foundry + fabless chip designer (TSMC; NVIDIA/AMD) & Accelerator and foundry layer (GPUs, TPUs, ASICs, custom silicon) & High (capital and process moat), cyclical \\
L2 & Microarchitecture vendor / OEM & Open serving runtimes (vLLM, SGLang) plus cloud inference platforms & Commoditizing (open-source middleware) \\
L3 & Memory OEM (DRAM/NAND) & Memory and retrieval middleware (vector databases, context systems) & Differentiated by integration, not standalone \\
L4 & Bus / interface standard (PCIe, USB, ISA) & Open protocols (MCP, A2A) as the agent I/O bus & Winner-take-most; low direct revenue \\
L5 & OS vendor (Microsoft, Apple) & Agent runtimes and agent-OS gateways (Codex, Claude Code, \ldots) & \textbf{Highest}---the traffic gateway, platform lock-in \\
L6 & Application ISV & Vertical agent apps and domain workflows & Long tail; winner-take-most per vertical \\
\bottomrule
\end{tabularx}
\end{table}

The mapping reveals where value concentrates and where it bleeds away. Returns accrue at the two ends of the stack: at L1, where the capital intensity and fabrication-process moats of accelerator production mirror the classical foundry barrier, and at L5, where the agent-OS gateway controls the interface between users and all underlying capability---replaying the Microsoft-versus-Intel dynamic of the PC era. The middle layers tell the opposite story. Open-source serving runtimes at L2, open protocols at L4, and largely interchangeable memory middleware at L3 erode standalone margins, pushing value either down to the silicon or up to the gateway. The copyability of model weights (Section~\ref{sec:copyability}) only accelerates the squeeze: unless weights are hardware-bound (Section~\ref{sec:hardware-binding}), the model layer drifts toward commodity, which is precisely the pressure that forces the industry toward either the closed-API strategy or the physical-moat strategy.

This layer-resolved view also clarifies the standardization prerequisite for disaggregation. In the PC era, the x86 ISA, the PCI bus, and USB were the interface contracts that allowed the stack to separate into competing specialists. In the AI stack, the analogous prerequisites are the L4 protocols (MCP, A2A) and, more fundamentally, the inter-layer interface contracts of ICA itself. Until those contracts stabilize, the stack cannot disaggregate and vertically integrated players retain their advantage; once they do, the fabless-AI structure above becomes structurally inevitable rather than merely predictive.

%% file: figures/fig_social_idm_fabless.tex
\begin{figure}[htbp]
  \centering
  \includegraphics[width=0.88\textwidth]{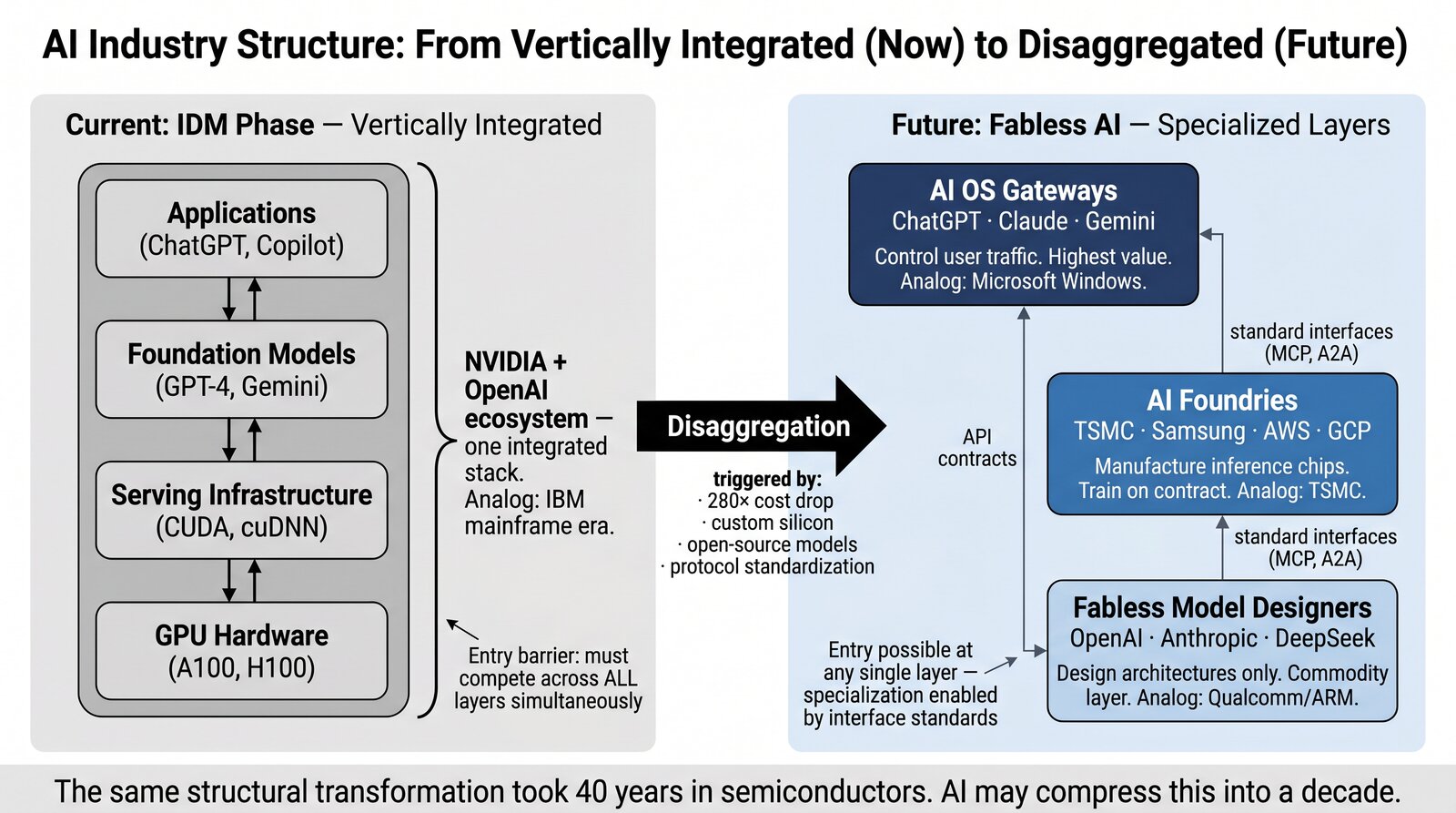}
  \caption{Parallel disaggregation trajectories: semiconductor industry (top) versus the emerging AI industry (bottom).
  The semiconductor industry evolved through four stages over 40 years:
  vertically integrated IDMs (Intel, TI) $\to$ pure-play foundry (TSMC, 1987) $\to$ fabless design (NVIDIA, Qualcomm) $\to$ IP licensing (ARM).
  The AI industry currently occupies the IDM phase (NVIDIA controls hardware, software, and ecosystem)
  and shows early signs of disaggregation through custom silicon (Google TPU, Amazon Trainium).
  The predicted end state mirrors semiconductor disaggregation: fabless model designers, AI foundries, and OS-layer gateways.}
  \label{fig:social:idm_fabless}
\end{figure}

%% file: figures/fig_social_hardware_binding.tex
\begin{figure}[htbp]
  \centering
  \includegraphics[width=0.88\textwidth]{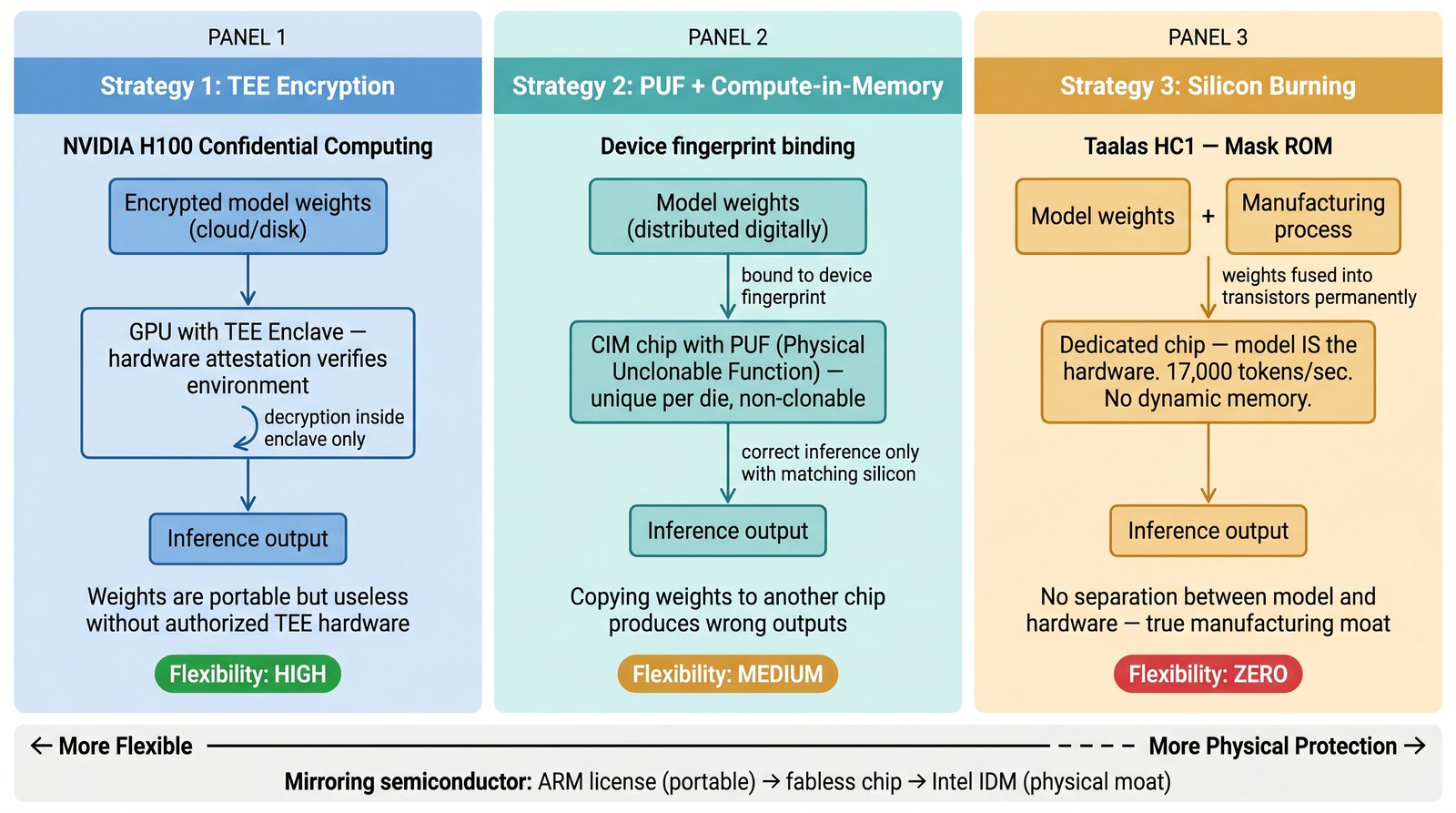}
  \caption{Three strategies for protecting model weights from copyability, arranged along a physicalization spectrum.
  TEE encryption (NVIDIA H100 Confidential Computing) decrypts weights only inside a hardware-verified enclave.
  PUF-based compute-in-memory binds correct inference to a device-specific silicon fingerprint.
  Silicon burning (Taalas HC1) permanently embeds weights into Mask ROM, creating an inseparable model--hardware unit
  with 17,000 tokens/sec and a true manufacturing moat.
  Each step trades flexibility for IP protection, mirroring the semiconductor industry's progression
  from licensed IP (ARM) to manufactured chips (Intel) to dedicated ASICs.}
  \label{fig:social:hardware_binding}
\end{figure}

%% file: figures/fig_social_fabless_ai.tex
\begin{figure}[htbp]
  \centering
  \includegraphics[width=0.88\textwidth]{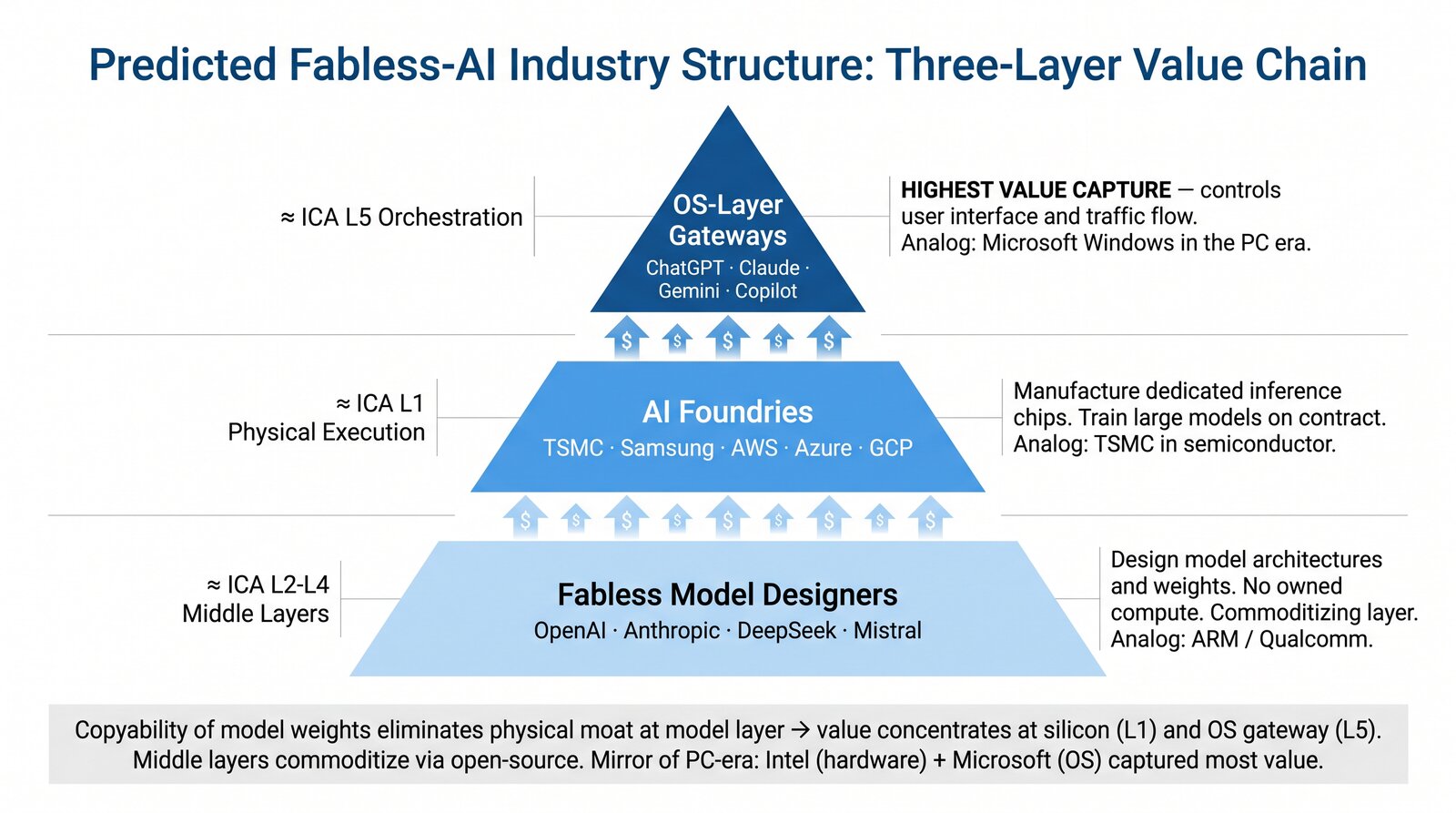}
  \caption{Predicted three-layer Fabless-AI industry structure (right) and its semiconductor analog (left).
  Fabless model designers (the future ARM/Qualcomm of AI) design architectures and weights without owning compute.
  AI foundries (TSMC/Samsung plus hyperscale cloud) manufacture dedicated inference chips and provide large-scale training.
  OS-layer gateways (ChatGPT, Claude, Gemini) control the user interface and capture disproportionate value---
  replaying the Microsoft-versus-Intel dynamic of the PC era, where the OS gateway
  captured more value than the chip manufacturer.
  The fundamental driver: copyability of weights eliminates the physical moat at the model layer,
  pushing value to the silicon (L1) and the gateway (L5).}
  \label{fig:social:fabless_ai}
\end{figure}

%% file: en_sections/section_implementation.tex
\section{Open-Source Implementation and Engineering Practice}
\label{sec:implementation}

Viewed through an engineering lens, today's open-source ecosystem already contains several projects that serve as de facto prototypes for the model-native computing architecture envisioned in this paper. Figure~\ref{fig:impl_map} maps these systems onto the six-layer ICA architecture. In this section, we survey representative systems along three dimensions of the ICA stack (the serving layer (L2), the agent layer (L5), and the protocol layer (L4)) and distill five concrete implementation recommendations grounded in both industrial practice and the design axioms introduced earlier.

\input{figures/fig_impl_map}

\subsection{Serving Layer (L2): Inference Serving and KV Cache Management}
\label{sec:impl-serving}

The serving layer corresponds to ICA's L2 and is responsible for model weight loading, KV cache management, continuous batching, and prefill--decode scheduling. This layer has reached a high degree of engineering maturity, with three distinct design philosophies emerging in production systems.

\textbf{vLLM} centers on PagedAttention, which organizes the KV cache into fixed-size blocks and maps them to non-contiguous physical GPU memory through a block table, effectively replicating the virtual memory management of a conventional operating system~\cite{kwon2023pagedattention,vllm_docs_2026}. Its Automatic Prefix Caching feature extends this analogy by enabling cross-request reuse of shared prompt prefixes, functionally equivalent to shared read-only code pages in an L2 hardware cache~\cite{vllm_prefix_2026}.

\textbf{SGLang} takes a complementary approach built around structured program execution and RadixAttention~\cite{sglang2024,sglang_docs_2026}. RadixAttention maintains a radix tree that indexes the token prefixes of all historical requests, automatically discovering and reusing the longest common prefix KV cache. This yields notably higher cache hit rates in multi-turn dialogue and batch inference workloads. Empirical evaluations show that SGLang achieves up to 29\% higher throughput than vLLM on certain workloads~\cite{sglang2024}.

\textbf{TGI} (Text Generation Inference) from Hugging Face and \textbf{TensorRT-LLM} from NVIDIA represent production-oriented deployment paths~\cite{tgi_docs_2026,tensorrtllm_docs_2026}. TGI emphasizes a unified multi-model interface and operational simplicity, while TensorRT-LLM pursues maximum throughput through hardware-specific optimizations such as FP8 quantization and Tensor Core acceleration on NVIDIA GPUs.

Together, these three families of systems, namely vLLM (virtual-memory-style management), SGLang (structured cache reuse), and TGI/TensorRT-LLM (hardware-deep optimization), constitute the full engineering spectrum of the L2 serving layer.

\subsection{Agent Layer (L5): Agent Runtimes and Operating Systems}
\label{sec:impl-agent}

The agent layer maps to ICA's L5 and is responsible for task planning, agent scheduling, permission approval, failure recovery, and state commitment. This is the most active and diverse area of the current open-source ecosystem, spanning three sub-directions.

\subsubsection{General-Purpose Agents}

\textbf{OpenHands} provides an open agent platform supporting multiple LLM backends, sandboxed execution, and extensible tool interfaces~\cite{openhands_paper_2025}. \textbf{SWE-agent} targets software engineering tasks, unifying code editing, testing, and debugging through a structured Agent-Computer Interface (ACI)~\cite{sweagent2024}. \textbf{AutoGPT} explores autonomous, long-running workflows with automatic goal decomposition~\cite{autogpt_platform_2026}. \textbf{Letta} (built by the original MemGPT team) focuses on stateful agents that treat long-term memory as a first-class citizen~\cite{letta_stateful_2026}.

\subsubsection{Agent Operating Systems}

Several projects have begun to treat the agent runtime itself as an operating system. \textbf{ArbiterOS} introduces a governance layer and proposes a hierarchical architecture in which a governor manages the ``probabilistic CPU''~\cite{arbiteros2025}. \textbf{AgentOS} constrains agent behavior through a reasoning kernel governed by structured OS logic~\cite{agentos2025}. \textbf{UFO$^2$} demonstrates the feasibility of a desktop-scale agent OS, organizing a HostAgent, AppAgents, and a GUI--API operation layer into a hierarchy~\cite{ufo2_2025}. \textbf{Aura} brings a security-first agent architecture to mobile devices, comprising a System Agent, sandboxed App Agents, and an Agent Kernel~\cite{aura2025}.

\subsubsection{Industrial Agent Systems}

Two industrial systems merit particular attention because they explicitly adopt OS-building conventions. \textbf{OpenAI Codex} elevates sub-agents, skills, MCP server integration, sandboxed execution, and approval policies to first-class citizens in its runtime~\cite{openai_codex_overview_2026,openai_codex_skills_2026,openai_codex_sandbox_2026,openai_codex_approvals_2026}. \textbf{Anthropic Claude Code} offers sub-agents, skills, hooks, MCP connections, and project memory files, defaulting to read-only permissions with explicit approval gates~\cite{anthropic_claudecode_overview_2026,anthropic_claudecode_security_2026}. Both systems signal that the industry has begun to construct agent runtimes as bona fide operating systems.

\subsection{Protocol Layer (L4): Standardized Tool Connection and Agent Interconnection}
\label{sec:impl-protocol}

The protocol layer corresponds to ICA's L4 and is undergoing a fundamental transition from point-to-point tool invocations to standardized bus protocols.

\textbf{MCP} (Model Context Protocol) provides a standardized protocol for connecting AI applications to external data sources, tools, and workflows through a JSON-RPC specification~\cite{mcp_intro_2026}. By the end of 2025, Anthropic reported over 10{,}000 active public MCP servers~\cite{mcpadoption2026}. In the analogy to traditional computer architecture, MCP plays the role of a peripheral bus such as PCIe or USB: a universal connector between the compute core and external capabilities.

\textbf{A2A} (Agent-to-Agent Protocol), introduced by Google in April 2025, targets inter-agent communication as a standardized protocol supporting capability discovery, task lifecycle management, and multimodal collaboration~\cite{google_a2a_2025,agentprotocols2025}. It has been donated to the Linux Foundation for vendor-neutral governance~\cite{a2alinux2025}. If MCP is the \textit{vertical bus} connecting an agent to its tools, A2A is the \textit{horizontal interconnect} linking agents to one another, analogous to TCP/IP in the traditional network stack. Together, MCP and A2A form complementary axes of the protocol layer.

\subsection{Five Implementation Recommendations}
\label{sec:impl-recommendations}

Drawing on the engineering practices surveyed above and the six design axioms, we distill five recommendations that can directly guide the construction of production model-native computing systems. They span the stack, ordered roughly from architecture down to operations. \textbf{R1} separates the deterministic control plane from the probabilistic execution plane, concentrating permission, audit, and recovery logic at L4--L5 rather than letting it leak into the inference path. \textbf{R2} tiers context into a hot working window and cold persistent storage, the L3 analogue of a memory hierarchy. \textbf{R3} versions every behavior-affecting object---prompts, tool schemas, skills, memory files---so that interfaces evolve compatibly, the way an ISA or ABI does. \textbf{R4} enforces explicit per-agent resource quotas (context size, tool-call count, inference time, monetary cost) as an operating system does through cgroups. \textbf{R5} treats failure recovery as a distributed-systems concern, with checkpoint, retry, and rollback for long-running tasks. Figure~\ref{fig:impl_recommendations} maps each recommendation to its primary ICA layer (\textbf{Key}) and its supporting roles (\textbf{Partial}), together with the governing axiom; the subsections below expand each in turn.

\input{figures/fig_impl_recommendations}

\subsubsection{Separate the Control Plane from the Data Plane}

Deterministic control logic (permission approval, audit logging, failure recovery) must be peeled away from the probabilistic inference path. These two paths correspond, respectively, to the deterministic control plane and the probabilistic execution plane in the dual-plane architecture introduced in Section~\ref{sec:framework}. Codex's approval policies and Claude Code's hooks are concrete instantiations of this principle~\cite{openai_codex_approvals_2026,anthropic_claudecode_security_2026}.

\subsubsection{Separate Short-Term Context from Long-Term Memory}

High-frequency ``hot context'' (the active dialogue window) and low-frequency ``cold memory'' (historical summaries, knowledge-base indices) should reside on different storage media with different management strategies. This maps directly to the hot/warm/cold tiering design of ICA's L3 layer. MemGPT's virtual context management exemplifies this principle~\cite{memgpt2023}. The business impact is tangible: Augment Code reported that a project a CTO estimated would take 4--8 months was completed in just 2 weeks, demonstrating how effective context management accelerates developer onboarding and project delivery in enterprise settings~\cite{anthropic2026agenticreport}.

\subsubsection{Version Every Behavior-Affecting Object}

Prompt templates, tool schemas, skill packages, memory files, and MCP server capability descriptions should all carry version numbers and changelogs~\cite{mcp_intro_2026,agentprotocols2025}. Traditional computers guarantee backward compatibility through the stability of ISA and ABI boundaries; model-native computing systems require an analogous ``interface contract'' to accommodate rapid evolution.

\subsubsection{Enforce Resource Quotas Like an Operating System}

Every agent or task should operate under an explicit resource budget: an upper bound on context window size, tool call count, inference time, and monetary cost~\cite{linux_cgroup_2026}. This recommendation instantiates Axiom~5 (Least Privilege) and Axiom~6 (Observability) at the resource management layer, mirroring the role of cgroups in Linux.

\subsubsection{Design Failure Recovery Like a Distributed System}

Long-running agent tasks may fail due to model timeouts, tool errors, or context overflow. The system must provide checkpoint, retry, and rollback mechanisms akin to those in distributed databases, ensuring that side-effect operations remain auditable and reversible~\cite{raft2014,spanner2012}. Rakuten's experience with Claude Code provides compelling evidence: an agent autonomously completed an activation-vector extraction task across 12.5 million lines of code in the vLLM project over 7 hours with 99.9\% numerical accuracy, demonstrating that long-running agents with implicit checkpoint and retry capabilities can already handle industrial-scale workloads~\cite{anthropic2026agenticreport}.

%% file: figures/fig_impl_map.tex
\begin{figure}[htbp]
    \providecolor{c1}{HTML}{234C84}
    \providecolor{c2}{HTML}{2F5F9E}
    \providecolor{c3}{HTML}{3E73B2}
    \providecolor{m4}{HTML}{E38431}
    \providecolor{m5}{HTML}{EE9D50}
    \providecolor{g6}{HTML}{7B8A9E}
    \centering
    \resizebox{0.98\linewidth}{!}{%
    \begin{tikzpicture}[
        x=1cm, y=1cm,
        font=\sffamily,
        >={Stealth[length=2mm,width=1.4mm]},
        servbox/.style={
            draw=classical!70, rounded corners=5pt,
            fill=white, text=classical!80!black,
            font=\small\bfseries, align=center,
            minimum width=3.0cm, minimum height=1.0cm,
            inner sep=4pt, line width=0.8pt
        },
        layerbox/.style={
            draw=none, rounded corners=5pt,
            text=white, font=\small\bfseries, align=center,
            minimum width=5.8cm, minimum height=0.9cm,
            inner sep=4pt
        },
        agentbox/.style={
            draw=modelnative!70, rounded corners=5pt,
            fill=white, text=modelnative!80!black,
            font=\small\bfseries, align=center,
            minimum width=3.2cm, minimum height=1.0cm,
            inner sep=4pt, line width=0.8pt
        },
        chiplab/.style={
            draw=gray!50, rounded corners=3pt,
            fill=gray!15, font=\tiny\bfseries,
            inner xsep=3pt, inner ysep=2pt,
            text=gray!65!black
        },
        arrow/.style={-{Stealth[length=2mm]}, thick, black!50},
        darrow/.style={-{Stealth[length=2mm]}, thick, dashed, black!40},
    ]
    
    \def\sxl{0.0} \def\sxr{3.3} \def\lxl{4.1} \def\lxr{9.9} \def\axl{10.7} \def\axr{14.0} \def\rxl{15.4}
    
    \def\yL{0.0} \def\yLa{1.3} \def\yLb{2.6} \def\ybnd{3.38} \def\yLc{4.15} \def\yLd{5.45} \def\yLe{6.75}
    
    \draw[draw=classical!50, rounded corners=8pt, fill=classical!5, line width=0.8pt]
        (\sxl-0.15, -0.65) rectangle (\sxr+0.15, 7.55);
    \node[font=\small\bfseries, text=classical!70!black] at ({(\sxl+\sxr)/2}, 7.25) {Serving Layer};
    
    \node[servbox] (vllm) at ({(\sxl+\sxr)/2}, 6.35) {vLLM\\{\scriptsize (PagedAttention)}};
    \node[servbox] (sgl)  at ({(\sxl+\sxr)/2}, 5.05) {SGLang\\{\scriptsize (RadixAttention)}};
    \node[servbox] (tgi)  at ({(\sxl+\sxr)/2}, 3.55) {TGI\\{\scriptsize (Unified multi-model}\\[-1pt]{\scriptsize interface)}};
    \node[servbox] (trt)  at ({(\sxl+\sxr)/2}, 2.0) {TensorRT-LLM\\{\scriptsize (FP8, Tensor-Core)}};
    \node[servbox] (kf)   at ({(\sxl+\sxr)/2}, 0.6) {kernel fusion};
    
    \node[layerbox, fill=c1] (L1) at ({(\lxl+\lxr)/2}, \yL) {L1\quad Physical Execution};
    \node[layerbox, fill=c2] (L2) at ({(\lxl+\lxr)/2}, \yLa) {L2\quad Inference Serving};
    \node[layerbox, fill=c3] (L3) at ({(\lxl+\lxr)/2}, \yLb) {L3\quad Context Management};
    
    \draw[dashed, thick, black!38, line width=0.8pt] (\lxl-0.2, \ybnd) -- (\axr+0.15, \ybnd);
    \node[font=\tiny\itshape, text=black!42, anchor=north]
        at ({(\lxl+\lxr)/2}, \ybnd+0.05) {L3--L4 Dual-plane boundary};
    
    \node[layerbox, fill=m4, minimum height=1.4cm, inner ysep=2pt] (L4) at ({(\lxl+\lxr)/2}, \yLc)
        {\raisebox{0.28cm}{L4\quad Semantic Interface}};
    \node[chiplab] at ({(\lxl+\lxr)/2 - 1.9}, \yLc - 0.42) {MCP};
    \node[chiplab] at ({(\lxl+\lxr)/2 - 0.4}, \yLc - 0.42) {JSON-RPC};
    \node[chiplab] at ({(\lxl+\lxr)/2 + 1.6}, \yLc - 0.42) {tool-server-bus};
    
    \node[layerbox, fill=m5] (L5) at ({(\lxl+\lxr)/2}, \yLd) {L5\quad Orchestration Layer};
    
    \node[layerbox, fill=g6, minimum height=1.4cm, inner ysep=2pt] (L6) at ({(\lxl+\lxr)/2}, \yLe)
        {\raisebox{0.28cm}{L6\quad Application Layer}};
    \node[chiplab] at ({(\lxl+\lxr)/2}, \yLe - 0.38) {agent-communication};
    
    \node[agentbox] (a2a) at ({(\axl+\axr)/2}, \yLe) {A2A\\{\scriptsize agent-to-agent}};
    \node[agentbox] (codex) at ({(\axl+\axr)/2}, \yLd) {Codex, Claude\\Code, OpenHands};
    \node[agentbox, minimum height=1.5cm] (arbiter) at ({(\axl+\axr)/2}, \yLc - 0.28) {ArbiterOS / AgentOS\\{\scriptsize governance · reasoning}\\{\scriptsize kernel · sandboxing}};
    
    \draw[draw=gray!40, rounded corners=6pt, fill=gray!5, line width=0.6pt]
        (\rxl-0.15, -0.65) rectangle (\rxl+5.25, 7.55);
    \node[font=\small\bfseries, text=black, anchor=south] at (\rxl+2.5, 6.65) {Five recs};

    \node[font=\small, align=left, text width=5.0cm] (R1) at (\rxl+2.5, 5.9) {\textbf{R1. Separate control \& data planes}\\[1pt]{\scriptsize\textcolor{gray!55}{Codex approvals, Claude Code hooks}}};
    \node[font=\small, align=left, text width=5.0cm] (R2) at (\rxl+2.5, 4.5) {\textbf{R2. Separate hot context \& cold memory}\\[1pt]{\scriptsize\textcolor{gray!55}{MemGPT virtual context, L3 tiering}}};
    \node[font=\small, align=left, text width=5.0cm] (R3) at (\rxl+2.5, 3.1) {\textbf{R3. Version every behavior object}\\[1pt]{\scriptsize\textcolor{gray!55}{MCP schemas, tool ABI versioning}}};
    \node[font=\small, align=left, text width=5.0cm] (R4) at (\rxl+2.5, 1.7) {\textbf{R4. Enforce resource quotas}\\[1pt]{\scriptsize\textcolor{gray!55}{cgroup-style budgets, Axiom 5 \& 6}}};
    \node[font=\small, align=left, text width=5.0cm] (R5) at (\rxl+2.5, 0.3) {\textbf{R5. Design failure recovery}\\[1pt]{\scriptsize\textcolor{gray!55}{Checkpoint/retry, distributed DB patterns}}};
    
    \draw[arrow] (tgi.east) to[out=0, in=180] (L3.west);
    \draw[arrow] (trt.east) to[out=0, in=180] (L2.west);
    \draw[arrow] (kf.east)  to[out=0, in=180] (L1.west);
    \draw[arrow] (a2a.west)     -- (L6.east);
    \draw[arrow] (codex.west)   -- (L5.east);
    \draw[arrow] (arbiter.west) -- (L4.east);
    \draw[darrow] (a2a.east)   to[out=0, in=180] (R1.west);
    \draw[darrow] (codex.east) to[out=0, in=180] (R1.west);
    \draw[darrow, rounded corners=6pt] (L3.east) -- (14.8, \yLb) -- (14.8, 4.5) -- (R2.west);
    \draw[darrow] (arbiter.east) to[out=0, in=180] (R3.west);
    \draw[darrow, rounded corners=6pt] (L5.east) -- (10.1, \yLd) -- (10.1, 1.7) -- (R4.west);
    \draw[darrow] (L2.east) to[out=0, in=180] (R5.west);
    
    \end{tikzpicture}%
    }
    \caption{Open-source ecosystem mapped to the six-layer ICA architecture.}
    \label{fig:impl_map}
    \end{figure}
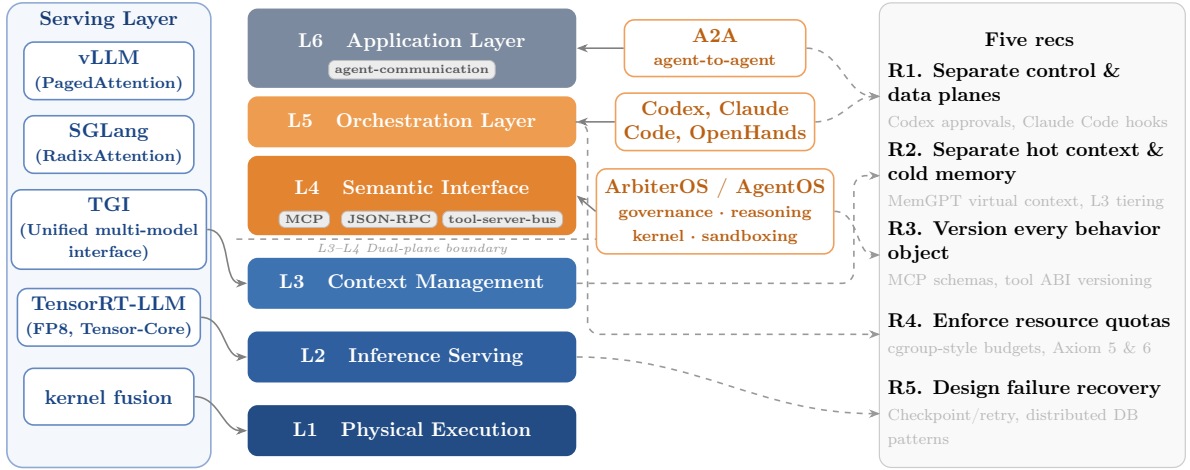

%% file: figures/fig_impl_recommendations.tex
\begin{figure}[htbp]
\centering
\providecolor{r1Color}{HTML}{4A78A8}
\providecolor{r2Color}{HTML}{5C9ED4}
\providecolor{r3Color}{HTML}{E07B39}
\providecolor{r4Color}{HTML}{C44E52}
\providecolor{r5Color}{HTML}{3A8A3A}
\providecolor{dotFull}{HTML}{333333}
\providecolor{dotPart}{HTML}{999999}
\resizebox{0.8\textwidth}{!}{%
\begin{tikzpicture}[
    font=\small, >=Latex,
    recbox/.style={draw=#1!65!black, fill=#1!14, rounded corners=5pt,
        minimum width=3.4cm, minimum height=1.2cm,
        align=center, font=\small\bfseries, text=#1!80!black, line width=0.7pt,
        inner sep=5pt},
    layercell/.style={draw=#1!50, fill=#1!12, rounded corners=3pt,
        minimum width=1.55cm, minimum height=0.75cm,
        align=center, font=\scriptsize, text=#1!75!black, line width=0.4pt},
    nabox/.style={fill=gray!8, draw=gray!25, rounded corners=3pt,
        minimum width=1.55cm, minimum height=0.75cm,
        align=center, font=\scriptsize, text=gray!50, line width=0.3pt},
    axiomtag/.style={draw=#1!55, fill=#1!10, rounded corners=2pt,
        font=\tiny\bfseries, text=#1!75!black, inner xsep=3pt, inner ysep=2pt,
        line width=0.4pt},
    hdr/.style={font=\small\bfseries, text=black!65},
]

\def\xR{-0.2}   
\def\xLone{2.8} \def\xLtwo{4.5} \def\xLthree{6.2}
\def\xLfour{7.9} \def\xLfive{9.6} \def\xLsix{11.3}
\def\xAxiom{12.8}

\node[hdr] at (\xR, 6.5)     {Recommendation};
\node[hdr, text=blue!60!black]   at (\xLone,   6.5) {L1};
\node[hdr, text=blue!60!black]   at (\xLtwo,   6.5) {L2};
\node[hdr, text=blue!60!black]   at (\xLthree, 6.5) {L3};
\node[hdr, text=orange!60!black] at (\xLfour,  6.5) {L4};
\node[hdr, text=orange!60!black] at (\xLfive,  6.5) {L5};
\node[hdr, text=orange!60!black] at (\xLsix,   6.5) {L6};
\node[hdr, text=red!60!black]    at (\xAxiom,  6.5) {Axiom};

\def\yRa{5.1}
\node[recbox=r1Color, text width=3.0cm] (r1) at (\xR, \yRa)
    {R1\\Separate control\\$\&$ data planes};
\node[nabox]          at (\xLone,   \yRa) {---};
\node[nabox]          at (\xLtwo,   \yRa) {---};
\node[nabox]          at (\xLthree, \yRa) {---};
\node[layercell=r1Color] at (\xLfour,  \yRa) {Key};
\node[layercell=r1Color] at (\xLfive,  \yRa) {Key};
\node[nabox]          at (\xLsix,   \yRa) {---};
\node[axiomtag=r1Color]  at (\xAxiom,  \yRa) {A3, A6};

\def\yRb{3.4}
\node[recbox=r2Color, text width=3.0cm] (r2) at (\xR, \yRb)
    {R2\\Hot context /\\cold memory};
\node[nabox]          at (\xLone,   \yRb) {---};
\node[layercell=r2Color] at (\xLtwo,   \yRb) {Partial};
\node[layercell=r2Color] at (\xLthree, \yRb) {Key};
\node[nabox]          at (\xLfour,  \yRb) {---};
\node[nabox]          at (\xLfive,  \yRb) {---};
\node[nabox]          at (\xLsix,   \yRb) {---};
\node[axiomtag=r2Color]  at (\xAxiom,  \yRb) {A1, A4};

\def\yRc{1.7}
\node[recbox=r3Color, text width=3.0cm] (r3) at (\xR, \yRc)
    {R3\\Version every\\behavior object};
\node[nabox]          at (\xLone,   \yRc) {---};
\node[nabox]          at (\xLtwo,   \yRc) {---};
\node[layercell=r3Color] at (\xLthree, \yRc) {Partial};
\node[layercell=r3Color] at (\xLfour,  \yRc) {Key};
\node[layercell=r3Color] at (\xLfive,  \yRc) {Partial};
\node[nabox]          at (\xLsix,   \yRc) {---};
\node[axiomtag=r3Color]  at (\xAxiom,  \yRc) {A2};

\def\yRd{0.0}
\node[recbox=r4Color, text width=3.0cm] (r4) at (\xR, \yRd)
    {R4\\Enforce resource\\quotas};
\node[nabox]          at (\xLone,   \yRd) {---};
\node[layercell=r4Color] at (\xLtwo,   \yRd) {Partial};
\node[layercell=r4Color] at (\xLthree, \yRd) {Partial};
\node[nabox]          at (\xLfour,  \yRd) {---};
\node[layercell=r4Color] at (\xLfive,  \yRd) {Key};
\node[nabox]          at (\xLsix,   \yRd) {---};
\node[axiomtag=r4Color]  at (\xAxiom,  \yRd) {A5, A6};

\def\yRe{-1.7}
\node[recbox=r5Color, text width=3.0cm] (r5) at (\xR, \yRe)
    {R5\\Design failure\\recovery};
\node[nabox]          at (\xLone,   \yRe) {---};
\node[layercell=r5Color] at (\xLtwo,   \yRe) {Partial};
\node[layercell=r5Color] at (\xLthree, \yRe) {Partial};
\node[nabox]          at (\xLfour,  \yRe) {---};
\node[layercell=r5Color] at (\xLfive,  \yRe) {Key};
\node[layercell=r5Color] at (\xLsix,  \yRe) {Partial};
\node[axiomtag=r5Color]  at (\xAxiom,  \yRe) {A3, A6};

\def\legy{-3.1}
\node[layercell=dotFull, minimum width=1.3cm, text=dotFull!70] at (0.6, \legy) {Key};
\node[font=\scriptsize, anchor=west] at (1.45, \legy) {Primary layer};
\node[layercell=dotPart, minimum width=1.3cm, text=dotPart!70] at (5.1, \legy) {Partial};
\node[font=\scriptsize, anchor=west] at (5.95, \legy) {Supporting role};
\node[nabox, minimum width=1.3cm] at (9.6, \legy) {---};
\node[font=\scriptsize, anchor=west] at (10.45, \legy) {Not applicable};

\end{tikzpicture}}%
\caption{Mapping of the five implementation recommendations (R1--R5) onto ICA layers (L1--L6) and their governing design axioms. \textbf{Key} cells indicate the primary ICA layer at which the recommendation must be enforced; \textbf{Partial} cells indicate supporting roles. R1 (control/data plane separation) targets L4--L5; R2 (context tiering) targets L3; R3 (versioning) targets L4; R4 (resource quotas) and R5 (failure recovery) both center on L5 with support from adjacent layers.}
\label{fig:impl_recommendations}
\end{figure}

%% file: en_sections/section_roadmap.tex
\section{Research Roadmap}
\label{sec:roadmap}

Table~\ref{tab:roadmap} presents a research roadmap organized around the ICA six-layer model, spanning three phases: short-term (1--2 years), mid-term (2--4 years), and long-term (4--8 years). Each entry specifies the problem to be solved, the proposed method, and the expected outcome, making every item a concrete, actionable research project. Figure~\ref{fig:roadmap_swimlane} provides a visual overview of the roadmap as a swim-lane diagram organized by ICA layer and time horizon.

\footnotesize
\renewcommand{\arraystretch}{1.35}
\begin{longtable}{>{\small}p{0.06\textwidth}>{\small}p{0.06\textwidth}>{\small}p{0.13\textwidth}>{\small}p{0.25\textwidth}>{\small}p{0.18\textwidth}>{\small}p{0.14\textwidth}}
\caption{Research agenda organized by ICA layer, phase, and associated heuristic or axiom.}
\label{tab:roadmap}\\
\toprule
\textbf{Phase} & \textbf{ICA Layer} & \textbf{Topic} & \textbf{Method} & \textbf{Metrics} & \textbf{Related Heuristic / Axiom} \\
\midrule
\endfirsthead
\toprule
\textbf{Phase} & \textbf{ICA Layer} & \textbf{Topic} & \textbf{Method} & \textbf{Metrics} & \textbf{Related Heuristic / Axiom} \\
\midrule
\endhead

Short-term & L2 & KV cache hit-rate optimization & Semantic eviction; KV quantization; prefix sharing & $H$, $S$, TTFT & Heur.~I (Semantic Locality) \\[3pt]
Short-term & L3 & Unified context compiler & Hot / warm / cold tiering; version stamps with TTL & $W_{\mathrm{eff}}$, compression ratio & Heur.~II (Context Budget) \\[3pt]
Short-term & L4 & Tool ABI standardization & Schema versioning; capability annotation & Invocation success rate; security violation rate & Axiom~5 (Least Privilege) \\[3pt]
Short-term & L5 & Software engineering agent kernel & Unified ACI; specialized sub-agents & Task completion rate; $E$ & Heur.~III (Agent Speedup) \\[3pt]
Mid-term & L3 & Semantic page replacement & Learned eviction; summary write-back retrieval & $\beta(L)$ curve & Heur.~II \\[3pt]
Mid-term & L5 & Consistency-controlled shared memory & Event sourcing; conflict detection & Memory correctness rate & Axiom~4 (Virtualization) \\[3pt]
Mid-term & L2 & Heterogeneous model collaboration & Speculative decoding; MoE routing & End-to-end latency; $H$ & Heur.~I \\[3pt]
Mid-term & L4 & Agent security architecture & Sandboxing; audit logging; capability security & Audit completeness & Axioms~5,~6 \\[3pt]
Long-term & L1--L6 & Intelligent ISA and programmable skill layer & Task DSL; controlled output formats; reusable skill packages & Generalization ability & Axioms~2,~3 \\[3pt]
Long-term & L3--L5 & Stateful individual agents & Long-term memory; experience abstraction and strategy learning & Experience utilization rate & Heur.~II,~III \\[3pt]
Long-term & L1--L6 & Distributed model-native compute fabric & Memory replication; consistency layers; load balancing & Cluster utilization & Heur.~I,~III \\[3pt]
Long-term & L5 & Governance-layer primitives & Probabilistic / deterministic dual-plane; audit logs; permission declarations & Governance coverage & Axioms~3,~5,~6 \\[3pt]
Long-term & L1--L2 & Substrate-independent compute stack & Non-silicon interface generalization; baseline intelligence score threshold & Interface compatibility rate & Axiom~2 \\[3pt]
Long-term & L4--L5 & Physical-world interface drivers & Sensor / actuator driver abstraction; simulation environment integration & Driver coverage rate & Axioms~2,~5 \\[3pt]
Mid-term & Full stack & Weakest-link evaluation framework & Worst-at-$k$; multi-dimensional capability profiling; compliance test suite & Minimum dimension score & Weakest-Link Principle \\[3pt]
\bottomrule
\end{longtable}
\renewcommand{\arraystretch}{1.0}
\normalsize

\input{figures/fig_roadmap_swimlane}

\subsection{Short-Term Research Topics}

The short-term items focus on optimizing and standardizing existing engineering prototypes, with the goal of bringing current systems to production-grade performance and reliability.

\paragraph{KV cache hit-rate optimization (L2).}
The decoding phase of LLM inference is a classic memory-bandwidth-bound problem~\cite{gholami2024memorywall}, and the KV cache hit rate $H$ directly governs the inference speedup ratio $S$ described by the Semantic Locality heuristic (Heuristic~I).
Three complementary techniques can be brought to bear.
First, \textit{semantic eviction} strategies, exemplified by H2O's heavy-hitter identification~\cite{h2o2023}, retain the KV entries most likely to be reused based on attention weight patterns rather than recency alone.
Second, \textit{KV quantization} compresses the KV cache with minimal quality loss: KVQuant achieves up to $8\times$ compression~\cite{hooper2024kvquant}, while KIVI introduces asymmetric 2-bit quantization~\cite{liu2024kivi}.
Third, \textit{prefix sharing} across requests, as implemented by PagedAttention~\cite{kwon2023pagedattention} and RadixAttention~\cite{sglang2024}, enables multiple requests with common system prompts to reuse the same KV blocks, directly increasing $H$.
The target outcome is to raise $H$ from the current 0.5--0.7 range to above 0.85 under typical agent workloads, yielding the 5--10$\times$ speedup consistent with Heuristic~I.

\paragraph{Unified context compiler (L3).}
Different agent frameworks manage context in idiosyncratic ways: MemGPT uses virtual context management~\cite{memgpt2023}, Letta builds stateful agents with explicit memory tiers~\cite{letta_stateful_2026}, and HiAgent employs hierarchical working memory~\cite{hiagent2024}.
What is missing is a unified \textit{context compiler}, analogous to an operating system's page replacement subsystem, that decides, for each piece of information, whether to load it verbatim, as a summary, or as a retrievable index entry.
We propose a hot / warm / cold tiered context management strategy, where each context block carries a version stamp and a time-to-live (TTL) annotation.
The expected result is a $\geq 50\%$ increase in the effective context utilization ratio $W_{\mathrm{eff}}/C$ without degrading task completion rates, thereby providing direct empirical grounding for the Context Budget heuristic (Heuristic~II).

\paragraph{Tool ABI standardization (L4).}
While MCP and A2A have made significant strides at the protocol level~\cite{mcp_intro_2026,agentprotocols2025}, tool schemas themselves still lack unified specifications for versioning, backward compatibility, and capability annotation.
We propose enriching tool schemas with semantic version numbers, input/output type constraints, side-effect declarations, and capability annotations, transforming L4 into a genuine Application Binary Interface (ABI) for the model-native computing stack.
The expected outcome is a measurable increase in tool invocation success rates and a decrease in security violations, providing the interface foundation required to operationalize Axiom~5 (Least Privilege).

\paragraph{Software engineering agent kernel (L5).}
SWE-bench and SWE-agent have demonstrated that current code agents still achieve limited resolution rates on real-world GitHub issues~\cite{swebench2023,sweagent2024}.
A key bottleneck is the lack of a unified Agent-Computer Interface (ACI) standard: different frameworks adopt inconsistent sub-agent delegation strategies.
We propose standardizing the ACI specification and designing specialized sub-agents (a code-search agent, an edit agent, and a test agent) whose orchestration efficiency $E$ can be characterized through the Agent Speedup heuristic (Heuristic~III).
The target is to increase task completion rates on SWE-bench-class benchmarks while raising $E$ from the current 0.3--0.5 range to above 0.6.
Figure~\ref{fig:roadmap:metrics} summarizes the current state and target for each short-term metric.
\input{figures/fig_roadmap_metrics}

\subsection{Mid-Term Research Topics}

The mid-term items shift focus from per-layer optimization to cross-layer coordination and system-level security, moving model-native computing from ``point optimizations'' to ``full-stack co-design.''

\paragraph{Semantic page replacement (L3).}
Classical operating systems replace pages using deterministic access patterns (LRU, CLOCK). In a model-native computing system, ``page replacement'' must be governed by \textit{semantic relevance}, deciding which context blocks to evict based on their pertinence to the current task.
We propose a learned eviction policy that scores each context block against the active task and generates compressed summaries for evicted blocks, enabling later retrieval.
The primary research output is an empirically measured $\beta(L)$ curve that characterizes how attention retention decays with position, providing the Context Budget heuristic with calibrated quantitative parameters.

\paragraph{Consistency-controlled shared memory (L5).}
When multiple agents collaborate, shared memory consistency management lacks a unified framework.
Concurrent modifications to the same memory block can cause conflicts and information loss, the analog of cache coherence problems in multiprocessor systems.
We propose introducing event sourcing and conflict detection mechanisms, combined with tunable consistency levels ranging from eventual consistency to strong consistency~\cite{raft2014,cap2002}.
The target is to ensure that multi-agent memory correctness rates match single-agent baselines, providing empirical grounding for Axiom~4 (Virtualization) at the memory layer.

\paragraph{Heterogeneous model collaboration (L2).}
Models of different sizes offer distinct trade-offs among latency, cost, and capability, yet current systems typically deploy a single model.
Heterogeneous model collaboration, using smaller draft models for speculative decoding~\cite{leviathan2023speculative} and mixture-of-experts (MoE) routing~\cite{switch2022,mixtral2024}, can direct simple queries to lightweight models and complex queries to large ones.
The expected result is a reduction in end-to-end latency at equivalent output quality, with a secondary benefit of increased KV cache hit rates $H$.

\paragraph{Agent security architecture (L4).}
As agents acquire increasingly powerful external capabilities, including reading and writing files, executing commands, and accessing the network, the security perimeter becomes dangerously blurred.
We propose embedding sandboxing, auditing, and capability-based security directly into the agent runtime~\cite{debenedetti2025camel,ironclaw2025,agenticsecurity2025}, ensuring that every side-effecting operation undergoes explicit approval and produces an auditable event record.
The target is 100\% audit completeness: every side-effecting operation across all ICA layers must be traceable, establishing the security infrastructure required to enforce Axioms~5 and~6 in practice.

\paragraph{Weakest-link evaluation framework (full stack).}
The Weakest-Link Principle introduced in Section~\ref{sec:paradigm} posits that system intelligence is bounded by its weakest capability dimension, yet current evaluation practice overwhelmingly relies on aggregate or average metrics.
We propose an evaluation suite built around worst-at-$k$ scoring and multi-dimensional capability profiling (covering reasoning, safety, tool use, multilingual understanding, and physical-world interaction) that serves as a compliance test for the intelligent ISA.
The expected outcome is a compliance-grade evaluation standard for intelligent compute cores, transforming substrate independence from an architectural principle into an enforceable contractual property.

\subsection{Long-Term Research Topics}

The long-term items target fundamental architectural innovations, with the goal of constructing a truly native model-based computing architecture rather than incrementally improving today's systems.

\paragraph{Intelligent ISA and programmable skill layer (L1--L6).}
Prompts, tool descriptions, and skill packages collectively form the ``soft instructions'' of model-native computing, yet they currently lack formal semantics and stability guarantees.
We envision developing task-specific domain-specific languages (DSLs), controlled output formats, and reusable skill packages that together constitute a stable interface layer, the model-native analog of a classical Instruction Set Architecture.
As argued in Section~\ref{sec:paradigm}, the stability of this interface layer is a prerequisite for substrate independence: whether the underlying hardware is a silicon GPU, a neuromorphic chip, or a biological compute substrate, any implementation satisfying the intelligent ISA contract should be capable of running the same agent framework.
The expected outcome is significantly improved agent generalization, enabling the same skill package to migrate seamlessly across different models and runtimes, thereby operationalizing Axioms~2 (Layered Abstraction) and~3 (Probabilistic Execution).

\paragraph{Stateful individual agents (L3--L5).}
Most agents today are stateless: each session starts from scratch, unable to accumulate or reuse experience.
We propose equipping agents with long-term memory and experience abstraction mechanisms, enabling them to learn strategies from historical executions, distill recurring patterns, and proactively optimize their own behavior~\cite{amem2025,agentmemorysurvey2025}.
The expected outcome is a measurable increase in \textit{experience utilization}, defined as the fraction of current task performance attributable to distilled prior experience, which opens new empirical dimensions for both the Context Budget heuristic (Heuristic~II) and the Agent Speedup heuristic (Heuristic~III).

\paragraph{Distributed model-native compute fabric (L1--L6).}
When multiple agent clusters collaborate across nodes, distributed systems problems such as memory replication, state consistency, and load balancing remain largely unsolved in the model-native computing context.
Drawing on distributed database techniques for memory replication and consistency management~\cite{spanner2012,raft2014}, we propose a unified resource scheduling and state management substrate for multi-node agent clusters.
The target is for cluster utilization to scale linearly (or at least near-linearly) with node count, empirically grounding the Semantic Locality heuristic (Heuristic~I) and the Agent Speedup heuristic (Heuristic~III) in distributed settings.

\paragraph{Governance-layer primitives (L5).}
Current governance mechanisms in agent systems (permissions, auditing, compliance) are retrofitted patches rather than first-class architectural primitives.
We propose elevating the dual-plane separation (probabilistic execution plane and deterministic control plane), audit logging, and permission declarations to core architectural primitives~\cite{arbiteros2025,ironclaw2025}, ensuring that every ICA layer has a corresponding governance mechanism.
The target is 100\% governance coverage: every side-effecting operation at every layer falls under governance constraints, fully realizing Axioms~3,~5, and~6 across the entire stack.

\paragraph{Substrate-independent intelligent compute stack (L1--L2).}
Section~\ref{sec:paradigm} discussed the rapid development of non-silicon substrates, including neuromorphic processors, biological computing, and compute-in-memory architectures, yet current ICA interface contracts implicitly assume a GPU-centric execution model.
We propose investigating how L1--L2 interface contracts can be generalized to accommodate diverse substrates: event-driven spiking computation, analog in-memory computation, and the asynchronous response patterns of biological neural networks.
A baseline intelligence score threshold would serve as the core admission criterion.
The expected outcome is a cross-substrate interface contract specification that enables future non-silicon compute cores to integrate seamlessly with existing agent frameworks.

\paragraph{Physical-world interface drivers (L4--L5).}
Section~\ref{sec:paradigm} identified the extension from purely digital interaction to the physical world, spanning robotics, autonomous driving, agriculture, and space navigation, as the next inflection point for intelligent operating systems, yet current L4 tool protocols (MCP, A2A) were designed for the digital domain.
We propose extending ICA's L4--L5 interfaces to support sensors, actuators, simulation environments, and telemetry data streams, following the same abstraction patterns as MCP tool servers.
The expected outcome is a tool driver specification for the physical world such that ``adding a robotic arm driver'' follows the same workflow as ``adding an MCP tool server.''

\subsection{A Note on Decoupling from Model Advances}

It is worth emphasizing that these research topics do not require stronger foundation models as a prerequisite.
Many of the problems, such as context compilation, Tool ABI standardization, and agent security architecture, are fundamentally systems engineering challenges that are orthogonal to improvements in base model capability.
This decoupling mirrors the history of classical computing: architectural breakthroughs have come not only from faster transistors but equally from more mature interfaces and runtimes.
The stability of the ISA enabled a thriving software ecosystem; the invention of virtual memory freed programs from physical memory constraints; the POSIX standard allowed applications to migrate across platforms.
Model-native computing systems stand in need of precisely the same kind of systems-level breakthroughs.

From an industry perspective, the analysis in Section~\ref{sec:sociotechnical} suggests that these systems-level standardization and decoupling efforts are co-evolving with structural changes in the industry itself.
Protocol standardization (MCP, A2A, Tool ABI) reduces inter-layer coupling, creating the technical conditions for the fabless AI industry structure predicted in Section~\ref{sec:sociotechnical}; conversely, growing industry specialization accelerates independent optimization at each layer.
This co-evolution of technology and industry structure recapitulates the central narrative of the semiconductor and personal computer industries over the past four decades.
Figure~\ref{fig:roadmap:dependency} maps the enabling dependencies across all three research phases.
\input{figures/fig_roadmap_dependency}

%% file: figures/fig_roadmap_swimlane.tex
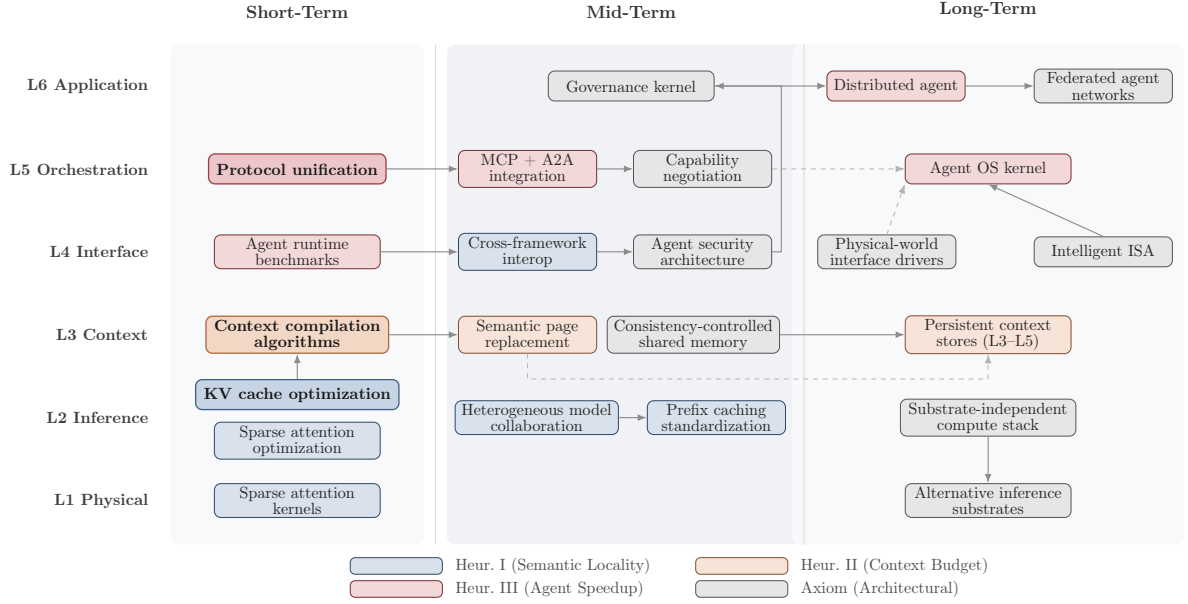
\begin{figure}[htbp]
\centering
\providecolor{lawI}{HTML}{4C78A8}
\providecolor{lawII}{HTML}{D9824B}
\providecolor{lawIII}{HTML}{C44E52}
\providecolor{axiom}{HTML}{8C8C8C}
\providecolor{laneBg}{HTML}{F8F9FA}
\providecolor{laneAlt}{HTML}{F0F2F5}
\resizebox{0.98\textwidth}{!}{%
\begin{tikzpicture}[
    >=Latex,
    font=\small,
    lane label/.style={
        font=\small\bfseries, anchor=east, text=black!75
    },
    phase header/.style={
        font=\normalsize\bfseries, anchor=south, text=black!80
    },
    phase sub/.style={
        font=\small, anchor=north, text=black!50
    },
    bar/.style={
        draw=#1!60!black, fill=#1!22, rounded corners=4pt,
        minimum height=0.65cm, minimum width=3.6cm,
        inner xsep=5pt, inner ysep=2pt,
        font=\small, align=center, line width=0.5pt,
        text=black!85
    },
    bar selected/.style={
        draw=#1!80!black, fill=#1!32, rounded corners=4pt,
        minimum height=0.65cm, minimum width=3.6cm,
        inner xsep=5pt, inner ysep=2pt,
        font=\small\bfseries, align=center, line width=0.8pt,
        text=black!90
    },
    barp/.style={
        draw=#1!60!black, fill=#1!22, rounded corners=4pt,
        minimum height=0.65cm, minimum width=3.0cm,
        inner xsep=4pt, inner ysep=2pt,
        font=\small, align=center, line width=0.5pt,
        text=black!85
    },
    dep arrow/.style={
        -{Latex[length=2mm]}, thick, black!45, line width=0.7pt
    },
    dep arrow dashed/.style={
        -{Latex[length=2mm]}, thick, black!30,
        line width=0.6pt, dashed
    },
]

\def\xShortL{3.5}   \def\xShortR{8.5}
\def\xMidL{9.5}     \def\xMidR{17.0}
\def\xLongL{17.0}   \def\xLongR{25.0}

\def\xShortM{6.0}
\def\xMidM{13.25}   
\def\xLongM{21.0}

\def\xMidA{11.0}
\def\xMidB{15.3}

\def\xLongA{18.5}
\def\xLongB{23.5}

\def\yLone{0.6}
\def\yLtwo{2.4}
\def\yLthree{4.2}
\def\yLfour{6.0}
\def\yLfive{7.8}
\def\yLsix{9.6}

\def\laneHalf{0.75}

\pgfmathsetmacro{\bandBot}{\yLone-\laneHalf-0.2}
\pgfmathsetmacro{\bandTop}{\yLsix+\laneHalf+0.15}

\fill[laneBg, rounded corners=5pt]
    (\xShortL-0.25, \bandBot)
    rectangle (\xShortR+0.25, \bandTop);
\fill[laneAlt, rounded corners=5pt]
    (\xMidL-0.25,   \bandBot)
    rectangle (\xMidR+0.25,   \bandTop);
\fill[laneBg, rounded corners=5pt]
    (\xLongL-0.25,  \bandBot)
    rectangle (\xLongR+0.25,  \bandTop);

\draw[black!15, line width=0.5pt]
    ({(\xShortR+\xMidL)/2},  \bandBot) -- ({(\xShortR+\xMidL)/2},  \bandTop);
\draw[black!15, line width=0.5pt]
    ({(\xMidR+\xLongL)/2},   \bandBot) -- ({(\xMidR+\xLongL)/2},   \bandTop);

\node[phase header] at ({(\xShortL+\xShortR)/2}, \bandTop+0.45) {Short-Term};

\node[phase header] at ({(\xMidL+\xMidR)/2},   \bandTop+0.45) {Mid-Term};

\node[phase header] at ({(\xLongL+\xLongR)/2}, \bandTop+0.45) {Long-Term};

\node[lane label] at (\xShortL-0.6, \yLsix)   {L6 Application};
\node[lane label] at (\xShortL-0.6, \yLfive)  {L5 Orchestration};
\node[lane label] at (\xShortL-0.6, \yLfour)  {L4 Interface};
\node[lane label] at (\xShortL-0.6, \yLthree) {L3 Context};
\node[lane label] at (\xShortL-0.6, \yLtwo)   {L2 Inference};
\node[lane label] at (\xShortL-0.6, \yLone)   {L1 Physical};



\node[bar=axiom]
    (L6gov)  at (\xMidM,  \yLsix) {Governance kernel};
\node[barp=lawIII]
    (L6dist) at (\xLongA+0.5, \yLsix) {Distributed agent};
\node[barp=axiom]
    (L6fed)  at (\xLongB, \yLsix) {Federated agent\\[-2pt]networks};

\node[bar selected=lawIII]
    (L5proto) at (\xShortM, \yLfive) {Protocol unification};
\node[barp=lawIII]
    (L5mcp)   at (\xMidA,   \yLfive) {MCP + A2A\\[-2pt]integration};
\node[barp=axiom]
    (L5cap)   at (\xMidB-0.5,   \yLfive) {Capability\\[-2pt]negotiation};
\node[bar=lawIII]
    (L5os)    at (\xLongM,  \yLfive) {Agent OS kernel};

\node[bar=lawIII]
    (L4bench)   at (\xShortM, \yLfour)
    {Agent runtime\\[-2pt]benchmarks};
\node[barp=lawI]
    (L4interop) at (\xMidA,   \yLfour)
    {Cross-framework\\[-2pt]interop};
\node[barp=axiom]
    (L4sec)     at (\xMidB-0.5,   \yLfour)
    {Agent security\\[-2pt]architecture};
\node[barp=axiom]
    (L4driver)  at (\xLongA+0.3,  \yLfour)
    {Physical-world\\[-2pt]interface drivers};
\node[barp=axiom]
    (L4isa)     at (\xLongB,  \yLfour)
    {Intelligent ISA};

\node[bar selected=lawII]
    (L3comp)    at (\xShortM, \yLthree)
    {Context compilation\\[-2pt]algorithms};
\node[barp=lawII]
    (L3tier)    at (\xMidA,   \yLthree)
    {Semantic page\\[-2pt]replacement};
\node[barp=axiom]
    (L3mem)     at (\xMidB-0.7,   \yLthree)
    {Consistency-controlled\\[-2pt]shared memory};
\node[bar=lawII]
    (L3persist) at (\xLongM,  \yLthree)
    {Persistent context\\[-2pt]stores (L3--L5)};

\node[bar selected=lawI]
    (L2kv)       at (\xShortM, \yLtwo+0.50)
    {KV cache optimization};
\node[bar=lawI]
    (L2sparse)   at (\xShortM, \yLtwo-0.50)
    {Sparse attention\\[-2pt]optimization};
\node[barp=lawI]
    (L2prefix)   at (\xMidA+0.2,   \yLtwo)
    {Heterogeneous model\\[-2pt]collaboration};
\node[barp=lawI]
    (L2semcache) at (\xMidB-0.2,   \yLtwo)
    {Prefix caching\\[-2pt]standardization};
\node[bar=axiom]
    (L2substrate) at (\xLongM,  \yLtwo)
    {Substrate-independent\\[-2pt]compute stack};

\node[bar=lawI]
    (L1sparse) at (\xShortM, \yLone)
    {Sparse attention\\[-2pt]kernels};
\node[bar=axiom]
    (L1alt)    at (\xLongM,  \yLone)
    {Alternative inference\\[-2pt]substrates};


\draw[dep arrow] (L2kv.north) -- (L3comp.south);

\draw[dep arrow] (L3comp.east) -- (L3tier.west);

\draw[dep arrow] (L3mem.east) -- (L3persist.west);

\draw[dep arrow dashed]
    (L3tier.south) -- ++(0,-0.55)
    -| (L3persist.south);

\draw[dep arrow] (L5proto.east) -- (L5mcp.west);

\draw[dep arrow] (L5mcp.east) -- (L5cap.west);

\draw[dep arrow dashed] (L5cap.east) -- (L5os.west);

\draw[dep arrow] (L4bench.east) -- (L4interop.west);

\draw[dep arrow] (L4interop.east) -- (L4sec.west);

\draw[dep arrow]
    (L4sec.east) -- ++( 0.20, 0)
    -- ++(0, \yLsix-\yLfour)
    -- (L6gov.east);

\draw[dep arrow] (L6gov.east) -- (L6dist.west);

\draw[dep arrow] (L6dist.east) -- (L6fed.west);

\draw[dep arrow] (L4isa.north) -- (L5os.south);

\draw[dep arrow dashed] (L4driver.north) -- (L5os.south west); 

\draw[dep arrow] (L2prefix.east) -- (L2semcache.west);

\draw[dep arrow] (L2substrate.south) -- (L1alt.north);

\pgfmathsetmacro{\legYA}{\bandBot - 0.65}
\pgfmathsetmacro{\legYB}{\bandBot - 1.15}
\def\legBoxW{2.0}
\def\legBoxH{0.36}
\pgfmathsetmacro{\diagMid}{(\xShortL+\xLongR)/2}
\pgfmathsetmacro{\legColL}{\diagMid - 7.1}
\pgfmathsetmacro{\legColR}{\diagMid + 0.4}

\fill[lawI!22, draw=lawI!60!black, rounded corners=3pt, line width=0.3pt]
    (\legColL, \legYA) rectangle ++(\legBoxW, \legBoxH);
\node[font=\small, anchor=west, text=black!70]
    at (\legColL+\legBoxW+0.15, \legYA+\legBoxH/2)
    {Heur.~I (Semantic Locality)};

\fill[lawII!22, draw=lawII!60!black, rounded corners=3pt, line width=0.3pt]
    (\legColR, \legYA) rectangle ++(\legBoxW, \legBoxH);
\node[font=\small, anchor=west, text=black!70]
    at (\legColR+\legBoxW+0.15, \legYA+\legBoxH/2)
    {Heur.~II (Context Budget)};

\fill[lawIII!22, draw=lawIII!60!black, rounded corners=3pt, line width=0.3pt]
    (\legColL, \legYB) rectangle ++(\legBoxW, \legBoxH);
\node[font=\small, anchor=west, text=black!70]
    at (\legColL+\legBoxW+0.15, \legYB+\legBoxH/2)
    {Heur.~III (Agent Speedup)};

\fill[axiom!22, draw=axiom!60!black, rounded corners=3pt, line width=0.3pt]
    (\legColR, \legYB) rectangle ++(\legBoxW, \legBoxH);
\node[font=\small, anchor=west, text=black!70]
    at (\legColR+\legBoxW+0.15, \legYB+\legBoxH/2)
    {Axiom (Architectural)};

\end{tikzpicture}}%
\caption{Research roadmap organized by ICA layer and time horizon.
  Horizontal lanes correspond to ICA layers L1--L6; columns represent
  short-term (1--2\,yr), mid-term (2--4\,yr), and long-term (4--8\,yr)
  research phases.  Bar colors indicate which heuristic or axiom each topic
  targets: {\color{lawI}Heur.~I} (Semantic Locality, blue),
  {\color{lawII}Heur.~II} (Context Budget, orange),
  {\color{lawIII}Heur.~III} (Agent Speedup, red), and
  {\color{axiom}Axiom} (architectural principles, gray).
  Arrows denote key dependencies between research topics.}
\label{fig:roadmap_swimlane}
\end{figure}

%% file: figures/fig_roadmap_metrics.tex
\begin{figure}[htbp]
  \centering
  \includegraphics[width=0.85\textwidth]{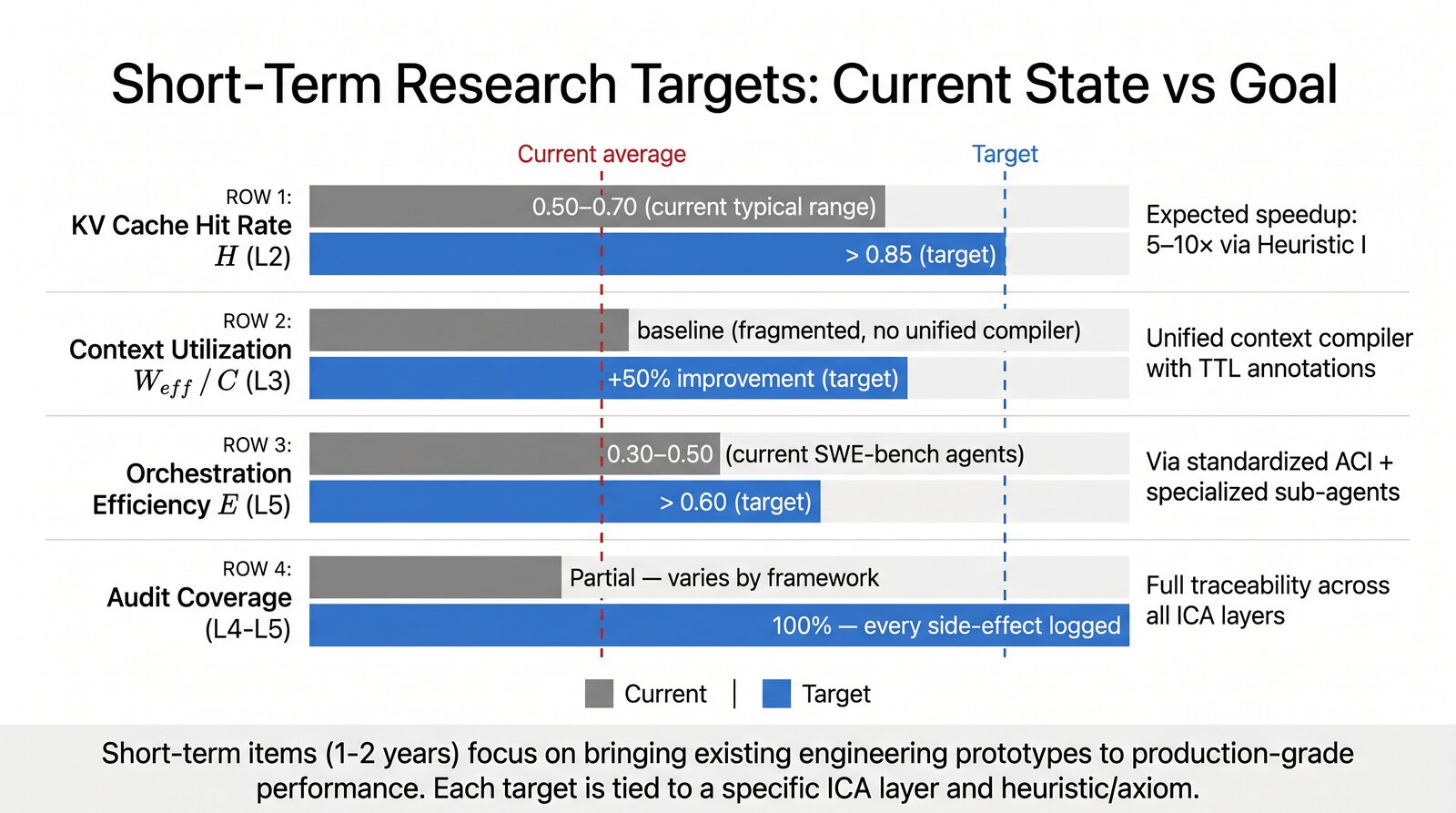}
  \caption{Short-term research targets: current state versus goal for the four key metrics.
  KV cache hit rate $H$ must rise from the current 0.50--0.70 range to above 0.85
  to realize the 5--10$\times$ speedup predicted by Heuristic~I (Semantic Locality).
  Context utilization $W_{\mathrm{eff}}/C$ must improve by at least 50\% through the unified context compiler.
  Orchestration efficiency $E$ must exceed 0.60 to yield meaningful agent speedup under Heuristic~III.
  Audit coverage must reach 100\% to operationalize Axioms~5 and~6 across all ICA layers.}
  \label{fig:roadmap:metrics}
\end{figure}

%% file: figures/fig_roadmap_dependency.tex
\begin{figure}[htbp]
  \centering
  \includegraphics[width=0.85\textwidth]{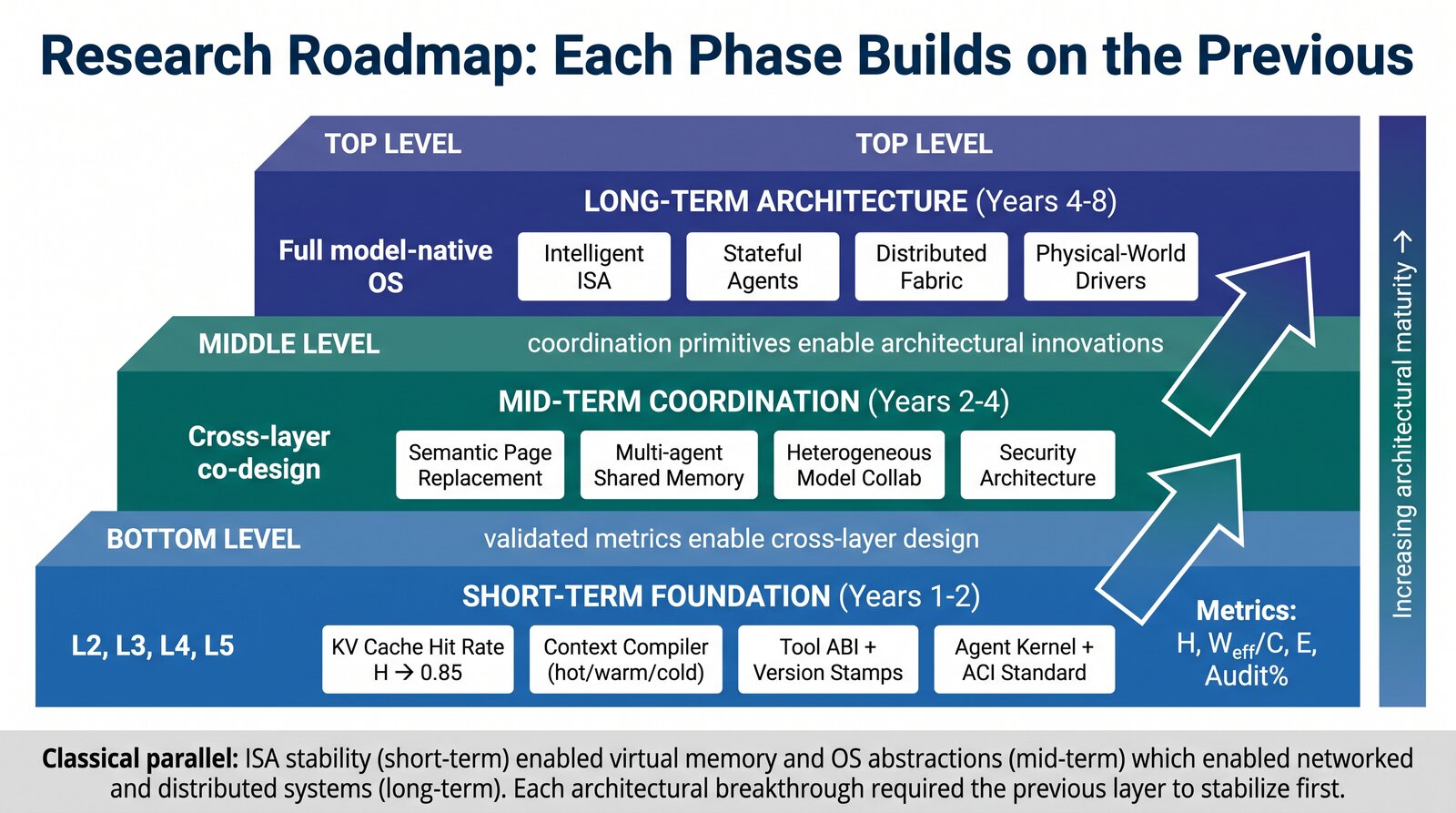}
  \caption{Research enabling dependencies across the three roadmap phases.
  Short-term items produce validated metrics ($H$, $W_{\mathrm{eff}}$, $E$) and interface contracts
  that mid-term work extends into cross-layer co-design.
  Mid-term results create the architectural substrate (semantic memory, shared-state consistency,
  security primitives) on which long-term innovations (Intelligent ISA, stateful agents,
  distributed fabric, physical-world drivers) depend.
  The graph makes explicit that long-term goals are not independent moonshots
  but cumulative extensions of near-term engineering progress.}
  \label{fig:roadmap:dependency}
\end{figure}

%% file: en_sections/section_ethics.tex
\section{Ethics, Safety, and Explainability}
\label{sec:ethics}

Treating large-model systems as computers does not make ethical concerns vanish.  If anything, the architectural lens brings certain ethical questions into sharper relief, because it maps each concern to a specific ICA layer and its corresponding governance mechanism.  In particular, the \emph{weakest-link principle} introduced in Section~\ref{sec:paradigm} implies that a system's overall ethical standing is likewise bounded by its lowest-scoring safety dimension: a system that scores only at the 20th percentile on safety compliance receives an overall ethical rating of~20, not the average of its other dimensions.

\subsection{Permissions and Isolation (L5 Orchestration Layer)}
\label{sec:ethics-permission}

If agents are the new processes, then the questions of what privileges they hold, what secrets they may observe, whether they may access the network, and whether they may mutate persistent state must be governed through explicit security policies, analogous to how an operating system mediates process permissions~\cite{openai_codex_approvals_2026,openai_codex_sandbox_2026,anthropic_claudecode_security_2026,anthropic_claudecode_sandbox_2026,debenedetti2025camel,ironclaw2025}.

Concretely, the governance mechanisms at ICA's L5 layer must address four questions:

\begin{itemize}[nosep]
\item \textbf{Who may create agents?}  This is the analog of process-creation privileges.  OpenAI Codex requires developers to declare agent behavioral boundaries through an explicit \texttt{AGENTS.md} manifest~\cite{openai_codex_agents_md_2026}.
\item \textbf{What resources may an agent access?}  This corresponds to file-system permission bits.  Claude Code adopts a default-read-only posture: every write operation requires explicit user approval~\cite{anthropic_claudecode_security_2026}.
\item \textbf{How are permissions scoped to individual operations?}  This mirrors capability-based security.  CaMeL~\cite{debenedetti2025camel} attaches \emph{capability tokens} to each tool invocation, ensuring that an agent can exercise only the rights explicitly granted for that particular call.
\item \textbf{How is a misbehaving agent terminated?}  This is the analog of process signals and sandbox isolation.  Codex employs OS-level sandboxing so that all agent write operations are confined to a controlled environment~\cite{openai_codex_sandbox_2026}.  IronClaw~\cite{ironclaw2025} extends this model by proposing hardware-enforced isolation boundaries for agentic workloads, drawing a direct parallel to hypervisor-based virtual-machine isolation.
\end{itemize}

Viewed through the lens of the \emph{dual-plane architecture}, these permission controls fall squarely within the remit of the \emph{deterministic control plane}: the \emph{probabilistic execution plane} determines what the model \emph{can} do (reasoning and generation), while the deterministic control plane determines what it \emph{is permitted} to do (approval and isolation).
Figure~\ref{fig:ethics:permissions} maps the four governance questions onto their classical OS counterparts.
\input{figures/fig_ethics_permissions}

\subsection{Privacy and Memory Governance (L3 Memory Layer)}
\label{sec:ethics-privacy}

Privacy concerns concentrate at L3, the memory layer.  Once a system supports long-term state, it must confront four governance questions that have analogs in classical memory management but become substantially more complex in a semantic system.

\paragraph{What should be persisted?}
Not every interaction warrants long-term retention.  Classical operating systems use page-protection bits to distinguish readable, writable, and executable pages; L3 must similarly distinguish between \emph{persistable} information and \emph{session-scoped} information.  A user's code-editing preferences may be safely persisted, whereas temporary credentials or sensitive conversation fragments should be tagged as session-scoped and automatically purged when the session ends.

\paragraph{For how long?}
This raises the question of retention policies.  Traditional file systems manage temporary files through Time-To-Live (TTL) metadata; L3 must assign a TTL and a version number to every memory block.  The design is further complicated by a direct policy conflict: the EU AI Act~\cite{euaiact2026} requires high-risk AI systems to retain operational logs for at least ten years, while the GDPR's Right to Be Forgotten~\cite{gdpr_righttoforget} empowers users to request the deletion of their personal data.  Reconciling these obligations is not a matter of model design; it is a governance decision that L3's retention-policy architects must address explicitly.

\paragraph{Who has the right to delete?}
In a multi-agent setting with shared memory, a memory block written by one agent may be referenced by others.  Deletion cannot proceed naively; it requires a reference-tracking mechanism akin to garbage collection, whereby a block may be physically freed only when all referencing agents have consented or have themselves terminated.  This extends the Observability Axiom (Section~\ref{sec:icam-axioms}): the act of deletion must itself be auditable.

\paragraph{Can summaries be reverse-engineered to reveal sensitive information?}
L3's context compiler routinely compresses raw inputs into summaries to conserve context-window capacity.  However, summaries may retain enough semantic signal that sensitive attributes, such as personal identities and financial data, can be inferred.  This demands that L3's semantic address contracts enforce four properties: \emph{minimal retention} (store only task-essential information), \emph{cascading deletion} (when raw data is removed, all derived summaries are purged together), \emph{access isolation} (memory belonging to different users must not be cross-accessible), and \emph{encrypted storage}~\cite{longmemeval2024,navaie2025privacy}.
Figure~\ref{fig:ethics:privacy} illustrates the four governance gates applied to every memory block entering L3.
\input{figures/fig_ethics_privacy}

\subsection{Explainability: From Explaining Models to Explaining Systems (Full-Stack)}
\label{sec:ethics-explainability}

Traditional LLM explainability research focuses on interpreting a model's internal representations, including the roles of attention heads and the semantics of individual neurons.  From ICA's perspective, however, the more operationally actionable unit of explanation for an agent system running atop a complete model-native computing architecture lies not inside the model but in the system's \emph{behavioral trace}.  This insight is the very core of the Observability Axiom (Section~\ref{sec:icam-axioms}).

ICA accordingly decomposes explainability into the following actionable tiers, each aligned with a specific layer:

\begin{itemize}[nosep]
\item \textbf{L2:} Cache hit/miss logs: which context fragments were reused, which were evicted, and what policy governed each eviction decision.
\item \textbf{L3:} Context-compilation decisions: which information was loaded verbatim versus in summary form, the provenance of each summary, and the compression ratio achieved.
\item \textbf{L4:} Tool-call parameters and results: which tool was invoked, with what arguments, what it returned, and whether a permission denial was triggered.
\item \textbf{L5:} Approval decisions and state commits: which agent requested which privilege at what time, whether the request was granted or denied, and what state mutation was ultimately committed.
\end{itemize}

This notion of \emph{system-level behavioral-trace explainability} is more directly useful than model-internal explainability.  It does not require prying open the model's black box; instead, it demands a faithful record of what the system did at every step, why it did so, and what the outcome was.  Viewed through the dual-plane architecture, this is precisely the core output of the deterministic control plane: an auditable, replayable event stream.
Figure~\ref{fig:ethics:explainability} shows what each ICA layer contributes to this trace.
\input{figures/fig_ethics_explainability}

\subsection{Fairness and Resource Allocation (L2 Inference Layer and L5 Orchestration Layer)}
\label{sec:ethics-fairness}

When multiple users or multiple agents share an inference-serving cluster, fair resource allocation becomes an ethical imperative.  Classical operating systems enforce fair sharing of CPU, memory, and I/O through control groups (cgroups)~\cite{linux_cgroup_2026}; model-native computing systems must analogously enforce per-agent or per-task budgets for context-window capacity, inference time, and monetary cost.

The specific fairness concerns include:

\begin{itemize}[nosep]
\item \textbf{Inference-resource fairness.}  Are GPU resources distributed equitably, or do high-volume users monopolize capacity?
\item \textbf{KV-cache contention.}  Do long-running agents hoard KV-cache entries, starving short tasks of memory, an analog of the classical memory-hog problem?
\item \textbf{Side-effect equity.}  When tool invocations produce side effects (e.g., modifying shared state), are those effects applied uniformly for all users, or do scheduling biases advantage some agents over others?
\end{itemize}

These fairness questions are a direct extension of the Least-Privilege Axiom (Section~\ref{sec:icam-axioms}) into the resource-management plane: an agent should consume no more than its fair share of shared resources, and the system must enforce this bound as a first-class invariant.


%% file: figures/fig_ethics_permissions.tex
\begin{figure}[htbp]
  \centering
  \includegraphics[width=0.88\textwidth]{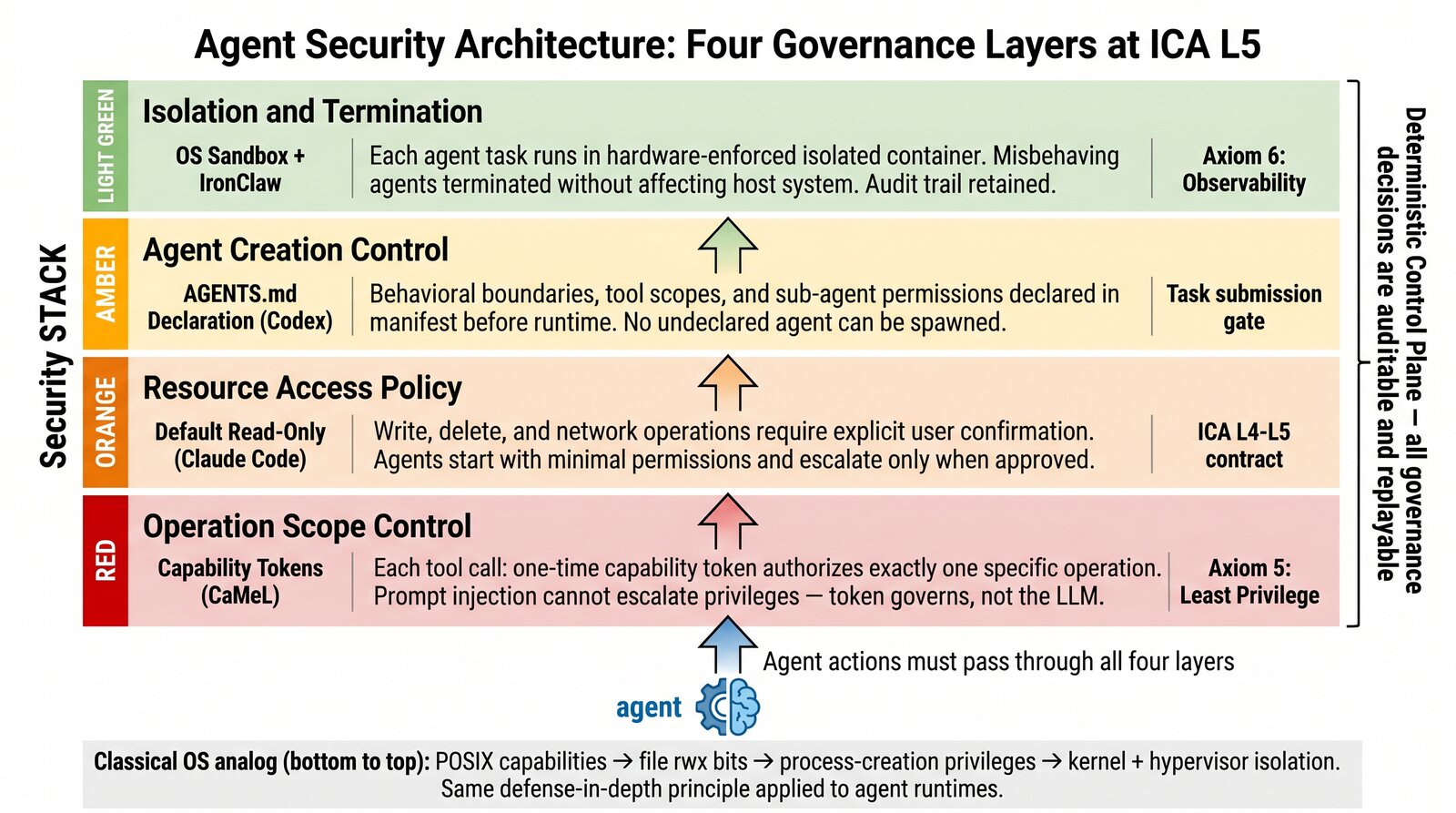}
  \caption{Agent permission model (right) mapped onto the classical OS process permission model (left).
  The four governance questions---who may create agents, what resources they may access,
  how permissions are scoped per operation, and how misbehaving agents are terminated---
  have direct OS analogs in process-creation privileges, file permission bits,
  POSIX capabilities, and sandbox/signal-based termination.
  Under the dual-plane architecture, these controls fall squarely within the
  deterministic control plane: it governs what agents \emph{are permitted} to do,
  while the probabilistic plane determines what they \emph{can} do.}
  \label{fig:ethics:permissions}
\end{figure}

%% file: figures/fig_ethics_privacy.tex
\begin{figure}[htbp]
  \centering
  \includegraphics[width=0.88\textwidth]{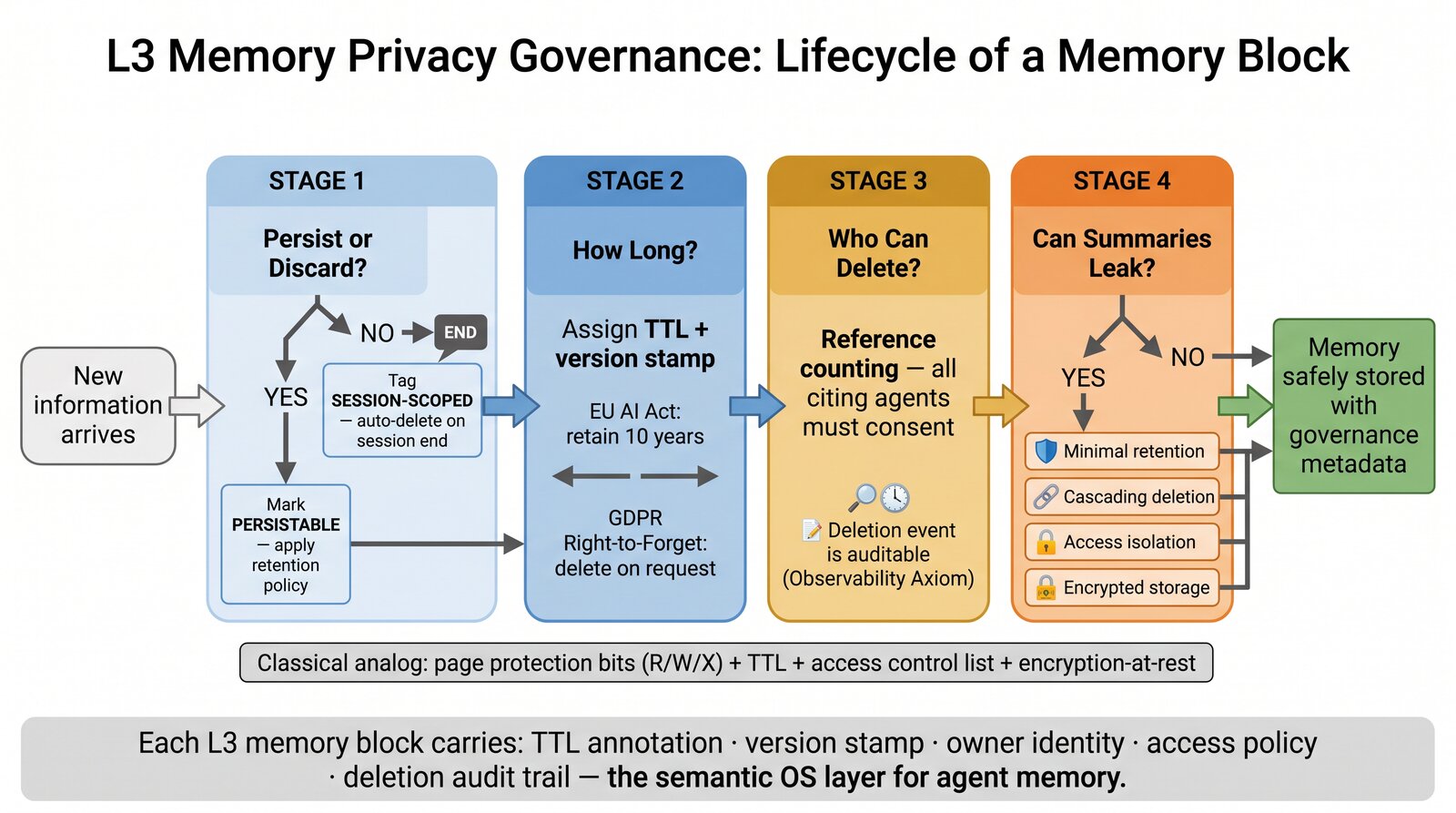}
  \caption{Four sequential governance gates applied to every memory block entering the L3 context layer.
  Gate~1 decides whether information should be persisted or scoped to the current session.
  Gate~2 assigns a Time-To-Live annotation balancing EU AI Act retention requirements
  against GDPR's Right to Be Forgotten.
  Gate~3 enforces reference-counted deletion: physical removal requires all referencing agents to consent.
  Gate~4 tests whether summaries could leak sensitive attributes, triggering cascading deletion,
  access isolation, and encrypted storage if necessary.
  These four gates are the semantic analog of OS page-protection bits.}
  \label{fig:ethics:privacy}
\end{figure}

%% file: figures/fig_ethics_explainability.tex
\begin{figure}[htbp]
  \centering
  \includegraphics[width=0.88\textwidth]{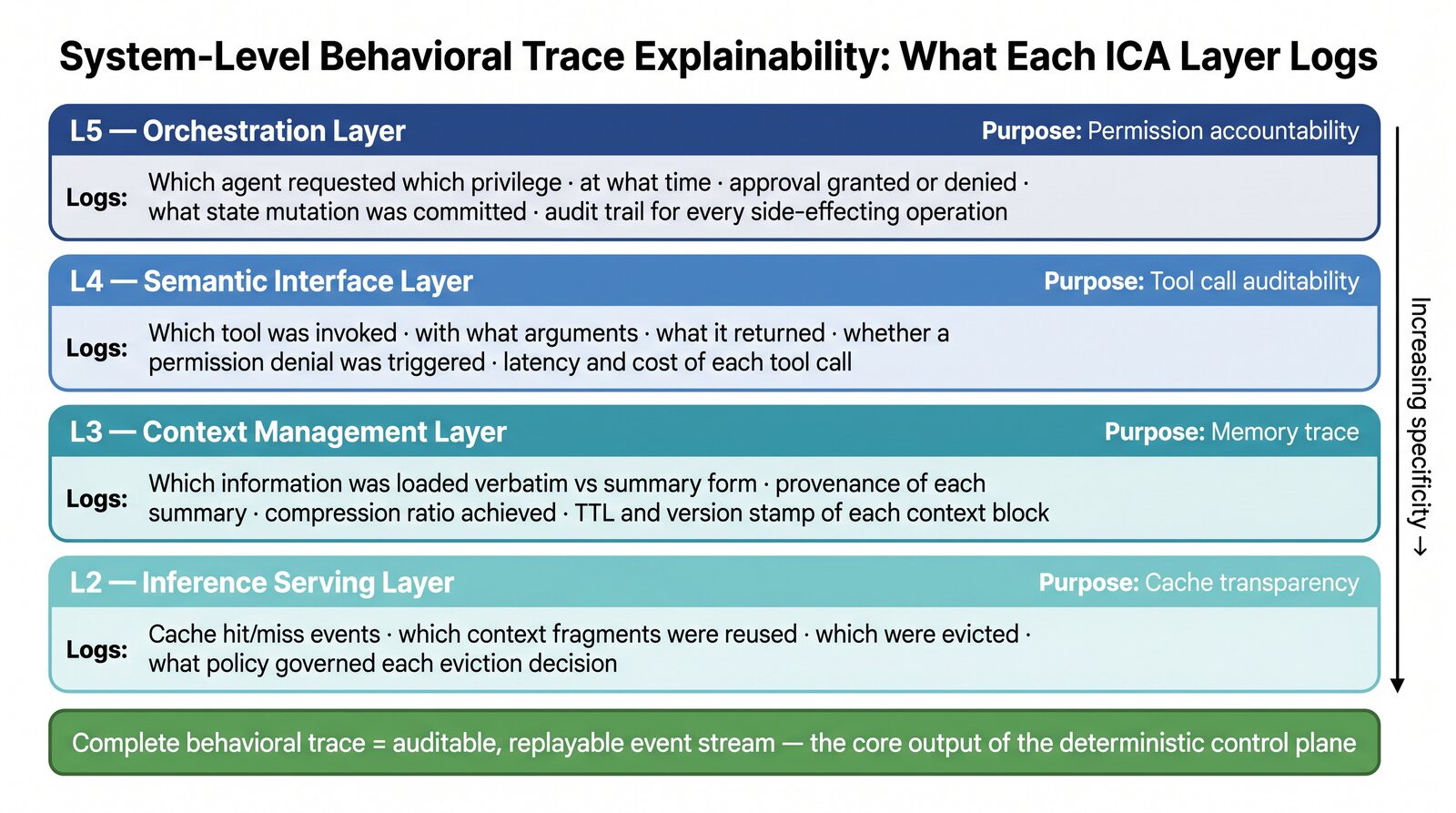}
  \caption{System-level behavioral-trace explainability decomposed by ICA layer.
  L5 (Orchestration) logs privilege requests, approval decisions, and state commits.
  L4 (Semantic Interface) logs tool invocations, arguments, results, and permission denials.
  L3 (Context Management) logs context-compilation decisions, summary provenance, and eviction events.
  L2 (Inference Serving) logs cache hits, misses, and the eviction policy applied to each block.
  Together these four tiers form a complete, replayable behavioral trace---
  the core output of the deterministic control plane---without requiring any
  access to the model's internal representations.}
  \label{fig:ethics:explainability}
\end{figure}

%% file: en_sections/section_conclusion.tex
\section{Conclusion and Outlook}
\label{sec:conclusion}

The central contributions of this paper can be distilled into a unified framework, an architectural resolution, a set of Amdahl-style design heuristics, a paradigm mapping, and a socio-technical analysis.

\subsection{Core Contributions}

\paragraph{Unified framework: the ICA six-layer model.}
We have proposed the Intelligent Computing Architecture (ICA), which defines a model-native computing system as six functional layers (L1--L6), each equipped with explicit interface contracts, design axioms, and performance metrics. ICA is not a mechanical transcription of classical computer architecture; rather, it is a layered abstraction purpose-built for probabilistic execution. Its primary value lies in unifying previously disparate efforts, from AIOS~\cite{mei2024aios} and MemGPT~\cite{memgpt2023} to MCP~\cite{mcp_intro_2026} and PagedAttention~\cite{kwon2023pagedattention}, under a single, shared vocabulary that makes their interdependencies visible and their design trade-offs explicit.

\paragraph{Architectural resolution: the dual-plane architecture.}
We have identified the root cause of the enduring ``Is an LLM a CPU or an OS?'' metaphor clash and resolved it through the dual-plane architecture. The \textit{probabilistic execution plane} (model inference) captures what the system \emph{can} do; the \textit{deterministic control plane} (schedulers, approvers, audit logs within the agent runtime) governs what the system \emph{should} do. This separation reconciles ArbiterOS's ``Probabilistic CPU''~\cite{arbiteros2025}, AIOS's ``kernel''~\cite{mei2024aios}, and AgentOS's ``reasoning kernel''~\cite{agentos2025}: they are not contradictory claims about a single component but complementary perspectives on two distinct planes within the same system.

\paragraph{Organizing heuristics: three Amdahl-style design models.}
The Semantic Locality ($S = 1/((1-H) + H \cdot \alpha^{-1})$), Context Budget ($W_{\mathrm{eff}} = C \cdot \bar{\beta} \leq C$), and Agent Speedup ($S_{\mathrm{agent}} = 1/((1-F) + F/(N \cdot E))$) heuristics provide model-native computing with back-of-envelope guardrails of the same \emph{form} that Amdahl's Law and the Roofline model have long offered classical architecture. These heuristics are isomorphic to Amdahl's Law; their contribution is not mathematical novelty---they are an explicit transplantation---but the provision of a shared \emph{order-of-magnitude design language} for a space that has thus far relied on intuition. We illustrate them with published data and are explicit that they are organizing models rather than validated scaling laws; systematic predict-then-measure verification remains the principal next step.

\paragraph{Paradigm mapping: from silicon to substrate.}
We have systematically mapped the canonical lessons from five decades of CPU evolution, including Dennard scaling and the power wall, big.LITTLE heterogeneous architectures and energy-aware scheduling, specialized coprocessor partitioning, and ISA abstraction and substrate independence, onto the evolutionary trajectory of LLM/agent systems. This mapping yields not only historical analogies but two architecturally significant principles: \textbf{substrate independence} (an intelligent operating system should be decoupled from the physical implementation of its compute core) and the \textbf{weakest-link principle} (the reliability of an intelligent system is bounded by its weakest capability dimension).

\paragraph{Socio-technical analysis: the industry structure of intelligent computing.}
Drawing on the disaggregation histories of the semiconductor and personal-computer industries, we have analyzed the structural dynamics of the emerging AI stack. The fundamental property of model weights as infinitely copyable software stands in sharp contrast to the physical moats of chip fabrication, driving a convergence toward a fabless AI industry structure comprising specialized model designers, AI foundries, and OS-level platform gateways.

\subsection{Positioning and Boundaries}

We explicitly acknowledge the following limitations.

\paragraph{Analogies are not isomorphisms.}
A foundation model is not a deterministic CPU, semantic retrieval is not exact addressing, long-term memory is not lossless paging, and an agent runtime is not a mature operating system. Yet these very gaps, the places where the analogy breaks down, define the most fertile ground for future research: new ABIs for semantic systems, new operating system abstractions for uncertain reasoning, new virtual memory designs for persistent state, and new commit protocols for multi-agent coordination.

\paragraph{Incremental, not inaugural.}
This paper does not claim to be the first to draw parallels between large language models and computer architecture. Ge et al.~\cite{ge2023llmos}, L2MAC~\cite{holt2024l2mac}, and Mi et al.~\cite{mi2025llmagentscomputersystems} have mapped several key routes, and MemGPT~\cite{memgpt2023} and MemOS~\cite{memos2025} have already systematized memory subsystems to a high degree. Our contribution lies in being the \emph{first to integrate these independently developed and sometimes mutually conflicting analogies into a single full-stack model-native computing architecture}---a six-layer stack with explicit inter-layer interface contracts, a dual-plane probabilistic/deterministic separation that resolves the CPU-versus-OS metaphor conflict, and Amdahl-style design heuristics. This goes beyond Mi et al.~\cite{mi2025llmagentscomputersystems}, whose von-Neumann-inspired framework maps agent modules but provides no layered interface contracts, design axioms, or quantitative heuristics, and beyond single-plane proposals such as ArbiterOS~\cite{arbiteros2025}.

\paragraph{Visionary survey, not final answer.}
This paper is a visionary survey that aims to provide a research framework for future model-native computing architectures, not a definitive specification. The six-layer decomposition of ICA and the parameterized forms of the three heuristics will almost certainly be revised and refined as the field matures. Our goal is to pose the right questions and offer falsifiable predictions, not to deliver the last word.

\paragraph{The collaboration paradox remains uncaptured.}
Our framework does not yet fully model the structural constraints of human--AI collaboration. Industry evidence reveals that engineers use AI for approximately 60\% of their work yet can fully delegate only 0--20\% of tasks~\cite{anthropic2026agenticreport}. This ``collaboration paradox'' suggests that the task-submission contracts at the L5--L6 interface must incorporate a \emph{delegation confidence} dimension, reflecting task verifiability and organizational context dependence, that lies beyond the current ICA model's scope.

\subsection{Future Directions}

If the past decade focused on training stronger foundation models, the next decade's central question may be: how to organize foundation models, contextual memory, tool interfaces, and agent runtimes into a truly programmable, scalable, and governable model-native computing architecture.

The answer is likely to emerge from the convergence of advances across multiple fronts simultaneously:
\begin{itemize}[nosep]
    \item At L2, inference serving optimizations such as KV cache management, quantization, and speculative decoding are making model serving ever more efficient.
    \item At L3, context compilation and memory management are beginning to give agents genuine working memory.
    \item At L4, protocol standardization (MCP, A2A) is shifting tool and agent interconnection from point-to-point integration to a bus-based paradigm.
    \item At L5, agent operating systems are moving task decomposition, permission control, and failure recovery from hand-coded scripts to platform-native primitives.
    \item At L6, workflow automation is enabling end users to define complex tasks in natural language.
\end{itemize}

The economic evidence is suggestive, though drawn from a single vendor survey of customer self-reports. TELUS reportedly created over 13,000 customized AI solutions, saved more than 500,000 hours, and accelerated engineering code delivery by 30\%. CRED reportedly doubled execution speed by shifting developers to higher-value work. Approximately 27\% of AI-assisted tasks reportedly consist of work that would not have been done at all without AI~\cite{anthropic2026agenticreport}. These self-reported figures suggest that model-native computing may not merely accelerate existing work but could expand the frontier of economically feasible tasks; systematic, independently audited economic studies are needed to confirm this.

At a more fundamental level, the paradigm discussion in Section~\ref{sec:roadmap} argues that the intelligent computing stack must prepare to transcend silicon, from neuromorphic chips and biological substrates to quantum processors, and to expand from purely code-based interaction to physical-world interfaces. These paradigm-level shifts will demand fundamental evolution in intelligent ISA contracts, evaluation benchmarks, and OS scheduling models. Meanwhile, the analysis in Section~\ref{sec:sociotechnical} shows that this technical evolution is accompanied by an industrial disaggregation: from vertically integrated AI platforms toward specialized layering (fabless model design, AI foundries, and OS-level platform gateways), a pattern that mirrors the historic disaggregation of the semiconductor and PC industries.

This paper offers a systematic research framework for the road ahead: a layered model (ICA), a set of design axioms, three Amdahl-style design heuristics, a paradigm mapping, and a research roadmap spanning near-term optimizations to long-term architectural vision. We hope this framework will help researchers and engineers identify system-level questions worth asking and discover verifiable design principles amid the rapidly evolving model-native computing ecosystem.